\documentclass[12pt]{ociamthesis}  % default square logo 
\usepackage{booktabs}
\usepackage{verbatim}
\usepackage{amsgen,amsmath,amstext,amsbsy,amsopn,tikz,amssymb,amsthm}
\usepackage{mathtools}
\usepackage{fancyhdr}
\usepackage{dsfont}
\usepackage{listings}
\usepackage{times}
\usepackage{epsfig}
\usepackage{graphicx}
\usepackage{natbib}
\usepackage{mleftright}

\usepackage{nicefrac} 
\usepackage{array,multirow}
\graphicspath{ {./figures/} }
\usepackage{caption}
\usepackage{subcaption}
\usepackage[ruled,noend]{algorithm2e}
\usepackage{algorithm}
\usepackage{algorithmic}
\usepackage{enumitem}
\usepackage{color,soul}
\usepackage{cancel}
\usepackage{csquotes}
\usepackage{etoc}
\usepackage{setspace}
\usepackage{thmtools}
\usepackage{refcount}  % For \getrefnumber
\usepackage{xparse} 
\usepackage{mathrsfs}
\usepackage{bbm}

\usepackage[breaklinks,hidelinks]{hyperref}
\usepackage[capitalise,noabbrev,nameinlink]{cleveref}

\usepackage{siunitx}
\usepackage{xcolor}
\usepackage{minitoc}
\dominitoc% <-- Get it ready

\definecolor{codegreen}{rgb}{0,0.6,0}
\definecolor{codegray}{rgb}{0.5,0.5,0.5}
\definecolor{codepurple}{rgb}{0.58,0,0.82}
\definecolor{backcolour}{rgb}{0.95,0.95,0.92}
\lstdefinestyle{mystyle}{
    backgroundcolor=\color{backcolour},   
    numberstyle=\tiny\color{codegray},
    basicstyle=\footnotesize,
    breakatwhitespace=false,         
    breaklines=true,                 
    captionpos=b,                    
    keepspaces=true,                 
    numbers=left,                    
    numbersep=6pt,                  
    showspaces=false,                
    showstringspaces=false,
    showtabs=false,                  
    tabsize=2
}
\theoremstyle{plain}
\newtheorem{theorem}{Theorem}[chapter]
\newtheorem{proposition}[theorem]{Proposition}
\newtheorem{lemma}[theorem]{Lemma}
\newtheorem{corollary}[theorem]{Corollary}
\theoremstyle{definition}
\newtheorem{definition}[theorem]{Definition}
\newtheorem{assumption}[theorem]{Assumption}
\newtheorem{condition}[theorem]{Condition}
\newtheorem{example}[theorem]{Example}
\theoremstyle{remark}
\newtheorem{remark}[theorem]{Remark}

\newcommand{\bill}[1]{\rm {\raggedright\color{red}\textsf{#1}}} 
\DeclarePairedDelimiter\abs{\lvert}{\rvert}%
\DeclarePairedDelimiter\norm{\lVert}{\rVert}%

\DeclareMathOperator*{\argmax}{arg\,max}
\DeclareMathOperator*{\argmin}{arg\,min}

\usepackage{latexsym}
\DeclareFontFamily{OT1}{pzc}{}
\DeclareFontShape{OT1}{pzc}{m}{it}{<-> s * [1] pzcmi7t}{}
\DeclareMathAlphabet{\mathpzc}{OT1}{pzc}{m}{it}

\title{Learning Optimal and Sample-Efficient Decision Policies with Guarantees}   %note \\[1ex] is a line break in the title
\author{Daqian Shao}             %your name
\college{Wolfson College}  %your college
\degree{DPhil in Computer Science}     %the degree
\degreedate{Trinity 2025}
\begin{document}
\maketitle                  % create a title page from the preamble info
\doublespacing

%this baselineskip gives sufficient line spacing for an examiner to easily
%markup the thesis with comments
% \baselineskip=18pt plus1pt

%set the number of sectioning levels that get number and appear in the contents
\setcounter{secnumdepth}{3}
\setcounter{tocdepth}{2}

% \include{dedication}        % include a dedication.tex file
% \include{acknowlegements}   % include an acknowledgements.tex file
% \include{abstract}          % include the abstract

% \hypersetup{colorlinks=false, linkcolor=black}

% \pagenumbering{roman}
% \include{chapters/abstract}
% % \include{chapters/Declaration}
% \include{chapters/acknowlegements}
% \tableofcontents            % generate and include a table of contents
% %now include the files of latex for each of the chapters etc
% \listoffigures
% % \listoftables

% % \include{chapters/abbreviations}
% % list of symbols

% \marta{Check and fix title page, compare with the official guidance}

\begin{romanpages}          % start roman page numbering
\chapter*{Acknowledgements}

First and foremost, I would like to express my greatest gratitude to my supervisor, Prof. Marta Kwiatkowska. Without her, this thesis would not have been possible. I am truly grateful for her continuous care and advice throughout my DPhil, while being an extremely responsible supervisor and mentor.

Secondly, I would like to thank my colleagues and collaborators who I am fortunate enough to work with during my DPhil journey. To the colleagues in the QAVAS group, thank you for making my DPhil memorable and for sharing this DPhil journey. To my talented collaborators, I have learnt so much from each of you. Thank you for all the insightful meetings and intriguing discussions that shaped my path as a researcher.

I gratefully acknowledge that this work was supported by the EPSRC Prosperity Partnership FAIR (grant number EP/V056883/1). I acknowledge funding from the Turing Institute and Accenture collaboration. 
The topics studied in this thesis are aligned with my supervisor's projects FUN2MODEL (grant agreement No.~834115, funded by the ERC under the European Union’s Horizon 2020 research and innovation programme) and 
ELSA: European Lighthouse on Secure and Safe AI project (grant agreement
No. 101070617 under UK guarantee).

Now, to my parents Dr Yan Shao and Ms. Wenhong Tian. I am forever grateful for all your unwavering support throughout my long studying life. You inspire me to be passionate about the things I do and to be fearless in pursuing my passion. To my other family members, thank you for all the love and joy, encouraging me to carry on.

Finally, to Dr Sichen Liu. I cannot imagine this DPhil journey without you. Thank you for being an important part of my life through the ups and downs, taking care of me and supporting me. It is my privilege to share this journey with you.
\begin{abstract}
The paradigm of decision-making has been revolutionised by reinforcement learning and deep learning due to their ability to learn intelligent agents in unknown environments. Although this has led to significant progress in domains such as robotics, healthcare, and finance, the use of reinforcement learning in practice is challenging, particularly when learning decision policies in high-stakes applications that may require guarantees. Traditional reinforcement learning algorithms rely on online interactions with the environment to gather data for policy learning and are often data-inefficient. This is problematic in scenarios where online interactions are costly, dangerous, or infeasible. In this thesis, we work in the setting of offline reinforcement learning, which assumes access to fixed sets of pre-collected data, and develop algorithms for learning effective policies with provable guarantees. However, learning from offline datasets is hindered by the presence of hidden confounders, that is, variables that influence the actions and the outcome but are not observed in the offline dataset. Such confounders can cause spurious correlations in the dataset and can mislead the agent into taking suboptimal or adversarial actions. Firstly, we address the problem of learning from offline datasets in the presence of hidden confounders within the context of reinforcement learning while ensuring sample efficiency and optimality guarantees. To achieve this, we work with instrumental variables (IVs) to identify the causal effect, which we show is an instance of a conditional moment restrictions (CMR) problem. Inspired by double/debiased machine learning, we derive a novel Neumann orthogonal score function and design a cross-fitting regime to obtain a sample-efficient algorithm for solving CMR problems with convergence and optimality guarantees, which outperforms state-of-the-art on IV regression and proximal causal learning problems. Secondly, we relax the conditions on the hidden confounders in the setting of (offline) imitation learning, and adapt our CMR estimator to derive an algorithm that can learn effective imitator policies with convergence rate guarantees, which outperforms the state of the art. Finally, we consider the problem of learning high-level objectives expressed in linear temporal logic (LTL) and develop a provably optimal learning algorithm that improves sample efficiency over existing methods. The derived methods have been implemented and are made available in open source. Through evaluation on reinforcement learning benchmarks and synthetic and semi-synthetic datasets, we demonstrate the usefulness of the methods developed in this thesis in real-world decision making.
\end{abstract}
\tableofcontents
\listoffigures
\listoftables
% list of symbols
\end{romanpages}            % end roman page numbering

\newcommand{\realNumber}{\mathbb{R}}
\newcommand{\naturalNumber}{\mathbb{N}}
\newcommand{\integerNumber}{\mathbb{Z}}
\newcommand{\probP}{\mathds{P}}
\newcommand{\expectE}{\mathds{E}}
\newcommand{\distribution}{\mathcal{D}}
\newcommand{\indep}{\perp \!\!\! \perp}

\newcommand{\vertices}{V}
\newcommand{\vertex}{v}
\newcommand{\edges}{E}
\newcommand{\edge}{e}

\newcommand{\MDP}{\mathcal{M}}
\newcommand{\states}{\mathcal{S}}
\newcommand{\state}{s}
\newcommand{\initstate}{\mu_0}
\newcommand{\actions}{\mathcal{A}}
\newcommand{\action}{a}
\newcommand{\confounders}{\mathcal{U}}
\newcommand{\transitions}{\mathcal{P}}
\newcommand{\reward}{r}
\newcommand{\ExpReward}{G}
\newcommand{\labFunc}{L}
\newcommand{\propositions}{\mathcal{AP}}
\newcommand{\alphabets}{\Sigma}
\newcommand{\alphabet}{\nu}
\newcommand{\policy}{\pi}
\newcommand{\discount}{\gamma}
\newcommand{\Value}{V}
\newcommand{\paths}{Paths^\MDP}
\newcommand{\Fpaths}{FPaths^\MDP}
\newcommand{\filtration}{\mathcal{F}}
\newcommand{\MDPpath}{\sigma}
\newcommand{\policySpace}{\Pi}

\newcommand{\productMDP}{\mathcal{M}^\times}
\newcommand{\productStates}{S^\times}
\newcommand{\productState}{s^{\times}}
\newcommand{\ill}{\nu}

\newcommand{\productTransitions}{T^\times}
\newcommand{\productAccept}{\mathcal{F}^\times}
\newcommand{\productActions}{A^\times}
\newcommand{\productAction}{a^\times}
\newcommand{\productReward}{\reward^\times}
\newcommand{\productDiscount}{\gamma^\times}

\newcommand{\automaton}{\mathcal{A}}
\newcommand{\autoStates}{\mathcal{Q}}
\newcommand{\autoState}{\mathpzc{q}}
\newcommand{\autoTransitions}{\Delta}
\newcommand{\autoAccept}{\mathcal{F}}
\newcommand{\LTL}{\varphi}
\newcommand{\Autorun}{\theta}
\newcommand{\word}{\mathcal{W}}
\newcommand{\potential}{\Phi}

\def\mcirc{\mathbin{\scalerel*{\circ}{j}}}

\newcommand{\shifts}{\mathcal{I}}
\newcommand{\shift}{i}
\newcommand{\testD}{\probP_{\textrm{test}}}
\newcommand{\dataset}{\mathcal{D}}
\newcommand{\orthoM}{\psi}

\newcommand{\Var}{\mathrm{Var}}
\newcommand{\Cov}{\mathrm{Cov}}

\newcommand{\fix}{\marginpar{FIX}}
\newcommand{\new}{\marginpar{NEW}}
\newcommand{\feat}{m}
\newcommand{\improve}[1]{{\color{red} #1}}
\newcommand{\nsr}{\textbf{\textsc{NSR}}}
\newcommand{\pr}[1]{ \mathbb{P} \left ( #1 \right )}

\def\[#1\]{\begin{align*}#1\end{align*}}

\NewDocumentCommand{\numberthis}{om}{%
  \IfNoValueTF{#1}{%
    \refstepcounter{equation}\tag{\theequation}%
  }{%
    \tag{#1}%
  }%
  \label{#2}%
}

%%% Repeat Theorems

\newcounter{savetheorem}

% \NewDocumentEnvironment{repeatthm}{m o}{%
%   \setcounter{savetheorem}{\value{theorem}}%
%   \edef\orignum{\getrefnumber{#1}}%
%   \setcounter{theorem}{\numexpr\orignum-1\relax}%
%   \begingroup
%     \renewcommand{\thetheorem}{\hyperref[#1]{\arabic{theorem}}}%
%     \IfValueTF{#2}
%       {\begin{theorem}[#2]}%
%       {\begin{theorem}}%
% }{%
%     \end{theorem}%
%   \endgroup
%   \setcounter{theorem}{\value{savetheorem}}%
% }

\NewDocumentEnvironment{repeatthm}{m o}{%
  \setcounter{savetheorem}{\value{theorem}}%
  \edef\orignum{\getrefnumber{#1}}%
  \begingroup
    \renewcommand{\thetheorem}{\orignum}%
    \IfValueTF{#2}
      {\begin{theorem}[#2]}%
      {\begin{theorem}}%
}{%
    \end{theorem}%
  \endgroup
  \setcounter{theorem}{\value{savetheorem}}%
}

\NewDocumentEnvironment{repeatlemma}{m o}{%
  \setcounter{savetheorem}{\value{theorem}}%
  \edef\orignum{\getrefnumber{#1}}%
  \begingroup
    \renewcommand{\thelemma}{\orignum}%
    \IfValueTF{#2}
      {\begin{lemma}[#2]}%
      {\begin{lemma}}%
}{%
    \end{lemma}%
  \endgroup
  \setcounter{theorem}{\value{savetheorem}}%
}

\NewDocumentEnvironment{repeatprop}{m o}{%
  \setcounter{savetheorem}{\value{theorem}}%
  \edef\orignum{\getrefnumber{#1}}%
  \begingroup
    \renewcommand{\theproposition}{\orignum}%
    \IfValueTF{#2}
      {\begin{proposition}[#2]}%
      {\begin{proposition}}%
}{%
    \end{proposition}%
  \endgroup
  \setcounter{theorem}{\value{savetheorem}}%
}

% Repeat Proposition
\NewDocumentEnvironment{repeatcoro}{m o}{%
  \setcounter{savetheorem}{\value{theorem}}%
  \edef\orignum{\getrefnumber{#1}}%
  \begingroup
    \renewcommand{\thecorollary}{\orignum}%
    \IfValueTF{#2}
      {\begin{corollary}[#2]}%
      {\begin{corollary}}%
}{%
    \end{corollary}%
  \endgroup
  \setcounter{theorem}{\value{savetheorem}}%
}
\chapter{Introduction}\label{chapter:intro}

Reinforcement learning (RL) and deep learning (DL) have revolutionised decision-making by enabling intelligent agents to learn optimal policies in unknown environments. This is achieved through either online interactions with the environment or offline pre-collected data, without relying on predefined rules or human supervision. RL algorithms allow agents to discover the sequence of actions that maximise expected cumulative rewards, making them well-suited for complex, sequential decision-making problems. The integration of deep neural networks (DNNs) in deep RL further enhances this capability by allowing agents to learn directly from high-dimensional and unstructured data such as images or sensor readings, without manual feature engineering. This has led to breakthroughs in domains such as robotics, healthcare, and finance, where deep RL agents can make decisions based on raw state inputs and learn intricate strategies that surpass human-level performance. However, many fundamental challenges remain in using RL for real-world decision making, including learning from \textit{offline datasets}, specifying \textit{rewards} that align with the learning objective, and designing \textit{sample-efficient} algorithms.

Traditional RL algorithms learn policies through trial and error,
which often require a vast amount of data from online interactions with the environment to learn an optimal policy, making them data-inefficient compared to other machine learning techniques. Although this problem is not severe for (video) games and simulated robotic environments, it is detrimental in many real-world applications where online interactions with the environment are costly, dangerous, or simply infeasible, such as autonomous driving in robotics, credit and loan applications in finance, and medical treatment in healthcare. In these scenarios, \textit{sample-efficient} RL algorithms, \bill{which can learn decision policies with a relatively small number of training samples,} are desirable. If interactions with the environment are infeasible, previously collected \textit{offline datasets}, such as execution trajectories and logs, can be used to learn optimal policies. This learning paradigm is known as \textit{offline RL}. Most of the offline RL literature focuses on tackling the \textit{distribution shift} problem, where the learnt policy becomes unreliable if it leads the agent outside the state-action distribution of the offline dataset. We instead focus on a less studied but very important problem, prevalent in real-world applications, where the agent may be confounded by \textit{hidden confounders}, which are variables that influence both the actions and the outcome (for example, the next state and reward), but are not observed in the offline dataset. Such hidden confounders cause spurious correlations between the action and the outcome, where the observed effect of an action on the outcome is not its true causal effect. These spurious correlations can mislead the agent to take suboptimal and even adversarial actions. For example, suppose we have aeroplane ticket sales and pricing data~\citep{Hartford2017DeepPrediction}, and we wish to learn a policy from this offline dataset that maximises revenue. During the holiday season, observational data may contain evidence of a concurrent surge in ticket sales and prices, which may cause the learning algorithm to learn an incorrect policy that higher ticket prices will always drive higher sales. 

In the presence of hidden confounders, the ability to learn optimal policies heavily depends on structural assumptions regarding the environment and the offline dataset. To learn the optimal policy by identifying the true causal effects of actions in the environment, additional assumptions are necessary for the environment and the offline dataset~\citep{Shpitser2008}. One practical and popular assumption is the existence of \textit{instrumental variables} (IVs)~\citep{wright1928}, which are heterogeneous random variables that affect the action, but not the outcome. These IVs have been observed and used extensively in many fields, such as econometrics, drug testing, and social sciences. In the aeroplane ticket example, supply cost shifters (e.g., fuel price) are instrumental variables, as their variations are independent of the
demand for aeroplane tickets and affect sales solely via
ticket prices. More broadly, a range of causal inference techniques, such as IV regression and proximal causal learning (PCL), can be framed within the general problem of solving \textit{conditional moment restrictions} (CMRs), and we derive algorithms that solve this problem with provable guarantees on sample efficiency and optimality.

However, despite the widespread use of causal inference methods such as IVs to identify the causal effect, these methods often require strong assumptions on the dataset, which are not guaranteed to be satisfied in real-world applications. Therefore, the causal effect of actions in these scenarios may not be identifiable. Fortunately, if the offline dataset is produced by an expert demonstrator, it might still be possible to imitate and match the performance of the expert by learning from their decision trajectories in the offline dataset. This problem is known as \textit{imitation learning} (IL). Like offline RL, most of the IL literature focuses on addressing distribution shifts through interactive expert queries and inverse RL. However, the hidden confounder problem is also very prevalent in IL, which can causally confuse the imitator policy, leading to unreliable performance. We consider the general setting, where the environment has hidden variables that are not recorded in the offline dataset, whereas the expert, on the other hand, can access parts of the hidden variables due to imperfect environment logging and expert knowledge. In this setting, it is desirable to learn a history-dependent policy that captures information regarding the hidden variables from trajectory histories while mitigating the causal confusion caused by the confounding noise. We will show that this complex IL problem can be cast into a CMR problem, where we develop sample-efficient algorithms with suboptimality guarantees.

The RL algorithms introduced above maximise quantitative rewards. However, the desired behaviour and objectives for the learning agent are often qualitative and high level. Therefore, it is desirable for a system, or for an agent within the system, to learn policies that satisfy these high-level specifications. Specifying appropriate reward functions that accurately capture the objectives in these scenarios is non-trivial and can lead to suboptimal and undesirable policies if not carefully designed. \textit{Linear temporal logic} (LTL) is a temporal logic language that can encode formulae regarding properties of an infinite sequence of logic propositions. LTL is widely used for
the formal specification of high-level objectives for robotics
and multi-agent systems, and it is highly expressive and versatile. For example, infinite-horizon reach-avoidance objectives, sequences of conditional tasks, periodic tasks, and many other safety objectives can all be specified using LTL.

In order to learn an optimal policy that maximises the probability of satisfying LTL objectives using RL, most methods first transform the LTL objective into an automaton and build a product \textit{Markov decision process} (MDP) using the original environment MDP and the automaton. A reward function needs to be constructed for this product MDP to determine when, where, and how much reward to give to the agent. Crucially, such a reward function should guide the agent towards the optimal satisfaction of LTL objectives with theoretical guarantees. However, existing works that provide such optimality guarantees are impractical for two reasons. Firstly, it is unclear how to choose the key hyperparameters that ensure the optimality of the algorithm in practice. Secondly, these algorithms are highly sample-inefficient, rendering them costly or intractable in practice. Therefore, it is important to develop algorithms for LTL learning that are sample-efficient, optimal, and practical.

In this thesis, we tackle the aforementioned problems encountered in RL- and DL-driven decision making in the real world. More specifically, we design practical algorithms that can learn optimal policies in the presence of hidden confounders under different levels of assumptions, and to learn optimal policies with respect to qualitative and high-level LTL objectives, all with optimality and sample efficiency guarantees.

\section{Summary of Contributions}

With the aim of this thesis outlined above, we provide a summary of the contributions made by each main chapter towards fulfilling this aim.

\begin{itemize}
    \item In~\cref{chapter:dmliv}, we consider offline RL and study the problem of learning optimal policies in the presence of hidden confounders when IVs are observed in the dataset. The problem of identifying causal effect through IVs is also known as IV regression, which at its core is a \textit{conditional moment restriction} (CMR) problem. We propose DML-CMR, a sample-efficient and novel estimator that solves general CMR problems and converges at $N^{-1/2}$ rate, where $N$ is the sample size. This leads to $O(N^{-1/2})$ suboptimality for the learnt policy when applying DML-CMR to the offline IV bandit problem, which is a single-step RL problem within a confounded environment using IVs. DML-CMR leverages the \textit{double/debiased machine learning} (DML) framework by deriving a novel, \textit{Neyman orthogonal}, score function for CMR problems, and designing a cross-fitting regime for the DML-CMR estimator. 
    % DML-CMR can be applied to both offline bandit problems and sequential offline RL problems, as long as IVs can be observed at each time step. 
    We experimentally demonstrate that DML-CMR\footnote{Link to codebase for \cref{chapter:dmliv}: \url{https://github.com/shaodaqian/DML-CMR}} outperforms other modern IV regression methods~\citep{Hartford2017DeepPrediction,Bennett2019DeepAnalysis,Singh2019,Xu2020} on a range of IV regression and offline IV bandit benchmarks, including two real-world datasets~\citep{Hill2011,Wyatt2020}. In addition, since DML-CMR can solve general CMR problems, we evaluate DML-CMR on the PCL task, demonstrating superior performance against modern PCL algorithms~\citep{Im2021,Mastouri2021,Xu2021,Kompa2022,Wu2024}.

    \item In~\cref{chapter:il}, we study the problem of learning decision policies from expert trajectories, which is known as \textit{imitation learning} (IL), in the presence of hidden confounders. In this chapter, we relax the explicit assumption in~\cref{chapter:dmliv} that additional IVs are observed in the dataset. We propose a novel and unifying framework for confounded IL, where the expert can have access to parts of the hidden confounders that are not recorded in the dataset. We show that IL under this setting can be transformed into a CMR problem by leveraging trajectory histories as instruments and derive the DML-IL algorithm based on DML-CMR to learn a history-dependent policy. The learnt imitator policy can match the expert policy if the expert trajectory history contains sufficient information regarding the hidden confounder observable to the expert. We prove that the performance gap between our imitator and the expert is of the rate $O(\epsilon T^2)$, where $\epsilon$ is some small optimisation error of the algorithm and $T$ is the horizon of the IL task. We experimentally demonstrate that DML-IL\footnote{Link to codebase for \cref{chapter:il}: \url{https://github.com/shaodaqian/Confounded-IL}} can effectively learn good imitator policies in our confounded IL setting on a range of environments including MuJoCo~\citep{Todorov2012}, and outperforms other confounded IL methods~\citep{Swamy2022,Swamy2022_temporal}.
    
    \item In~\cref{chapter:ltl}, we consider the problem of learning LTL objectives optimally and sample-efficiently using online RL. As explained above, existing methods typically define a product MDP with an automaton, where a reward function is constructed to learn the policy. We propose a novel RL algorithm that is guaranteed to converge to the optimal policy for satisfying LTL objectives utilising a novel product MDP design with a generalised reward structure that improves sample efficiency. We provide theoretical optimality guarantees and theoretical analysis for choosing the key hyperparameters to ensure the algorithm returns an optimal policy in practice. To further improve sample efficiency, we adopt counterfactual imagining, a method to exploit the known structure of the LTL objective by creating imagination experiences through counterfactual reasoning. We directly evaluate our algorithms\footnote{Link to codebase for \cref{chapter:ltl}: \url{https://github.com/shaodaqian/rl-from-ltl}} through a novel integration of probabilistic model checkers, which are formal verification tools that analyse stochastic systems to compute the probability of satisfying formal specifications, within the evaluation pipeline. Our algorithms demonstrate better sample efficiency and training convergence against other RL algorithms that learn LTL objectives~\citep{Hahn2019Omega-regularLearning,Bozkurt2019ControlLearning,Hasanbeig2020DeepLogics} on a range of grid-world environments~\citep{Brockman2016OpenAIGym,Icarte2022RewardLearning}.
\end{itemize}

\section{Thesis Outline}

We now outline the structure of this thesis. In~\cref{chapter:background}, we introduce the preliminary background related to reinforcement learning, paradigms in policy learning, and hidden confounders. In~\cref{chapter:lit}, we provide a literature review, which includes an overview of RL, various aspects of RL relevant to this thesis, and causality frameworks. In~\cref{chapter:dmliv}, we propose a novel algorithm, DML-CMR, to solve CMR problems, which we use to learn policies from confounded datasets with IVs. In~\cref{chapter:il}, we present a unifying framework for imitation learning from confounded datasets and transform this problem into a CMR problem, which can be solved by adopting the DML-CMR algorithm. In~\cref{chapter:ltl}, we present a novel RL algorithm to learn policies that satisfy high-level LTL objectives with optimality guarantees. Finally, in~\cref{chapter:conclusion}, we provide a summary of the thesis findings and propose viable avenues for future research. Formal proofs of theorems and other technical details can be found in the Appendix.

\section{Publications and Statement of Authorship}

This thesis is based on several papers that were published~\citep{Shao2023SampleGuarantees,Shao2024,Shao2025}
or under submission~\citep{Shao2025cmr} during my DPhil studies. In addition, I contributed to other works that are not included in this thesis~\citep{Shao2023STR-Cert:Transformers}. In this section, my contributions to the aforementioned papers in the context of this thesis are outlined.

\cref{chapter:dmliv} is primarily based on~\citet{Shao2024} (published at ICML 2024) and a journal extension of that paper~\citep{Shao2025cmr}, which is currently under submission. In~\citet{Shao2024}, the DML-IV algorithm is proposed, the theoretical results are proven, and experiments are conducted for IV regression and policy learning. I formalised the problem, proposed the Neyman orthogonal score and the DML-IV algorithm, derived and proved the theoretical results, implemented the algorithm, and performed all experimental evaluations. In the journal extension, the algorithm and theory are generalised for solving CMR problems, where additional experiments on the PCL task are carried out. I generalised the Neyman orthogonal score and the algorithm to the CMR framework, extended and proved new theoretical results, and performed additional experimental evaluations.

\cref{chapter:il} is primarily based on~\citet{Shao2025} (SCSL workshop at ICLR 2025 and under review at NeurIPS 2025), where a unifying framework for causal imitation learning is proposed, the corresponding CMR problem for causal IL is derived, theoretical results are proved, and experimental evaluations are performed. I proposed the unifying framework, derived the CMR problem, proved the theoretical results, designed and implemented the algorithm, and conducted the experiments.

\cref{chapter:ltl} is primarily based on~\citet{Shao2023SampleGuarantees} (published at IJCAI 2023), where the novel product MDP and the RL algorithm for LTL learning are proposed with theoretical guarantees and a practical implementation. I proposed the product MDP and the algorithm, implemented it, proved the theoretical results, and conducted experiments.
\chapter{Background}\label{chapter:background}

\minitoc

In this chapter, we cover the essential technical background that underpins the work
in this thesis. The chapter is divided into three sections. The first section introduces reinforcement learning, which builds the foundation for the entire thesis and formalises the concept of policy learning. The next two sections introduce concepts that motivate the concrete problems that we solve in the following chapters. Specifically, the second section covers hidden confounders in the context of structural causal models, with a particular focus on instrumental variables, which provide a way to handle hidden confounders when learning policies. The third section concerns high-level objectives as learning targets for RL.

To provide structure to the concepts introduced in this chapter and clarify the relationships between the subsequent chapters, we categorise the concrete policy learning problems in~\cref{fig:intro/overview} based on their learning objectives and data sources for learning. This maps the characteristics of practical problems encountered in policy learning to different clusters of solutions.

\begin{figure}
    \centering
    \includegraphics[width=1\linewidth]{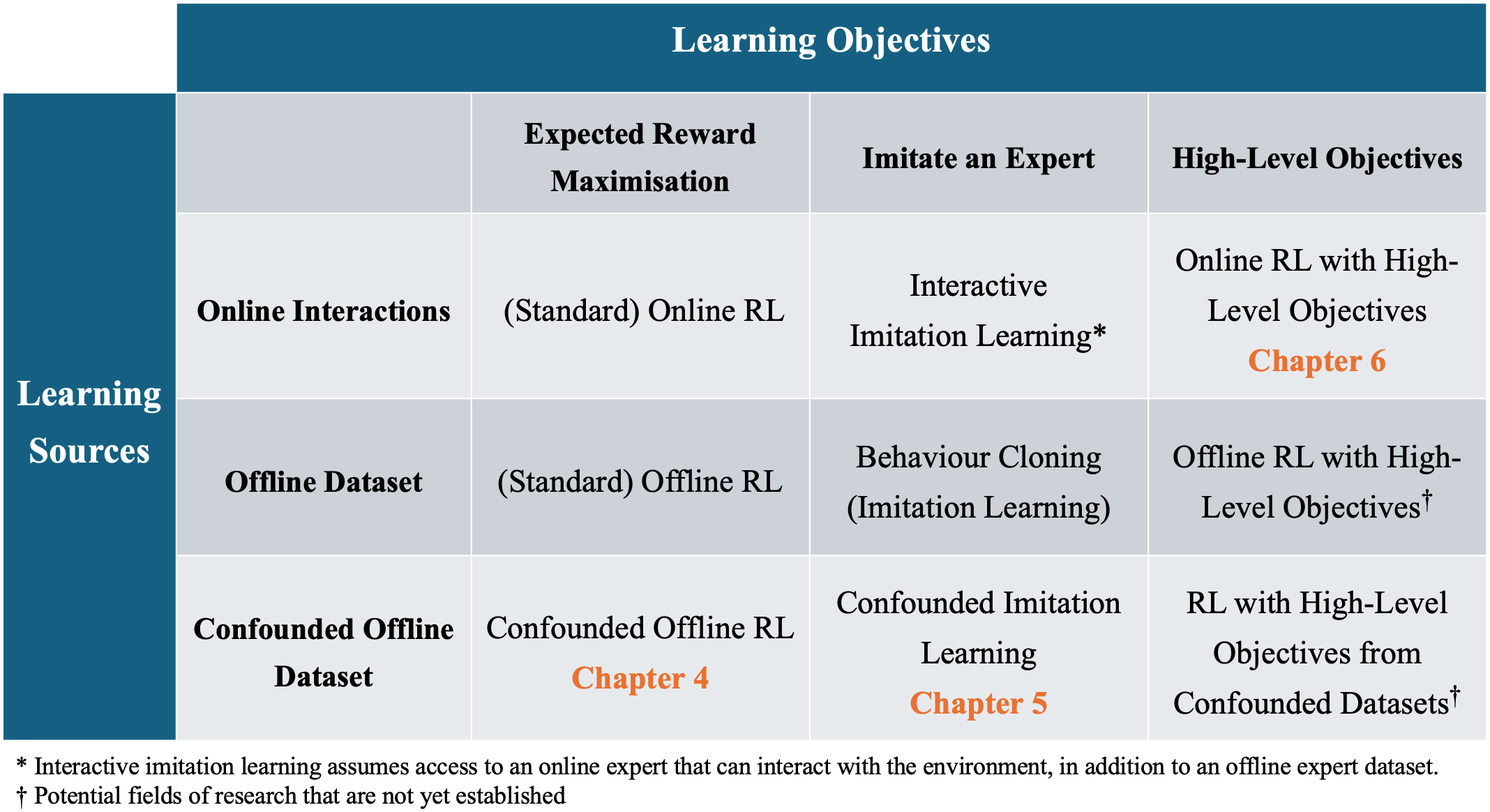}
    \caption[An overview of the types of problems considered in policy learning.]{An overview of the types of problems considered in policy learning, categorised by the learning objective and the data source from which the policy learns. Problems considered in this thesis are highlighted with the corresponding chapters.}
    \label{fig:intro/overview}
\end{figure}

\section*{Notations}
We write uppercase letters such as $X$ to denote random variables and use the corresponding calligraphic letter such as $\mathcal{X}$ to denote the set from which the random variable takes its value. For example, $X\in\mathcal{X}\subseteq\realNumber^{d}$ is a $d$-dimensional real-valued random variable in $(\mathcal{X},\mathcal{B}_\mathcal{X})$, where $\mathcal{B}_\mathcal{X}$ is the Borel algebra on $\mathcal{X}$. We refer to the probability distribution of $X$ given a probability space $(\Omega,\mathcal{F},\probP)$ as the probability measure denoted as $F_X:\mathcal{B}_\mathcal{X}\rightarrow \realNumber$. The support of $X$ is denoted as $supp(X)$, and the set of all distributions supported on $\mathcal{X}$ is denoted as $D(\mathcal{X})$. An observed realisation of $X$ is denoted by a lowercase letter $x$. We abbreviate $\expectE[Y \lvert X=x]$, a realisation of the conditional expectation $\expectE[Y \lvert X]$, as $\expectE[Y \lvert x]$. $[N]$ denotes the set $\{1,...,N\}$ for $N\in\naturalNumber$. We write $\expectE[X\lvert do(A=a)]$ for the expectation of $X$ under \emph{do}  intervention~\citep{Pearl2000causality} of setting $A=a$. We use $\norm{\cdot}_p$ to denote the functional norm, defined as $\norm{f}_p\coloneqq\expectE[\abs{f(X)}^p]^{1/p}$, where the measure is implicit from the context. For a function $f$, we use $f_0$ to denote the true function and $\hat{f}$ an estimator of the true function. We use $O$ and $o$ to denote big-O and little-o notations~\citep{Weisstein2023}, respectively.

\section{Reinforcement Learning}

\bill{In this section, we cover the basics of reinforcement learning (RL)~\citep{Sutton2018ReinforcementIntroduction} and imitation learning (IL). We begin by introducing Markov decision processes, which formulate the potentially stochastic environment in which RL takes place. Next, we elucidate the different paradigms of RL relevant to this thesis, including online RL, offline RL, and IL. We also provide an overview of the key developments in reinforcement learning (RL) algorithms for solving online RL (\cref{lit:online_rl}) and offline RL (\cref{lit:offline_rl}). While online RL and offline RL represent two problem formulations for RL in different settings, the algorithms designed to address these problems can generally be classified into two categories: model-free and model-based. Model-free RL learns policies or value functions directly from experience, without an explicit model of the environment, and relies on trial and error. In contrast, model-based RL first learns a model or representation of the environment's dynamics and subsequently uses this model for policy learning. We will introduce and discuss both classes of algorithms for online RL and offline RL, respectively.}

\subsection{Markov Decision Processes}
\label{prelim:mdp}

We first define Markov decision processes (MDPs), which will later be extended with atomic proposition labels and reward functions. The definitions presented draw from both formal verification and RL literature.

\begin{definition}[Markov decision processes~\citep{Puterman1994}]
\label{def:mdp}
An MDP $\MDP$ is a tuple $(\states,\initstate,\actions,\transitions)$, where $\states$ is a set of states, $\initstate\in D(\states)$ is the initial state distribution, $\actions$ is a set of actions, and $\transitions:\states\times\actions\times\states\rightarrow [0,1]$ is the probabilistic transition function. Let $A(\state)$ denote the set of available actions in state $\state$. Then, for all $\state\in\states$, it holds that $\transitions(\state,\action,\cdot)$ is a valid probability distribution if $\action\in\actions(\state)$ and $\transitions(\state,\action,\cdot)=0$ otherwise.
\end{definition}

An infinite path is a sequence of states $\MDPpath=\state_0,\state_1,\state_2...$, where for all $i\geq 0$, there exists $\action_{i+1}\in\actions(\state_{i})$ such that $\transitions(\state_i,\action_{i+1},\state_{i+1})>0$, and a finite path is a finite such sequence. We denote the set of infinite and finite paths of the MDP $\MDP$ as $\paths$ and $\Fpaths$, respectively. We use $\MDPpath[i]$ to denote $\state_i$, and $\MDPpath[:i]$ and $\MDPpath[i:]$ to denote the prefix and suffix of the path, respectively. Furthermore, we assume self-loops: if $A(\state)=\varnothing$ for some state $\state$, we let $\transitions(\state,\action,\state)=1$ for some $\action\in\actions$ and $A(\state)={\action}$ such that all finite paths can be extended to an infinite one.

A finite-memory policy $\policy$ for $\MDP$ is a function $\policy:\Fpaths\rightarrow D(\actions)$ such that $supp(\policy(\MDPpath))\subseteq \actions(\MDPpath[\text{-1}])$,
where $\MDPpath[\text{-1}]$ is the last state of a finite path $\MDPpath$. A policy $\policy$ is memoryless if it only depends on the current state, that is, $\MDPpath[\text{-1}]=\MDPpath^\prime[\text{-1}]$ implies $\policy(\MDPpath)=\policy(\MDPpath^\prime)$, and a policy is deterministic if $\policy(\MDPpath)$ is a point distribution for all $\MDPpath\in\Fpaths$. For a memoryless policy, we let $\policy(\state)$ represent $\policy(\MDPpath)$ where $\MDPpath[\text{-1}]=\state$.

Let $\paths_\policy\subseteq\paths$ denote the subset of infinite paths that follow policy $\policy$, and we define the probability space $(\paths_\policy,\filtration_{\paths_\policy},\probP_\policy)$ on $\paths_\policy$. Then, for any function $f:\paths_\policy\rightarrow\realNumber$, let $\expectE_\policy[f]$ be the expectation of $f$ over the infinite paths of $\MDP$ following $\policy$.

\subsubsection{Markov Decision Processes with Atomic Proposition Labels}\label{prelim:mdp_ap}

MDP states can also be labelled with atomic propositions, such that we can specify logic properties and specifications regarding the paths generated by the MDP.

\begin{definition}[MDP with atomic proposition labels]
\label{def:mdp_ap}
A Markov decision process (MDP) with atomic proposition labels is a tuple $(\states,\initstate,\actions,\transitions,\propositions,\labFunc)$ that extends a standard MDP~\cref{def:mdp} with $\propositions$, the set of atomic propositions, and $\labFunc: \states\rightarrow2^{\propositions}$, the proposition labelling function that maps each state to a subset of atomic propositions.
\end{definition}

A Markov chain (MC) induced by MDP $\MDP$ with atomic proposition labels and deterministic memoryless policy $\policy$ is a tuple $\MDP_\policy=(\states,\initstate,\transitions_\policy,\propositions,\labFunc)$, where $\transitions_\policy(\state,\state^\prime)=\transitions(\state,\policy(\state),\state^\prime)$. A bottom (sink) strongly connected component (BSCC) of a Markov chain is a set of states $C \subseteq \states$ such that, for all pairs $\state_1,\state_2 \in C$, there exists a path from $\state_1$ to $\state_2$ following the transition function $\transitions_\policy$ (strongly connected), and there is no state $\state^\prime\in S\setminus C$ such that $\transitions_\policy(\state,\state^\prime)>0$ for all $\state\in C$ (sink).

\subsection{Online Reinforcement Learning}\label{background:onlineRL}
Online reinforcement learning~\citep{Sutton2018ReinforcementIntroduction} teaches an agent to select an action from its action space in an unknown environment, in order to maximise rewards over time. The RL agent interacts with the unknown environment and learns from the rewards it receives in the environment. The environment is typically modelled as an MDP $\MDP=(\states,\initstate,\actions,\transitions)$, for which a reward function $\reward:\states\times\actions\times\states\rightarrow\realNumber$ and a discount factor function $\discount:\states\rightarrow (0,1]$ are defined. It is important to note that, in the context of RL, MDPs are sometimes defined with $r$ and $\discount$ included in the tuple for simplicity. Given a memoryless policy $\policy$, at each time step $t$, let the agent's current state be $\state_t$. Then, the action $\action$ following the distribution $\policy(\state_t)$ is chosen, and the next state $\state_{t+1}\thicksim\transitions(\state_t,\action,\cdot)$ together with the immediate reward $\reward(\state_t,\action,\state_{t+1})$ is received from the environment. Then, starting at $\state\in\states$, the expected discounted reward following policy $\policy$ is
\begin{align}
\ExpReward^\policy(\state)=\expectE_\policy[\sum^\infty_{i=0}(\prod_{j=0}^{i-1}\discount(\state_{j}))\cdot\reward(\state_{i},\action_{i},\state_{i+1})\mid \state_0=\state]
\end{align}
where $\prod_{j=t}^{t-1}\coloneqq1$. The agent's goal is to learn the optimal policy $\policy^*$ that maximises the expected discounted reward $\ExpReward$ over the initial state distribution. This quantity is sometimes defined as the value of the policy $\pi$, defined formally as
\begin{align}
    J(\pi)=\expectE_{\policy,\initstate}[\sum^\infty_{i=0}(\prod_{j=0}^{i-1}\discount(\state_{j}))\cdot\reward(\state_{i},\action_{i},\state_{i+1})].
\end{align}

In the various reinforcement learning methods, for which we provide a comprehensive survey in~\cref{lit:rl}, the state-action value function, or the Q function~\citep{Watkins1992Q-learning}, is widely used. The Q function $Q_\policy(\state,\action)$ is defined as the expected discounted reward of taking action $\action$ in state $\state$ and following policy $\policy$ thereafter:
\begin{align}
    Q_\pi(s,a)=\expectE_{\policy}[\sum^\infty_{i=0}(\prod_{j=0}^{i-1}\discount(\state_{j}))\cdot\reward(\state_{i},\action_{i},\state_{i+1})\mid s_0=s, a_0=a].
\end{align}

For example, in Q-learning~\citep{Watkins1992Q-learning}, which is a classic RL algorithm for MDPs with discrete state and action spaces, the optimal policy $\policy^*$ is derived from the optimal Q function $Q^*\coloneqq Q_{\policy^*}$ by selecting the action $\action$ with the highest state-action pair value in each state $\state$. The Q-learning algorithm learns the optimal Q function through the agent's experiences within the environment by adopting an exploration policy. Typically, the $\epsilon$-greedy policy,  which selects the action with the highest Q value $\arg\max_{\action\in\actions}Q(\state,\action)$ with probability $1-\epsilon$ and selects a random action with probability $\epsilon$, is used. At each iteration, the agent's experiences, which include the next state $\state^\prime$ and immediate reward $\reward(\state,\action,\state^\prime)$, are used to update the Q function:
\begin{equation}
    Q(\state,\action)\overset{\alpha}{\gets}\reward(\state,\action,\state^\prime)+\discount(\state)\max_{\action^\prime\in\actions}Q(\state^\prime,\action^\prime)
\end{equation}
where $\alpha$ is the learning rate and $x\overset{\alpha}{\gets}y$ represents $x\gets x+\alpha(y-x)$. It is proven that Q-learning converges to the optimal Q function $Q^*$ in the limit, provided that each state-action pair is visited infinitely often~\citep{Watkins1992Q-learning}, thereby learning the optimal policy.

\subsubsection{Finite-horizon Reinforcement Learning}

In certain areas, specifically for imitation learning, RL is sometimes defined with a finite time horizon $T$ and without a discount factor $\gamma$. In this context, the environment is also modelled as an MDP $\MDP=(\states,\initstate,\actions,\transitions)$, where a reward function $\reward:\states\times\actions\times\states\rightarrow\realNumber$ and a maximum time horizon $T$ are given. Because of the finite time horizon, the sum of the total rewards of a trajectory is finite, so there is no need for a discount factor. With the maximum time horizon $T$, the expected total reward that the agent optimises is
\begin{align}
\ExpReward^\policy(\state)=\expectE_\policy[\sum_{i=0}^{T}\reward(\state_{i},\action_{i},\state_{i+1})\mid \state_0=\state],
\end{align}
and the $Q$-function is defined as
\begin{align}
    Q_\pi(s_t,a_t)=\expectE_\pi[\sum_{i=t}^T \reward(\state_{i},\action_{i},\state_{i+1})].
\end{align}
The finite-horizon formulation is sometimes preferred to the infinite-horizon discounted reward formulation if the RL task is an episodic task with a known endpoint or fixed duration. Moreover, finite-horizon formulation avoids discounting bias, which is the systematic tendency to prioritise short-term rewards over long-term rewards due to the exponential discounting factor in the discounted reward formulation.

\bill{\subsubsection{Online Model-free RL Algorithms}\label{lit:online_rl}}

Online reinforcement learning~\citep{Sutton2018ReinforcementIntroduction} teaches an agent in an unknown environment to select an action from its action space to maximise rewards over time. The online RL agent interacts with the unknown environment and learns from the rewards it receives within that environment. With the development of deep neural networks~\citep{Lecun2015}, these powerful function approximators have been incorporated into RL, and together with effective training algorithms have enabled RL to achieve exceptional results in numerous applications, including playing Atari games~\citep{Mnih2013PlayingLearning,Schrittwieser2020MasteringModel}, the game of Go~\citep{Silver2017}, robotic control and manipulation~\citep{Levine2015End-to-EndPolicies,Kalashnikov2018QT-Opt:Manipulation}, and resource allocation in business processes~\citep{Huang2011ReinforcementManagement}, among others.

We first discuss the various model-free algorithms used for online RL. The first and most direct approach to optimise the RL objective, which is the maximisation of expected rewards, is by employing policy gradient methods~\citep{Sutton1999PolicyApproximation,Kakade2001AGradient}. These methods directly estimate the gradient of the expected reward with respect to the model parameters and perform gradient descent~\citep{Robbins1951AMethod}. \citet{Schulman2015TrustOptimization} then proposed a robust policy gradient variant that guarantees monotonic improvement of the policy, while \citet{Thomas2014BiasAlgorithms} modified the classic policy gradient algorithm to work in the average reward setting. These methods are applicable for general RL problems and can handle continuous state and action spaces, but generally suffer from high variance in gradient estimates, leading to noisy and unstable convergence with poor sample efficiency.

The second class of algorithms involves approximate dynamic programming~\citep{Powell2007ApproximateEdition}, which estimates the expected reward from a state or a state-action pair following a policy. These expected rewards are also referred to as value or Q functions, and with accurate value functions one can recover a near-optimal policy. In fact, the value functions must satisfy the Bellman equation~\citep{Bellman1957AProcess} for dynamic programming, and \citet{Watkins1992Q-learning} proposed Q-learning to find the Q value functions by minimising the error between both sides of the Bellman equation. \citet{Lagoudakis2003Least-SquaresIteration} used temporal-difference learning to improve the performance of the Q function approximation and proved convergence of the algorithm to the optimal policy. A group of algorithms known as fitted Q-iteration~\citep{Ernst2005Tree-BasedWehenkel,Riedmiller2005NeuralMethod} fully minimises the difference between both sides of the Bellman equation at each iteration. \citet{Lin1992Self-improvingTeaching}, instead of performing a single update of the Q function at each iteration, employed a replay buffer to store past agent experiences. This is also known as off-policy RL, as the replay buffer includes data collected from all past policies, which differs from the current policy being updated. To handle high-dimensional and continuous state spaces, \citet{Mnih2013PlayingLearning} proposed Deep Q-Network (DQN), which uses deep neural networks (DNNs) and gradient descent to estimate the Q function by minimising the Bellman error from a replay buffer. For improved training stability, many approaches used target networks for the Q function~\citep{Hausknecht2015,Hasselt2018DeepTriad,Wang2015}, which are less frequently updated, to calculate the Bellman error. 

% The most famous among them is known as Double DQN~\citep{Hausknecht2015}.
% Dueling DQN~\citep{Wang2015} proposed to use separate estimators for the state value function and for the action advantage function, which estimates the value advantage of an action over the average state value. This factorisation allows for generalised learning across actions and leads to good performance on Atari games. Deep Recurrent Q-Learning~\citep{VanHasselt2015} proposed the use of LSTMs to handle time-dependent states and actions in a partially observable MDP. Although these approaches violate the assumptions for the convergence proof of Q-learning, good convergence results have been empirically shown~\citep{Fu2019DiagnosingAlgorithms}.

The third class of algorithms, actor-critic algorithms~\citep{Konda2000Actor-CriticAlgorithms}, combines the basic ideas of the above two approaches by approximating the Q function to provide a better estimate of the policy gradient for policy optimisation. The algorithm Deterministic Policy Gradient (DPG)~\citep{Silver2014DeterministicAlgorithms} is an off-policy actor-critic algorithm
that learns a deterministic target policy from Q function estimations, which are learnt by minimising the Bellman error. It was later extended to use DNNs~\citep{Lillicrap2015} and to a distributed framework~\citep{Barth-Maron2018}. Advantage actor-critic (A2C)~\citep{Mnih2016} proposed to use advantage values, which measure the improvement by taking a particular action compared to the expected state value, for more stable policy gradient updates. Soft actor-critic~\citep{Haarnoja2018SoftActor} extended this approach for continuous action tasks by also maximising the entropy of the policy. Recent advances include a better theoretical sample complexity for global convergence of actor-critic algorithms~\citep{Kumar2024}, and a value-improved actor-critic algorithm~\citep{Oren2024}.

% \citet{Ma2020} extended soft actor-critic to distributional RL with a focus on risk-sensitive RL, where, beyond maximising expected rewards, three specific risk-related metrics are considered: percentile, mean variance, and distorted expectation. 

% Deep Deterministic Policy Gradient (DDPG)~\citep{Lillicrap2015} extends DPG to use DQN with a target network and replay buffer for the estimation of the Q function and can handle continuous action spaces. D4PG~\citep{Barth-Maron2018} further extends DDPG to a distributed framework for large-scale off-policy learning, with a distributional version of the critic update for more stable training.

% \citet{Heess2015LearningGradients} extended the classic actor-critic algorithm to the off-policy setting by adopting a replay buffer and used stochastic value gradients to update the actor. 

Lastly, we discuss methods that constrain the policy update process, which improves the stability and efficiency of RL training. Trust Region Policy Optimisation (TRPO) \citep{Schulman2015TrustOptimization} is a policy optimisation method that ensures each policy update does not change the policy too drastically, thereby preventing performance collapse. To overcome the computational challenges of TRPO, Proximal Policy Optimization (PPO) \citep{Schulman2017ProximalAlgorithms} proposes to use a clipped surrogate objective, which directly restricts how much the new policy can deviate from the old one. More recent improvements of PPO and TRPO include extensions to multi-agent RL~\citep{Yu2021,Liu2024}, extensions to model-based RL~\citep{Janner2019}, and safe RL~\citep{Yao2023,Milosevic2024}.

\subsubsection{Online Model-based RL Algorithms}\label{lit:model-based_RL}

The above methods find the optimal policy without learning the explicit transition function or dynamic model of the environment, where the agents interact with the environment and select actions without knowing the next state after the transition. We refer to this group of algorithms as model-free reinforcement learning. Naturally, there exists another class of RL algorithms that uses explicit estimates of the dynamic model for policy learning, namely model-based reinforcement learning. \citet{Tassa2012SynthesisOptimization} proposed a method to learn only the dynamic model of the environment for planning at test time, using trajectory optimisation. Improvements on this method have also been developed by adopting neural networks as dynamic models~\citep{Nagabandi2017NeuralFine-Tuning}. \citet{Chua2018DeepModels} further improved the performance of model-based RL using a method called probabilistic ensembles with trajectory sampling, while an end-to-end extension~\citep{Amos2018DifferentiableControl} to continuous state and action spaces via a differentiable policy has been proposed. \citet{Hafner2018LearningPixels} developed the PlaNet algorithm that accurately learns a latent dynamic model for image tasks and uses fast online planning algorithms in the latent space.

Another group of online model-based RL methods employs the dynamic model to generate synthetic data, also called model rollouts, which augment the procedure of policy optimisation. The classic Dyna algorithm~\citep{Sutton1991DynaReacting} combines model learning and data generation with Q-learning for policy learning. Many model-free RL algorithms incorporate this data generation functionality to improve sample complexity and accelerate learning~\citep{Gu2016ContinuousAcceleration,Kurutach2018Model-EnsembleOptimization,Clavera2018Model-BasedOptimization}.
% including accelerated deep reinforcement learning for continuous control tasks with learnt models~\citep{Gu2016ContinuousAcceleration}, ensemble dynamic models with the classic trust-region policy optimisation algorithm~\citep{Kurutach2018Model-EnsembleOptimization} and meta-policy optimisation~\citep{Clavera2018Model-BasedOptimization}. 
\citet{Janner2019} later formalised a model-based RL approach with a monotonic improvement guarantee that uses short model-generated rollouts to avoid the model generalisation error for long-term rollouts. 
% In image domains, \citet{Watter2015EmbedImages} adopted a latent dynamics model to generate long-term image trajectories from the latent space for control optimisation. \citet{Kaiser2019Model-BasedAtari} exploited video prediction models to simulate Atari game-plays that dramatically reduced the agent's interaction with the game environment, while 
\citet{Hafner2020MasteringModels} proposed DreamerV2, that learns the policy purely from synthetic data generated from a learnt world model with a discrete latent space. \citet{Schiewer2024} studied hierarchical world models that are abstractions of the environment model across hierarchical levels. 
% New model architectures are also explored for model-based RL. \citet{Wang2024Mamba} proposed to use state space models with Mamba~\citep{Gu2024} architecture to achieve linear memory and computational complexity with respect to the time horizon, as well as improved long-term dependency modelling. 
Theoretical works also exist for model-based RL, with the most recent progress by~\citet{Li2023} who proposed a conservative model-based algorithm that achieved minimax-optimal sample complexity for infinite-horizon MDPs.

Last but not least, the learnt dynamic model can also improve the value estimates used as targets~\citep{Buckman2018Sample-EfficientExpansion} in temporal difference learning and as the critic~\citep{Feinberg2018Model-BasedLearning} in actor-critic algorithms. This is achieved by estimating the uncertainty through model rollouts while measuring the impact of an imperfect dynamics model, ensuring that the performance of the learning algorithm is not degraded.

\subsection{Offline Reinforcement Learning}\label{background:offlineRL}

Although online RL has demonstrated outstanding performance across various fields~\citep{Silver2017,Levine2015End-to-EndPolicies,Kalashnikov2018QT-Opt:Manipulation}, the intrinsic online nature of RL algorithms remains a significant barrier to their wider adoption. Online RL involves repeatedly interacting with the environment using the current policy to collect new experiences, which are then used to improve the policy. However, in many cases, online interaction is impractical due to the high cost of data collection (e.g., robotics, educational agents, healthcare) or safety risks (e.g., autonomous driving, healthcare). Even when online interaction is feasible, relying on previously collected data can be preferable, particularly in complex domains where effective generalisation necessitates large datasets. These challenges motivate the development of offline RL.

The offline RL problem can be considered as a \textit{data-driven} formulation of the RL problem. The goal remains to learn the optimal policy $\policy^*$ that maximises the expected discounted reward $\ExpReward$, but the agent no longer has the ability to interact with the environment and collect additional transitions. An offline RL algorithm, instead, is given a static, or offline, dataset $\dataset=\{(s_t^i,a_t^i,s_{t+1}^i,r_{t}^i)\}$ of transitions from different trajectories, where $i$ is the trajectory index and $t$ represents the time-step. All trajectories are generated following the transition function $\transitions$ and a data collection policy, which we denote as the behaviour policy $\policy_\beta$. An offline RL algorithm is required to learn a policy $\pi$ solely from this fixed offline dataset $\dataset$ that achieves the highest expected reward when evaluated online in the environment.

The terms off-policy RL, batch RL, and offline RL can all describe algorithms that learn from a pre-collected dataset rather than interacting with the environment in real time. Off-policy RL allows learning from data collected by any policy, but it often involves additional online data collection. Batch RL is sometimes used interchangeably with offline RL, but this can lead to confusion, as it may refer to iterative batch learning that still gathers new data. To avoid ambiguity, offline RL is used in this thesis to describe learning exclusively from a fixed offline dataset, without any further data collection.

\bill{\subsubsection{Offline Model-free RL Algorithms}\label{lit:offline_rl}}

The classic paradigm of online RL fundamentally involves online learning, where the agent iteratively interacts with the environment to collect experience and uses that experience to improve its policy. However, such interactions may be costly and dangerous in many settings, and online RL algorithms can suffer from exponential sample complexity~\citep{Kakade2003OnLearning}.
% such interaction may not be practical in many settings because it is expensive (e.g., robotics, healthcare, and business operations) and dangerous (e.g., autonomous driving and healthcare). Furthermore, even if online interactions are feasible, online RL algorithms can suffer from exponential sample complexity~\citep{Kakade2003OnLearning}, especially for complex tasks that require a large amount of data. 
This problem has driven research in various directions, including building high-fidelity simulators to learn policies for tasks like autonomous driving and robotics control, and then transferring these policies to the real world~\citep{Sadeghi2016CAD2RL:Image,Chebotar2018ClosingExperience}. On the other hand, offline RL is a direct approach that provides a paradigm to use previously collected offline data~\citep{Fu2020D4RL:Learning,Kumar2019StabilizingReduction}, without any online interaction with the environment. This enables the exploitation of modern scalable machine learning algorithms that can potentially achieve better performance in the RL setting if more data is used for training. Examples of successful offline RL applications include robotic manipulation~\citep{Kalashnikov2018QT-Opt:Manipulation}, robotic navigation skills~\citep{Kahn2020BADGR:System}, and human dialogue generation~\citep{Jaques2019WayDialog}, among others.

However, as discussed in~\cref{chapter:intro}, offline RL presents major challenges including \textit{distribution shift}~\citep{Levine2020OfflineProblems} and \textit{hidden confounders}. Since the hidden confounder problem is fundamentally a causal inference problem, it is often not considered in conjunction with offline RL in the literature. In fact, most literature considers the offline RL problem exclusively as a distribution shift problem. Therefore, in this section, we mainly introduce offline RL algorithms that address the distribution shift problem. Later, we dedicate a separate section in~\cref{lit:confoundedRL} to the literature on the intersection between RL and hidden confounders.

% The most widely studied problem is the \textit{distribution shift}~\citep{Levine2020OfflineProblems} problem. It describes the difference between the distribution of the state and action pairs in the offline dataset, also known as the behaviour policy distribution, and the distribution of the state and action pairs in the real environment. Regardless of the RL method being used, the policy, Q value function, and the learnt model are trained under the behaviour policy distribution and evaluated on a different real-world distribution, causing the agent in practice not only to move beyond the state distribution in the offline dataset, but also to take risky out-of-distribution actions.

% In addition, a fundamental and challenging problem that offline RL encounters is the \textit{hidden confounder} problem. In online RL, since the agent can directly interact (or, in causal inference terms, intervene) with the environment, the causal effect between actions and the feedback from the environment can be directly learnt by the agent. In offline RL, when the offline dataset is confounded, the agent cannot learn the true effects of its actions due to the spurious correlations between the state and action pairs in the dataset, leading to suboptimal and potentially adversarial policies. 

Before introducing algorithms specifically designed for offline reinforcement learning (RL), it is important to note that any off-policy RL algorithm, such as Q-learning~\citep{Watkins1992Q-learning}, allows the replay buffer to differ from the distribution used for evaluation and, in principle, can be applied directly to the offline setting. However, off-policy RL~\citep{Harutyunyan2016QCorrections} typically assumes that new in-distribution experience can be added to the dataset through online interactions with the environment. This naturally limits~\citep{Munos2016SafeLearning} the degree of distribution shift in the dataset. However, most of these methods are not nearly as effective in the offline setting.

Most offline RL algorithms involve approximate dynamic methods with various techniques to limit the distribution shift. The first group of works focuses on imposing constraints on the policy to limit the degree of distribution shift from the behaviour policy. \citet{Jaques2019WayDialog} proposed to explicitly add the Kullback–Leibler (KL) divergence between the behaviour policy and the current policy as a penalty to the Q values, and \citet{Wu2019BehaviorLearning} proposed Behaviour Regularised actor-critic (BRAC), where both the policy improvement step and the policy evaluation step can be penalised. However, these explicit f-divergence methods~\citep{Nowozin2016F-GAN:Minimization} (a superset of KL-divergence) require explicit estimation of the behaviour policy, which is difficult to estimate accurately, especially for multimodal distributions. Another line of implicit f-divergence constraint methods, such as AWR~\citep{Peng2019Advantage-WeightedLearning}, AWAC~\citep{Nair2020AWAC:Datasets} and ABM~\citep{Siegel2020}, finds a nonparametric representation of the optimal policy updates for f-divergence constraints. Integral probability metrics~\citep{Sriperumbudur2009OnClassification} are also used for policy constraints, with BEAR~\citep{Kumar2019StabilizingReduction} adopting the maximum mean discrepancy distance to represent the policy constraint and \citet{Wu2019BehaviorLearning} proposing a first-order Wasserstein distance policy constraint. 
% However, \citet{Kumar2019StabilizingReduction} argued that these divergence constraints effectively limit the optimality of the learnt policy by restricting it from concentrating around a small number of good actions, while constraining the supports of the behaviour policy and the learnt policy to allow the learnt policy to concentrate around in-distribution actions while preventing out-of-distribution actions. They showed that a finite sample estimate of the maximum mean discrepancy distance can approximate support constraints, while \citet{Sachdeva2020Off-policySupport} imposed the exact support constraint in the off-policy bandit setting. Finally, \citet{Fujimoto2021ALearning} proposed a minimalist approach by introducing a behaviour cloning regulariser to penalise action mismatches between the learnt policy and the behaviour policy during the policy improvement steps.

The second group of works focuses on finding a conservative estimate of the actual Q function to mitigate the effect of out-of-distribution actions. 
% Explicit estimation of the epistemic uncertainty can provide a conservative Q value by subtracting the uncertainty from the value estimate. 
The early work of \citet{Jaksch2010Near-optimalLearning} maintains an explicit confidence set of upper and lower confidence bounds to capture uncertainty. \citet{Osband2016DeepDQN} used bootstrap ensembles of Q functions to represent samples from the distribution of Q functions to measure the uncertainty, while \citet{ODonoghue2017TheExploration} parameterised the distribution of Q functions using a Gaussian distribution with learnt parameters. However, high precision in uncertainty estimation is required, and the uncertainties must faithfully capture the reliability of the Q function. To avoid the difficulty in computing accurate uncertainties of the Q function, \citet{Kumar2020ConservativeLearning} proposed regularising the value function directly by penalising high Q values, particularly for out-of-distribution actions, to achieve conservative Q functions. 
% This method is practical and provides pessimistic value guarantees, but it could suffer from overly pessimistic values for out-of-distribution states that limit the performance of the learnt policy.

The third group of offline RL algorithms uses importance sampling~\citep{Kloek1978BayesianCarlo} to perform offline policy evaluation (OPE) and policy optimisation. \citet{Precup2000EligibilityEvaluation} first proposed the use of per-action importance sampling to estimate the expected reward. However, the variance of this estimator is too high to be effective, leading to the proposal of a doubly robust estimator~\citep{Thomas2016Data-EfficientLearning} that uses Q-value approximations alongside methods that optimise the trade-off between bias and variance~\citep{Wang2016OptimalBandits} to reduce variance. \citet{Thomas2015High-ConfidenceEvaluation} presented a method that guarantees the performance of the learnt policy with high probability for safety-critical applications. Alternative approaches~\citep{Precup2000EligibilityEvaluation,Precup2001Off-PolicyApproximation,Peshkin2002LearningExperience} have been proposed to directly estimate the policy gradient using similar per-action importance sampling techniques. To address the high variance of the policy gradient estimates, various approaches have been proposed including doubly robust estimators for the policy gradient~\citep{Gu2017DeepUpdates,Huang2019FromGradient,Cheng2019Trajectory-wiseMethods}, regularisation of the shift~\citep{Levine2013GuidedSearch,Schulman2017ProximalAlgorithms}, estimating the state-marginal importance ratio~\citep{Sutton2015AnLearning,Hallak2017ConsistentEvaluation,Gelada2019Off-PolicyShift,Liu2019BreakingEstimation}, and using the backward Bellman equation~\citep{Lee2019,Nachum2019DualDICE:Corrections,Nachum2020ReinforcementDuality}.

Other works in offline RL include the use of one-step methods~\citep{Brandfonbrener2021OfflineEvaluation,Kostrikov2022OfflineQ-Learning} that perform a single policy update step from the behaviour policy for out-of-distribution actions, and imitation learning (see~\cref{lit:il} for more details) that mimics the behaviour policy by finding and exploiting the well-performing parts of the behaviour policy.

% on-sample RL~\citep{Kostrikov2022OfflineQ-Learning}, which evaluates only in-sample state-actions while allowing policies to differ from the behaviour policy through one-step updates. 
% These methods have demonstrated strong results on benchmarks, but when the behaviour policy is highly suboptimal yet provides good coverage of the state-action space, standard multi-step methods can generally learn better policies. In addition, imitation learning approaches 

% Methods like BAIL~\citep{Chen2019BAIL:Learning} fitted a value function of the behaviour policy and then used it to select the best actions for training. \citet{Emmons2022RvS:Learning} proposed to perform offline RL through supervised learning using learning policies conditioned on the goals or rewards of the samples in the offline dataset.

\subsubsection{Offline Model-based RL Algorithms}\label{lit:offline_model-based_rl}

As discussed above, there exists an active field of research on model-based RL that uses learnt predictive transition models. These models can also be powerful in the offline RL setting, as they can leverage supervised learning to learn the dynamic model from large datasets. 

In fact, online model-based RL already suffers from the distribution shift induced by model errors, and previous work in the online setting addresses the distribution shift problem with uncertainty estimation.
% In low-dimensional settings~\citep{Deisenroth2011PILCO:Search}, Bayesian models, such as Gaussian processes, can provide such epistemic uncertainty estimates, while in high-dimensional settings, bootstrap ensembles of Bayesian probabilistic neural networks, whose output parameterises a Gaussian distribution on the next state and reward, are adopted~\citep{Chua2018DeepModels,Janner2019}. 
In the offline setting, some online model-based RL methods are applied directly to robotics problems~\citep{Kahn2018ComposableNavigation,Kahn2020BADGR:System}, yet they provide few theoretical guarantees of performance. \citet{Berkenkamp2017SafeGuarantees} proposed a method to identify safe regions in the state and action space and imposed constraints to ensure Lyapunov stability. Based on model-based policy optimisation methods that use uncertainty estimates~\citep{Janner2019}, two concurrent approaches, MOReL~\citep{Kidambi2020MOReLLearning} and MOPO~\citep{Yu2020}, have been proposed to produce conservative value estimates by modifying the learnt MDP model with reward penalties based on epistemic uncertainty estimates. In both cases, it is shown that the policy value estimates under the modified MDP provide a lower bound on the policy's true value.
% However, the uncertainty estimates obtained through the bootstrap ensemble are not guaranteed to be consistent under sampling error.
\citet{Yu2021COMBO:Optimization} proposed COMBO that extends the work on offline RL conservative Q functions~\citep{Kumar2020ConservativeLearning} to the model-based setting that provides a guaranteed bound on the true policy value with model-based policy value and model errors.
% Instead of requiring explicit uncertainty estimates, COMBO regularises the value function on out-of-support state-action tuples that provide a guaranteed bound on the true policy value with model-based policy value and model errors. This means that if the policy improves by more than the model error under model-based value estimates, its true value must also improve. 
% \citet{Matsushima2021Deployment-EfficientOptimization} proposed an offline RL method that can learn from an offline dataset with many distinct data collection policies. They called it deployment efficient offline RL, and it is able to improve policy during successive deployments when new data are added to the dataset.

A different line of work, called sequential trajectory optimisation, is proposed to learn a model of the trajectory distribution induced by the behaviour policy. The model can then produce trajectories that we can use to plan an optimal set of actions from the initial state. \citet{Janner2021OfflineProblem} proposed a method called the Trajectory Transformer that used Transformer models~\citep{Vaswani2017AttentionNeed} to learn the trajectory distribution, and finally adopted beam search for planning. Concurrently, \citet{Chen2021DecisionModeling} proposed Decision Transformer, which used a similar formulation of trajectory optimisation as \citet{Janner2021OfflineProblem}, but during training, minimised trajectory error only for actions. These approaches leverage large supervised learning models to exploit large offline datasets, but they are costly to train.

\subsection{Imitation Learning}\label{prelim:IL}

Imitation learning (IL) offers a powerful alternative to RL that simplifies the challenge of maximising rewards and iteratively optimising policies. Instead of designing complex reward functions and relying on extensive trial-and-error interactions with the environment, IL learns directly from expert demonstrations. This is particularly useful in scenarios where human experts can easily demonstrate the desired behaviour but cannot encode it programmatically. IL is well suited for tasks that require human-like behaviour, making it an attractive approach in robotics, autonomous driving, and other domains involving complex decision-making.

The main purpose of imitation learning (IL) is to enable agents to learn a policy that performs a specific task or behaviour by imitating an expert through the provision of expert demonstrations in a static dataset. The dataset $\dataset=\{(s_t^i,a_t^i)\}$ typically consists of state and action pairs, but does not include the reward. These demonstrations are used to train agents to perform a task by learning a mapping between states and expert actions. We will introduce three broad categories of IL approaches: behavioural cloning (BC), interactive IL, and inverse reinforcement learning (IRL).

\subsubsection{Behavioural Cloning}
Behavioural cloning (BC) is an IL technique that treats the problem of learning expert behaviour as a supervised learning task. BC algorithms use the expert demonstration dataset $\dataset$ of state and action pairs to train a policy $\pi(s)$ that mimics an expert's behaviour by learning the mapping between the state and action. The supervised learning target of BC is $\expectE[s_t\lvert a_t]$, which is the expectation of the states conditional on the actions. This objective is typically minimised using log-likelihood or least squares methods.

The main advantage of BC is its simplicity. BC does not require knowledge of the environment dynamics or the reward function, as it relies solely on expert demonstrations to learn the required behaviour. Moreover, BC is computationally efficient, as it involves supervised learning, a problem that is extensively researched in machine learning.

\subsubsection{Other Imitation Learning Paradigms}

Existing BC approaches, however, suffer from drawbacks. The main drawback is the covariate shift problem~\citep{Pomerleau1988}, which occurs when the learnt policy encounters states that are not well represented in the training dataset at test time, leading to high test-time errors. Furthermore, unlike RL, BC does not explicitly correct errors once the learnt policy deviates from the expert, and this lack of feedback correction results in compounding errors.

A popular area of research that mitigates the covariate shift problem is interactive IL. This class of algorithms operates on the premise that the agent can access an online expert throughout the training process. The first interactive imitation learning method, Dataset aggregation (DAgger)~\citep{Ross2011}, addresses the discrepancy between training and testing environments by instructing the agent on its self-generated state distribution. Subsequent interactive IL algorithms extend DAgger in various ways~\citep{Kelly2018,Mandlekar2020,Hoque2021}, but the main concept of using an interactive agent remains unchanged.

Another related research domain is inverse reinforcement learning (IRL). IRL infers the reward function that underlies the observed demonstrations, which are assumed to come from an expert who acts optimally~\citep{Piot2017}. The IRL agent then uses the inferred reward function to optimise an imitator policy iteratively through standard RL. It is noteworthy that IRL, unlike BC, assumes that the agent can interact with the environment online for the final reinforcement learning step after a reward function has been learned.

In~\cref{chapter:il}, we consider the problem of IL from offline confounded expert demonstrations, without online access to the environment or an interactive expert. Among the IL paradigms introduced, our setting is most similar to BC.

\section{Hidden Confounders}\label{background:hiddenconf}

In this section, we introduce the problem of hidden confounders within the context of \textit{structural causal models} (SCMs) and discuss instrumental variables, which we will use to address the hidden confounder issue in the remainder of this thesis.

\textit{Hidden confounders}~\citep{Pearl2000causality} are unobserved variables that influence multiple observable variables simultaneously, confounding them and introducing spurious correlations between them. In causal inference literature, it is often assumed that the hidden confounder affects two specific variables: the \textit{action} (or \textit{intervention}) and the \textit{outcome}, where the goal is to understand the causal relationships between them. To properly account for the hidden confounder and understand the true causal effect of actions, we need to model the causal (or structural) relationship between the action and the outcome, which is expressed through a \textit{causal function}. However, learning the causal function in the presence of hidden confounders is known to be challenging and sometimes infeasible~\citep{Shpitser2008}. To formalise the concept of hidden confounders and provide a framework for specifying the underlying causal mechanisms in a data-generating process, we introduce structural causal models.

\subsection{Structural Causal Models}\label{background:SCM}

\begin{definition}[Structural Causal Model]
An SCM $M$ is a tuple $(U,V,F,P(U))$, where $U$ is a set of exogenous (i.e., outside the model) random variables, which are
typically unobserved; $V$ is a set of observed endogenous (i.e., inside the model) variables; $F=\{f_i\}$ is a set of deterministic functions such that, for each $V_i\in V$, $f_i(PA_i,U_i)=V_i$, where $PA_i$ denotes the endogenous parents of $V_i$; and $P(U)$ is the joint distribution of exogenous variables.
\end{definition}

In this definition, endogenous variables are variables for which we would like to study the causal effects (e.g., price and revenue). Exogenous variables serve as external sources of noise (e.g., seasonality) that can confound the causal relationships among endogenous variables. To better illustrate the relationships between variables in an SCM, a graphical abstraction, known as a causal graph, can be used to represent the causal relationships between variables within an SCM. A causal graph is a directed acyclic graph, where the nodes represent variables in the SCM. A solid directed edge indicates a direct causation between two endogenous variables: $V_i \rightarrow V_j$; and a dashed directed edge implies direct causation between an exogenous variable and an endogenous variable: $U_j \rightarrow V_j$.

% \begin{definition}[Causal Graph]
% A causal graph $G=(V,E)$ associated with M is directed acyclic graph of a graph of endogenous variables $V$, where a directed edge $V_i \rightarrow V_j$ implies $V_i\in PA_j$, and a bidirected edge means that they share an exogenous variable.
% \end{definition}

We now introduce the concept of causal interventions, which are tools that allow us to study causal effects between variables. Interventions are defined through
a mathematical operator called $do(x)$~\citep{Pearl2000causality}. An intervention, denoted by $do(X=x)$, simulates a physical intervention by removing the natural dependencies of $X$ on its parent variables in the SCM and forcing it to take a specific value $x$, while keeping the rest of the model unchanged. The resulting causal model after the intervention is denoted $M_x$. The post-intervention distribution resulting from $do(X = x)$ is given by the equation:
\begin{align}
 \probP_M(y\lvert do(x)) = \probP_{M_x}(y),
\end{align}
where the post-intervention distribution of a variable $Y$ is defined as the distribution of $Y$ in the intervened model $M_x$. Causal interventions allow researchers to analyse the causal effect of $X$ on other variables while eliminating confounding factors. 

For example, in a causal model where $A$ (treatment) affects $Y$ (outcome), an observational study may show correlation, but an intervention $do(A = a)$ would simulate a randomised experiment, ensuring that changes in $Y$ are caused by $A$ and not by other confounding factors.

\begin{figure}
    \centering
    \includegraphics[width=0.4\linewidth]{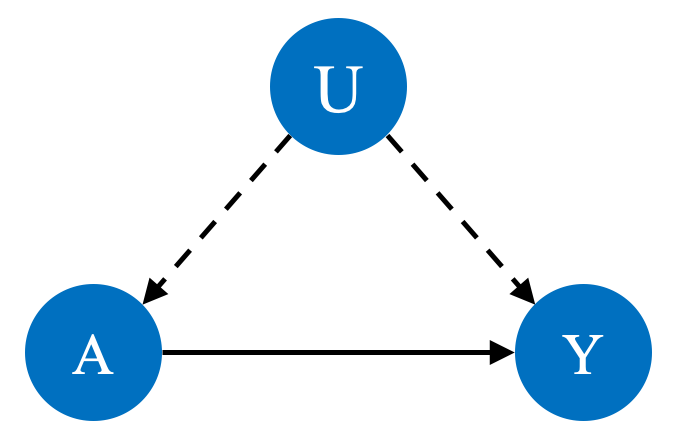}
    \caption{The causal graph of outcome $Y$, treatment $A$ and hidden confounder $U$.}
    \label{fig:intro/causal_graph}
\end{figure}

In the SCM formulation, an exogenous random variable $U$ is considered to be a \textit{hidden confounder} if it affects two or more endogenous variables, such as $V_i$ and $V_j$, and is unobserved. Consider an SCM that specifies two endogenous variables: the outcome $Y\in\mathcal{Y}$ and the treatment $A\in\mathcal{A}$,
\begin{align}
    Y=f(A,U)\label{eq:scm},
\end{align}
where $U\in\mathcal{U}$ is a hidden confounder that affects both $A$ and $Y$, as illustrated in the causal graph depicted in~\cref{fig:intro/causal_graph}. Due to the presence of this hidden confounder, with only observational data, standard regression methods (e.g., ordinary least squares) generally fail to produce consistent estimates of the causal relationship between $A$ and $Y$~\citep{Pearl2000causality}, that is, $\expectE[Y \mid do(A)]$. Therefore, the ability to identify the causal relationship between $A$ and $Y$ requires additional assumptions. Next, we introduce techniques that allow for the identification of $\expectE[Y \mid do(A)]$ under the assumption of observing additional variables.

\subsection{Conditional Moment Restrictions}\label{background:CMRs}

To begin with, we introduce the problem of \textit{conditional moment restrictions} (CMRs). CMRs describe a family of problems in causal inference, including \textit{instrumental variable} (IV) regression and \textit{proximal causal learning} (PCL), that aim to identify the causal effect $\expectE[Y \mid do(A)]$ in the presence of hidden confounders.

\begin{definition}[Conditional Moment Restriction]\label{defn:cmr}
Let $X\in\mathcal{X}\subseteq\realNumber^d$, $C\in\mathcal{C}\subseteq\realNumber^p$ and $Y\in\mathcal{Y}\subseteq\realNumber$ be random variables with their corresponding distributions. A CMR problem aims to estimate a function of interest $f\in\mathcal{F}\subseteq(\mathcal{X}\rightarrow\realNumber)$ in a hypothesis function space $\mathcal{F}$, where the true function $f_0$ satisfies the following condition:
\begin{align*}
    \expectE[Y-f_0(X)\lvert C=c]=0, \text{ for all } c\in\mathcal{C}.
\end{align*}
For simplicity, we denote the CMR problem as the following equation:
\begin{align}
    \expectE[Y-f(X)\lvert C]=0. \label{eq:cmr}
\end{align}
\end{definition}

CMR problems are central to many problems in various fields, including statistical learning~\citep{Vapnik1998}, integral equations~\citep{Honerkamp1990}, deconvolution~\citep{Carrasco2007} and econometrics~\citep{Carrasco2007} (see~\citealt{Carrasco2007}, Section 1.3 for more examples).

However, solving CMRs analytically is ill-posed~\citep{Nashed1974} because it is an inverse problem for definite integrals that requires the derivation of a function inside conditional expectation. Many CMR estimators (e.g.,~\citealt{Angrist1996,Newey2003,Hartford2017DeepPrediction,Singh2019}) estimate $f_0$ with $\widehat{f}$ in some function space $\mathcal{F}$ by solving the following objective function using a two-stage approach:
\begin{equation}
    \hat{f} \in \arg\min_{f\in\mathcal{F}} \expectE[(Y-\expectE[f(X)\lvert C])^2]\label{eq:optim}.
\end{equation}
Specifically, the method of \citet{Newey2003} employs a minimax approach that indirectly optimises this objective by solving a minimax unconditional moment problem in two stages. \citet{Angrist1996,Singh2019,Hartford2017DeepPrediction}, on the other hand, directly optimise the above objective in two stages. The first stage involves learning the conditional expectation $\expectE[f(X)\lvert c]$ through density estimation or kernel methods from observations. In the second stage, the objective in~\cref{eq:optim} is minimised using the estimates obtained in the first stage. For both stages, linear regression, sieve methods, and deep neural networks (DNNs) are used for estimation, respectively, for each work. In~\cref{chapter:dmliv}, we point out that two-stage estimators often suffer from slow convergence with a lack of guarantees, and we propose a novel two-stage CMR estimator that mitigates this problem with fast convergence rate guarantees. Next, we introduce two specific instances of CMR problems in causal inference that enable causal identification under hidden confounders.

\subsection{Instrumental Variable Regression}\label{background:IV}

\begin{figure}
    \centering
    \includegraphics[width=0.6\linewidth]{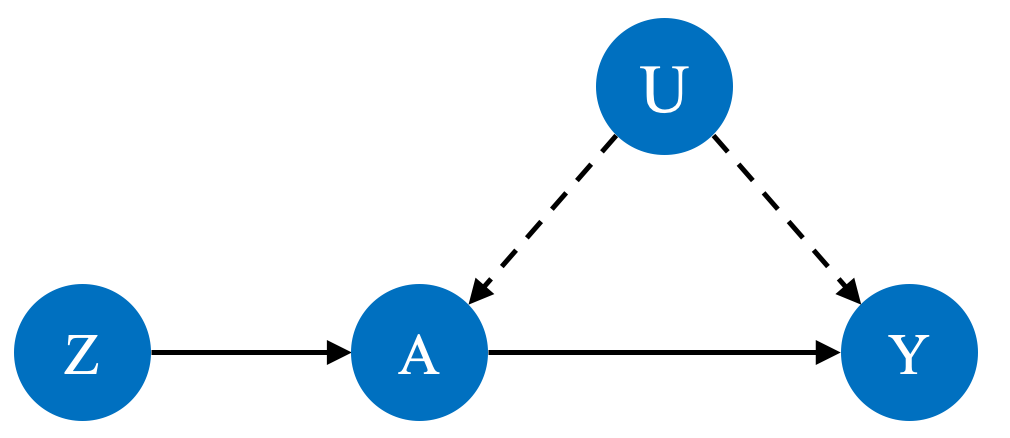}
    \caption{The causal graph of outcome $Y$, treatment $A$, hidden confounder $U$ and an instrumental variable $Z$.}
    \label{fig:intro/causal_graph_iv}
\end{figure}

\textit{Instrumental variable} (IV) regression~\citep{Newey2003} is a CMR problem that relies on the observation of additional IVs to identify the causal effect $\expectE[Y \mid do(A)]$. To begin with, we introduce the concept of IVs.

\begin{definition}[Instrumental Variable]\label{def:iv}
Under the SCM of outcome $Y\in\mathcal{Y}$, treatment $A\in\mathcal{A}$ and hidden confounder $U\in\mathcal{U}$ defined in~\cref{eq:scm}, an instrumental variable $Z\in\mathcal{Z}$ is an observable variable that satisfies the following conditions~\citep{Newey2003}: 
\begin{itemize}[leftmargin=40pt, topsep=5pt]
\label{assump:iv}
    \item \textit{Unconfounded Instrument}: $Z\indep U$, that is, $Z$ is independent of $U$;
    \item \textit{Relevance}: $\probP(A\lvert Z)$ is not constant in $Z$;
    \item \textit{Exclusion}: $Z$ does not directly affect $Y$: $Z\indep Y \mid (A,U)$,
\end{itemize}
where a causal graph with an instrumental variable $Z$ is depicted in~\cref{fig:intro/causal_graph_iv}.
\end{definition}

Intuitively, IVs are heterogeneous
random variables that only directly affect the action, but not the outcome. IVs have been used extensively to identify the causal effect of actions in many applications, including econometrics~\citep{Reiersol1945,Angrist2009}, drug testing~\citep{Angrist1996}, and social sciences~\citep{Angrist1990}.

However, in order to identify the causal effect $\expectE[Y \mid do(A)]$, an additional assumption of \textit{additive noise} is required, where we assume that
\begin{align}
    Y=f(A)+\epsilon(U) \ \text{ with } \  \expectE[\epsilon(U)]=0\label{eq:iv_reg}.
\end{align}
The additive noise assumption, in conjunction with the IV conditions, is standard for the IV settings~\citep{Newey2003,Xu2020,Shao2024} and ensures the minimal condition to identify the causal effect. Another observation we can make here is that, following~\cref{eq:iv_reg}, 
\begin{align}
\expectE[Y \mid do(A)]&=\expectE[f(A)\mid do(A)+\expectE[\epsilon(U)\mid do(A)]\\
&=f(A)+\expectE[\epsilon(U)]=f(A),
\end{align}
which means the task of identifying the \textit{causal effect} $\expectE[Y \mid do(A)]$ is equivalent to learning the \textit{causal function} $f(A)$.

In the IV setting, since the hidden confounder $U$ affects both $A$ and $Y$, it is generally the case that $\expectE[\varepsilon(U) \mid A]\neq 0$. This makes standard regression methods, such as ordinary least squares, fail to estimate the correct causal effect. However, there exist many IV regression methods that use IV to estimate the causal function $f(A)$. In order to identify $f(A)$, a key observation~\citep{Newey2003} is that,  by taking the expectation on both sides of~\cref{eq:iv_reg} conditional on $Z$, we have
\begin{align}
\expectE[Y\lvert Z]&=\expectE[f(A)+\epsilon(U)\lvert Z\Big]\nonumber\\
&=\expectE[f(A)\lvert Z]+\expectE[\epsilon(U)]\nonumber\\
&=\expectE[f(A)\lvert Z]=\int f(A) \probP(A\lvert Z) dA,\label{eq:h_exp}
\end{align}
where the expectation $\expectE[Y\lvert Z]$ and the distribution $\probP(A\lvert Z)$ are both observable. Therefore, it becomes possible to solve for $f$ that satisfies the above equality. From~\cref{eq:h_exp}, we can rearrange and conclude that IV regression can be reduced to the following CMRs:
\begin{align}
\expectE[Y-f(A)\lvert Z]=0,
\end{align}
where the variables $X$ and $C$ in the general CMR formulation~\eqref{eq:cmr} are renamed to $A$ and $Z$ in IV regression.

\subsection{Proximal Causal Learning}\label{background:PCL}

\begin{figure}
    \centering
    \includegraphics[width=0.6\linewidth]{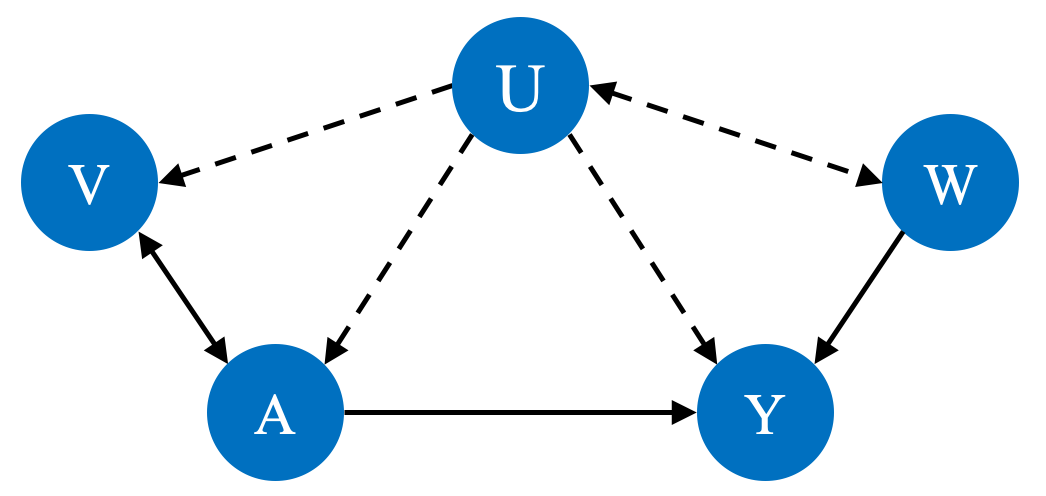}
    \caption{The causal graph of outcome $Y$, treatment $A$, hidden confounder $U$ and proxies $V$ and $W$.}
    \label{fig:intro/causal_graph_proxy}
\end{figure}

\textit{Proximal causal learning} (PCL) is also a CMR problem that allows for the identification of the causal effect between variables under hidden confounders. Following the SCM of outcome $Y\in\mathcal{Y}$, treatment $A\in\mathcal{A}$ and hidden confounder $U\in\mathcal{U}$ defined in~\cref{eq:scm}, PCL uses two proxy variables to identify the causal effect of the treatment $A$ on the outcome $Y$, i.e., $\expectE[Y \mid do(A)]$. The first proxy $V\in\mathcal{V}$ is a treatment-inducing proxy, and the second proxy $W\in\mathcal{W}$ is an outcome-inducing proxy. For $V$ and $W$ to be valid proxies, they must satisfy the following \textit{conditional independence conditions}:
\begin{itemize}[leftmargin=40pt, topsep=5pt]
\label{assump:proxy}
    \item $Y\indep V \mid (A,U)$;
    \item $W\indep (A,V) \mid U$,
\end{itemize}
where a causal graph with proxies $V$ and $W$ is depicted in~\cref{fig:intro/causal_graph_proxy}.

In addition, for the identifiability of the causal effect, proxies should satisfy \textit{completeness assumptions}. Let $l:\mathcal{U}\rightarrow \realNumber$ be any square-integrable function, that is, $\norm{l}_2\leq \infty$. The following conditions hold for any $a\in \mathcal{A}$:
\begin{align*}
    \expectE[l(U)\lvert A=a,W=w]=0 \quad\forall w\in \mathcal{W} &\iff l(u)=0 \text{ \textit{almost everywhere} on } \probP(U);\\
    \expectE[l(U)\lvert A=a,V=v]=0 \quad\forall v\in \mathcal{V} &\iff l(u)=0 \text{ \textit{almost everywhere} on } \probP(U).
\end{align*}
It has been shown that, when the conditional independence conditions and the completeness assumptions are satisfied, it is possible to identify $\expectE[Y\lvert do(A)]$.

\begin{proposition}[\citet{Miao2018}]
Let the conditional independence and completeness assumptions hold, then there exists at least one solution to the following equality,
\begin{align}
\expectE[Y\lvert A,V]&=\expectE[h(A,W)\lvert A,V]\label{eq:proximal}\\
&=\int h(A,W)\probP(A,W\lvert A,V) dW,
\end{align}
for all $(A,V)\in \mathcal{A}\times \mathcal{V}$. Let $h^*$ be a solution of~\cref{eq:proximal}, then the \textit{causal effect} $\expectE[Y \mid do(A)]$ can be estimated by $\expectE_W[h(A,W)]$.
\end{proposition}

From this proposition, we can see that the problem of estimating the causal effect $\expectE[Y \mid do(A)]$ can be reduced to estimating $h^*$, which we denote as the bridge function following~\citet{Miao2018}. Therefore, estimating the causal effect in the PCL setting can be reduced to the following CMR problem:
\begin{align}
\expectE[Y-h(A,W)\lvert A,V]=0.
\end{align}

In~\cref{chapter:dmliv}, we additionally evaluate our proposed CMR estimator on the PCL task, where state-of-the-art performance is demonstrated.

\begin{remark}
For both IV and PCL, it is possible to include additional observed confounders $\mathbf{C}$ that affect both the treatment and the outcome as additional information or context. The resulting CMRs would be $\expectE[Y-f(A,\mathbf{C})\lvert Z,\mathbf{C}]=0$ for IV regression and $\expectE[Y-h(A,W,\mathbf{C})\lvert A,V,\mathbf{C}]=0$ for PCL. Note that, for IV regression, we can estimate $f(A,\mathbf{C})=\expectE[Y\lvert do(A),\mathbf{C}]$,
 which is the conditional causal effect. This can be used to derive a decision policy that depends on the context $\mathbf{C}$ to maximise the expected outcome, as we do in~\cref{chapter:dmliv}. However, for PCL, we can only estimate the expected causal effect $\expectE[Y\lvert do(A)]=\expectE[h(A,W,X)]$, which implies that a context-dependent policy cannot be derived.
\end{remark}

\section{High-Level Objectives}\label{prelim:high-level}

In this section, we introduce the various ways to specify high-level objectives as targets for RL. While the target for standard RL is to maximise the expected reward with respect to some reward function $\reward$, specifying the RL target in terms of reward functions for objectives such as patrolling between points while avoiding obstacles can be challenging, both in terms of human interpretability and the risk of the reward mismatching the true objective. Therefore, we consider specifying RL objectives directly using high-level objectives, specifically in terms of logic languages, since they are concise and intuitive to humans. In addition, we discuss automata-based objectives, such as reward machines~\citep{ToroIcarte2018UsingLearning}, which are widely used in the literature to specify high-level objectives.

\subsection{Linear Temporal Logic}\label{prelim:ltl}
A very commonly used and expressive logic is linear temporal logic (LTL). LTL provides a high-level description for the specifications of a system. In LTL, ``temporal'' means that the logic includes temporal operators expressing the order of events and can encode formulae regarding the future behaviour of a system. ``Linear'' means that it considers a linear, or a single, sequence of time points. LTL is very expressive and can describe specifications with an infinite horizon.

\begin{definition}[\citep{Baier2008}]
An LTL formula over atomic propositions $\propositions$ is defined iteratively by the grammar:
\begin{equation*}
    \LTL \vcentcolon\vcentcolon= true\mid p\mid\LTL_1\land\LTL_2\mid\neg\LTL\mid \textsf{\upshape X }\LTL\mid\LTL_1 \textsf{\upshape U } \LTL_2, \quad p\in\propositions,
\end{equation*}
where $\textsf{\upshape X}$ represents \textit{next} and $\textsf{\upshape U}$ represents \textit{until}. Other Boolean and temporal operators are derived as follows: or: $\LTL_1\lor\LTL_2=\neg(\neg\LTL_1\land\neg\LTL_2)$; implies: $\LTL_1\rightarrow\LTL_2=\neg\LTL_1\lor\LTL_2$; \textit{eventually}: $\textsf{\upshape F }\LTL=true \textsf{\upshape U }\LTL$; and \textit{always}: $\textsf{\upshape G }\LTL=\neg(\textsf{\upshape F } \neg\LTL)$.
\end{definition}

From the definition of an LTL formula, a language of infinite sequences of atomic propositions can be generated iteratively. In the context of RL with MDP and atomic proposition labels $(\states,\initstate,\actions,\transitions,\propositions,\labFunc)$, the satisfaction of an LTL formula $\LTL$ by an infinite MDP path $\MDPpath\in\paths$, denoted by $\MDPpath\models\LTL$, is defined by induction on the structure of $\LTL$:
\begin{align*}
    \MDPpath\text{ satisfies $\LTL$ if }\quad p\in\labFunc(\MDPpath[0])\quad &\text{ for } p\in\propositions;\\ 
    \MDPpath[1:]\models\LTL\quad &\text{ for } \textsf{\upshape X }\LTL;\\
    \exists i, \MDPpath[i]\models\LTL_2 \text{ and } \forall j<i, \MDPpath[j]\models\LTL_1\quad &\text{ for } \LTL_1\textsf{\upshape U }\LTL_2 ,
\end{align*}
with the satisfaction of Boolean operators $\land$ and $\neg$ defined by their default meaning. Note that we will use $\neg$ and $\land$ as well as ! and \& interchangeably in this thesis. To better explain the inductive satisfaction of LTL formulae, we provide some examples of LTL formulae and the infinite paths that satisfy them.

\begin{example}
Examples of LTL formulae and the infinite paths satisfying them in increasing complexity. For $a,b,c\in \propositions$:
\begin{enumerate}
    \item LTL formula \enquote{a}: any infinite path where the first element is labelled $a$ by $\labFunc$.
    \item LTL formula \enquote{\textsf{\upshape X }a}: any infinite path where the second element is labelled $a$.
    \item LTL formula \enquote{!a \textsf{\upshape U }b}: any infinite path where there exist an element labelled $b$ with all previous elements not labelled $a$.
    \item LTL formula \enquote{\textsf{\upshape G }a}: any infinite path where no element is not labelled $a$, meaning all elements in the path must be labelled $a$. 
    \item LTL formula \enquote{\textsf{\upshape FG }a}: any infinite path where, after some finite index $i$, all elements are labelled $a$.
    \item LTL formula \enquote{\textsf{\upshape F}\textsf{\upshape G }a \& \textsf{\upshape G }!c}: any infinite path where no element is labelled $c$, and, after some finite index $i$, all elements are labelled $a$.
\end{enumerate}
\end{example}

\subsection{Other Temporal Logics}\label{prelim:temporal}

Beyond LTL, there also exist other temporal logics that can specify conditions regarding the future of a system. We are specifically considering temporal logics because we would like to focus on objectives that reason about systems in terms of time. Before we justify the choice of LTL, we briefly overview other temporal logics.

Computation tree logic (CTL) is a branching-time temporal logic, meaning that time is viewed as a tree of possible futures. It allows for reasoning about different possible execution paths of a system and includes two path quantifiers: \textsf{\upshape A } means all possible future paths, and \textsf{\upshape E } means there exists at least one possible future path. By extending CTL with temporal operators such as \textsf{\upshape X } and \textsf{\upshape U } as defined in LTL, we have the full computation tree logic (CTL*), which is a more expressive logic that generalises both LTL and CTL. This extension allows for the combination of path quantifiers with LTL-style operators in a more flexible way.

Interval temporal logic (ITL) reasons about entire intervals of time and includes a duration operator \textsf{\upshape D } that specifies time constraints on intervals. Signal temporal logic (STL) incorporates time-constrained temporal operators, which allow STL to specify inequalities over real-valued functions of time that can evolve continuously or discretely (these functions are called signals). These temporal logics are suitable for specifying properties for systems where time duration matters.

Finally, we mention additional variants of the previously introduced temporal logics. While LTL assumes infinite traces, which means that temporal operators such as \textit{always} \textsf{\upshape G } extend indefinitely, there are also finite variants of LTL. LTLf is a variant of LTL that only considers finite traces, while truncated LTL considers LTL with a finite horizon. There are also probabilistic versions of temporal logic, such as PLTL and PCTL, which extend temporal logic by incorporating probability. This allows for specifications such as ``The system eventually reaches a failure state with probability at most 0.1''.

In this thesis, we use LTL to specify high-level objectives for RL for the following reasons. Firstly, under the discounted reward setting in RL, each environment run produces a single infinite-length trajectory, which is well suited for linear-time logics with an infinite horizon. Secondly, the goal of RL is to maximise an objective, and with LTL objectives, the natural goal would be to maximise the probability of satisfying LTL objectives. Therefore, probabilistic LTL's ability to specify objectives with a specific probability is less useful and introduces additional complexity when providing optimality guarantees. Last but not least, LTL can be translated into $\omega$-automata (finite state automata on infinite sequences), which we will introduce later in~\cref{sec-ltl:automata}. This property enables RL with temporal logic objectives and is essential for our algorithm proposed in \cref{chapter:ltl}.

\subsection{Reward Machines}

Another popular approach to specify high-level objectives is through reward machines~\citep{ToroIcarte2018UsingLearning}, which are finite state automata that define how rewards are assigned in a RL environment based on the trajectory history of the agent. Instead of hardcoding the reward function into the RL environment, a reward machine defines its transitions based on the atomic propositions associated with the environment states (recall \cref{prelim:mdp_ap}), while its states specify the corresponding rewards.

% For example, consider the task of a robot reaching a goal while avoiding obstacles. The reward machine would have three states: start, goal reached, and obstacle hit. the start state is the initial state for reward machine, and it would transition to goal reached if the RL environment state labeled goal is reached, and it would transition to obstetrical hit if an obstacle is hit in the RL environment. Rewards are given if 

Comparing to LTL, which provides declarative high-level objectives specifying what should happen, a reward machine provides a procedural representation of the high-level objective that defines how the agent should be rewarded. Functionally, it plays a similar role to an $\omega$-automaton, which also provides a procedural specification of the objective using finite state automata. However, reward machines are less expressive than $\omega$-automata and LTL because they cannot express specifications over an infinite sequence. In fact, reward machines are equivalent to regular expressions in terms of the expressivity of objectives. In contrast, LTL not only allows for more expressive objectives, but also provides a more concise and intuitive declarative objective for RL.

% \section{Summary}

% To summarise this chapter, we first introduced RL to formalise the problem of policy learning and introduced various types of Rl problems including online RL, offline RL, and imitation learning. Next, we introduced hidden confounders and high-level objectives, which motivates the concrete problems considered in this thesis.

\chapter{Literature Review}\label{chapter:lit}

\minitoc

In this chapter, we provide a comprehensive review of the literature on reinforcement learning, hidden confounders, and the connection between the two, in order to provide further context for the research questions that we address in the upcoming chapters.

\section{Reinforcement Learning}\label{lit:rl}

In this section, we discuss the literature on RL that is closely related to this thesis. \cref{lit:confoundedRL} discusses RL with hidden confounders, which is related to \cref{chapter:dmliv} and \cref{chapter:il}. \cref{lit:il} discusses imitation learning with and without hidden confounders, which is related to \cref{chapter:il}. \cref{lit:ltl} discusses the problem of learning high-level objectives with RL, which is related to \cref{chapter:ltl}. In \cref{lit:robust_rl}, we provide a brief review of the literature on robust and safe RL, which, although not directly related to the individual chapters, is generally pertinent to the overarching theme of this thesis.

\subsection{RL with Hidden Confounders}\label{lit:confoundedRL}

As discussed in~\cref{background:hiddenconf}, since the hidden confounder problem is fundamentally a causal inference problem, most of the RL literature does not consider hidden confounders in the environment. For the works that lie at the intersection of hidden confounders and RL, the first group of papers learns a policy from scratch using the least number of online interactions, with the help of a static offline dataset~\citep{Zhang2020DesigningApproach,Subramanian2022CausalInterventions,Lu2020RegretKnowledge,Wang2020ProvablyData,Gasse2023}. The second group focuses on learning policy purely from a confounded offline dataset, which is more relevant to this thesis and will be our primary focus.

For confounded offline RL, most work performs causal inference (e.g., instrumental variable regression~\citep{Murphy1985EstimationModels}) to mitigate the confounding effect and maximise the RL objective. \citet{Liao2024} considered a confounded Markov decision process (MDP), where the next state, current state, and action are confounded, with IVs observed at each time step. They applied IV regression (specifically Dual IV~\citep{Muandet2020}, which is discussed in~\cref{lit:iv}) to learn the transition dynamics of the MDP, and then they used the classical value iteration algorithm on the learnt transition dynamics to obtain a policy. \citet{Li2021} also considered RL with a confounded dataset, where only the reward is confounded with the state-action pair, and IVs are observed at each time step. They argued that the bias due to this confounding effect (named R-bias, which stands for reinforcement-bias) is amplified during the interaction of the agent with the environment. They proposed to mitigate this bias through IV regression, namely the classic 2SLS algorithm~\citep{Angrist1990} to learn the reward function before applying actor-critic methods to learn a policy. \citet{Xu2023} and \citet{Fu2022OfflineProcesses} considered the off-policy policy evaluation (OPE) problem with hidden confounders, where the current state, reward, action, and next state are all confounded, with IVs observed at each time step. They both derived doubly robust estimators for the policy values when instrumental variables exist. \citet{Xu2023} did this by designing efficient influence functions as estimator adjustments, and \citet{Fu2022OfflineProcesses} achieved double robustness by combining the value function estimator with the marginalised importance sampling estimator. \citet{Chen2021} considered the OPE problem in a standard MDP without hidden confounders. Their novel contribution is to consider the previous state-action pair as an instrument for the Bellman residual estimation problem of the current state-action pair. Hence, they can employ any IV regression method for this purpose, and in particular, they applied DFIV~\citep{Xu2020}, Deep IV~\citep{Hartford2017DeepPrediction}, and DeepGMM~\citep{Bennett2019DeepAnalysis}, among others. \citet{Bennett2021} considered the OPE problem in a specific setting of an infinite-horizon ergodic MDP that is confounded by hidden confounders. They proposed a method that uses states and actions as proxies for the hidden confounders to identify policy values. Their formulation involves the estimation of the stationary density ratio, and therefore it is not directly generalisable to the general nonlinear IV regression problem. When exact causal inference is unattainable due to the lack of IVs or proxies and the causal effect remains unidentifiable, \citet{Pace2024} developed a pessimistic algorithm based on Delphic uncertainty. This algorithm quantifies the bias arising from hidden confounders by considering variations across plausible world models consistent with observed data. This pessimistic RL algorithm did demonstrate empirically that it accounts for the uncertainty caused by hidden confounders, but without identifiability there are no theoretical guarantees on the performance of the learnt policy. 

In~\cref{chapter:dmliv}, we propose a novel DML-CMR estimator that solves IV regression with theoretical convergence rate guarantees. We point out that DML-CMR can be directly applied in many of the aforementioned methods (e.g., \citealt{Liao2024,Li2021,Chen2021})
by replacing the IV regression component in their algorithms with DML-CMR, granting them the convergence rate guarantees of DML-CMR.

\subsection{Imitation Learning}\label{lit:il}

Learning from demonstrations has been studied in applications such as autonomous driving~\citep{Pomerleau1988,Lecun2005} and robotics~\citep{Schaal1999}. Standard imitation learning (IL) methods include Behaviour Cloning~\citep{Pomerleau1988}, Inverse RL~\citep{Russell1998}, and adversarial methods~\citep{Ho2016}. IL can suffer from a mismatch between the expert trajectory distribution and the imitator roll-out distribution due to the accumulation of errors. This is first formalised by~\citet{Ross2011}, who demonstrated that interactive experts that suggest actions in states generated by imitator policy roll-outs allow the imitator to recover from mistakes. \citet{Swamy2021} provided theoretical upper and lower bounds on the imitation gap of IL algorithms that only have access to the trajectories (off-policy), and compared their performance against algorithms that additionally have access to an interactive expert and/or a simulator (on-policy). As expected, on-policy algorithms can produce better imitator policies.

\subsubsection{Causal and Confounded Imitation Learning}\label{lit:causal_il}

Recently, IL from offline trajectories has been shown to suffer from the existence of latent variables~\citep{Ortega2021}, which cause causal delusion. This issue can be resolved by learning an interventional policy. Following this discovery, various methods~\citep{Vuorio2022,Swamy2022} considered IL when the expert has access to the full hidden context that is fixed throughout each episode, but the imitator does not observe the hidden context. 
These methods learn an interventional policy through on-policy IL algorithms that require an interactive demonstrator and/or an interactive simulator (e.g., DAgger~\citep{Ross2011}). 

Orthogonal to these works,~\citet{Swamy2022_temporal} considered latent variables unknown to the expert that act as confounding noise that affects the expert policy, but not the transition dynamics. To address this challenge, the problem is then cast into an IV regression problem. 
In~\cref{chapter:il}, we combine and generalise the above works \citep{Vuorio2022,Swamy2022,Swamy2022_temporal} to allow the latent variables to be only partially known to the expert, evolving over time in each episode and directly affecting both the expert policy and the transition dynamics. Solving this generalisation implies solving the above problems simultaneously. 

Causal confusion~\citep{deHaan2019,Pfrommer2023} considers the situation where the expert's actions are spuriously correlated with non-causal features of the previous observable states. While it is implicitly assumed that there are no latent variables present in the environment, we can still model this spurious correlation as the existence of hidden confounders that affect both previous states and current expert actions. Slight variations of this setting have been studied in~\citet{Wen2020,Spencer2021,Codevilla2019}.

From the perspective of causal inference~\citep{Kumor2021,Zhang2020}, there have been studies of the theoretical conditions on the causal graph such that the imitator can exactly match the performance of the expert through backdoor adjustments (\textit{imitability}). Similarly, \citet{Ruan2023} extended imitation conditions and backdoor adjustments to inverse RL. Beyond backdoor adjustments, imitability has also been theoretically studied using context-specific independence relations~\citep{Jamshidi2023}.

\subsection{Learning High-Level Objectives with RL}\label{lit:ltl}

As mentioned above, Linear Temporal Logic (LTL) is a temporal logic language that can specify high-level objectives and is widely used for robotics and multi-agent systems. For unknown MDP environments, it is desirable for an agent to learn policies with respect to these high-level specifications, thus adopting RL. However, it is not straightforward to use RL to learn the optimal policy that maximises the probability of satisfying the LTL specifications~\citep{Alur2022} as discussed in~\cref{chapter:intro}.

Regarding LTL learning with RL, most existing approaches learn a policy within a product MDP, that is, a product of the environment MDP and an automaton translated from the LTL specification. In earlier work, \citet{Sadigh2014ASpecifications} used a deterministic Rabin automaton to create the product and a simple discounted reward design to learn LTL, but it was later proven~\citep{Hahn2019Omega-regularLearning} that this approach does not provide optimality guarantees. Subsequent work is motivated by a new automaton design, called the limit-deterministic Büchi automaton (LDBA)~\citep{Sickert2016Limit-deterministicLogic}, which is more compatible with reinforcement learning because the limited nondeterminism can be handled by adding additional actions for the agent. \citet{Hahn2019Omega-regularLearning} adopted a product MDP with LDBA and augmented it with sink states to reduce the LTL satisfaction problem into a limit-average reward problem with optimality guarantees. \citet{Hahn2020FaithfulObjectives} later modified this approach by including two discount factors with similar optimality guarantee results, and these algorithms are later integrated into the Mungojerrie~\citep{Hahn2021Mungojerrie:Objectives} tool. \citet{Bozkurt2019ControlLearning} proposed a discounted reward learning algorithm on the product MDP with optimality guarantees, where the discount factor is chosen based on certain assumptions of the unknown environment MDP. \citet{Hasanbeig2019ReinforcementGuarantees} proposed an accepting frontier function for the product MDP as the reward for the agent, which works best with multiple goal states that must be visited in succession. \citet{Cai2021ModularLogic} extended this reward frontier to continuous motion control tasks. In~\cref{chapter:ltl}, we propose a novel and generalised product MDP with an RL algorithm that can learn LTL objectives with optimality guarantees through a more in-depth theoretical analysis.

% We propose a generalised product MDP and counterfactual imagining, which allows our algorithm to achieve significantly better sample efficiency compared to the existing work. 
% To the best of our knowledge, these approaches are the only ones that provide optimality guarantees for full LTL language learning. However, many methods have nevertheless demonstrated empirical results for LTL and logic specification learning.

Due to the difficulty of learning the infinite-horizon properties of full LTL language, many approaches focus on learning restricted finite variants of the LTL language. \citet{DeGiacomo2013LinearTraces} and \citet{GiuseppeDeGiacomo2019FoundationsScheduling} introduced the LTLf variant and a corresponding reinforcement learning algorithm for it. \citet{Littman2017Environment-IndependentGLTL} proposed a learning algorithm for the GLTL variant, \citet{Aksaray2016Q-LearningSpecifications} proposed an algorithm to learn Signal Temporal Logic (STL), \citet{Kapoor2020Model-basedSpecifications} developed a model-based approach using model predictive control to learn STL, and \citet{Li2016ReinforcementRewards} introduced a truncated LTL variant for robotics applications. \citet{Cohen2021Model-basedSpecifications} developed a method to learn approximately optimal policies for co-safe LTL formulae in a continuous-time uncertain control system using a Lyapunov-based method to provide guarantees. Tractable extensions to learn STL objectives from RL include~\citet{He2024}, which optimises the value function space, and \citet{Wang2024}, which incorporates the idea of counterfactual experience replay inspired by~\citet{Shao2023SampleGuarantees}'s counterfactual imagining to improve tractability. These finite variants of LTL are still very expressive but are unable to represent properties of an infinite proposition sequence.

Another related line of work leverages automata to deal with non-Markovian rewards. \citet{Icarte2022RewardLearning,ToroIcarte2018UsingLearning} proposed a reward machine automaton to represent high-level non-Markovian rewards, and, by exploiting the structure of the known reward machine, better sample efficiency is achieved. \citet{Camacho2019LTLLearning} introduced a method to learn finite LTL specifications by transforming them into reward machines. \citet{Gaon2019ReinforcementRewards} used L* to learn a standard DFA from non-Markovian rewards to facilitate learning. Note that the expressiveness of reward machines and DFAs is equivalent to that of regular expressions, which can only express finite-horizon properties and is strictly weaker than LTL. However, recent theoretical work~\citep{Le2024} showed that it is possible to translate LTL objectives into a limit average reward paradigm when the reward can be specified with finite memory reward machines. The limit average reward paradigm is different from the classic RL setting, where discounted reward is considered, but it opened a new avenue for future theoretical and empirical work on learning LTL with RL.

Moreover, many works exploit high-level logic specifications, such as LTL, to guide hierarchical and compositional learning, which require less complex samples. \citet{Andreas2016ModularSketches} proposed to modularise the task into subtasks and learn subpolicies for them separately. \citet{Jiang2021Temporal-Logic-BasedTasks} proposed to use reward shaping~\citep{Ng1999PolicyShaping} based on LTL specifications to guide learning. \citet{Jothimurugan2020ATasks} and \citet{Jothimurugan2021CompositionalSpecifications} first proposed a logic language to specify high-level tasks, and used this language to guide the learning of compositional tasks represented on a high-level graph. \citet{Jothimurugan2022Specification-GuidedWelfare} further extended this specification-guided learning to the multi-agent setting, learning joint policies that form a Nash equilibrium. Multi-task RL where each task is specified by an LTL specification is also considered by~\citep{Vaezipoor2021}, where they exploited the compositional syntax of LTL to efficiently generalise to a large number of tasks. \citet{Jackermeier2025} improved multi-task RL with LTL specifications by proposing a novel algorithm that can exploit the structure of Büchi automata, instead of greedily or short-sightedly completing the LTL tasks.

\subsection{Robust and Safe RL}\label{lit:robust_rl}

\bill{In RL, safety concerns preventing the agent from taking actions that lead to undesirable or catastrophic outcomes, while robustness refers to maintaining reliable performance under model uncertainty or adversarial perturbations. Both are essential for the deployment of RL in safety-critical domains such as healthcare and autonomous driving.}

Early works in robust RL has demonstrated that RL agents are vulnerable to adversarial attacks~\citep{Lin2017}, which add small perturbations to the state and reward the agent sees. \citet{Huang2017} also showed that, by adversarially perturbing the policy, the performance of the policy can experience dramatic drops. To combat this, a group of works focuses on learning safe policies using RL. \citet{Mason2017} proposed assured RL, which is a framework that constrains agent behaviour using formally verified policies to achieve safety and optimality requirements. \citet{Berkenkamp2017SafeGuarantees} proposed to constrain the policy to stay within pre-identified safe regions in the state-action space with ensured Lyapunov stability. \citet{Chow2018ALearning} further developed an approach to construct Lyapunov functions as constraints under the constrained Markov decision problem framework to effectively guarantee global safety. \citet{Lutjens2019CertifiedLearning} proposed to compute guaranteed lower bounds on state-action values to choose optimal actions under worst-case adversarial perturbations in the input space, whilst \citet{Zhang2020RobustObservations} further developed a theoretically principled policy regularisation to ensure safety for a wider range of RL algorithms. \citet{Bethell2024} proposed adaptive post-shielding with a contrastive autoencoder to identify and prevent unsafe agent actions.

Another line of work directly verifies whether a given policy is safe in the RL environment. \citet{Wang2019} proposed to integrate neural network certification tools with robust control theory to find invariant sets in the dynamical control loop system that the agent will never escape. \citet{Gupta2020SafetyControllers} proposed a safety verification framework for model-based RL policies by finding all reachable states under the policy and verifying if the reachable set is safe. \citet{Bacci2021VerifyingInfinity} proposed a method to verify the RL agent up to infinite-horizon through abstract interpretations of the state space. Reachable set analysis was performed to ensure that the agent will remain in the safe regions forever. \citet{Bacci2022VerifiedLearning} further extended the abstract interpretation approach to verify probabilistic policies for deep RL with probabilistic guarantees on the execution of that policy.

\subsection{Causal Reinforcement Learning}\label{lit:causal_rl}

Beyond RL with hidden confounders, there is also a variety of research on causal RL, which is an area of research that generally aims to leverage and understand the causal mechanisms and consequences of actions to make better and informed decisions. Usually, additional assumptions or prior knowledge are required for causal RL methods to account for causal relationships beyond associative correlations in the data. Note that causal RL is not a specific problem formulation, but rather a cluster of RL methods with a causality mindset, aimed at solving various problems. Broadly speaking, there are three general challenges that causal RL aims to solve: spurious correlations and hidden confounders, generalisability and knowledge transfer, and efficient online learning. A review of RL methods dealing with spurious correlations and hidden confounders is already provided in~\cref{lit:causal_il} and~\cref{lit:confoundedRL}. In this section, we will introduce causal RL methods that focus on the other challenges. For these methods, it is generally assumed that no hidden confounders are present.

Firstly, we consider causal RL algorithms that leverage causal inference to generalise the knowledge acquired in the training environment to unseen scenarios and tasks. By learning the causal relationships between variables and identifying the invariance in the environment, agents are able to ignore irrelevant variables and generalise to varying environment dynamics. \citet{Zhang2020BlockMDP} proposed to use invariant causal prediction to learn causal representation under a block MDP framework that allows the agent to generalise to diverse observation spaces. 
% In a block MDP, observations are grouped into disjoint blocks, which enables efficient representation learning. 
A similar approach~\citep{Bica2021} proposed to use invariant causal representation to generalise IL across environments. \citet{Wang2022causal} focused on learning causal dynamics in the environment, which helps to eliminate irrelevant variables and dependencies. This allows the policy to focus on the true causal dynamics, which is generalisable. \citet{Saengkyongam2023} proposed to iteratively check invariance conditions for subsets of variables to find a subset of environment-invariant variables. There also exist algorithms~\citep{Sontakke2021,Lee2021,ding2022} that generalise to environments with parts of the causal mechanism changed, such as different gravitational constants. They all try to identify the relevant causal variables and discard irrelevant variables, either through causal representation learning or causal reasoning. For causal RL algorithms that generalise to different tasks, i.e. reward functions, \citet{Eghbal-Zadeh2021} proposed to use an attention module for the agent to incorporate disentangled environmental features into the causal representations as signals for adapting to new tasks. \citet{Pitis2022} considered learning a causal dynamic model to generate counterfactual transitions in a dataset, which are transitions predicted by a causal model (e.g., SCM) when the agent is put into some new environment counterfactually. 
% By training agents on counterfactual data, better generalisation to new tasks is achieved.

Secondly, we discuss causal RL algorithms that improve sample efficiency. The algorithms in this category mainly use causal representation learning, directed exploration, and counterfactual data augmentation. Causal representation learning aims to learn a good representation of the underlying causal structure of the environment, which can help the agent to identify significant and causal features while avoiding irrelevant ones, thereby improving its sample efficiency. Therefore, algorithms~\citep{Sontakke2021,Lee2021,Wang2022causal,Huang2022} that use causal representation learning for generalisation also benefit from improved sample efficiency for this reason. 
% In addition,~\citet{Huang2022} considered high-dimensional and noisy environments, proposing a method that first learns an action-sufficient state representation, which is a minimal set of state variables containing sufficient information for decision-making, through random exploration, and then uses it for sample-efficient learning.
For directed exploration, \citet{Zhang2017} proposed to derive upper and lower bounds for causal effects when the true causal effects are unidentifiable and use the bounds to guide exploration towards more promising actions. \citet{Seitzer2021} first quantifies the causal influence of actions from environment interactions and uses this information for guided exploration. For counterfactual data augmentation, as the name suggests, it includes methods that use counterfactual reasoning to generate additional data to improve sample efficiency. \citet{Buesing2019} introduced counterfactually-guided policy search, which uses an SCM framework to evaluate the outcomes of counterfactual actions using online interactions. Similarly, \citet{Lu2020} used an SCM framework to evaluate the potential outcome of counterfactual actions for agents, which reduces the number of interactions with the environment. \citet{Pitis2020} proposed to use a locally factored dynamics model, where it is assumed that the state-action space can be partitioned into disjoint subsets, to generate counterfactual transitions.

\section{Hidden Confounders}\label{lit:causal}

The problem of \textit{hidden confounders}~\citep{Pearl2000causality} has been studied in many areas of research. In this section, we first focus on reviewing the literature that identify the causal effect between action and outcome in the presence of hidden confounders through instrumental variables (\cref{lit:iv}) and proximal causal learning (\cref{lit:proxy}), which is closely related to the work in~\cref{chapter:dmliv} and~\cref{chapter:il}. Next, in~\cref{lit:dml}, we provide an overview of the work related to double machine learning, which is a statistical technique that we use to tackle the problem of hidden confounders in~\cref{chapter:dmliv}.

\subsection{Instrumental Variable Regression via Conditional Moment Restriction Estimation}\label{lit:iv}

Instrumental variable (IV) regression is a typical example of a conditional moment restriction (CMR) problem, and CMR estimators can be generally applied to solve IV regression directly. Therefore, we first introduce CMR estimators and then introduce algorithms that are proposed specifically for IV regression.

The classic framework of the generalised method of moments (GMM)~\citep{Hansen1982} was first proposed to translate conditional moment restrictions into unconditional moments, which can thus be estimated under the GMM framework. However, finding the set of unconditional moments that fully capture the conditional moments is challenging. Sometimes, for nonlinear CMRs and functions of interest $f$, an infinite set of unconditional moments may be required to represent the conditional moments, and a misspecification of the unconditional moments can bias the results~\citep{Domnguez2004}. To mitigate some of these issues, \citet{Angrist1996} considered linear functions of interest and proposed the classic two-stage least squares (2SLS) algorithm as a special case of GMMs. Following this, many efforts extend 2SLS to solve CMRs, where the function of interest is nonlinear or nonparametric. One common approach is the sieve method, which uses nonlinear basis functions. Sieve minimum distance (SMD) estimator~\citep{Newey2003,Ai2003} performs regression in two stages using an increasing set of nonlinear
basis functions as the number of samples increases. Later, a penalised version of SMD~\citep{Chen2012} was proposed to generalise SMD by allowing non-smooth residuals and high-dimensional function spaces. These sieve-GMM methods and later works~\citep{Blundell2007,Chen2018,Singh2019,Muandet2020} that consider different dictionaries of basis functions enjoy strong consistency and efficiency guarantees, %from the sieve method literature, 
but their flexibility is limited by the set of basis functions, and they can be sensitive to the choice of such functions and regularisation hyperparameters. Note that these works also perform the estimation in two stages.

More recently, various  CMR estimation methods based on DNNs %machine learning (ML) 
have been proposed, since machine learning methods can model highly nonlinear and high-dimensional relationships with greater flexibility. DeepIV~\citep{Hartford2017DeepPrediction} extended the classic 2SLS algorithm to the nonlinear setting by adopting DNNs for both stages with conditional density estimation~\citep{Darolles2011} in the first stage. GMM methods are also extended to use DNNs~\citep{Bennett2019DeepAnalysis,Liao2020,Dikkala2020,Bennett2020}, where a minimax criterion is optimised adversarially. Minimax approaches solve a two-player zero-sum game where the players play adversarially. Specifically, player one chooses a hypothesis function to minimise moment violation, and player two chooses a test function that maximises moment violation. These DNN-based GMM methods require minimax optimisation, which is similar to the training of Generative Adversarial Networks~\citep{Goodfellow2014} and could be experimentally unstable. In~\cref{chapter:dmliv}, we propose a novel CMR estimator, DML-CMR, that uses the double machine learning~\citep{Chernozhukov2018Double/debiased} framework, which we will introduce in~\cref{sec:dml}, and DNNs to provide fast convergence rate guarantees for the estimator.

% In this thesis, we propose a doubly robust CMR estimator, DML-CMR, in~\cref{chapter:dmliv}. It uses the double machine learning~\citep{Chernozhukov2018Double/debiased} framework and DNNs to provide fast convergence rate guarantees for the estimator. To our knowledge, this is the first work to use Neyman orthogonality for the general nonlinear IV regression problem where we proved a fast semiparametric $N^{1/2}$ convergence rate for DML-CMR under mild assumptions over the function classes of nuisance parameters. In addition, our approach empirically compares favourably to Deep IV, DeepGMM, Kernel IV, and DFIV.

For algorithms that are developed specifically for IV regression, DFIV~\citep{Xu2020} proposed to use basis functions parameterised by DNNs, which remove restrictions on the functional form. Kernel IV~\citep{Singh2019} and Dual IV~\citep{Muandet2020} used different dictionaries of basis functions in reproducing kernel Hilbert spaces (RKHS) to solve the IV regression problem. DeepGMM~\citep{Bennett2019DeepAnalysis} is a DNN-based method that was inspired by GMM to solve IV regression using a minimax approach. \citet{Kremer2024} improved GMM-based IV regression methods in settings where the data manifold is not uniform through data-derivative information. \citet{Chen2023} presented two semiparametric efficient procedures to estimate a weighted average derivative, and therefore nonparametric IV regression. Their procedures use neural network sieves to approximate the unknown IV functions. They also showed that, with a good choice of sieves, their approach can achieve the semiparametric convergence rate. In addition, other scenarios where the hidden confounder is correlated with some observed confounders, or context, that also affect both the action and the outcome, have been studied by~\citet{wu2022}. The correlation will cause the second stage in two-stage IV regression to generate imbalanced observed confounders, and they propose a method to balance the generated confounders that can improve the consistency of the algorithm. Extending this setting, \citet{Cheng2024} considered nonlinear IV regression with IVs that can be correlated with the observed confounder, called conditional IVs. They proposed an IV regression approach leveraging conditional IVs through confounding balancing and representation learning.

\subsection{Proximal Causal Learning}\label{lit:proxy}

\bill{Proximal Causal Learning (PCL), which considers the problem of learning causal effects with hidden confounders, is also an instance of CMR problems}. PCL was first proposed by~\citet{Miao2018} to leverage two \textit{proxy variables} for causal identification in estimating the causal function. This was extended by~\citet{Shi2020} to a general semiparametric framework, where \citet{Tchetgen2020} introduced a two-stage procedure for linear causal models based on ordinary least squares regression. \citet{Mastouri2021} resolved how to handle nonlinear causal models by replacing linear regression with kernel ridge regression. To extend the kernel-based PCL methods, \citet{Xu2021} used DNNs as feature maps instead of fixed kernels. This improves the flexibility of the method, especially for highly nonlinear models. 
\citet{Kompa2022} proposed a single-stage PCL method based on %the formulation of 
maximum moment restrictions, where they train neural networks to minimise a loss function derived to satisfy the maximum moment restrictions. \citet{Cui2023} introduced a treatment bridge function and incorporated it into the Proximal Inverse Probability Weighting (PIPW) estimator. They considered only binary treatments and derived the Proximal Doubly Robust (PDR) estimator via influence functions. A similar approach by \citet{Wu2024} derived a doubly robust estimator for PCL with continuous treatment through influence functions.

The algorithms proposed specifically for IV regression and PCL often require additional problem-specific assumptions about variables and the functional form. We instead provide a general method for solving CMRs, in~\cref{chapter:dmliv}, that can be directly applied to a range of problems, including IV regression and PCL.

\subsection{Double Machine Learning for Hidden Confounders}\label{lit:dml}

Double machine learning (DML) was originally proposed for semiparametric regression~\citep{Robinson1988}. It relies on the derivation of a Neyman orthogonal~\citep{Neyman1965} score function that describes the regression problem as the learning objective. DML was then extended by adopting DNNs for generalised linear regressions~\citep{Chernozhukov2021AutomaticRegression}. Its strength is that it provides unbiased estimates for two-stage estimations~\citep{Jung2021,Chernozhukov2022RieszNetForests} under certain identifiability conditions and offers $N^{-1/2}$ convergence rate guarantees, where $N$ is the sample size.

There are previous works on combining DML with CMR estimation, specifically for the IV regression problem, but they are mainly focused on linear and partially linear functions of interest. \citet{belloni2012sparse} proposed a method to use Lasso and Post-Lasso methods for the first-stage estimation of linear IV to estimate the optimal instruments. To avoid selection biases, \citet{belloni2012sparse} leveraged techniques from weak identification robust inference. In addition, \citet{chernozhukov2015post} proposed a Neyman-orthogonalised score for the linear IV problem with control and instrument selection to potentially be robust to regularisation and selection biases of Lasso as a model selection method. Neyman orthogonality for partially linear models with IVs was mainly discussed in the work of \citet{Chernozhukov2018Double/debiased}. For an additional discussion, we refer to the book \citep{chernozhukov2024applied}.

DML for semiparametric models~\citep{Chernozhukov2022LocallyEstimation,Ichimura2022} has been previously applied to solve the nonparametric IV (NPIV) problem. However, their methods require that the average moment of the Neyman orthogonal score is affine (linear) in the nuisance parameters. Therefore, when applied to solve NPIV, functional assumptions regarding the IV set and the residual function were made. Such assumptions are not required in~\cref{chapter:dmliv} since we are considering a different problem setting and we formulate a novel Neyman orthogonal score. Furthermore, in contrast to our work in~\cref{chapter:dmliv}, no implementation and empirical evaluation are performed by these works, and thus it is unclear whether any practical estimator that enjoys these theoretical properties can be derived from their score.

\subsubsection{Double Machine Learning for Causal Inference}\label{lit:causal_dml}

For completeness, we also briefly survey works that use DML for estimating the causal effect without hidden confounders. Doubly robust estimation for causality problems revolved predominantly around the estimation of average treatment effects (ATE)~\citep{robins1994estimation, funk2011doubly,benkeser2017doubly,bang2005doubly,sloczynski2018general}. Recently, there has been a surge in doubly robust identification of causal structures beyond the ATE settings. \citet{korth,quinzan2023drcfs} focus on finding direct causes of the target variable using orthogonalised scores. Furthermore, DML techniques for identifying the local average treatment effects (LATE) for nonlinear models with a binary instrument and treatment (action) have been explored before \citep{chernozhukov2024applied}.

\section{Summary}\label{lit:summary}

In this section, we summarise the literature reviewed in this chapter and discuss the strengths and limitations of the current state-of-the-art (SOTA) methods. We then position the contributions of this thesis within the existing literature, highlighting the gaps in the literature that this thesis addresses.

The works on online and offline RL, introduced in \cref{lit:online_rl} and \cref{lit:offline_rl}, respectively, have demonstrated significant progress in developing practical and high-performing policy learning algorithms with theoretical advancements. However, these approaches do not consider the presence of hidden confounders in the RL environment. They are included in this thesis to provide an overview of the policy learning literature and background for policy learning under hidden confounders. Similarly, the literature on robust RL and causal RL, discussed in \cref{lit:robust_rl} and \cref{lit:causal_rl}, respectively, is included for completeness due to its relevance to this thesis's aim of developing practical and optimal policies with theoretical guarantees.

The works pertaining to RL with hidden confounders, reviewed in \cref{lit:confoundedRL}, are related to \cref{chapter:dmliv}. However, as explained in \cref{lit:confoundedRL}, these works focus on applying existing IV regression and causal inference techniques to the confounded RL setting, which involves sequential decision-making. In contrast, \cref{chapter:dmliv} proposes a novel CMR estimator, DML-CMR, that solves the IV regression and the offline IV bandit problem, which is a single-step decision-making problem with hidden confounders and IVs.

In \cref{lit:causal}, we reviewed the literature on CMR estimation, IV regression, and PCL. As discussed in that section, classical approaches that use the GMM framework with sieve and kernel methods offer strong theoretical guarantees, particularly in terms of consistency. However, their practicality is often limited when dealing with high-dimensional or complex function classes, leading to degraded estimation performance or intractable computational costs. Modern CMR estimators leverage DNNs to address these challenges, but often lack theoretical guarantees on consistency, especially on convergence rates. In \cref{chapter:dmliv}, our proposed DML-CMR estimator mitigates both theoretical and estimation challenges by employing DNNs for flexible function estimation while providing a fast convergence rate of $O(N^{-1/2})$ through the DML framework. To the best of our knowledge, no existing work adopts the DML framework with DNNs for nonlinear IV regression or general CMR problems. Furthermore, many methods developed specifically for IV regression and PCL are not generalised to solve the full class of CMR problems. In contrast, DML-CMR provides a solution applicable to general CMR problems.

In \cref{lit:il}, we reviewed the literature on IL, with a particular focus on confounded IL in \cref{lit:causal_il}. While prior work has explored IL with confounded datasets, as discussed, these approaches only address isolated aspects of the problem under specific assumptions and settings. A holistic treatment of the issue, along with a discussion of the connections between these settings, is still lacking in the literature. In \cref{chapter:il}, we propose a unifying framework that generalises existing methods and allows the solution of a broader class of confounded IL problems with optimality guarantees.

In \cref{lit:ltl}, we reviewed RL approaches for learning high-level objectives, with recent works showing good empirical results for LTL and other logic-based specifications. However, many of these methods lack optimality guarantees or are restricted to finite variants of LTL. Among the works that provide optimality guarantees for full LTL, as discussed in \cref{lit:ltl}, their theoretical results rely on key hyperparameter choices, either by proving the existence of such hyperparameters without any constructive methods to obtain them or by providing unnecessarily harsh bounds. This presents a practical challenge --- without appropriate guidance for selecting these hyperparameters, user misspecification can result in suboptimal policies or inefficient learning. In \cref{chapter:ltl}, we propose a novel RL algorithm for learning full LTL objectives that not only provides optimality guarantees, but also achieves substantially improved sample efficiency over existing methods. Moreover, our improved theoretical analysis includes an explicit procedure for selecting key hyperparameters to ensure optimality in practice.
\chapter{Learning Policies from Confounded Datasets with Instrumental Variables}\label{chapter:dmliv}

\minitoc

\section{Introduction}

In this chapter, we focus on learning sample-efficient and
optimal policies from offline datasets in the presence of hidden confounders, where IVs
are observed in the dataset. Specifically, we learn the causal function using IVs (also known as \textit{IV regression}), in order to learn a decision policy that maximises the expected reward (which we refer to as the \textit{offline IV bandit} problem, described in~\cref{sec:offlineivbandit}) with suboptimality guarantees. As introduced in~\cref{background:CMRs}, IV regression involves solving a CMR problem, but solving it analytically is ill-posed~\citep{Nashed1974,kress1999linear} because it is an inverse problem. Therefore, various techniques have been proposed to estimate the solutions to CMR problems. 

Recall from~\cref{lit:causal} that most existing CMR methods, including those that adopt DNNs, take a two-stage approach~\citep{Angrist1996,Newey2003,Chen2018,Singh2019,Muandet2020}. In the first stage, they estimate some nuisance parameters, which are parameters or infinite-dimensional functions of no direct interest, but are necessary for the second-stage estimation. However, in these settings, regularisation is often employed to trade off overfitting with the induced regularisation bias, especially for high-dimensional inputs. This is problematic because both regularisation and overfitting can cause heavy bias~\citep{Chernozhukov2018Double/debiased} in two-stage estimations when the first-stage estimator is naively plugged in for the second-stage estimation, which results in a slow convergence rate of the estimator.

To address this challenge, we take inspiration from \textit{double/debiased machine learning}~\citep{Chernozhukov2018Double/debiased} (DML), which is a statistical framework that provides unbiased estimators with strong convergence rate guarantees for general two-stage regression problems. DML relies on having a Neyman orthogonal~\citep{Neyman1965} score function, which describes the estimation problem, to deal with regularisation bias. In addition, it uses cross-fitting, that is, an efficient form of (randomised) data splitting, to address overfitting bias. The DML framework is well-suited for CMR estimation as it allows the use of flexible and powerful estimators such as DNNs, while still providing fast convergence guarantees, addressing key limitations in the literature (recall ~\cref{lit:summary}).
% These properties address key limitations in the literature, as discussed in~\cref{lit:summary}. 

% However, the use of DML for CMR estimation involving DNNs has not been explored to the best of our knowledge.

In this chapter, we propose DML-CMR, a novel CMR estimator with fast convergence rate guarantees based on the DML framework. We derive a novel Neyman orthogonal score for CMR problems and design a cross-fitting regime such that, under mild regularity conditions, our estimator is guaranteed to converge at the rate of $N^{-1/2}$, where $N$ is the sample size. We then adopt DML-CMR to solve the offline IV bandit problem, where we derive a policy from the DML-CMR estimator and provide a $O(N^{-1/2})$ suboptimality bound with high probability that matches the suboptimality bounds of \textit{unconfounded} offline bandit algorithms~\citep{Jin2021,Nguyen-Tang2022}. Finally, we evaluate DML-CMR on multiple benchmarks for IV regression, offline IV bandit, and proximal causal learning (PCL).
% where superior results are demonstrated compared to state-of-the-art (SOTA) methods. 
% In addition, we provide evaluation of DML-CMR on another CMR problem, proximal causal learning, where SOTA performance is demonstrated.

\paragraph{Main Contributions of this Chapter.} 
\begin{itemize}[leftmargin=10pt, topsep=2pt, itemsep=2pt, topsep=2pt]
\item We propose DML-CMR, % in~\cref{sec:3}, 
a novel CMR estimator that leverages the DML framework to provide unbiased estimates of the solutions to CMR problems, which we use to solve the offline IV bandit problem.
\item We derive a novel, Neyman orthogonal, score function for CMR problems in~\cref{sec:neyman},
and design a cross-fitting regime for the DML-CMR estimator to mitigate the bias in \cref{sec:dmliv}.
\item In \cref{sec:theory}, we show an asymtoptic convergence rate for the DML-CMR estimator at the rate of $N^{-1/2}$, which is minimax optimal under parameterisation and mild regularity conditions, leading to $O(N^{-1/2})$ suboptimality for the derived policy proved in \cref{sec:ivbandit_subopt}.
\item On a range of IV regression, offline IV bandit, and PCL benchmarks, including two real-world datasets, we empirically demonstrate that DML-CMR outperforms other SOTA methods in \cref{sec:exp}.
\end{itemize}

Work reported in this chapter first appeared in~\citet{Shao2024,Shao2025cmr}.

\subsection{Double Machine Learning}\label{sec:dml}

Before introducing the problem studied in this chapter, we first describe the DML framework. DML considers the problem of estimating a function of interest $f$ as a solution to an equation of the form
\begin{equation}
\label{eq:score_function}
    \expectE[\orthoM(\dataset;f_0, \eta) ] =0,
\end{equation}
where $\orthoM$ is referred to as a score function and $f_0$ is the true function. Here, $\eta$ is a nuisance parameter, which can be of parametric form or infinite-dimensional functions. It is of no direct interest, but must be estimated to obtain an estimate of $f_0$. For example, in a two-stage CMR estimator, nuisance parameters such as conditional density are estimated in the first stage, and in the second stage, they are used to estimate $f_0$. DML provides a set of tools to derive an unbiased estimator of $f_0$ with convergence rate guarantees, even when the nuisance parameter $\eta$ suffers from regularisation, overfitting, and other types of biases present in the training of ML models, which typically cause slow convergence when learning $f_0$.
% induces a bias in the model, due to regularisation and a slow convergence rate.

In order to estimate $f_0$, DML reduces biases by using score functions $\orthoM$ that are Neyman orthogonal~\citep{Neyman1965} in $\eta$. This requires the Gateaux derivative, which defines the directional derivative for functionals, of the score function $\orthoM$ w.r.t. the nuisance parameters at $f_0,\eta_0$ to be zero:
\begin{align}
\label{eq:neyman}
\frac{\partial}{\partial r}\Big\lvert_{r=0} \expectE[\orthoM(\dataset;f_0,\eta_0+ r\eta)] = 0 \quad \text{for all } \eta.
\end{align}
Here, $f_0$ and $\eta_0$ are the true parameters where $\expectE[\orthoM(\dataset;f_0,\eta_0)] = 0$. Intuitively, the condition in~\cref{eq:neyman} is met if small changes in the nuisance parameter do not significantly affect the score function around the true function $f_0$. Neyman orthogonality is key in DML, as it allows fast convergence for estimating $f_0$, even if the estimator for the nuisance parameter $\eta$ is biased. For score functions that are Neyman orthogonal, we define DML with \textit{K-fold cross-fitting} as follows.
\begin{definition}[DML, Definition 3.2~\citep{Chernozhukov2018Double/debiased}]\label{defn:dml}
Given a dataset $\mathcal{D}$ of $N$ observations, consider a score function $\orthoM$ as in~\cref{eq:score_function}, and suppose that $\orthoM$ is Neyman orthogonal that satisfies~\cref{eq:neyman}. Take a \textit{K-fold} random partition $\{I_k\}^K_{k=1}$ of observation indices $[N]$ each with size $n=N/K$, and let $\mathcal{D}_{I_k}$ be the set of observations $\{d_i:i\in I_k\}$. Furthermore, define $I^c_k\coloneqq [N]\setminus I_k$ for each fold $k$, and construct estimators $\widehat{\eta}_k$ of the nuisance parameter using $\mathcal{D}_{I^c_k}$. Then, construct an estimator $\widehat{f}$ as a solution to the following equation,
\begin{align}
\label{eq:sol_model}
    \frac{1}{K}\sum_{k=1}^K \widehat{\expectE}_{k}[\orthoM(\dataset_{I_k};\widehat{f},\widehat{\eta}_k)]=0,
\end{align}
where $\widehat{\expectE}_k$ is the empirical expectation over $\mathcal{D}_{I_k}$.
\end{definition}
In the above definition, $\widehat{f}$ is defined as an exact solution to the empirical expectation equation in~\cref{eq:sol_model}. In practice, we can also define the estimator $\widehat{f}$ as an approximate solution to~\cref{eq:sol_model}.\footnote{This approximation error is different from the estimation error. The estimation error measures the difference between $\widehat{f}$ and $f_0$, whereas the approximation error concerns the error of minimising the empirical risk. In fact, the approximation error contributes to the estimation error, which is analysed in~\cref{sec:theory}.}

\section{Problem Formulation}

We can now formally state the problem considered in this chapter. We first introduce the contextual IV setting with the corresponding offline IV bandit problem and discuss their connection to CMR problems.

\subsection{Contextual IV Setting}\label{sec:contextualIV}

\begin{figure}[tb]
\centering
\includegraphics[width=0.6\textwidth]{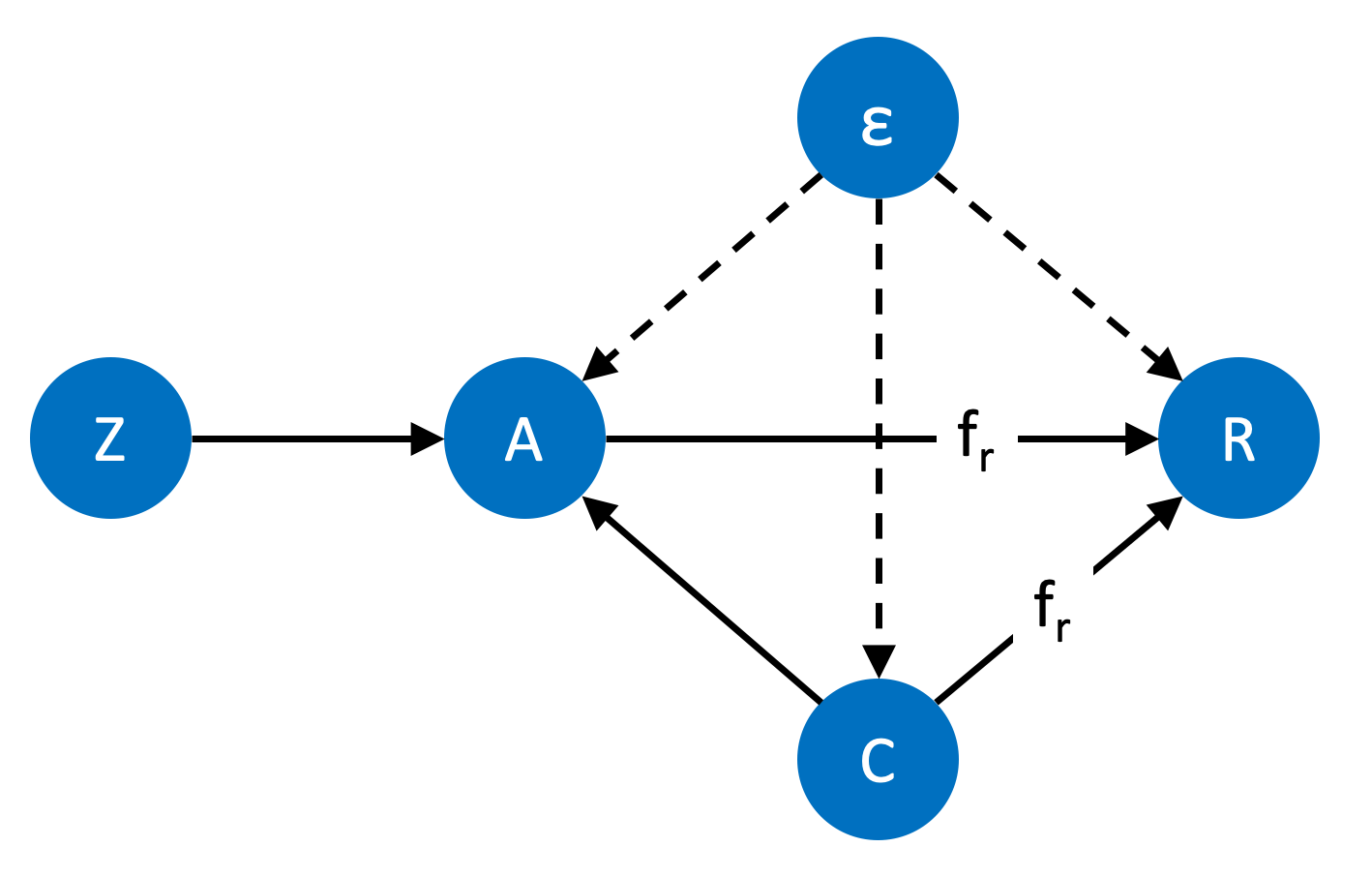}
    \caption[The causal graph of the contextual IV setting]{The causal graph of the contextual IV setting, where $R=f_r(C,A)+\epsilon$ and $Z$ is an \textit{instrumental variable} that affects $R$ only through $A$.}
    \label{fig:SCM}
    % \vskip -0.2in
\end{figure}

We begin with a description of the contextual IV setting~\citep{Hartford2017DeepPrediction} that is considered in this chapter. We observe an \textit{action} $A\in\mathcal{A}\subseteq\realNumber^{d_A}$, a \textit{context} $C\in\mathcal{C}\subseteq\realNumber^{d_C}$, an \textit{IV}  $Z\in\mathcal{Z}\subseteq\realNumber^{d_Z}$ and an \textit{outcome} $R\in\realNumber$, where there exist \textit{unobserved confounders} that affect all of $A$, $C$ and $R$ through a hidden variable (or \textit{noise}) $\epsilon$. IV directly affects the action $A$, does not directly affect the outcome $R$, and is not correlated with the hidden confounder $\epsilon$. These causal relationships are illustrated in~\cref{fig:SCM} and are represented by the following structural causal model (recall \cref{background:SCM}):
\begin{align}
    &R\coloneqq f_r(C,A)+\epsilon,\label{eq:reward}\quad\expectE[\epsilon]=0, \quad\expectE[\epsilon\lvert A,C]\neq0,
\end{align}
where $f_r$ is an unknown, continuous, and potentially nonlinear causal function. Denote the set of observations $\{(c_i,z_i,a_i,r_i)\}_{i\in[N]}$ generated from this model as the \textit{offline dataset} $\dataset$ of size $N$. In this chapter, we consider the problem of learning the \textit{counterfactual prediction function}~\citep{Hartford2017DeepPrediction},
\begin{align}\label{eq:IV-CATE}
f_0(C,A):=f_r(C,A)+\expectE[\epsilon \lvert C]=\expectE[R\lvert do(A), C]
\end{align}
The term $\expectE[\epsilon \lvert C]$ is typically nonzero\footnote{In the setting where $\expectE[\epsilon\lvert C]=0$ is assumed~\citep{Bennett2019DeepAnalysis,Xu2020}, $f_0=f_r$ and all our results apply.}, but learning $f_0$ still allows us to compare between different actions when given a context as $f_0(C,a_1)-f_0(C,a_2)=f_r(C,a_1)-f_r(C,a_2)$ for all $a_1,a_2\in\actions$, and in particular, $\argmax_a f_0(C,a)=\argmax_a f_r(C,a)$.

As discussed in~\cref{background:IV}, learning $f_0$ in this setting is an IV regression problem with observed context $C$, provided that $Z$ satisfies the IV conditions in~\cref{def:iv}. Furthermore, it can be formulated as a CMR problem of the form $\expectE[R-f_0(C,A)\lvert Z,C]=0$. Therefore, we will be able to use the novel CMR estimator with fast convergence rate guarantees proposed in~\cref{sec:3} to learn the counterfactual prediction function $f_0$.

\subsection{Offline IV Bandit}\label{sec:offlineivbandit}

The learnt estimator of $f_0$ from the offline dataset $\dataset$ can be used to solve the \textit{offline bandit problem in the contextual IV setting}~\citep{Zhang2022}, that is, to identify a (deterministic) policy $\policy:\mathcal{C}\rightarrow\actions$ that maximises the value,
$$\Value(\pi)\coloneqq\expectE_{c\sim\testD}[R\lvert do(A=\pi(c)),c]=\expectE_{c\sim\testD}[f_0(c,\pi(c))],$$ which is the expected outcome when performing actions following $\policy$. $\testD$ is a test context distribution that can potentially differ from the distribution of $\dataset$. The optimal policy $\policy^*$ should satisfy $\Value(\policy^*)=\max_\policy \Value(\policy)$, and suboptimality of a policy $\policy$ is defined as $\textrm{subopt}(\policy)\coloneqq\Value(\policy^*)-\Value(\policy)$. Furthermore, the optimal policy $\policy^*$ can be retrieved from $f_0$ by selecting $\policy^*(c)=\argmax_{a\in\mathcal{A}} f_0(c,a)$.

% It is natural to extend the IV setting to the offline contextual bandit problem, which performs an action, represented by a do intervention $(do(A=a))$, given a context that will produce the highest expected outcome (or reward). Formally, given a offline dataset $\dataset$ that is generated by~\cref{eq:reward} that satisfies~\cref{assump:struc}, the task of offline contextual bandit is to find a (deterministic) policy $\policy:\realNumber^{d_C}\rightarrow\actions$ that maximises the value $\Value(\pi)\coloneqq\expectE_{C\sim\testD}[R\mid C,do(A=\policy(C))]$, which is the expected reward following policy $\policy$, on some test context distribution $\testD$ that can be potentially different to the distribution of $\dataset$. We define the optimal value as $\Value^*=\Value(\policy^*)=\max_\policy \Value(\policy)$, where $\policy^*$ denotes the optimal policy, and the suboptimality is defined as
% \begin{align}
% subopt(\policy)=\Value^*-\Value(\policy)\label{eq:subopt}.
% \end{align}

\section{Solving Conditional Moment Restrictions with DML}\label{sec:3}

We now present the main contribution of this chapter --- an estimator that solves CMR problems using the DML framework. We first propose a novel Neyman orthogonal score for the estimator and a novel two-stage algorithm that estimates the solutions to CMR problems. Our estimator can use DNN estimators in both stages and provides guarantees on the convergence rate by leveraging the DML framework. Then, we adopt this CMR estimator to solve the offline IV bandit problem.

To solve the CMR problem defined in~\cref{eq:cmr} using the DML framework, we first need a Neyman orthogonal score. As introduced in~\cref{background:CMRs}, the optimisation objective for many two-stage CMR estimators~\citep{Angrist1996,Hartford2017DeepPrediction,Singh2019} is $\hat{f} \in \arg\min_{f\in\mathcal{F}} \expectE[(Y-\expectE[f(X)\lvert C])^2]$ (recall \cref{eq:optim}). Let $g_0(f,c)\coloneqq\expectE[f(X)\lvert c]$ and $\mathcal{G}$ be some function space that includes $g_0$ and its potential estimators $\widehat{g}$. Then, these two-stage CMR estimators estimate $g_0$ with $\widehat{g}$ in the first stage, and then use $\widehat{g}$ to optimise the following loss function, $\ell=(Y-\widehat{g}(f,c))^2$, in the second stage. However, as we show in \cref{prop:standard_score} with proof deferred to~\cref{appen:score}, this objective, or the score function, is not Neyman orthogonal. 

\begin{proposition}\label{prop:standard_score}
The score (or objective) function for standard two-stage CMR estimators, 
$\ell=(Y-\widehat{g}(f,c))^2$, 
is not Neyman orthogonal at $(f_0, g_0)$.
\end{proposition}

This means that small misspecifications or estimation biases of $\widehat{g}$ can lead to significant changes to the score function, and there are no guarantees on the convergence rate if the first-stage estimator $\widehat{g}$ is naively plugged into the second stage to estimate $f_0$. To address this, we first derive a novel Neyman orthogonal score function for the CMR problem and then design a CMR algorithm with K-fold cross-fitting that uses the DML framework.

\subsection{Neyman Orthogonal Score}\label{sec:neyman}

Typically, to construct a Neyman orthogonal score from a non-orthogonal score, additional nuisance parameters need to be estimated~\citep{Chernozhukov2018Double/debiased}. These additional nuisance terms adjust the score in a way that makes it orthogonal, where the error in estimating $f_0$ due to errors in the nuisance parameters becomes second order in the Taylor expansion~\citep{Foster2019OrthogonalLearning}. In our case, to construct a Neyman orthogonal score for CMR problems from the standard objective in~\cref{eq:optim}, we first select relevant functions that should be estimated as nuisance parameters. Following two-stage IV regression approaches~\citep{Hartford2017DeepPrediction}, estimating $g_0$ is essential for identifying $f_0$, so we will estimate it as a nuisance parameter. We find that, by additionally estimating $s_0(c)\coloneqq\expectE[Y\lvert c]$ inside some function space $\mathcal{S}$, we can construct the score function:
\begin{equation}
\orthoM(\dataset;f,(s,g))=(s(c)-g(f,c))^2.\label{eq:neyman_score}
\end{equation}
Here, the nuisance parameters are $\eta=(s,g)$. For this to be a valid Neyman orthogonal score function, we check, with the following theorem, that
\begin{itemize}
    \item $\expectE[\orthoM(\dataset;f_0,(s_0,g_0))]=0$ with the true functions $(s_0,g_0)$;
    \item The Gateaux derivative of $\orthoM(\dataset;f,(s,g))$ vanishes at $(f_0,(s_0,g_0))$.
\end{itemize}
The proof is deferred to~\cref{appen:neyman}.
\begin{theorem}[Neyman orthogonality]\label{thm:neyman}
The score function $\orthoM(\dataset;f,(s,g))=(s(c)-g(f,c))^2$ obeys the Neyman orthogonality conditions at $(f_0,(s_0,g_0))$.
\end{theorem}
This Neyman orthogonal score function is abstract in the sense that it allows for general estimation methods for $g_0$ and $s_0$, as long as they satisfy certain convergence conditions, which are introduced in the next section. In addition, having a Neyman orthogonal score is useful in general to debias two-stage estimators~\citep{Foster2019OrthogonalLearning}, beyond the DML framework.

\subsection{A DML Estimator for Solving CMR Problems --- DML-CMR}\label{sec:dmliv}

With the Neyman orthogonal score, we now propose a novel DML estimator, DML-CMR, that solves~\cref{eq:cmr}. Note that, in general, $f_0$ is allowed to be infinite-dimensional, as commonly seen in the nonparametric IV literature~\citep{Newey2003}. We also allow $f_0$ to be infinite-dimensional for the Neyman orthogonal score introduced in~\cref{sec:neyman}. For the theoretical analysis of DML-CMR, while it is possible to provide a general analysis following~\citet{Foster2019OrthogonalLearning} for nonparametric $f_0$ with the Neyman orthogonal score, the analysis would require more assumptions and the convergence rate will depend on the complexity of the function classes involved in the estimation. For our analysis, since we propose a concrete estimator, we would like to provide a concrete analysis following the DML framework~\citep{Chernozhukov2018Double/debiased}, which is designed for semiparametric estimation, to show an optimal parametric rate for DML-CMR. Therefore, we assume that $f_0$ is finite-dimensional and parameterised for the theoretical analysis of DML-CMR.

\begin{assumption}[Parameterisation]\label{assump:parameter}
Let $f_0=f_{\theta_0}$ and $\Theta\subseteq \realNumber^{d_\theta}$ be a compact space of parameters of $f$, where the true parameter $\theta_0\in\Theta$ is in the interior of $\Theta$.
\end{assumption}
From this assumption, we can define $\mathcal{F}\coloneqq\{f_\theta:\theta\in\Theta\}$ as the function space of $f$.

\paragraph{The DML-CMR Estimator.} 
The procedure of our DML-CMR estimator with K-fold cross-fitting is outlined in~\cref{alg:dml-iv-kf}. Given a dataset $\dataset=(y_i,x_i,c_i)_{i\in[N]}$ of size $N$, we first split the dataset using a random partition $\{I_k\}^K_{k=1}$ of dataset indices $[N]$ such that the size of each fold $I_k$ is $N/K$, 
and let $\mathcal{D}_{I_k}$ denote the set of observations $\{d_i:i\in I_k\}$.

\begin{algorithm}[tb]
   \caption{DML-CMR with K-fold cross-fitting}
   \label{alg:dml-iv-kf}
\begin{algorithmic}[1]
   \STATE {\bfseries Input:} Dataset $\dataset$ of size $N$, number of folds $K$ for cross-fitting, mini-batch size $n_b$
   \STATE Get a partition $(I_k)^K_{k=1}$ of dataset indices $[N]$
   \FOR{$k=1$ {\bfseries to} $K$}
   \STATE $I^c_k\coloneqq[N]\setminus I_k$
   \STATE Learn $\widehat{s}_k$ and $\widehat{g}_k$ using $\{(\dataset_i):{i\in I^c_k}\}$
   \ENDFOR
   \STATE Initialise $f_{\widehat{\theta}}$
   \REPEAT
   \FOR{$k=1$ {\bfseries to} $K$}
   \STATE Sample $n_b$ data $c_i^k$ from $\{(\dataset_i):{i\in I_k}\}$
   \STATE $\mathcal{L}=\widehat{\expectE}_{c_i^k}\left[(\widehat{s}_k(c)-\widehat{g}_k(f_{\widehat{\theta}},c))^2\right]$
   \STATE Update $\widehat{\theta}$ to minimise loss $\mathcal{L}$ 
   \ENDFOR
    \UNTIL{convergence}
    \STATE {\bfseries Output:} The DML-CMR estimator $f_{\widehat{\theta}}$
\end{algorithmic}
\end{algorithm}

As introduced in~\cref{sec:dml}, our DML estimator will be a two-stage procedure. In the first stage (lines 4-7 in~\cref{alg:dml-iv-kf}), for each fold $k\in [K]$, we learn $\widehat{s}_k$ and $\widehat{g}_k$ using data $\dataset_{I^c_k}$ with indices $I^c_k\coloneqq[N]\setminus I_k$. Then $\widehat{s}_k\approx\expectE[Y\lvert C]$ can be learnt through standard supervised learning using a neural network with inputs $C$ and labels $Y$. For $\widehat{g}_k$, we follow~\citet{Hartford2017DeepPrediction} to estimate $F_0(X\lvert C)$, the conditional distribution of $X$ given $C$, with $\widehat{F}$, and then estimate $\widehat{g}$ via
\begin{equation*}
\widehat{g}(f_\theta,c)=\sum_{\dot{X}\sim \widehat{F}(X\lvert C)} f_\theta(\dot{X})\approx \int f_\theta(X)\widehat{F}(X\lvert C=c) dX\approx\expectE[f_\theta(X)\lvert c].
\end{equation*}
For example, if the action space is discrete, $\widehat{F}$ can be a categorical model, e.g., a DNN with softmax output. For a continuous action space, a mixture of Gaussian models can be adopted to estimate the distribution $F_0(X\lvert C)$, where a DNN is used to predict the mean and standard deviation of the Gaussian distributions.

In the second stage (lines 8-15 in~\cref{alg:dml-iv-kf}), we estimate $\widehat{\theta}$ using our Neyman orthogonal score function $\orthoM$ in~\cref{eq:neyman_score}. The key is to optimise $\widehat{\theta}$ with data from the $k$-th fold $\mathcal{D}_{I_k}$ using nuisance parameters $\widehat{s}_k$, $\widehat{g}_k$ that are trained with $\dataset_{I^c_k}$, the complement of $\mathcal{D}_{I_k}$. This is important to fully debias the estimator $\widehat{\theta}$. The DML estimator $f_{\widehat{\theta}}$ is then defined as
\begin{align}
    f_{\widehat{\theta}}\coloneqq \min_{f_\theta\in \mathcal{F}}\frac{1}{K}\sum_{k=1}^K \widehat{\expectE}_{k}[(\widehat{s}_k(c)-\widehat{g}_k(f_{\widehat{\theta}},c)^2],\label{eq:dml_estimator}
\end{align}
where $\widehat{\expectE}_k$ is the empirical expectation over $\mathcal{D}_{I_k}$. In practice, we alternate between the $K$ folds while sampling a mini-batch $c_i^k$ of size $n_b$ from each fold $\mathcal{D}_{I_k}$ to update $\widehat{\theta}$ by minimising the empirical loss on the mini-batch following our Neyman orthogonal score $\orthoM$,
\begin{equation*}
\mathcal{L}=\widehat{\expectE}_{c_i^k} \left[(\widehat{s}_k(c)-\widehat{g}_k(f_{\widehat{\theta}},c))^2\right] = \sum_{c_i^k}\frac{1}{n_b}\left((\widehat{s}_k(c)-\widehat{g}_k(f_{\widehat{\theta}},c))^2\right ).
\end{equation*}
When the second stage converges, we return the DML-CMR estimator $f_{\widehat{\theta}}$.

\bill{The two hyperparameters for this algorithm are the number of folds $K$ and the mini-batch size $n_b$. In theory, the convergence guarantee derived below holds for any choice of $K\in[2...N]$, where $K=N$ would be leave-one-out cross-fitting. However, there exists a trade-off between estimation error and computational complexity. While a larger $K$ allows more data to train the nuisance parameters $(\widehat{s}_k,\widehat{g}_k)$ for each fold, which implies a lower estimation error, the nuisance parameters must be trained $K$ times, increasing the computation cost. Empirically, good performance can often be observed from $K=5$ to $K=10$. The mini-batch size $n_b$ is a standard hyperparameter in deep learning that can be determined based on the training dataset, the neural network architecture, and the computation hardware.}

\subsection{Theoretical Analysis}\label{sec:theory}

In this section, we provide a theoretical analysis on the convergence of DML-CMR. The key benefit of DML is its debiasing effect for two-stage regressions, and crucially, it is possible to leverage the DML framework~\citep{Chernozhukov2018Double/debiased} to show a fast asymptotic convergence rate of $O(N^{-1/2})$, i.e., the DML estimator $\widehat{\theta}$ converges to the true parameters $\theta_0$ at the rate of $O(N^{-1/2})$ with high probability. To provide a road map of this section, we first list all the technical conditions required for a general DML estimator to converge at the fast rate of $O(N^{-1/2})$ following~\citet{Chernozhukov2018Double/debiased}. Later, in~\cref{thm:dml}, we will prove that all these conditions hold for our DML-CMR estimator. The full proofs in this section are deferred to \cref{appen:dml}.

\begin{condition}[Technical conditions of DML $N^{-1/2}$ rate proved later in~\cref{thm:dml}]\label{condition:dml}
For sample size $N\geq3$: 
\begin{enumerate}[label=(\alph*)]
\item The map $(\theta,(s,g))\mapsto \expectE[\orthoM(\dataset;f_{\theta},(s,g))]$ is twice continuously Gateaux-differentiable.
\item  The score $\orthoM$ obeys the Neyman orthogonality conditions in \cref{eq:neyman}.
\item The true parameter $\theta_0$ obeys $\expectE[\orthoM(\dataset;f_{\theta_0},(s_0,g_0))]=0$ and $\Theta$ contains a ball of radius $c_1 N^{-1/2}\log N$ centered at $\theta_0$.
\item For all $\theta\in\Theta$, the identification relationship
\begin{align*}
2\norm{\expectE[\orthoM(\dataset;f_{\theta},(s_0,g_0))]}\gtrsim \norm{J_0(\theta-\theta_0)}
\end{align*}
is satisfied, where $J_0\coloneqq\partial_{\theta^\prime}\{\expectE[\orthoM(\dataset;f_{\theta^\prime},(s_0,g_0))]\}|_{\theta^\prime=\theta_0}$ is the Jacobian matrix, with singular values bounded between $c_0>0$ and $c_1>0$.
\item Let $K$ be a fixed integer. Given a random partition $\{I_k\}_{k=1}^K$ of indices $[N]$, each of size $n=N/K$, the nuisance parameter estimator $\widehat{s}_k$ and $\widehat{g}_k$ learnt using data with indices $I^c_k$ belongs to shrinking realisation sets $\mathcal{S}_N$ and $\mathcal{G}_N$, respectively, and the nuisance parameters should be estimated at the $o(N^{-1/4})$ rate, e.g., $\norm{\widehat{s}-s_{0}}_2=o(N^{-1/4})$.
\end{enumerate}
\end{condition}

Among these conditions, (a), (b), and (c) are conditions regarding the Neyman orthogonal score $\orthoM$. The Neyman orthogonality in (b) is shown in~\cref{thm:neyman} and the other conditions (b) and (c) are mild regularity conditions, standard for moment problems. (d) is an identification condition that ensures sufficient identifiability of $\theta_0$. This condition also implies bounded ill-posedness, which we will discuss in detail in~\cref{sec:ill-posedness}. Finally, (e) is a key condition that states the nuisance parameters should converge to their true values at the crude rate of $o(N^{-1/4})$, where a shrinking realisation set $\mathcal{S}_N$ is a decreasing set of possible estimators $\widehat{s}$ as the sample size $N$ increases.

\bill{In~\citet{Chernozhukov2018Double/debiased}, an additional condition that all eigenvalues of the matrix $\expectE[\orthoM(\dataset;f_{\theta_0},(s_0,g_0))\orthoM(\dataset;f_{\theta_0},(s_0,g_0))^T]$ are strictly positive (bounded away from zero) is required, which means that the score function has nondegenerate variance at true parameters. This additional condition allows them to show that $\widehat{\theta}$ is asymptotically normal around $\theta_0$, in addition to the $N^{-1/2}$ convergence rate. However, since our score function's variance at true parameters is 0 almost surely, we omit this condition and instead provide $N^{-1/2}$ convergence rate guarantee with high probability.}

In \cref{lemma:nuisances}, following recent works~\citep{Chernozhukov2018Double/debiased,Chernozhukov2021AutomaticRegression,Chernozhukov2022RieszNetForests}, we show that the convergence condition for the nuisance parameters in \cref{condition:dml} (e) can be transformed into a condition on the critical radius~\citep{bartlett2005local} of the realisation sets, which is a quantity that describes the complexity of estimation and has been widely studied for various function classes. We start by assuming the realisability of the true functions $g_0, s_0$ and $f_0$ in their corresponding function classes, and further assuming that they are bounded. We formalise these assumptions in \cref{assump:dml}.

\begin{assumption}
[Realisable and bounded function classes]\label{assump:dml}
We assume that $g_0, s_0, f_0$ are realisable in the function classes $\mathcal{G},\mathcal{S},\mathcal{F}$, that is, $g_0, s_0, f_0\in\mathcal{G},\mathcal{S},\mathcal{F}$, respectively, and furthermore, $\norm{f}_\infty, \norm{s}_\infty  \leq B$ for all $f, s \in \mathcal{F}, \mathcal{S}$, where $B$ is a positive constant. Moreover, we assume that the random variable $|Y| \leq B$ \textit{almost surely}.
\end{assumption}

 As discussed in \cref{sec:neyman}, DML-CMR has two nuisance parameters that are required to be estimated: $\widehat{s} \in \mathcal{S}$ and $\widehat{g} \in \mathcal{G}$. As we saw in \cref{sec:dmliv}, the estimation of $\widehat{s}$ is made through standard supervised learning algorithms that we can directly analyse. However, the estimation of $\widehat{g}$ has two steps: 
 (i) we estimate the conditional distribution $\widehat{F}(X|C) \in \mathcal{P}$, where the density sieve $\mathcal{P}$ is defined as
\[
\mathcal{P} \subset \mleft\{ F : \int F(x | C = c) dx = 1 \quad \forall c \in \mathcal{C}\mright\};
\]
and (ii) we plug in the functional $f_\theta$ into the conditional expectation estimator,
\[
\widehat{g}_{\widehat{F}}(f_\theta, c) \coloneqq \int f_\theta(x) \widehat{F}(x | C = c) dx,
\]
for all candidate test functions $f_\theta \in \mathcal{F} = \mleft\{ f_\theta \;:\; \theta \in \Theta \mright\}.$
% \[
% f_\theta \in \mathcal{F} = \mleft\{ f_\theta \;:\; \theta \in \Theta \mright\}.
% \]
From the realisability assumption of $g_0 \in \mathcal{G}$ in \cref{assump:dml}, it follows that $F_0(X|C) \in \mathcal{P}$, and the hypothesis space for the estimand $\widehat{g}$ and the true parameter $g_0$ is defined as $\mathcal{G} \coloneqq \mleft\{ g_{F} \;:\; F \in \mathcal{P}
 \mright\}$.

\begin{lemma}[Convergence of nuisance parameters]\label{lemma:nuisances}
     Under \cref{assump:dml}, let $\mathcal{S}^*_N$ be the star-hull of the realisation set $\mathcal{S}_N$ of function class $\mathcal{S}$ centred at $s_0$, defined as
    \[
    \mathcal{S}^*_N = \mleft\{ C \mapsto \gamma(s(C) - s_0(C)) \;:\; s \in \mathcal{S}_N, \gamma \in [0, 1] \mright\},
    \]
    $\mathcal{P}^*_N$ be the star-hull of the realisation set $\mathcal{P}_N$ of the function class $\mathcal{P}$ centred at $F_0$,
    \[
    \mathcal{P}^*_N = \mleft\{ C \mapsto \gamma(F(\cdot|C) - F_0(\cdot|C)) \;:\; F \in \mathcal{P}_N, \gamma \in [0, 1] \mright\},
    \]
    and $\mathcal{G}^*_N$ be the star-hull of the realisation set $\mathcal{G}_N$ of the function class $\mathcal{G}$ centred at $g_0$,
    \[
    \mathcal{G}^*_N = \mleft\{C, f \mapsto\gamma(g(C,f)-g_0(C,f))\;:\; g\in\mathcal{G}_N,  \gamma\in[0,1]\mright\},
    \]
     where $\mathcal{S}_N$, $\mathcal{P}_N$ and $\mathcal{G}_N$ are properly shrinking neighbourhoods of the true functions $s_0$, $F_0$ and  $g_0$. Then, there exist universal constants $c_1$ and $c_2$, for which we have that, with probability at least $1 - \xi$, the estimation errors are bounded as
    \[
    \norm{\widehat{s}-s_0}_2^2 & \leq c_1 \mleft( \delta_N(\mathcal{S}_N^*)^2+\sqrt{\frac{\log(1/\zeta)}{N} }
    \mright) ; \\
     \norm{\widehat{g} - g_0}_2^2 & \leq c_2 \mleft( \delta_N(\mathcal{P}^*_N)^2+\sqrt{\frac{\log(1/\zeta)}{N}} \mright),
    \]
    where $\delta_N(\mathcal{S}_N^*)$ denotes the critical radius of $\mathcal{S}_N^*$. The critical radius is defined as the minimum $\delta$ that satisfies $\mathcal{G(\mathcal{S}_N^*, \delta)} \leq {\delta^2}/2$, where
    \begin{align*}
    \mathcal{G(\mathcal{S}_N^*, \delta)} \coloneqq \expectE_{\epsilon}\left[\sup_{h \in \mathcal{H}^*: \norm{h}_N \leq \delta}  \langle \epsilon, h \rangle \right]
    \end{align*}
\end{lemma}

% \begin{definition}
%     The critical radius denoted by $\delta_N(\mathcal{H}^*)$ is defined as the minimum $\delta$ that satisfies the following upper bound on the local Gaussian complexity of a star-shaped function class $\mathcal{H}^*$
    
%     $\mathcal{G(\mathcal{H}^*, \delta)} \leq {\delta^2}/2$, where local Gaussian complexity is defined as
%     \begin{align*}
%     \mathcal{G(\mathcal{H}^*, \delta)} = \expectE_{\epsilon}\left[\sup_{h \in \mathcal{H}^*: \norm{h}_N \leq \delta}  \langle \epsilon, h \rangle \right], \numberthis{eq;fixed_point}
%     \end{align*}
%     with $\epsilon$ being a random i.i.d. zero-mean Gaussian vector.
% \end{definition}

This lemma shows that, if we can upper bound the critical radii $\delta_N(\mathcal{S^*})$ and $\delta_N(\mathcal{P^*})$ by $o(N^{-1/4})$, then $\norm{\widehat{s}-s_0}_2= o(N^{-1/4})$ and $\norm{\widehat{g}-g_0}_2 = o(N^{-1/4})$, meaning that nuisance parameters converge to their true values at the rate of $o(N^{-1/4})$ as required by \cref{condition:dml} (e).
% Next, we provide an analysis and concrete examples of estimators that satisfy this requirement on the critical radius.

For the critical radius, it is typically shown that $\delta_N=O(d_N^{1/2} N^{-1/2})$~\citep{Chernozhukov2022RieszNetForests,Chernozhukov2021AutomaticRegression}, where $d_N$ is the effective dimension of the hypothesis space, which is another notion to bound the estimation error in the regression
problem (for further details on the effective dimension, see~\cref{appen:critical_radius}.). This, together with \cref{lemma:nuisances}, implies that $\norm{\widehat{s}-s_0}_2=O(d_N(\mathcal{S}^*)^{1/2} N^{-1/2})$. Therefore, we can also see that, if the effective dimension satisfies $d_N(\mathcal{S}^*)=o(N^{1/4})$, then $\norm{\widehat{s}-s_0}_2 = o(N^{-1/4})$ as required by \cref{condition:dml} (e) (and similarly for $\widehat{g}$ and $d_N(\mathcal{P}^*)$).

Therefore, we can refer to results in the literature that analyse the effective dimension and critical radius of various estimators to provide examples of estimators that satisfy \cref{condition:dml} (e). For the estimation of $\widehat{s}$, we have a regression problem and \cref{condition:dml} (e) is satisfied by many supervised learning estimators such as parametric generalised linear models~\citep{van1996weak}, Lasso~\citep{bickel2009simultaneous}, random forests~\citep{syrgkanis2020estimation}, boosting~\citep{luo2016high}, Sobolev kernel regression with $\alpha$-smooth RKHS ($\alpha > d /2$, where $d$ is the dimension of $X$)~\citep{caponnetto2007optimal,christmann2008support} and neural networks~\citep{chen1999improved,yarotsky2018optimal,schmidt2020nonparametric,farrell2021deep}. For the conditional density estimator $\widehat{g}$, the above estimators also satisfy \cref{condition:dml} (e) if the conditional distribution can be parameterised accordingly. Otherwise, this condition is also satisfied by Gaussian mixtures density networks~\citep{ho2022convergence}, polynomial sieve with Hölder smoothness $\alpha > \frac{d + 1}{2}$~\citep{Ai2003}, and categorical-logistic models~\citep{zhao2022asymptotic}, among others.

Leveraging \cref{lemma:nuisances}, we derive the following theorem on the convergence of the DML-CMR estimator.

\begin{theorem}[Convergence of DML-CMR]\label{thm:dml}
Let $f_{\theta_0}\in\mathcal{F}$ be a solution that satisfies the CMRs in~\cref{eq:cmr}, let $\orthoM$ be the Neyman orthogonal score defined in~\cref{eq:neyman_score} and let $J_0\coloneqq\partial_{\theta^\prime}\{\expectE[\orthoM(\dataset;f_{\theta^\prime},(s_0,g_0))]\}|_{\theta^\prime=\theta_0}$ be the Jacobian matrix of $\expectE[\orthoM]$ w.r.t. $\theta$. Suppose that the upper bound of the critical radius $\delta_N=o(N^{-1/4})$ for $\widehat{s}$, $\widehat{g}$, and $J_0$ has bounded singular values. Then, if \cref{assump:parameter} and \ref{assump:dml} hold, our DML estimator $f_{\widehat{\theta}}$ satisfies that $\widehat{\theta}$ is concentrated in a $N^{-1/2}$ neighbourhood of $\theta_0$ with probability $1-o(1)$. Specifically,
for all $\zeta\in(0,1]$, there exist $K>0$ such that for all integer $N\geq K$, 
\begin{align*}
    \norm{\widehat{\theta}-\theta_0}=O(N^{-1/2})
\end{align*}
with probability $1-\zeta$.
\end{theorem}
\begin{proof}[Proof sketch]

From ~\cref{prop:thm3.3}, we have that if all technical conditions stated in \cref{condition:dml} are satisfied, the DML estimator converges at the required rate (formally stated and proved in \cref{prop:thm3.3}). Therefore, we need to verify that, under \cref{assump:dml}, all components of \cref{condition:dml} are satisfied. Condition (a) is satisfied since $(s-g)^2$ is twice continuously differentiable with respect to $s$ and $g$. Condition (b) and (c) are satisfied by \cref{thm:neyman} and \cref{assump:parameter}, since the true parameter $\theta_0\in\Theta$ is in the interior of $\Theta$. 

Condition (d) and (e) are technical conditions that ensure sufficient identifiability and non-degeneracy. Condition (d) can be proved to hold using the first-order Taylor series for the score function $\expectE[\orthoM(\dataset;f_{\theta},(s_0,g_0))]$ around the point $\theta_0$ and the fact that $J_0 J_0^T$ is non-singular since the singular value of $J_0$ is bounded, as stated in the Theorem. Finally, condition (e) is satisfied by \cref{lemma:nuisances} and the fact that the critical radius $\delta_N=o(N^{-1/4})$. Therefore, all components of \cref{condition:dml} are satisfied, and the proof follows by \cref{prop:thm3.3}.
\end{proof}

\cref{thm:dml} states that, with adequately trained nuisance parameter estimators in terms of their critical radius and identifiability conditions in terms of the non-singularity of the Jacobian matrix $J_0$, the estimator error $\widehat{\theta}-\theta_0$ is normally distributed, where its variance shrinks at the rate of $N^{-1/2}$. This implies that $\widehat{\theta}$ converges to $\theta_0$ at the rate $O(N^{-1/2})$ with high probability, which allows us to bound the estimation error $\norm{f_{\widehat{\theta}}-f_{\theta_0}}_2$ of the DML-CMR estimator with high probability, under a Lipschitz condition of $f_\theta$.

\begin{corollary}[Estimation error bounds]\label{coro:function_convergence}
Let $f_{\widehat{\theta}}$ be the DML estimator for CMRs. If all assumptions for~\cref{thm:dml} hold and there exists a constant $L>0$ such that $\norm{f_\theta(x)-f_{\theta_0}(x)}_2\leq L \norm{\theta-\theta_0}_2$ for all $x\in\mathcal{X}$ and $\theta\in\Theta$,
then for all $\zeta\in(0,1]$, we have that for all $\zeta\in(0,1]$, there exist $K>0$ such that for all integer $N\geq K$,
\begin{align*}
\norm{f_{\widehat{\theta}}-f_{\theta_0}}_2=O\left(L\cdot N^{-1/2}\right),
\end{align*}
with probability $1-\zeta$.
\end{corollary}

\begin{proof}
From \cref{thm:dml}, we have that the parameters $\widehat{\theta}$ for our DML estimator $f_{\widehat{\theta}}$ learnt from a dataset of size $N$ satisfy that for all $\zeta\in(0,1]$, there exist $M,K>0$ such that for all integer $N\geq K$, 
\begin{align*}
    \probP\left(\norm{\widehat{\theta}-\theta_0}\leq M\cdot N^{-1/2}\right)\geq 1-\zeta.
\end{align*}

If we let $L$ to be a constant such that $\norm{f_\theta(x)-f_{\theta_0}(x)}\leq L \norm{\theta-\theta_0}$ for all $x\in \mathcal{X}$ and $\theta\in\Theta$, we have that, for all $\zeta\in(0,1]$, there exist $M,K>0$ such that for all integer $N\geq K$,
\begin{align*}
\probP\left(\norm{f_{\widehat{\theta}}(x)-f_{\theta_0}(x)}\leq L\cdot M \cdot N^{-1/2}\right)\geq 1-\zeta,
\end{align*}
which implies $\norm{f_{\widehat{\theta}}-f_{\theta_0}}=O\left(L\cdot N^{-1/2}\right)$ with probability 1-$\zeta$ as required.
\end{proof}

Here, we assume a local Lipschitz condition of $f_\theta$ w.r.t. $\theta$ around $\theta_0$: $\norm{f_\theta(x)-f_{\theta_0}(x)}_2\leq L\norm{\theta-\theta_0}_2$ for all $x\in\mathcal{X}$ and $\theta\in\Theta$ for some $L>0$. Since $\Theta$ is compact, we see that it is enough for this Lipschitz condition to hold locally in a neighbourhood of $\theta_0$.

\subsection{A Computationally Efficient Variant of DML-CMR}\label{appen:cedml}

In this section, we propose a computationally efficient variant of DML-CMR. The standard DML-CMR with $K$-fold cross-fitting (\cref{alg:dml-iv-kf}) trains the nuisance parameters $\widehat{s}$ and $\widehat{g}$ $K$ times on different subsets of the dataset to mitigate overfitting bias. While this approach improves statistical robustness and stability, it is computationally expensive. To address this, we propose CE-DML-CMR, a variant of DML-CMR described in \cref{alg:ce-dml} that does not use $K$-fold cross-fitting. It trains $\widehat{s}$ and $\widehat{g}$ only once (instead of $K$ times) using the entire dataset, significantly reducing computational cost. CE-DML-CMR uses the same Neyman orthogonal score as the standard DML-CMR, so it still enjoys the partial debiasing effect~\citep{Mackey2018} from the Neyman orthogonal score. However, without $K$-fold cross-fitting, it lacks the theoretical convergence rate guarantees provided by~\cref{thm:dml}. Nonetheless, CE-DML-CMR can be viewed as a trade-off between computational complexity and theoretical guarantees, and it also serves as an ablation study on $K$-fold cross-fitting.

\begin{algorithm}[t]
   \caption{Computationally Efficient CE-DML-CMR}
   \label{alg:ce-dml}
\begin{algorithmic}
   \STATE {\bfseries Input:} Dataset $\dataset$ with size $N$, mini-batch size $n_b$
   \STATE Learn $\widehat{s}$ and $\widehat{g}$ using $\dataset$
   \STATE Initialise $f_{\widehat{\theta}}$
   \REPEAT
   \STATE Sample $n_b$ data $c_i$ from $\dataset$
   \STATE $\mathcal{L}=\widehat{\expectE}_{c_i}\left[(\widehat{s}(c)-\widehat{g}(f_\theta,c))^2\right]$
   \STATE Update $\widehat{\theta}$ to minimise loss $\mathcal{L}$ 
    \UNTIL{convergence}
    \STATE {\bfseries Output:} The CE-DML-CMR estimator $f_{\widehat{\theta}}$
\end{algorithmic}
\end{algorithm}

\subsection{DML Identifiability Condition and Ill-posedness}\label{sec:ill-posedness}

In this section, we provide a discussion regarding the relationship between our identifiability condition, \cref{condition:dml} (d), and a more common notion of identifiability for CMR problems, the \textit{ill-posedness}~\citep{Chen2012}. To begin with, we define the ill-posedness of CMR problems.

\begin{definition}[Ill-posedness~\citep{Chen2012,Dikkala2020}]\label{def:ill-posed}
Given a CMR problem as in~\cref{eq:cmr}, the \emph{ill-posedness} $\ill$ of the function space $\mathcal{F}$ is given by
\begin{align*}
    \ill=\sup_{f\in\mathcal{F}} \frac{\norm{f_0-f}_{2}}{\norm{\expectE[f_0(X)-f(X) \lvert C]}_{2}}.
\end{align*}
\end{definition}

 Intuitively, ill-posedness describes how well a small CMR error (the projected error under conditional expectation) implies a small $L_2$ error (root mean squared error) for $f_0$. For identifiability, it is usually assumed that the ill-posedness $\ill$ is bounded. Otherwise, even solving the CMRs with very small error does not guarantee a solution $\widehat{f}$ that is close to $f_0$. In our case, we demonstrate that the identification condition of a DML estimator actually implies bounded ill-posedness. Specifically, \cref{condition:dml} (d) 
% together with a local Lipschitz condition for $f_\theta$ around $\theta_0$, as required by \cref{coro:function_convergence},
implies that the ill-posedness is bounded, as shown by the following proposition.

\begin{proposition}[Identifiability and Ill-posedness]\label{prop:ill-posed}
For all $\theta\in\Theta$, if there exists a constant $L>0$ such that $\norm{f_\theta(x)-f_{\theta_0}(x)}_2\leq L \norm{\theta-\theta_0}_2$ for all $x\in\mathcal{X}$ and $\theta\in\Theta$, then the identifiability condition in \cref{condition:dml} (d), which states 
\begin{align*}
2\norm{\expectE[\orthoM(\dataset;f_{\theta},(s_0,g_0))]}\geq \norm{J_0(\theta-\theta_0)}
\end{align*}
and the Jacobian matrix $J_0$ have singular values bounded between $c_0>0$ and $c_1>0$, implies that the ill-posedness is bounded by $\ill\leq L/\sqrt{c_0}$.
\end{proposition}

The proof of \cref{prop:ill-posed} is deferred to~\cref{appen:illposed}. This interesting finding explains why, in~\cref{thm:dml}, there are no explicit assumptions about the ill-posedness of the problem. The only identifiability assumption for~\cref{thm:dml} is the non-singularity of the Jacobian matrix $J_0$, which for DML-CMR is sufficient to ensure \cref{condition:dml} (d) holds. Therefore, by \cref{prop:ill-posed}, this ensures a bounded ill-posedness of the problem, which in turn allows us to identify the solution $f_0$ with a small error.

\section{Suboptimality for Offline IV Bandit}\label{sec:ivbandit_subopt}

In this section, we apply DML-CMR (\cref{alg:dml-iv-kf}) to solve the offline IV bandit problem and provide an upper bound on the suboptimality of the learnt policy. Recall from~\cref{sec:contextualIV} that the CMR problem we solve for the offline IV bandit problem is $\expectE[R-f_0(C,A)\lvert Z,C]=0$. Using DML-CMR, we can provide an estimator $f_{\widehat{\theta}}(C,A)$ for $f_0$, and we retrieve (an estimate of) the induced optimal policy as $\widehat{\policy}(c)\coloneqq\argmax_a h_{\widehat{\theta}}(c,a)$. The suboptimality of a policy is $\textrm{subopt}(\widehat{\policy})\coloneqq\Value(\policy^*)-\Value(\widehat{\policy})$. Next, we show a suboptimality bound for the DML-CMR policy in terms of the sample size $N$ based on \cref{thm:dml}.

\begin{corollary}[Suboptimality Bounds]\label{coro:subopt}

Let the learnt policy from a dataset of size $N$ be $\widehat{\pi}(c)\coloneqq\argmax_a f_{\widehat{\theta}}(c,a)$, where $\widehat{\theta}$ is the DML-CMR estimator. Let $L$ be a constant such that $\norm{f_\theta(C,A)-f_{\theta_0}(C,A)}_2\leq L \norm{\theta-\theta_0}_2$ for all $C$ in the support of $\testD$, $A\in\mathcal{A}$, and $\theta\in\Theta$. Then, for all $\zeta\in(0,1]$, we have that there exist $K>0$ such that for all integer $N\geq K$, the suboptimality of $\widehat{\pi}$ satisfies
\begin{align*}
\textrm{subopt}(\widehat{\policy}_N)=O(L\cdot N^{-1/2})
\end{align*}
with probability $1-\zeta$.
\end{corollary}
%A more detailed version of this Theorem and 
The proof is deferred to~\cref{appen:subopt}. To the best of our knowledge, this is the first time that the suboptimality bounds of $O(N^{-1/2})$ have been proved for confounded IV bandit, matching the suboptimality bounds of the unconfounded bandit.

\section{Experimental Results}\label{sec:exp}

In this section, we empirically evaluate DML-CMR for IV regression, offline IV bandit, and PCL tasks. We also evaluate the computationally efficient CE-DML-CMR, finding CE-DML-CMR empirically performs as well as standard DML-CMR on low-dimensional datasets, where overfitting bias is not prevalent.

\paragraph{Experimental Setup}

The experiments are carried out on a Linux server (Ubuntu 18.04.2) with two Intel Xeon Gold 6252 CPUs and six NVIDIA GeForce RTX 2080 Ti GPUs. We use DNN estimators for both stages of DML-CMR (\cref{alg:dml-iv-kf}), with network architectures and hyperparameters detailed in~\cref{appen:networks}. Our evaluation considers both low- and high-dimensional synthetic datasets, as well as semi-synthetic real-world datasets. The synthetic datasets are based on an airline ticket demand scenario introduced in~\cref{chapter:intro}. The semi-synthetic datasets are from the healthcare domain, studying the effect of treatments on infants and the impact of air pollution on human health. Details of the datasets are provided in the following sections and in \cref{appen:DML-IV-dataset}. We run each method 20 times and plot the median, 25th, and 75th percentiles of the results. All algorithms are implemented using PyTorch~\citep{Paszke2019}, and the code is available on GitHub\footnote{\url{https://github.com/shaodaqian/DML-CMR}}.

\subsection{IV Regression}\label{exp:iv_reg}

For IV regression, we compare our methods against leading modern approaches: Deep IV~\citep{Hartford2017DeepPrediction}, DeepGMM~\citep{Bennett2019DeepAnalysis}, KIV~\citep{Singh2019} and DFIV~\citep{Xu2020}. As reviewed in~\cref{lit:iv}, Deep IV, DeepGMM, and DFIV are DNN-based estimators, whereas KIV is a kernel-based estimator that leverages the GMM framework. Therefore, this selection of methods provides a representative and diverse benchmark, covering both neural and kernel-based methods, and reflects the current best-performing algorithms commonly used as baselines in the IV regression literature.

Results of DML-CMR using tree-based estimators such as Random Forests and Gradient Boosting are provided in~\cref{appen:tree-based}, where comparable performance to DNN-based DML-CMR is demonstrated. In addition, we provide a sensitivity analysis against hyperparameter changes in~\cref{appen:sensitivity} and an evaluation of algorithms when the IV is weakly correlated with the treatment, representing  higher ill-posedness of the CMR problem, in~\cref{appen:weak_iv}. \bill{The training time per run for DML-CMR with $K=10$ is less than one hour, whereas CE-DML-CMR takes less than 10 minutes. The runtimes of other algorithms range from minutes to tens of minutes.}

\subsubsection{Ticket Demand Dataset}

\begin{figure*}[t]
\centering
% \begin{subfigure}[t]
% {1\textwidth}
% \centering
% \includegraphics[width=0.6\textwidth]{figures/dmliv/legend_flat.png}
% \end{subfigure}
\begin{subfigure}[c]
{0.49\textwidth}
\centering
\includegraphics[width=1\textwidth]{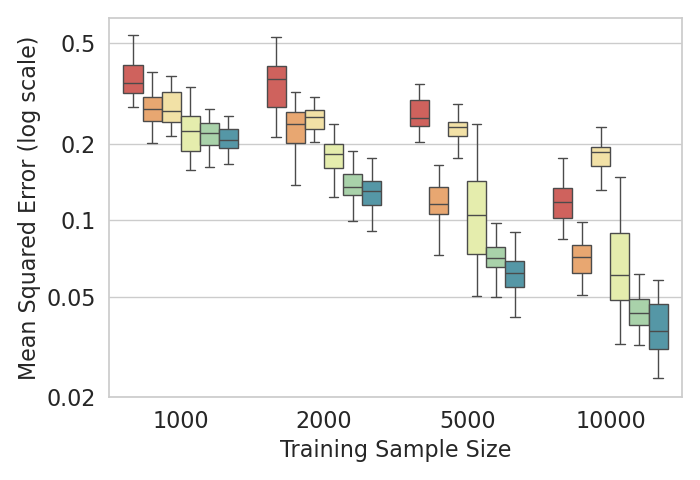}
    \caption{The mean squared error of $\widehat{f}$.}
    \label{fig:lowd_mse}
\end{subfigure}
\begin{subfigure}[c]
{0.49\textwidth}
\centering
\includegraphics[width=0.35\textwidth]{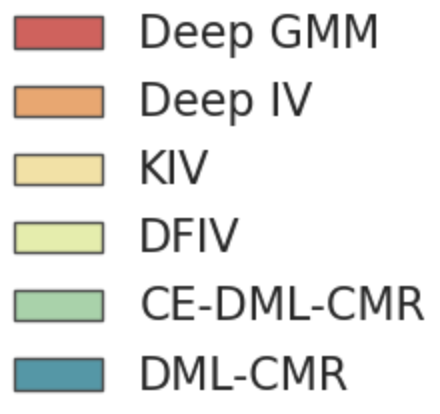}
\end{subfigure}
\begin{subfigure}[c]{0.49\textwidth}
\centering
\captionsetup{width=0.9\linewidth}
\includegraphics[width=1\textwidth]{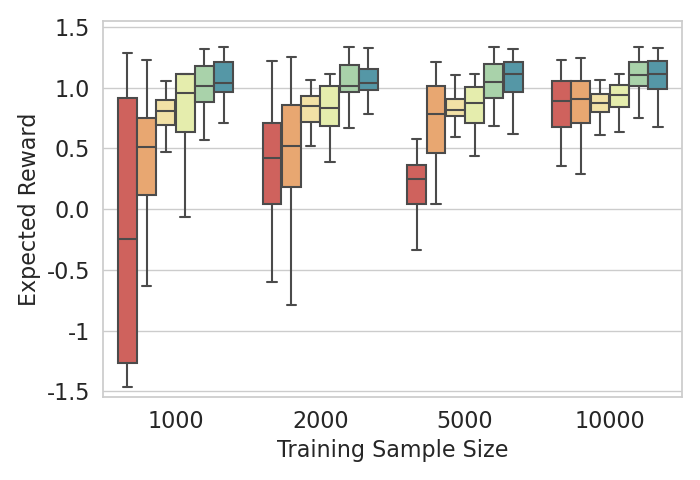}
\caption{The average reward following the policy $\widehat{\policy}$ derived from $\widehat{f}$.}
    \label{fig:lowd_r}
\end{subfigure}
\begin{subfigure}[c]{0.49\textwidth}
\centering
\captionsetup{width=0.9\linewidth}
\includegraphics[width=1\textwidth]{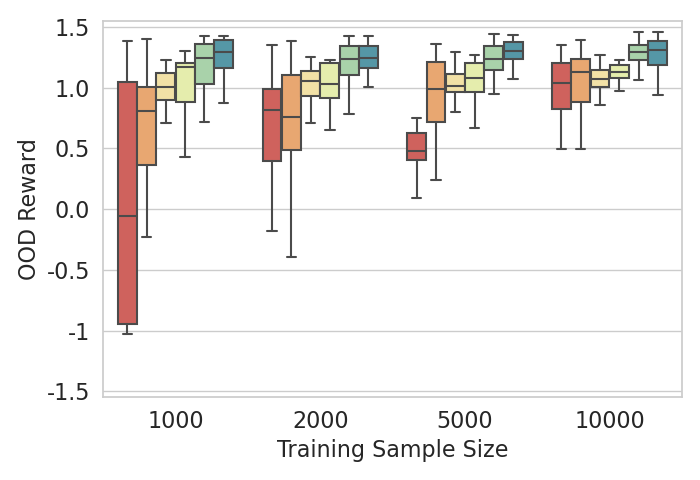}
\caption{The average reward following $\widehat{\policy}$ with out of training distribution context.}\label{fig:lowd_ood_r}
\end{subfigure}
\caption{Results on the ticket demand dataset with low-dimensional context.}
\end{figure*}

We first conduct experiments for IV regression on the ticket demand dataset, which is a synthetic dataset introduced by~\citet{Hartford2017DeepPrediction} that is now a standard benchmark for nonlinear IV methods. In this dataset, the goal is to understand how ticket prices $p$ affect ticket sales $r$. We observe two context variables, which are the time of year $t\in[0,10]$ and customer type $s\in[7]$ variables, the latter categorised by the level of price sensitivity. Price and context affect sales through $f_0((t,s),p)=100+(10+p)\cdot s \cdot \psi(t)-2p$, where $\psi(t)$ is a complex nonlinear function. However, the noise of $r$ and $p$ is correlated, which indicates the existence of unobserved confounders. The fuel price $z$ is introduced as an instrumental variable. Details of this dataset are included in~\cref{appen:demand}.

The mean squared errors (MSE) for learning $f_0$ with this dataset of various sizes are provided in~\cref{fig:lowd_mse}. It can be seen that DML-CMR performs better than other IV regression methods for all dataset sizes. CE-DML-CMR, which requires significantly less computation, matches the performance of DML-CMR in this case.

\subsubsection{High-Dimensional Dataset} 

\begin{figure*}[t]
\centering
\begin{subfigure}[c]
{0.49\textwidth}
\centering
\includegraphics[width=1\textwidth]{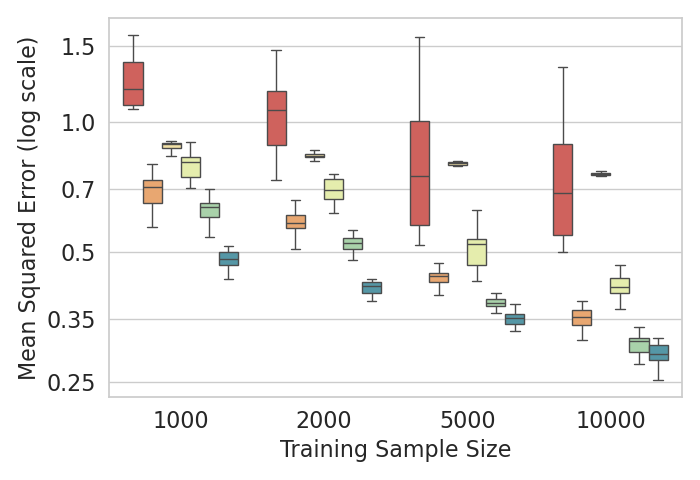}
    \caption{The mean squared error of $\widehat{f}$.}
    \label{fig:mnist_mse}
\end{subfigure}
\begin{subfigure}[c]
{0.49\textwidth}
\centering
\includegraphics[width=0.35\textwidth]{figures/dmliv/legend_iv.png}
\end{subfigure}
\begin{subfigure}[c]{0.49\textwidth}
\centering
\captionsetup{width=0.9\linewidth}
\includegraphics[width=1\textwidth]{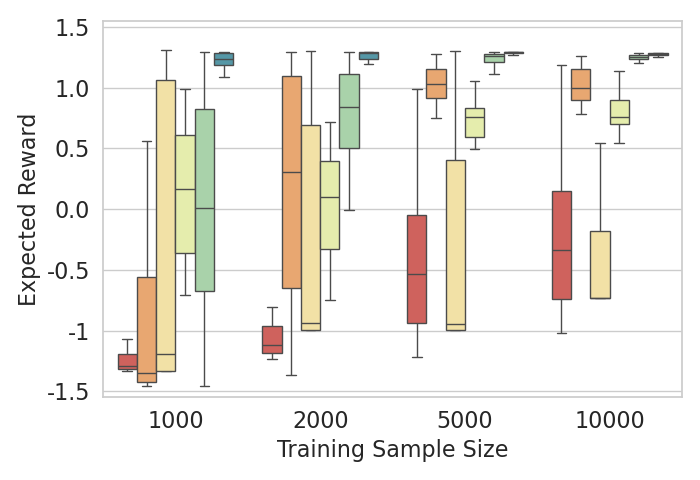}
\caption{The average reward following the policy $\widehat{\policy}$ derived from $\widehat{f}$.}
    \label{fig:mnist_r}
\end{subfigure}
\begin{subfigure}[c]{0.49\textwidth}
\centering
\captionsetup{width=0.9\linewidth}
\includegraphics[width=1\textwidth]{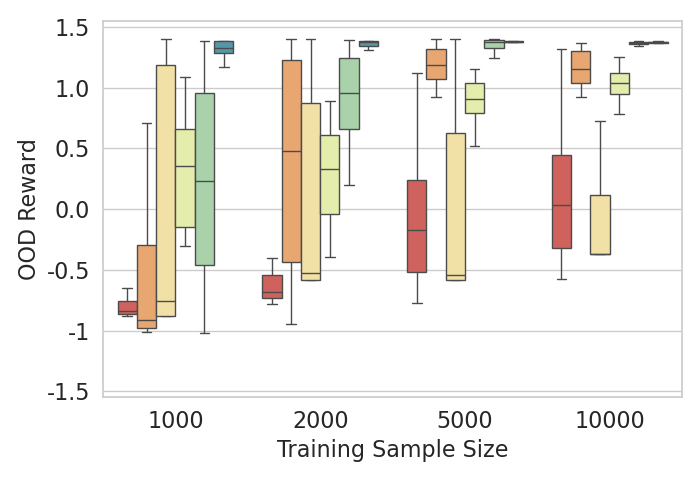}
\caption{The average reward following $\widehat{\policy}$ with out of training distribution context.}\label{fig:mnist_ood_r}
\end{subfigure}
\caption{Results on the ticket demand dataset with high-dimensional context. }
% \vspace{-0.1cm}
\end{figure*}

In real applications, we
typically do not observe variables such as the customer type as explicit categories. Therefore, we follow~\citet{Hartford2017DeepPrediction} and consider the case where the customer type $s\in[7]$ is replaced by images of the corresponding handwritten digits from the MNIST dataset~\citep{LeCun2010} to evaluate our methods with high-dimensional ($28^2$=784 dimensions) inputs. The task remains to learn $f_0$, but the algorithms are no longer explicitly given the 7 customer types, and instead have to infer the relationship between the image data and the outcome. Results for IV regression are plotted in~\cref{fig:mnist_mse}, where DML-CMR and CE-DML-CMR outperform all other methods. In these high-dimensional settings, regularisation is heavily used to avoid overfitting. DML-CMR demonstrates the benefits of using DML to reduce both the regularisation and overfitting bias caused by learning the nuisance parameters.

\subsubsection{Real-World Datsets}

\begin{figure}[t]
\centering
\begin{subfigure}[c]{0.4\textwidth}
% \centering
\includegraphics[width=1\textwidth]{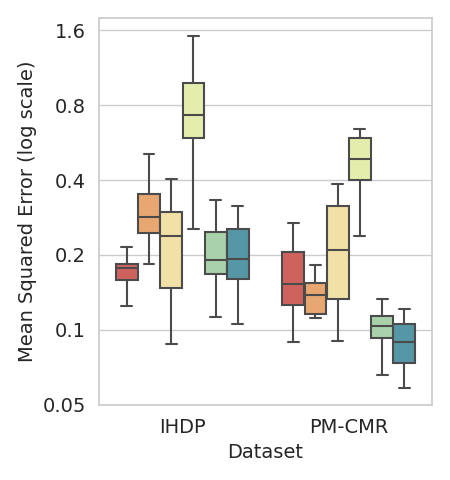}
    % \caption{The mean squared error for estimating $h_0$.}
    % \label{fig:real_mse}
\end{subfigure}
\begin{subfigure}[c]{0.4\textwidth}
% \centering
\includegraphics[width=1\textwidth]{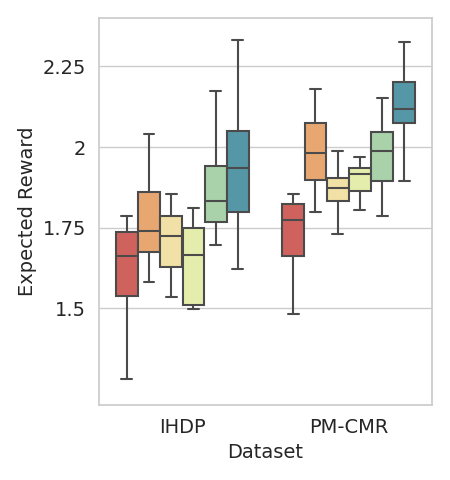}
% \caption{The average reward following the policy $\widehat{\policy}$ derived from $\widehat{f}$.}
\end{subfigure}
\begin{subfigure}[c]
{0.175\textwidth}
\centering
\includegraphics[width=1\textwidth]{figures/dmliv/legend_iv.png}
\end{subfigure}
\caption{The mean squared error of $\widehat{f}$ and average reward following $\widehat{\policy}$ for the real-world datasets.}
\label{fig:real_dataset}
% \vspace{-0.1cm}
\end{figure}

Next, we test the performance of DML-CMR on real-world datasets. The true counterfactual prediction function is rarely available for real-world data. %which makes evaluating different methods very challenging. Similar 
Therefore, in line with previous approaches~\citep{Shalit2017,Wu2023,Schwab2019,Bica2020}, we instead consider two semi-synthetic real-world datasets IHDP\footnote{IHDP: \url{https://www.fredjo.com/}.}~\citep{Hill2011} and PM-CMR\footnote{PM-CMR: \url{https://doi.org/10.23719/1506014}.}~\citep{Wyatt2020}. We directly use the continuous variables from IHDP and PM-CMR as context variables and generate the outcome variable with a nonlinear synthetic function following~\citet{Wu2023}. There are 470 and 1350 training samples in IHDP and PM-CMR, respectively (for details see~\cref{appen:demand}). As shown in~\cref{fig:real_dataset}, DML-CMR and CE-DML-CMR demonstrate comparable, if not lower, MSE of fitting $\widehat{f}$ than the other methods. This shows that our algorithm can reliably learn the counterfactual prediction function from real-world data.

\subsection{Offline IV Bandit}

We also evaluate the ability of DML-CMR to learn effective policies in the offline IV bandit setting, where the goal is to maximise expected reward. This evaluation is conducted using the same datasets and benchmarking methods as in \cref{exp:iv_reg}. From the learnt $\widehat{f}$, for each context sampled from the test distribution, we retrieve the best action by uniformly sampling actions from the action space $\actions$ and selecting the action for which $\widehat{f}$ returns the highest value. Using this induced policy $\widehat{\policy}$, we compare the expected reward following $\widehat{\policy}$ over the test distribution.

For the low-dimensional ticket demand dataset, we first set the test distribution to be the same as the training distribution and plot the average rewards in~\cref{fig:lowd_r}. In~\cref{fig:lowd_ood_r}, we shift the test distribution out of the training distribution by incrementing the distribution of $t$ by $1$. For the high-dimensional setting, \cref{fig:mnist_r} and~\cref{fig:mnist_ood_r} demonstrate the expected rewards for test distributions in and out of the training distribution, respectively. There is a clear trend that a better fitted (low MSE) $\widehat{f}$ leads to an induced policy with higher expected reward. In all cases, DML-CMR outperforms all other methods, especially in the high-dimensional setting, where DML-CMR consistently learns the near-optimal policy with only 2000 samples. CE-DML-CMR, on the other hand, only matches the performance of DML-CMR for the low-dimensional setting, but still outperforms the other methods in the high-dimensional setting. For the real-world dataset, DML-CMR also outperforms all other methods in \cref{fig:real_dataset}.

We only compare with other IV regression methods because there are no offline bandit methods that consider the IV setting, and standard offline bandit algorithms (e.g.,~\citep{Valko2013,Jin2021,Nguyen-Tang2022}) fail to learn meaningful policies when the dataset is confounded, as demonstrated in~\cref{appen:offline_bandit}.

\subsection{Proximal Causal Learning}

For the PCL task, recall that the CMR problem to solve is $\expectE[Y-h(A,W)\lvert A,V]=0$ as shown in~\cref{background:PCL}. We compare our methods with modern PCL methods: CEVAE~\citep{Im2021}, PMMR~\citep{Mastouri2021}, KPV~\citep{Mastouri2021}, DFPV~\citep{Xu2021}, NMMR U~\citep{Kompa2022}, NMMR V~\citep{Kompa2022} and PKDR~\citep{Wu2024}. As reviewed in~\cref{lit:proxy}, these methods also include both DNN-based and kernel-based methods, forming a high-performing and diverse benchmark. As discussed in \cref{background:PCL}, in PCL, we can only estimate the average treatment effect $\expectE[R\lvert do(A)]$ rather than the conditional average treatment effect $\expectE[R\lvert do(A),C]$, as in the case of IV regression. Therefore, a context-dependent policy cannot be derived using PCL. Nonetheless, we can still evaluate DML-CMR for solving the CMRs and estimating the average treatment effect in PCL. \bill{The training time per run for DML-CMR with $K=10$ and CE-DML-CMR is also around one hour and 10 minutes, respectively. The runtimes of other algorithms range from minutes to tens of minutes.}

\subsubsection{Ticket Demand Dataset}

\begin{figure*}[t]
\centering
\begin{subfigure}[c]
{0.6\textwidth}
\centering
\includegraphics[width=0.9\textwidth]{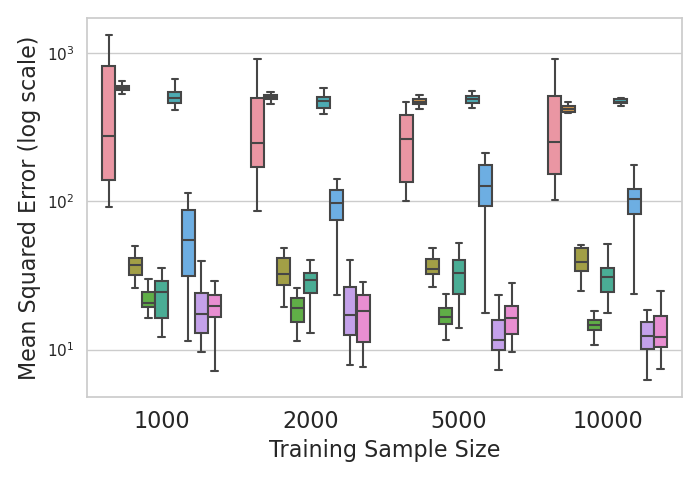}
\end{subfigure}
\begin{subfigure}[c]
{0.3\textwidth}
\centering
\includegraphics[width=0.6\textwidth]{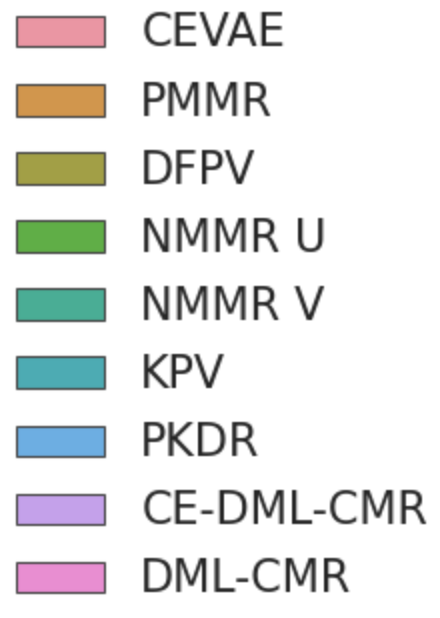}
\end{subfigure}
\caption{The mean squared error of $\widehat{f}$ on the ticket demand dataset for the PCL task.}
\label{fig:pcl_demand}
\end{figure*}

Similarly to IV regression, we start with the ticket demand dataset~\citep{Hartford2017DeepPrediction} which has been adopted to the PCL setting~\citep{Xu2021}, where the goal is to understand how ticket prices affect ticket sales and learn the causal function $f_0$. The hidden confounder in this case is the varying demand $U$, while the cost of fuel $V$ is the treatment proxy, which directly impacts the ticket price, and the number of views on the airline's reservation website $W$ is the outcome proxy. Details of this dataset are included in~\cref{dataset:pcl}.

The results for learning $f_0$ with this dataset of various sizes are provided in~\cref{fig:pcl_demand}. It can be seen that DML-CMR and CE-DML-CMR achieved state-of-the-art (SOTA) performance with very similar performance to each other. NMMR U can match DML-CMR at smaller dataset sizes, but DML-CMR achieves lower MSE at 5000 and 7500 sample sizes. The performance gap between CE-DML-CMR and DML-CMR is small in this case, as expected, because the variables are low-dimensional. 

\subsubsection{High-dimensional dSprites Dataset}

\begin{figure*}[t]
\centering
\begin{subfigure}[c]
{0.6\textwidth}
\centering
\includegraphics[width=0.95\textwidth]{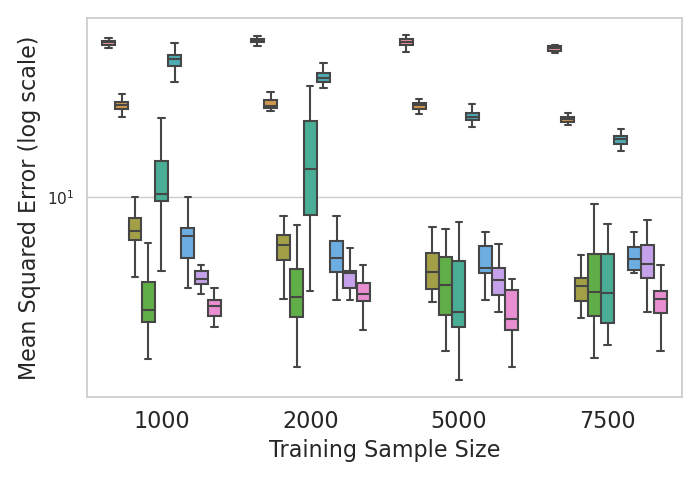}
\end{subfigure}
\begin{subfigure}[c]
{0.3\textwidth}
\centering
\includegraphics[width=0.6\textwidth]{figures/dmliv/pcl/legend_pcl.png}
\end{subfigure}
\caption{The mean squared error of $\widehat{f}$ on the dSprites dataset with high dimensional treatment for the PCL task.}
\label{fig:pcl_dsprite}
\end{figure*}

For a high-dimensional dataset, we adopt the dSprites dataset~\citep{dsprites17} for PCL, first introduced by~\citet{Xu2021}. dSprites is an image (64 $\times$ 64) dataset where each image is described by five parameters: \textit{shape}, \textit{scale}, \textit{rotation}, \textit{posX} and \textit{poxY}. The treatments are these high-dimensional dSprites images, the hidden confounder is posY, the proxies are noisy observations of scale, rotation, and posX, and the outcome is defined by a nonlinear causal function. Details of this dataset are provided in~\cref{dataset:pcl}.

The results are presented in~\cref{fig:pcl_dsprite}. DML-CMR achieved similar performance to the SOTA methods while outperforming CE-DML-CMR. In addition, a lower variance can be observed when using DML-CMR compared to other methods, especially for smaller data sizes. The NMMR methods, in some cases, outperform CE-DML-CMR. This is in line with our previous observations that, without K-fold validation, performance can be worse for high-dimensional datasets and the full debiasing effect of the DML framework is required to achieve the best results.

\section{Discussion}

In this chapter, we proposed a novel estimator for solving CMRs, DML-CMR. Using the DML framework and our novel Neyman orthogonal score, DML-CMR can effectively estimate solutions to CMR problems with fast convergence rate guarantees of $O(N^{-1/2})$ by mitigating the regularisation and overfitting biases in two-stage estimations. We theoretically analysed DML-CMR and proved a convergence rate of $O(N^{-1/2})$ with high probability under mild regularity and parametric assumptions. We also demonstrated interesting connections between the notion of ill-posedness for CMRs and DML's identifiability condition. Using DML-CMR, we are able to learn causal functions and effective decision policies for the offline IV bandit problem with suboptimality guarantees. We evaluated DML-CMR on IV regression and offline IV bandit problems with standard benchmarks and semi-synthetic real-world datasets, experimentally showing that it outperforms SOTA IV regression methods with lower estimation error and better stability. In addition, we evaluated DML-CMR on proximal causal learning, which also demonstrated SOTA performance.

The main limitation of this chapter is the requirement that IVs need to be observed in the dataset for each action in order to identify the causal effect of actions. However, finding IVs in real-world applications is a well-studied problem in the literature and it is well documented that IVs can be observed in many real-world scenarios including econometrics, drug testing, social sciences, and recommender systems (see \citet{WuAnpeng2023} Section 8 for more real-world datasets and applications). Moreover, even when IVs are not identified in a dataset by experts, there exist empirical methods that can test and find good candidates for IVs in the dataset (e.g., \citealt{Yuan2021,WuAnpeng2023} Section 6). That being said, it is not guaranteed that IVs can be found in all real-world datasets, so in \cref{chapter:il}, we propose a method to imitate expert demonstrations with hidden confounders without explicit IVs being observed in the dataset.

Another limitation is the parametric assumption on $f_0$ for the theoretical analysis following the DML framework. As discussed in \cref{sec:theory}, a more generic analysis can be performed for infinite-dimensional $f_0$ following \citet{Foerster2018}, but we provide a concrete analysis with an explicit convergence rate for DML-CMR in this chapter. Nevertheless, it would be interesting future work to analyse our Neyman orthogonal score for estimating nonparametric functions of interest. In addition, our offline IV bandit setting can be extended to sequential decision settings~\citep{Namkoong2020,Liao2024} with IVs at each time step, which is an interesting future direction to explore.
\chapter{Imitating Expert Policies from Confounded Datasets}\label{chapter:il}

\minitoc

\section{Introduction}\label{sec-il:intro}

In this chapter, we consider the problem of imitation learning (IL) in the presence of hidden confounders with suboptimality guarantees, without the explicit assumption relied upon in \cref{chapter:dmliv} that IVs can be observed in addition to the MDP trajectories in the dataset. While classical IL theory implies that, with infinite data, the IL error should converge to zero and the imitator should be value-equivalent to the expert~\citep{Ross2011,Spencer2021}, it has been observed in practice that IL algorithms often produce incorrect estimates of the expert policy, leading to suboptimal and unsafe behaviours~\citep{Lecun2005,Codevilla2019,Bansal2018,Kuefler2017}.

Most existing work attributes the underlying causes of this problem to hidden confounders and spurious correlations. However, as discussed in \cref{lit:causal_il}, they only consider specific aspects of the problem in different settings, and a holistic treatment of the problem is lacking. In this chapter, we propose a unifying framework that generalises existing work by explicitly formulating hidden confounders, which, in the context of IL, refers to variables present in the environment but not recorded in the demonstrations. Crucially, we distinguish between hidden confounders that are observable to the expert and those that are not\footnote{We always assume that the imitator does not observe the hidden confounders.}. When the expert can observe the hidden confounders, they represent privileged information. When the expert cannot observe the hidden confounders, they act as confounding noise that contaminates the demonstrations, causing spurious correlations and causal delusions.

% In our novel and general framework, we propose algorithms that can correctly and efficiently imitate the expert policy. In other settings, it has been shown that the application of an interactive IL algorithm such as DAgger~\citep{Ross2011}, which allows us to directly query the expert, can be effective in dealing with hidden confounders (e.g.,~\citep{Swamy2022,Vuorio2022,Swamy2022_temporal}). However, an interactive expert is not a realistic assumption in many domains and applications. Therefore, in our general framework, we develop approaches that solely rely on a fixed set of demonstrations. 

Building on our unifying framework, we develop algorithms that can correctly and efficiently imitate the expert policy in the presence of hidden confounders. To break the spurious correlations introduced by the expert-unobservable hidden confounders without explicit IVs at each time step, our key observation is that the trajectory histories themselves can serve as IVs for state-action pairs within our confounded IL framework. Moreover, by conditioning on the trajectory history, we can infer information about observable hidden confounders and learn a history-dependent policy that accurately mimics the expert's behaviour. Importantly, we demonstrate that, within our general framework, IL under hidden confounders can be reformulated as a conditional moment restriction (CMR) problem. This reformulation allows us to design practical algorithms based on CMR estimators (e.g., DML-CMR proposed in~\cref{chapter:dmliv}) with theoretical guarantees on the imitation gap.

\paragraph{Main Contributions of this Chapter.} 
\begin{itemize}[leftmargin=10pt, topsep=2pt, itemsep=2pt, topsep=2pt]
    \item We introduce a unifying framework for causal IL (\cref{sec-il:setting}) incorporating both expert-observable and expert-unobservable confounding variables to unify and generalise prior work (e.g., \citealt{Swamy2022_temporal,Swamy2022,Ortega2021,Vuorio2022}). 
    \item We reformulate the problem of confounded IL in our unifying framework as solving a CMR problem as in \cref{chapter:dmliv}, where we learn a history-dependent policy by leveraging trajectory histories as instruments to break the confounding (\cref{sec-il:method}).
    \item We propose DML-IL, a novel IL algorithm in our framework, for which we prove an upper bound on the imitation gap that recovers the theoretical results in prior work as special cases (\cref{thm:gap}).
    \item We empirically validate our algorithm in both custom and MuJoCo environments with both expert-observable and expert-unobservable confounders and demonstrate that DML-IL outperforms existing causal IL baselines (\cref{sec-il:exps}). This highlights the need to explicitly account for both types of hidden confounders. 
\end{itemize}

Work reported in this chapter first appeared in \citet{Shao2025}.

% The key idea is to leverage the trajectory history as Instrumental Variables (IVs) to break the spurious correlations caused by the expert-unobservable hidden confounders, and to infer information about the expert-observable hidden confounders from the expert's past trajectory by learning a history-dependent policy conditioned on the trajectory history. 
% Crucially, in our general framework, we show that IL in the presence of hidden confounders can be reduced to a set of Conditional Moment Restrictions (CMR), which is a problem widely studied in the econometrics and causal inference literature. This allows us to design practical algorithms with theoretical guarantees on the imitation gap based on efficient algorithms from causal inference for solving CMRs.

% \vspace{-5pt}

\section{Causal Imitation Learning}

Before formalising our framework and the corresponding IL problem in \cref{sec-il:setting}, we first discuss existing causal IL formulations and clarify the objective of this chapter. As discussed above, previous work on IL with hidden confounders, which we refer to as causal IL, extends standard IL (recall \cref{prelim:IL}) under varying assumptions in different settings, each adopting its own specific formalism.

Previous work has studied specific problem settings in isolation, including spurious correlations without explicit hidden confounders~\citep{deHaan2019,Codevilla2019,Pfrommer2023}, temporal confounding noise~\citep{Swamy2022_temporal}, expert-specific privileged information~\citep{Swamy2022,Vuorio2022,Chen2019,Choudhury2017}, causal delusions~\citep{Ortega2008,Ortega2021}, and covariate shifts~\citep{Spencer2021}. However, in real-world scenarios, these problems often exist simultaneously (e.g., confounding noise and expert-privileged information co-occurring), rendering partial or independent treatments insufficient, as demonstrated in \cref{sec-il:exps}. Our framework proposed in \cref{sec-il:setting} unifies these prior settings (see \cref{appendix:reduce}) and allows a much larger family of problems that are more realistic in practice to be considered.

More specifically, some of these works~\citep{Swamy2022,Vuorio2022,Swamy2022_temporal} assume that the imitator can interactively query the expert for additional demonstrations, and interactive IL algorithms (recall \cref{prelim:IL}), such as DAgger~\citep{Ross2011}, are used to address the hidden confounders. However, such an assumption is impractical in many real-world scenarios, where only a fixed set of demonstrations is available. For this reason, we focus on developing methods that mitigate the effects of hidden confounders while relying solely on a fixed set of non-interactive expert demonstrations.

\section{A Unifying Framework for Causal Imitation Learning}\label{sec-il:setting}

In this section, we introduce a unifying framework for causal IL in the presence of hidden confounders and formally state the problem setting. We then show in \cref{appendix:reduce} that this framework subsumes and recovers the specific settings considered in prior work.

\subsection{MDPs with Hidden Confounders.} 
We begin by introducing a Markov decision process (MDP) framework with hidden confounders, $(\states, \actions, \confounders, \transitions, \reward, \mu_0,T)$, where $\states$ is the state space, $\actions$ is the action space, and $\confounders$ is the confounder space. Importantly, parts of the hidden confounder $u_t$ at time $t$ may be available to the expert but not to the imitator due to imperfect data logging or expert knowledge. 
We model this by segmenting the hidden confounder at time $t$ into two parts $u_t=(u^o_t,u^\epsilon_t)$, where $u^o_t$ is observable to the expert and $u^\epsilon_t$ is not. Intuitively, $u^o_t$ corresponds to the additional information that only the expert observes, and $u^\epsilon_t$ acts as confounding noise in the environment that affects both the state and action.\footnote{In our framework, we allow the actual actions taken in the environment to be affected by the noise. Noise that only perturbs data records can be considered as a special case of our framework.}

\begin{figure}[t]
    \centering
\includegraphics[width=0.9\textwidth]{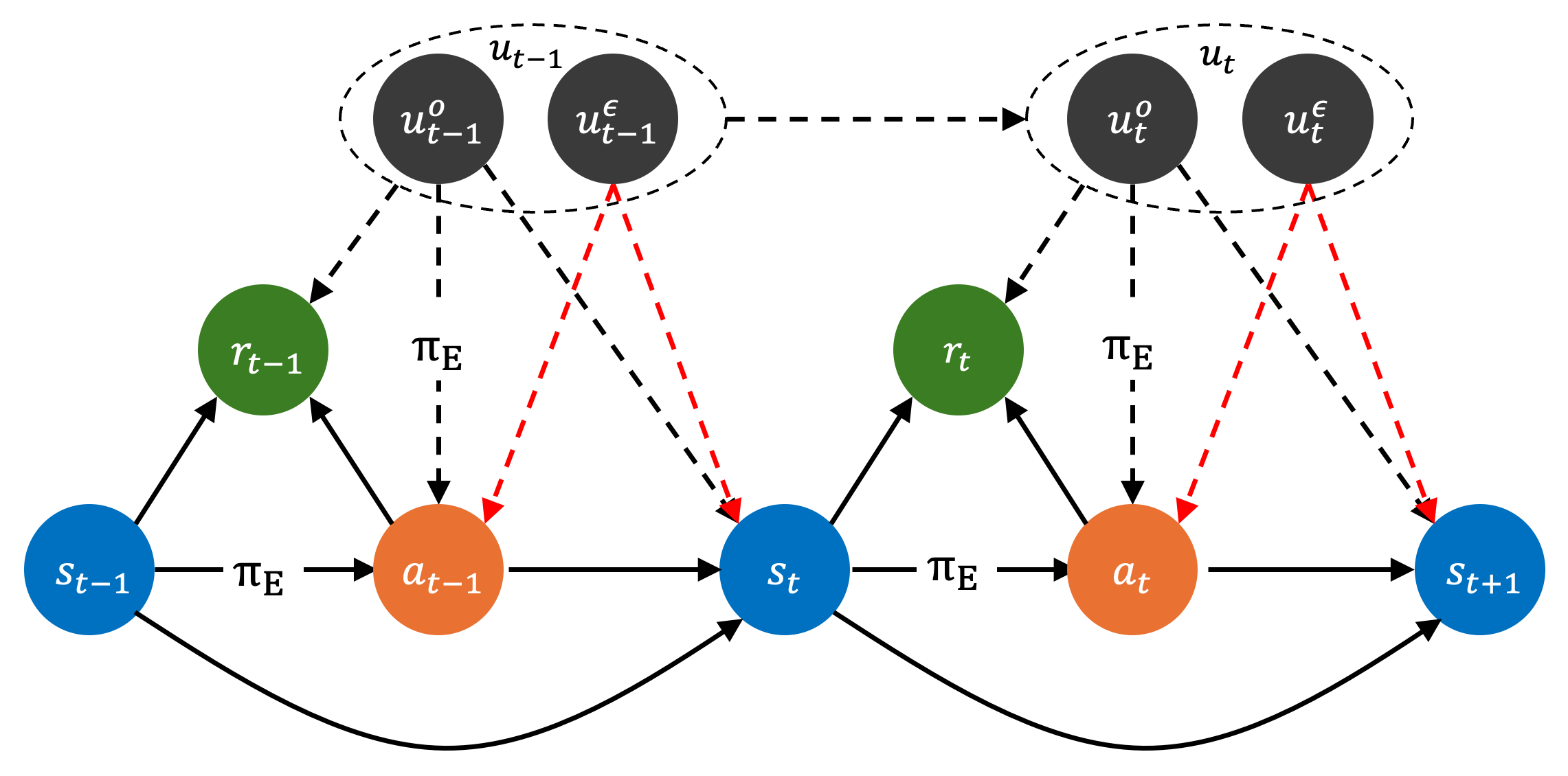}
    \caption[A causal graph of MDPs with hidden confounders $u_t=(u^o_t,u^\epsilon_t)$.]{A causal graph of MDPs with hidden confounders $u_t=(u^o_t,u^\epsilon_t)$. The black dotted lines represent the causal effect of the expert-observable confounder $u^o_t$, which directly affects the expert action $a_t$. It also directly affects $s_{t+1}$ and $r_t$. 
    The red dotted lines represent the causal effect of the expert-unobservable $u^\epsilon_t$, which acts as confounding noise and directly affects the states and actions. $u^\epsilon_t$ does not directly affect $r_t$ (following~\citet{Swamy2022_temporal}) because the expert policy does not take $u^\epsilon_t$ into account, and letting $u^\epsilon_t$ directly affect $r_t$ would only add noise to the expected return.}
    \label{fig:MDPUC}
\end{figure}

As a result, the transition function $\transitions(\cdot\mid s_t,a_at,(u^o_t,u^\epsilon_t))$ at time $t$ depends on both hidden confounders, but the reward function $\reward(s_t,a_t,u^o_t)$ only depends on the state, action, and the observable confounder $u^o_t$ since the confounding noise only directly affects the state and actions. Finally, $\mu_0$ is the initial state distribution, and $T$ is the time horizon. A causal graph illustrating these relationships is provided in~\cref{fig:MDPUC}. This nuanced distinction between $u^o_t$ and $u^\epsilon_t$ is crucial for determining the appropriate method for IL, and we begin with an example to motivate our setting and illustrate the importance of considering $u_t=(u^o_t,u^\epsilon_t)$.

\begin{example}\label{eg:plane}
% Consider a dynamic aeroplane ticket pricing scenario~\citep{wright1928}, where we would like to learn a ticket pricing policy by imitating actual airline prices based on the profit margins set by experts. We have access to information such as destinations, flight time, previous sales, and aeroplane records. However, seasonal demand patterns and events are known to the experts, but are not logged in the dataset. Hence, these hidden confounders are included in $u^o_t$. In contrast, the experts only determine profit margins as an action because the observed price is confounded (additively) by fluctuations in operating costs such as fuel price and maintenance costs, which are unknown to the expert when they set the profit margin. These hidden confounders act as $u^\epsilon_t$ in our setting. Without an explicit consideration of $u^o_t$ and $u^\epsilon_t$ separately, learning algorithms cannot distinguish between $u^o_t$ and $u^\epsilon_t$ and fail to correctly imitate the expert. We perform experiments on this toy example in~\cref{sec:exps}.
Consider an airline ticket pricing scenario~\citep{wright1928}, where the goal is to learn a pricing policy by imitating actual airline pricing based on expert-set profit margins. Suppose that seasonal patterns and external events are known only to experts, but are absent from the dataset. Hence, these latent variables serve as expert-observable confounders $u^o_t$. Meanwhile, actual airline prices are confounded (additively) by fluctuating operating costs, which are unknown to the experts when they set the profit margin and are not included in the dataset. Consequently, such fluctuating operating costs act as confounding noise $u^\epsilon_t$. We conduct experiments in a toy environment inspired by this example in~\cref{sec-il:exps}, and demonstrate that IL algorithms that do \emph{not} distinguish between $u^o_t$ and $u^\epsilon_t$ fail to correctly imitate the expert.
\end{example}

\subsection{Causal Imitation Learning in Our Framework} 
We assume that an expert is demonstrating a task following some expert policy $\pi_E$ (which we will specify later) and we observe a set of $N \geq 1$ expert demonstrations $\{d_1, d_2,...,d_N\}$. Each demonstration is a state-action trajectory $(s_1,a_1,...,s_{T},a_{T})$, where, at each time step $t$, we observe the state $s_t$ and the action $a_t$ taken in the environment. The next state is then sampled from the transition function $\transitions(\ \cdot\mid s_t, a_t, (u^o_t, u_t^\epsilon))$.

Let $h_t=(s_{1},a_{1},...,s_{t-1},a_{t-1},s_{t})\in\mathcal{H}$ denote the trajectory history at time $t$, where $\mathcal{H}\subseteq \bigcup_{i=0}^{T-1} (\states\times\actions)^{i}\times \states$ is the set of all possible trajectory histories. % at different time steps.
% and $h_t^k=(s_{t-k+1},a_{t-k+1},...,s_{t})$ for $k\leq t$ as the $k$ step history from $t$.
Importantly, we observe neither the reward nor the confounders $(u^o_t,u^\epsilon_t)$ at time $t$. Given the observed trajectories, our goal is to learn a history-dependent policy $\pi_h:\mathcal{H}\rightarrow \Delta(A)$, where we assume that our policy class $\Pi$ is convex and compact. The $Q$-function of a policy $\pi_{h}\in\Pi$ is given by $Q_{\pi_{h}}(s_t,a_t,u_t^o)=\expectE_{\tau\sim\pi_{h}}[\sum_{t^\prime=t}^T \reward(s_{t^\prime},a_{t^\prime},u^o_{t^\prime})]$, the value of a policy is $J(\pi_{h})=\expectE_{\tau\sim\pi_{h}}[\sum_{t^\prime=1}^{T}\reward(s_{t^\prime},a_{t^\prime},u^o_{t^\prime})]$, and the imitation gap between $\pi_{h}$ and the expert $\pi_E$ is defined as $J(\pi_E)-J(\pi_{h})$.

In order to learn a policy $\pi_h$ that matches the performance of $\pi_E$, we need to break the spurious correlation between states and expert actions by inferring what the expert would do if we intervened and placed them in state $s_t$ when observing $u^o_t$. 
However, the causal inference literature~\citep{Shpitser2008} tells us that, without further assumptions, it is generally impossible to identify $\pi_E$.
To determine the minimal assumptions that allow $\pi_E$ to be identifiable, we first observe that $u^\epsilon_t$ can be correlated for all time steps $t$, making it impossible to distinguish between the intended actions of the expert and the confounding noise. 
\bill{However, in practice, the confounding noise at far-apart time steps is often independent. For example, the effect of the confounding noise $u^\epsilon_t$ at time $t$ on future states and actions often diminishes over time. This is typically the case for random environment noise such as wind, which acts as confounding noise $u^\epsilon_t$. In addition, when the confounding noise $u^\epsilon_t$ at time $t$ becomes observable at a future time $t^\prime$, e.g., previous operating costs are eventually observed as in \cref{eg:plane}, the unobservable confounding noise at times $t$ and $t^\prime$ become independent. If this condition holds, it becomes possible to decouple the spurious correlation between the state and action pairs. Motivated by this observation, we formalise the following assumption on the \emph{confounding noise horizon} $k$.}

\begin{assumption}[Confounding noise horizon]\label{assump:horizon}
For every $t$, the confounding noise $u^\epsilon_t$ has a horizon of $k$ where $1\leq k< T$. More formally, $u^\epsilon_t\indep u^\epsilon_{t-k}\ \forall t>k$.
\end{assumption}

We also assume that the confounding noise is additive to the action, which is standard in causal inference~\citep{Pearl2000causality,Shao2024}. Without this assumption, the causal effect becomes unidentifiable (see, e.g.,~\citep{Balke1994}) and the best we can do is to upper/lower bound it.

\begin{assumption}[Additive noise]\label{assump:additive}
The structural equation that generates the actions in the observed trajectories is
\begin{align}
a_t&=\pi_E(s_t,u^o_t)+u^\epsilon_t,\label{eq:action}
\end{align}
where w.l.o.g.\ $\expectE[u^\epsilon_t]=0$ as nonzero expectation of $u^\epsilon_t$ can be included as a constant in $\pi_E$.
\end{assumption}

In \cref{sec-il:method}, we show that, with the above two assumptions, it becomes possible to identify the true causal relationship between states and expert actions, and to imitate $\pi_E$.

% \begin{remark}
% Our MDP setting is different to contextual MDPs (CMDP) since the unobserved context there is fixed thorughout an episode, and CMDP can be considered a very special case of MDPUC. MDPUC is also different to POMDPs since POMDPs doesn't have any markovian properties and doesn't imply confounding.
% \end{remark}

% application examples: 

% 1. helping people living with type-1 diabetes to time their
% insulin injections by monitoring their blood glucose level using some wearable device. observations of individuals’ blood glucose levels over time and the timing of insulin injections. However, there may in fact be events not recorded in the data, such as food intake and exercise, which may affect both the timing of injections and blood glucose.

% 2. Trading system, external factors such as market sentiments, economic indicators, and news events, which can be modelled as confounding noises.

% 3. Driving or robotics, invisible noises such as vibration or wind in the environment that can affect both current state and action

\subsection{Reducing Our Framework to Prior Work}\label{appendix:reduce}

In this section, we discuss how the various previous works can be obtained as special cases of %reduced from 
our unifying framework.

\subsubsection{Temporally Correlated Noise}

The temporally correlated noise (TCN) proposed in~\citet{Swamy2022_temporal} is a special case of our setting where $u^o=0$ and only the confounding noise $u^\epsilon$ is present. Following Equation 14-17 of~\citet{Swamy2022_temporal}, their setting can be summarised as
\begin{align*}
s_t &= \mathcal{T}(s_{t-1}, a_{t-1})\\
  &= \mathcal{T}(s_{t-1}, \pi_E(s_{t-1}) + u_{t-1} + u_{t-2});\\
a_t &= \pi_E(s_t) + u_t + u_{t-1},
\end{align*}
where $\mathcal{T}$ is the transition function and $u_t$ represents the TCN. It can be seen that TCN acts as the confounding noise $u^\epsilon$ since the expert policy does not take it into account, and it affects (or confounds) both the state and action. Therefore, this is a special case of our framework when $u^o_t=0$, where $a_t=\pi_E(s_t)+\epsilon(u^\epsilon_t)$ from~\cref{eq:action}, and more specifically when the confounding noise horizon $k=2$. In addition, the theoretical results in~\citet{Swamy2022_temporal} can be deduced from our main results, as shown later in \cref{corollary:unconfounded}.

\subsubsection{Unobserved Contexts}
The setting considered by~\citet{Swamy2022} is a special case of our setting when $u^\epsilon=0$ and only $u^o$ are present. Following Section 3 of~\citet{Swamy2022}, their setting can be summarised as
\begin{align*}
\mathcal{T}&:\states\times\actions\times C \rightarrow D(S);\\
\mathcal{r}&:\states\times\actions\times C \rightarrow [-1,1];\\
a_t&=\pi_E(\state_t,c),
\end{align*}
where $c\in C$ is the context, which is assumed to be fixed throughout an episode. There are no hidden confounders in this setting and the context $c$ is included in $u^o$ under our framework. In addition, our framework allows $u^o$ to vary throughout an episode. Furthermore, the theoretical results in~\citet{Swamy2022} can be deduced from our main results, as shown later in \cref{corollary:noUo}.

\subsubsection{Imitation Learning with Latent Confounders}

The setting considered by~\citet{Vuorio2022} is also a special case of our setting when $u^\epsilon=0$ and only $u^o$ are present, which is very similar to~\citet{Swamy2022}. In Section 2.2 of~\citet{Vuorio2022}, they introduced a latent variable $\theta\in\Theta$ that is fixed throughout an episode and $a_t=\pi_E(\state_t,\theta)$. Therefore, the latent variable $\theta$ is included in $u^o$ in our framework. In~\citet{Vuorio2022}, no theoretical imitation gap bounds are provided. However, \cref{corollary:noUo} can be applied directly to their setting to bound their imitation gap.

\subsubsection{Causal Delusion and Confusion}

The concepts of \textit{causal delusion}~\citep{Ortega2021} and \textit{causal confusion} have been extensively studied from various perspectives~\citep{deHaan2019,Pfrommer2023,Spencer2021,Wen2020}. A classic example arises in autonomous driving: when the states include dashboard and road images, the brake indicator light acts as a confounding variable since it correlates with the braking action in subsequent steps. As a result, the imitator might incorrectly learn to brake whenever the brake indicator light is already on, effectively latching onto spurious correlations between the current action and the trajectory history.

In~\citet{Ortega2021}, such causal delusion is explicitly modelled through hidden confounders, which correspond to the confounding noise $u^\epsilon$ in our framework. Other works~\citep{deHaan2019,Pfrommer2023,Spencer2021,Wen2020} do not explicitly model unobserved confounders, but the spurious correlation between previous states and actions can be modelled as the existence of confounding noise $u^o$ in our framework. In the autonomous driving example, the actual road hazard serves as $u^o$ that simultaneously causes the expert to brake and activates the brake indicator light.

However, while these settings can be viewed as special cases of our framework, we stress that the practical problems studied in~\citet{deHaan2019,Pfrommer2023,Spencer2021,Wen2020} differ from ours. In particular, they implicitly assume that the hidden confounders are either embedded in the observations or directly observable, and focus on designing practical algorithms that identify and exploit such confounders to improve IL performance.

\section{Causal IL as a CMR Problem}\label{sec-il:method}

In this section, we demonstrate that performing causal IL in our framework is possible using trajectory histories as instruments. We show that the problem can be reformulated as a CMR problem and propose an effective algorithm to solve it.

The typical target for IL would be the expert policy $\pi_E$ itself. However, since the expert has access to privileged information, namely $u^o_t$, which the imitator does not, the best thing an imitator can do is to learn a history-dependent policy $\pi_h$ that is the closest to the expert. A natural choice for a learning objective is the conditional expectation of $\pi_E(s_t,u^o_t)$ on the history $h_t$:
\begin{align}
\pi_h(h_t)\coloneqq \expectE_{\probP(u^o_t\mid h_t)}[\pi_E(s_t,u^o_t)]=\expectE[\pi_E(s_t,u^o_t)\mid h_t],\nonumber
\end{align}
because the conditional expectation minimises the least squares criterion~\citep{hastie01statisticallearning} and $\pi_h$ is the best predictor of $\pi_E$ given $h_t$. In $\pi_h$, the distribution $\probP(u^o_t\mid h_t)$ captures the information about $u^o_t$ that can be inferred from trajectory histories.
\begin{remark}
\emph{Learning $\pi_h$ is not trivial. Policies learnt naively using behaviour cloning (i.e., $\expectE[a_t\mid h_t]$) fail to match $\pi_E$. To see this, note that, in view of~\cref{eq:action}, we have 
\begin{align} 
\expectE[a_t\mid h_t]&=\expectE[\pi_E(s_t,u^o_t) \mid h_{t}]+\expectE[u^\epsilon_t\mid h_{t}]\nonumber\\
&=\pi_h(h_t)+\expectE[u^\epsilon_t\mid h_{t}],\label{eq:history_policy}
\end{align}
where $\expectE[u^\epsilon_t\mid h_{t}]\neq 0$ due to the spurious correlation between $u^\epsilon_t$ and the trajectory history $h_t$. As a result, $\expectE[a_t\mid h_t]$ becomes biased, which can lead to arbitrarily worse performance compared to $\pi_E$.   }
\end{remark}

% \vspace{-5pt}
\subsection{Derivation of the CMR Problem} 
Leveraging the confounding horizon from \cref{assump:horizon}, we are able to break the spurious correlation using the independence of $u^\epsilon_t$ and $u^\epsilon_{t-k}$. We propose to use the $k$-step trajectory history $h_{t-k}=(s_{1},a_{1},...,s_{t-k})$ as an instrument for the current state $s_t$. Taking the expectation conditional on $h_{t-k}$ for both sides of~\cref{eq:history_policy} yields:
\begin{align*}
    \expectE[a_t\mid h_{t-k}]  = \expectE\left[\expectE[a_t\mid h_{t}]\mid h_{t-k}\right]  & = \expectE[\pi_h(h_t)\mid h_{t-k}]+\expectE[\expectE[u^\epsilon_t\mid h_{t}]\mid h_{t-k}] \\
    & = \expectE[\pi_h(h_t) \mid h_{t-k}]+\expectE[u^\epsilon_t\mid h_{t-k}]\\
    &= \expectE[\pi_h(h_t) \mid h_{t-k}]+\expectE[u^\epsilon_t]=\expectE[\pi_h(h_t) \mid h_{t-k}],
\end{align*}
where we use the fact that $h_{t-k}$ is $\sigma(h_t)$-measurable because $h_{t-k}\subseteq h_t$, $u^\epsilon_t\indep u^\epsilon_{t-k}$ and $\expectE[u^\epsilon_t] = 0$ by \cref{assump:horizon}. As a result, the problem of learning $\pi_h$ reduces to solving for $\pi_h$ that satisfies the following identity:
\begin{align}
    \expectE[a_t-\pi_h(h_t)\mid h_{t-k}]=0,\label{eq:CMR}
\end{align}
which is a CMR problem as defined in~\cref{background:CMRs}. In this case, both $a_t$ and $h_t$ are observed in the confounded expert demonstrations, and $h_{t-k}$ acts as the instrument.

To ensure that the instrument $h_{t-k}$ is valid, we verify the three conditions of IV in \cref{background:IV}: firstly, $u^\epsilon_t\indep h_{t-k}$ as explained above.  Secondly, the environment and expert policy are nontrivial since $\probP(h_t\mid h_{t-k})$ is not constant in $h_{t-k}$. Finally, $h_{t-k}$ affects $a_t$ only through $s_t$ by the Markov property. However, the strength of the instrument $h_{t-k}$, representing its correlation with $h_{t}$, influences how well $\pi_h(h_t)$ can be identified by solving the CMR in~\cref{eq:CMR}. As the confounding horizon $k$ increases, this correlation weakens, making $h_{t-k}$ a less effective instrument. This relationship is formally analysed in \cref{IL-prop:ill-posed} and validated in experiments in~\cref{sec-il:exps}.

% Note this problem is equivalent to solving an IV regression on~\cref{eq:history_policy}, where $Y=\expectE[a_t\lvert h_t]$, $f(x)=\pi_h(h_t)$, $\epsilon=\expectE[u^\epsilon_t$ and the instrument $Z=h_{t-k}$.

\subsection{Practical Algorithms for Solving the CMR Problem}\label{sec:dml_il_algo}

% \vspace{-0.05cm}

% \begin{algorithm}[tb]
% % \setstretch{1.2}
% \DontPrintSemicolon
% \SetNlSty{textbf}{}{:}
% \caption{DML-IL}
% \label{alg:DML-IL}
% \BlankLine
% \SetKw{KwIn}{in}
% \SetKwInOut{Input}{Input}
% % \SetKwInOut{Output}{Output}
% % \SetKwFunction{proc}{LTLQ($\LTL$,$\MDP$)}
% \Input{Dataset $\dataset_E$ of expert demonstrations, confounding noise horizon $k$}
% % \Output{Game reward at $root$ node: $R(s_0,\sigma)$}
% % \SetKwProg{myproc}{Procedure}{:}{}
%   {
%   \nl Initialize the roll-out model $\widehat{M}$ as a Gaussian mixture model\label{algo:roll_out_1}\\
%   \nl\Repeat {Sample $(h_{t},a_t)$ from data $\dataset_E$}\\
%   sdf\\
%   \nl construct product MDP $\productMDP$ using $\MDP$ and $\automaton$\\
%   \nl   \Return induced policy on $\MDP$ by removing $\epsilon$-actions
%   }
% \end{algorithm}

\begin{algorithm}[tb]
   \caption{DML-IL}
   \label{alg:DML-IL}
\begin{algorithmic}[1]
   \STATE {\bfseries Input:} Dataset $\dataset_E$ of expert demonstrations, confounding noise horizon $k$
   \STATE Initialise the roll-out model $\widehat{M}$ as a Gaussian mixture model\label{algo:roll_out_1}
    \REPEAT
   \STATE Sample $(h_{t},a_t)$ from data $\dataset_E$
   \STATE Fit the roll-out model $(h_t,a_t)\sim\widehat{M}(h_{t-k})$ to maximise the log likelihood 
\UNTIL{convergence}\label{algo:roll_out_2}
   \STATE Initialise the expert model $\widehat \pi_h$ as a neural network
   \REPEAT
   % \FOR{$k=1$ {\bfseries to} $K$}
   \STATE Sample $h_{t-k}$ from $\dataset_E$
   \STATE Generate $\widehat{h}_t$ and $\widehat{a}_t$ using the roll-out model $\widehat{M}$
   \STATE Update $\widehat \pi_h$ to minimise the loss $\ell:= \norm{\widehat{a}_t - \widehat{\pi}_h (\widehat h_t)}_2$
   % \ENDFOR
    \UNTIL{convergence}
    \STATE {\bfseries Output:} A history-dependent imitator policy $\widehat{\pi}_h$
\end{algorithmic}
% \vspace{-0.1cm}
\end{algorithm}

There are various techniques~\citep{Bennett2019DeepAnalysis,Xu2020,Shao2024} for solving the CMR $\expectE[a_t\lvert h_{t-k}]=\expectE[\pi_h(h_t) \lvert h_{t-k}]$. Here, the \textit{CMR error} that we minimise is given by 
\begin{align*}
\sqrt{\expectE\big[\expectE[a_t-\widehat{\pi}_h(h_t)\lvert h_{t-k}]^2\big]}=\norm{\expectE[a_t-\widehat{\pi}_h(h_t)\lvert h_{t-k}]}_{2}.    
\end{align*}
In~\cref{alg:DML-IL}, we introduce DML-IL, an algorithm that applies our CMR estimator DML-CMR, proposed in~\cref{chapter:dmliv}, to solve the CMR problem we derived for causal IL. The first part of the algorithm (lines 3-7) learns a Gaussian mixture roll-out model $\smash{\widehat{M}}$ that predicts the transition dynamics and generates a trajectory $k$ steps ahead given $h_{t-k}$. Then, 
the DNN-based expert model $\smash{\widehat{\pi}_h}$ takes the generated trajectory $\smash{\widehat{h}_t}$ from $\smash{\widehat{M}(h_{t-k})}$ as input and minimises the mean square error to the next action $\widehat{a}_t$, that is, $\ell:= \norm{\widehat{a}_t - \widehat{\pi}_h (\widehat h_t)}_2$, until convergence (lines 8-13).
% the roll-out model $\widehat{M}$ is used to generate trajectory $\widehat{h}_t$ to train the policy model $\widehat{\pi}_h$ (line 8-13) as inputs and minimises the mean squared error to the next action. 
% $\widehat{\pi}_h$ takes the generated trajectory $\widehat{h}_t$ from $\widehat{M}(h_{t-k})$ as inputs, and minimises the mean squared error to the next action. 
Using generated trajectories $\smash{\widehat{h}_t}$ is crucial in breaking the spurious correlation caused by $u^\epsilon_t$, and the trajectory history before $h_{t-k}$ allows the imitator to infer information about $u^o_t$.

Note that~\cref{alg:DML-IL} requires the confounding noise horizon $k$ as input. Although the exact value of $k$ can be difficult to obtain in reality, any upper bound $\bar{k}$ of $k$ is sufficient to guarantee the correctness of ~\cref{alg:DML-IL}, since $h_{t-\bar{k}}$ is also a valid instrument. Ideally, we would like a data-driven approach to determine $k$. However, it is generally intractable to empirically verify whether $h_{t-k}$ is a valid instrument from a static dataset, especially the unconfounded instrument condition (i.e., $h_{t-k}\indep u^\epsilon_t$). Therefore, we rely on the user to provide a reasonable choice of $\bar{k}$ based on the environment that does not substantially overestimate $k$.

\begin{algorithm}[tb]
   \caption{DML-IL with $K$-fold cross-fitting}
   \label{alg:DML-IL-kfold}
\begin{algorithmic}
   \STATE {\bfseries Input:} Dataset $\dataset_E$ of expert demonstrations, Confounding noise horizon $k$, number of folds $K$ for cross-fitting
    % \STATE {\bfseries Output:} A history-dependent imitator policy $\widehat{\pi}_h$
   \STATE Get a partition $(I_i)^K_{i=1}$ of dataset indices $[N]$ of trajectories
   \FOR{$i=1$ {\bfseries to} $K$}
   \STATE $I^c_i\coloneqq[N]\setminus I_i$
   \STATE Initialise the roll-out model $\widehat{M}_i$ as a mixture of Gaussians model
   \REPEAT
   \STATE Sample $(h_{t},a_t)$ from data $\{(\dataset_{E,j}):{j\in I^c_i}\}$
   \STATE Fit the roll-out model $(h_t,a_t)\sim\widehat{M}_i(h_{t-k})$ to maximise log likelihood
   \UNTIL{convergence}
    \ENDFOR
   \STATE Initialise the expert model $\widehat \pi_h$ as a neural network
   \REPEAT
\FOR{$i=1$ {\bfseries to} $K$}
   \STATE Sample $h_{t-k}$ from $\{(\dataset_{E,j}):{j\in I_i}\}$
   \STATE Generate $\widehat{h}_t$ and $\widehat{a}_t$ using the roll-out model $\smash{\widehat{M}_i}$
   \STATE Update $\widehat \pi_h$ to minimise the loss $\ell:= \norm{\widehat{a}_t - \widehat{\pi}_h (\smash{\widehat{h_t}})}_2$
   \ENDFOR
    \UNTIL{convergence}
    \STATE {\bfseries Output:} A history-dependent imitator policy $\widehat{\pi}_h$
\end{algorithmic}
\end{algorithm}

DML-IL can also be implemented with $K$-fold cross-fitting (recall \cref{alg:dml-iv-kf}, \cref{chapter:dmliv}) to reduce estimation bias and improve training stability. As shown in~\cref{alg:DML-IL-kfold}, the expert dataset $\dataset_E$ with indices $[N]$ is partitioned into $K$ equal-sized folds $\{I_i\}^K_{i=1}$. For each fold $i\in[1..K]$, we use the remaining folds with indices $I^c_i\coloneqq[N]\setminus I_i$ to train a roll-out model $\widehat{M}_i$. Then, a single expert model $\widehat{\pi}_h$ is trained using samples from each fold $I_i$, where the trajectory rollouts $\widehat{h_t}$ are generated using the corresponding roll-out model $\widehat{M}_i$, by minimising the same mean squared error objective as in \cref{alg:DML-IL}. The CMR estimator we use---DML-CMR with $K$-fold cross-fitting---is theoretically shown in~\cref{chapter:dmliv} to converge at the rate of $O(N^{-1/2})$ under standard regularity conditions. Consequently, DML-IL with $K$-fold cross-fitting will also inherit this theoretical guarantee.

\subsection{Theoretical Analysis}\label{sec-il:theory}

In this section, we derive theoretical guarantees for our algorithm, focusing on the imitation gap and its relationship to existing work. All proofs in this section are deferred to~\cref{appendix:proofs}.

On a high level, in order to bound the imitation gap of the learnt policy $\widehat{\pi}_h$, i.e., $J(\pi_E)-J(\widehat{\pi}_h)$, we need to control:
\begin{enumerate}[leftmargin=25pt]
    \item[($i$)] The amount of information about the hidden confounders that can be inferred from trajectory histories;
    \item[($ii$)] The ill-posedness (or identifiability) of the induced CMR problem, which intuitively measures the strength of the instrument $h_{t-k}$;
    \item[($iii$)] The disturbance of the confounding noise to the states and actions at test time.
\end{enumerate}
These factors are all determined by the environment and the expert policy. To control ($i$), we measure how much information about $u^o_t$ is captured by the trajectory history $h_t$ by analysing the total variation (TV) distance between the distribution of $u^o_t$ and $\expectE[u^o_t\lvert h_t]$ along the trajectories of $\pi_E$. To control ($ii$) and ($iii$), we need to introduce the following two key concepts.

\begin{definition}[The ill-posedness of Confounded IL]
Given the derived CMR problem in~\cref{eq:CMR}, 
% for a policy $\pi\in\Pi$, $\norm{\pi_E-\pi}_2$ is the root mean squared error to the expert and $\norm{\expectE[a_t-\pi(s_t)\lvert s_{t-k}]}_2$ is the CMR error we minimise. Then, 
the \emph{ill-posedness} $\ill(\Pi,k)$ of the policy space $\Pi$ with confounding noise horizon $k$ is
\begin{align*}
    \ill(\Pi,k)=\sup_{\pi\in\Pi} \frac{\norm{\pi_E-\pi}_{2}}{\norm{\expectE[a_t-\pi(h_t)\lvert h_{t-k}]}_{2}}.
\end{align*}
\end{definition}
The ill-posedness $\ill(\Pi,k)$ measures the strength of the instrument, where a higher $\ill(\Pi,k)$ indicates a weaker instrument. It bounds the ratio between the $L_2$ error of the imitator to the expert policy, and the learning error of the imitator following our CMR objective.

As discussed previously, intuitively, the strength of the instrument would decrease as the confounding horizon $k$ increases. This is confirmed by the following proposition.
\begin{proposition}[Monotonicity of the ill-posedness]\label{IL-prop:ill-posed}
$\ill(\Pi,k)$ is monotonically increasing as the confounded horizon $k$ increases.
\end{proposition}

Next, we introduce the notion of c-total variation stability for our theoretical analysis.
\begin{definition}[c-total variation stability~\citep{Bassily2021,Swamy2022_temporal}]
Let $P(X)$ be the distribution of a random variable $X:\Omega\rightarrow \mathcal{X}$. $P(X)$ is c-TV stable if for $a_1,a_2\in \mathcal{X}$ and $\Delta>0$,
\begin{align*}
\norm{a_1-a_2}\leq\Delta \implies \delta_{TV}(a_1+X,a_2+X)\leq c\Delta,
\end{align*}
where $\norm{\cdot}$ is some norm defined on $\mathcal{X}$ and $\delta_{TV}$ is the TV distance.
\end{definition}
A wide range of distributions are c-TV stable. For example, standard normal distributions are $\tfrac{1}{2}$-TV stable. We apply this notion to the distribution over $u^\epsilon_t$ to bound the disturbance it induces in the trajectory and the expected return.

With the notion of ill-posedness and c-TV stability, we can now analyse and upper bound the imitation gap $J(\pi_E)-J(\widehat{\pi}_h)$ by controlling the three previously discussed components $(i)$ -- $(iii)$.

\begin{theorem}[Imitation Gap Bound]\label{thm:gap}
Let $\widehat{\pi}_h$ be the learnt policy with CMR error $\epsilon$ and let $\ill(\Pi,k)$ be the ill-posedness of the problem. Assume that $\delta_{TV}(u^o_t,\expectE_{\pi_E}[u^o_t\lvert h_t])\leq\delta$ for $\delta\in\realNumber^+$, $P(u^\epsilon_t)$ is c-TV stable and $\pi_E$ is deterministic. Then, the imitation gap is upper bounded by 
\begin{align*}
    J(\pi_E)-J(\widehat{\pi}_h)\leq T^2\big(c\epsilon\ill(\Pi,k)+2\delta\big)=\mathcal{O}\big(T^2(\delta+\epsilon)\big).
\end{align*}
\end{theorem}

\begin{proof}[Proof sketch]
We begin by using the Performance Difference Lemma~\citep{Kakade2002} to separate the imitation gap into two parts. Let $Q_{\widehat{\pi_h}}(s_t,a_t,u^o_t)$ be the Q-function of $\widehat{\pi_h}$, then
\begin{align*}
J(\pi_E)-J(\widehat{\pi_h})=\sum_{t=1}^T \expectE_{\tau\sim\pi_E}[\tilde{Q}-\expectE_{a\sim\widehat{\pi_h}}[\tilde{Q}]]+\sum_{t=1}^T \expectE_{\tau\sim\pi_E}[Q_{\widehat{\pi_h}}-\tilde{Q}-\expectE_{a\sim\widehat{\pi_h}}[Q_{\widehat{\pi_h}}-\tilde{Q}]],
\end{align*}
for any Q-function $\tilde{Q}(h_t,a_t)$ that takes in the trajectory history $h_t$ and action $a_t$. For this proof, we choose $\tilde{Q}$ as follows:
\begin{equation*}
\tilde{Q}(h_t,a_t)\coloneqq Q(s_t,a_t,\expectE_{\tau\sim \pi_E}[u^o_t\lvert h_t]).
\end{equation*}

The first part can be bounded with the ill-posedness $\ill(\Pi,k)$ and $c$ from the $c-$TV stability of the hidden confounder $U$. Specifically, $\expectE_{\tau\sim \pi_E}[\tilde{Q}_{\widehat{\pi_h}}-\expectE_{a\sim\widehat{\pi_h}}[\tilde{Q}_{\widehat{\pi_h}}]]\leq Tc\epsilon\ill(\Pi,k)$ by first applying the total variation distance bound for expectations of $\tilde{Q}_{\widehat{\pi_h}}$ over different distributions of action $a_t$ and Jensen's inequality.

For the second part, we first bound
\begin{align*}
\abs{\expectE_{\tau\sim\pi_E,\dot{a}\sim\pi}[Q(s_t,\dot{a},u_t)-\tilde{Q}(h_t,\dot{a})]}&\leq \norm{\expectE_{\pi_E,\pi}[Q(s_t,\dot{a},u^o_t)\lvert u^o_t]}_{\infty}\delta_{TV}(u^o_t,\expectE_{\pi_E}[u^o_t\lvert h_t])\\
&\leq T \cdot \delta_{TV}(u^o_t,\expectE_{\pi_E}[u^o_t\lvert h_t])\leq T\delta
\end{align*}
for any action $\dot{a}\sim \pi$ following some policy $\pi$ (in our case, it can be $\pi_E$ or $\widehat{\pi_h}$) using the tower property of expectations and total variation distance bound. Next, we apply this inequality twice to bound $\expectE_{\tau\sim\pi_E}[Q_{\widehat{\pi_h}}-\tilde{Q}_{\widehat{\pi_h}}-\expectE_{a\sim\widehat{\pi_h}}[Q_{\widehat{\pi_h}}-\tilde{Q}_{\widehat{\pi_h}}]]\leq 2T\delta$.

Finally, we use the bounds on the two parts of the imitation gap and deduce that 
\begin{align*}
    J(\pi_E)-J(\widehat{\pi}_h)\leq T\cdot (Tc\epsilon\ill(\Pi,k) + 2T\delta)= T^2(c\epsilon\ill(\Pi,k)+2\delta)=\mathcal{O}(T^2(\epsilon+\delta)).
\end{align*}
\end{proof}

This upper bound scales at the rate of $T^2$, which aligns with the expected behaviour of imitation learning without an interactive expert~\citep{Ross2010}.
Next, we show that the upper bounds on the imitation gap from prior work~\citep{Swamy2022_temporal, Swamy2022} are special cases of
% of  subsumed by the unifying causal IL framework introduced in \cref{sec:setting} are special cases of 
\cref{thm:gap}. The proofs are deferred to~\cref{appendix:corollaries}.
\begin{corollary}\label{corollary:noUo}
In the special case that $u^o_t = 0$, i.e., there are no expert-observable confounders, or $u^o_t=\expectE_{\pi_E}[u^o_t\lvert h_t]$, i.e., $u^o_t$ is $\sigma(h_t)$ measurable (all information about $u^o_t$ is contained in the history), the imitation gap is upper bounded by
\begin{align*}
    J(\pi_E)-J(\widehat{\pi}_h)\leq T^2\big(c\epsilon\ill(\Pi,k)\big)=\mathcal{O}\big(T^2\epsilon\big),
\end{align*}
which coincides with Theorem 5.1 of~\citet{Swamy2022_temporal}.
\end{corollary}

When there are no hidden confounders, i.e., $u^\epsilon_t=0$, our framework is reduced to that of~\citet{Swamy2022}. However, \citet{Swamy2022} provided an abstract bound that directly uses the supremum of key components in the imitation gap over all possible Q-functions to bound the imitation gap. We further extend and concretise the bound using the learning error $\epsilon$ and the TV distance bound $\delta$ instead of relying on the supremum.

\begin{corollary}\label{corollary:unconfounded}
In the special case that $u^\epsilon_t=0$, if the learnt policy has optimisation error $\epsilon$,  the imitation gap is upper bounded by
\begin{align*}
    J(\pi_E)-J(\widehat{\pi}_h)\leq T^2\left(\frac{2}{\sqrt{\dim(A)}}\epsilon+2\delta \right),
\end{align*}
which is a concrete bound that extends the abstract bound in Theorem 5.4 of~\citet{Swamy2022}.
\end{corollary}

\begin{remark}
\emph{If both $u^\epsilon_t$ and $u^o_t$ are zero, we then recover the classic setting of IL without confounders~\citep{Ross2010}, and the imitation gap bound is $T^2\epsilon$, where $\epsilon$ is the optimisation error of the algorithm.}
\end{remark}

\section{Experimental Results}\label{sec-il:exps}

\begin{figure*}[t]
\begin{subfigure}[t]{1\textwidth}
\centering\includegraphics[width=0.5\textwidth]{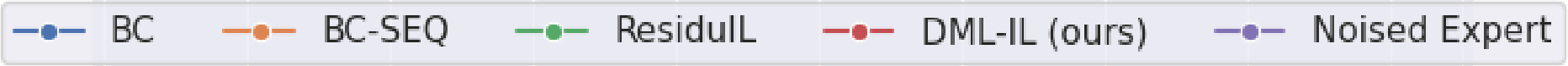}
\end{subfigure}
\centering
\begin{subfigure}[t]{0.49\textwidth}
\centering
\includegraphics[width=1\textwidth]{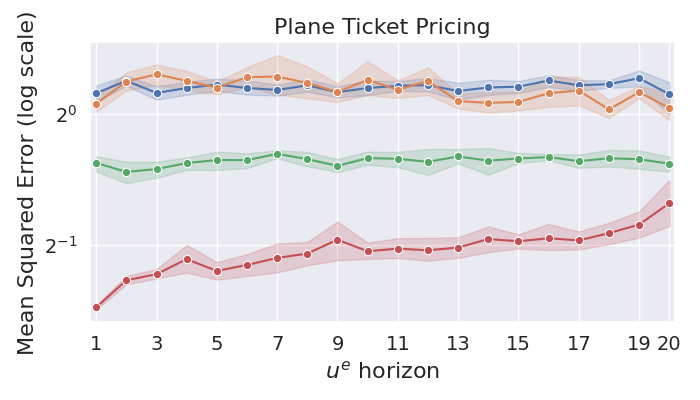}
\caption{MSE in log scale, lower is better.}
\end{subfigure}
\begin{subfigure}[t]{0.49\textwidth}
\centering
\includegraphics[width=1\textwidth]{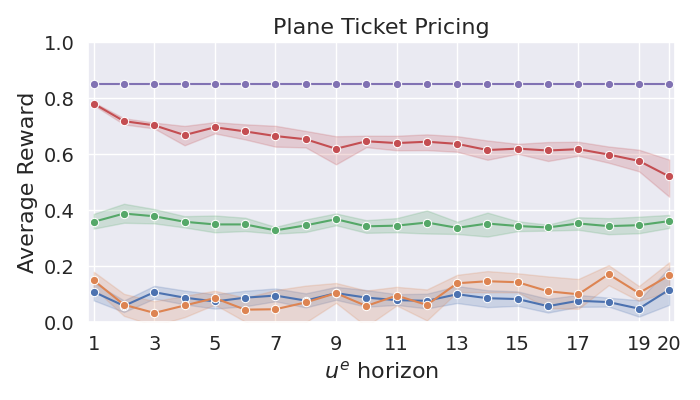}
\caption{Average reward, higher is better.}
\end{subfigure}
% \vspace{-3pt}
\caption[Experimental results for DML-IL in the plane ticket environment.]{\textbf{Plane Ticket Environment} (\cref{eg:plane}): On the left, the MSE between the learnt policy and the expert. On the right, the average reward of our approach and baselines.}
\label{fig:toy}
\end{figure*}

In this section, we empirically evaluate the performance of DML-IL (\cref{alg:DML-IL-kfold}) on the toy environment with continuous state and action spaces introduced in \cref{eg:plane} and Mujoco environments: Ant, Half Cheetah, and Hopper. We compare with the following existing methods reviewed in \cref{lit:il}: Behavioural Cloning (BC), which naively minimises $\expectE[-\log\pi(a_t\lvert s_t)]$; BC-SEQ \citep{Swamy2022}, which learns a history-dependent policy to handle hidden contexts observable to the expert; ResiduIL \citep{Swamy2022_temporal}, which we adapt to our setting by providing $h_{t-k}$ as instruments to learn a history-independent policy; 
%DML-IL is our proposed method described in~\cref{alg:DML-IL}; 
and the noised expert, which is the performance of the expert when put in the confounded environment, and corresponds to the maximally achievable performance. In \cref{sec-il:otheriv}, we also include additional evaluations of using other CMR estimation algorithms, including DFIV~\citep{Xu2020} and DeepGMM~\citep{Bennett2019DeepAnalysis}, as the core CMR estimator, but we found inconsistent and subpar performance.

\paragraph{Experimental Setup}

The experiments are carried out on a Linux server (Ubuntu 18.04.2) with two Intel Xeon Gold 6252 CPUs and six NVIDIA GeForce RTX 2080 Ti GPUs. We train imitators with 20000 samples (40 trajectories of 500 steps each) of the expert trajectory using each algorithm and report the average reward when tested online in their respective environments. \bill{Each training run on the toy environment takes around 20 minutes and a few hours on MuJoCo tasks.} The reward is scaled such that 1 is the performance of the un-noised expert, and 0 is a random policy. We also report the mean squared error (MSE) between imitator's and expert's actions. The purpose of evaluating the MSE is to assess how well the imitator learnt from the expert, and importantly whether the confounding noise problem is mitigated. When the confounding noise $u^\epsilon$ is not handled, we should expect to observe a much higher MSE. All results are plotted with one standard deviation as a shaded area. In addition, we vary the confounding noise horizon $k$ from 1 to 20 in order to increase the difficulty of the problem with weaker instruments $h_{t-k}$. Further training details, network architecture, and hyperparameters can be found in \cref{appendix:implement}. All algorithms are implemented using PyTorch~\citep{Paszke2019}, and the code is available on GitHub\footnote{\url{https://github.com/shaodaqian/Confounded-IL}}.

\begin{figure*}[t!]
\begin{subfigure}[t]{1\textwidth}
\centering\includegraphics[width=0.7\textwidth]{figures/confoundedIL/exp_results/legend.png}
\end{subfigure}
\begin{subfigure}[t]{0.49\textwidth}
\centering\captionsetup{width=0.9\linewidth}\includegraphics[width=1\textwidth]{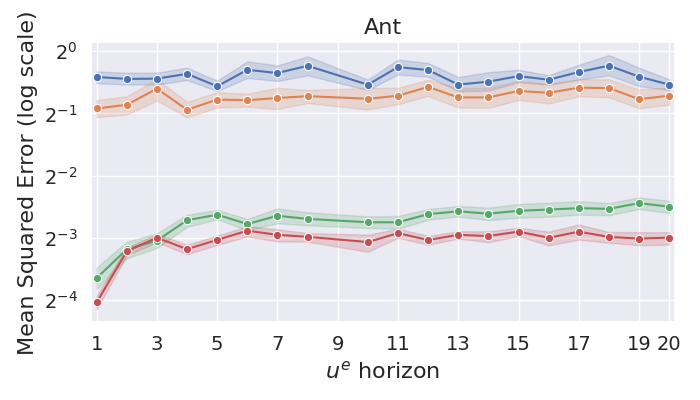}
\caption{MSE for Ant, lower is better.}
\end{subfigure}
\begin{subfigure}[t]{0.49\textwidth}
\centering\captionsetup{width=0.9\linewidth}\includegraphics[width=1\textwidth]{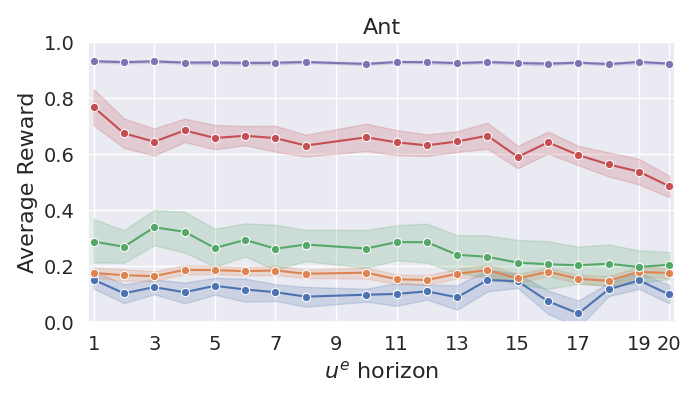}
\caption{Average reward for Ant, higher is better.}
    \label{fig:ant_rew}
\end{subfigure}

\begin{subfigure}[t]{0.49\textwidth}
\centering\captionsetup{width=0.9\linewidth}\includegraphics[width=1\textwidth]{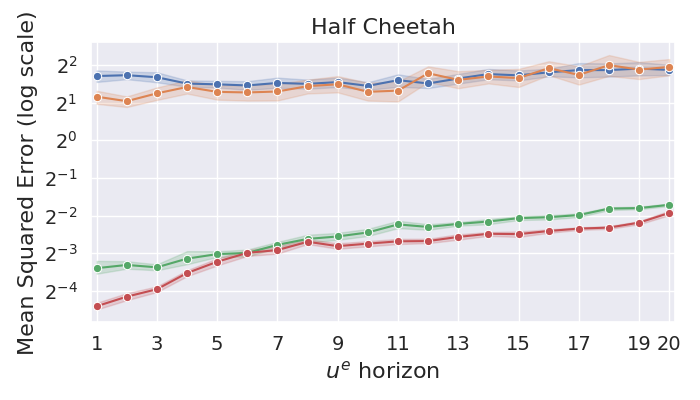}
\caption{MSE for Half Cheetah.}
\end{subfigure}
\begin{subfigure}[t]{0.49\textwidth}
\centering\captionsetup{width=0.9\linewidth}\includegraphics[width=1\textwidth]{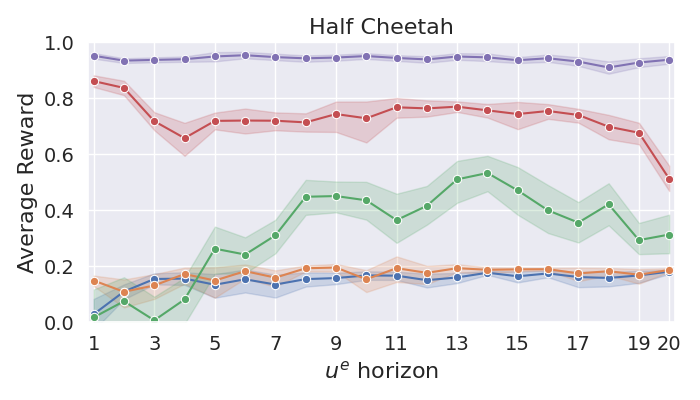}
\caption{Average reward for Half Cheetah.}
    \label{fig:hc_rew}
\end{subfigure}

\begin{subfigure}[t]{0.49\textwidth}
\centering\captionsetup{width=0.9\linewidth}\includegraphics[width=1\textwidth]{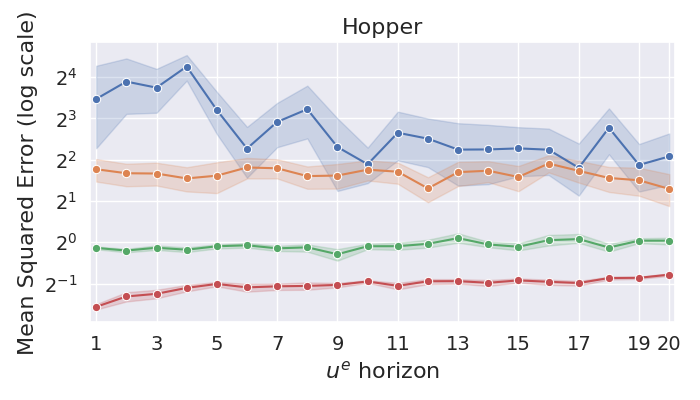}
\caption{MSE for Hopper.}
\end{subfigure}
% \vspace{.5cm}
\begin{subfigure}[t]{0.49\textwidth}
\centering\captionsetup{width=1\linewidth}\includegraphics[width=1\textwidth]{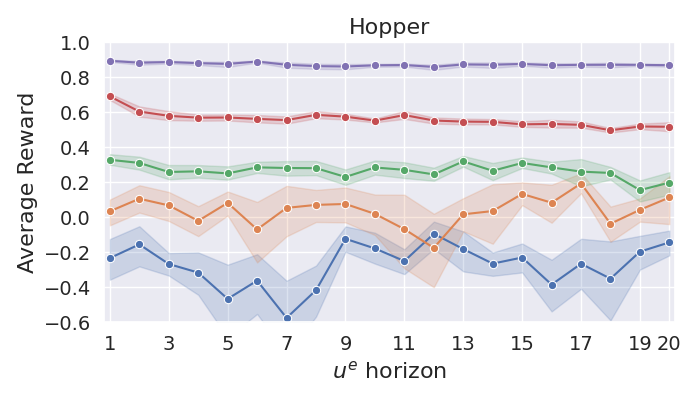}
\caption{Average reward for Hopper. Note that BC performs worse in Hopper than a random policy.}
\label{fig:hopper_rew}
\end{subfigure}
% \vspace{-3pt}
\caption[Experimental results for DML-IL in the MuJoCo environment.]{\textbf{MuJoCo:} On the left, the MSE between learnt policy and the expert. On the right, the average reward in the MuJoCo environments Ant, Half Cheetah and Hopper.}
\label{fig:gym}
\end{figure*}

\subsection{Ticket Pricing Environment}

We first consider the plane ticket pricing environment described in \cref{eg:plane}. The confounding noise $u^\epsilon$ are operational costs and $u^o_t$ are seasonal demand patterns and events. We set $u^o_t$ to continuously vary with a rate of change of approximately every 30 steps. Details on this environment are provided in~\cref{appendix:ticket}.

The results are presented in~\cref{fig:toy}. DML-IL performed the best with the lowest MSE and the highest average reward that is closest to the expert, especially when the $u^\epsilon_t$ horizon is 1. This implies that DML-IL is successful in handling both $u^\epsilon_t$ and $u^o_t$. ResiduIL is able to reduce the confounding effect of $u^\epsilon_t$, evident by the lower MSE compared to the two other methods that do not deal with $u^\epsilon_t$. However, since it does not explicitly consider $u^o_t$, the imitator has no information on $u^o_t$, and the best it can do is to assume some average value (or expectation) of $u^o_t$. Therefore, while ResiduIL still achieves some reward, its performance gap to DML-IL is due to the lack of consideration of $u^o_t$. Both BC and BC-SEQ fail completely in the presence of confounding noise $u^\epsilon_t$, with much higher MSE and average reward close to a random policy. From the similar performance of BC-SEQ and BC, we see that the use of trajectory history to infer $u^o_t$ is not helpful when the confounding noise is not handled explicitly. This demonstrates that considering the effect of $u^\epsilon_t$ and $u^o_t$ partially is insufficient to learn a well-performing imitator under the general setting.

In addition, as the confounding noise horizon $k$ increases (x-axis), the performance of DML-IL drops. This coincides with the fact that the instrument weakens and less information regarding $u^o_t$ can be captured from $h_{t-k}$ as $k$ increases. When $k=20$, it can be seen that the performance of DML-IL is close to that of ResiduIL, which does not consider the effect of $u^o_t$, because very limited information about the current expert-observable confounder $u^o_t$ can be inferred from states 20 steps prior.

\subsection{Mujoco Environments}

In~\cref{fig:gym}, we consider the Mujoco tasks. While the original tasks do not have hidden variables, we modify the environment to introduce $u^\epsilon_t$ and $u^o_t$. Specifically, instead of controlling the Ant, Half Cheetah, and Hopper, respectively, to travel as fast as possible, the goal is to control the agent to travel at a target speed that is varying throughout an episode. This target speed is $u^o_t$, which is observed by the expert but not recorded in the dataset. In addition, we add confounding noise $u^\epsilon_t$ to $s_t$ and $a_t$ to mimic the confounding noise such as wind. Additional details about the modifications made to the environments are provided in~\cref{appendix:mujoco}.

DML-IL outperforms other methods in all three Mujoco environments as shown in~\cref{fig:gym}. Similarly to the plane ticket environment, ResiduIL is effective in removing the confounding noise but fails to match the average reward of DML-IL due to the lack of knowledge of $u^o_t$. BC and BC-SEQ have much higher MSE and fail to learn meaningful policies. As $u^\epsilon_t$ horizon increases, the performance of DML-IL drops as expected due to weaker instruments and limited inferable information regarding $u^o_t$, especially for the Ant and Half Cheetah tasks.

% It can also be seen by the closer gap of MSE between DML-IL and ResiduIL compared to other methods that confounding noise $u^\epsilon_t$ fundamentally breaks the regression when not handled properly, and $u^o_t$ when not considered will yield an average performance by assuming an expectation of $u^o_t$.

\subsection{Adopting Other CMR Estimators}\label{sec-il:otheriv}

We also experimented with other CMR estimators that have been previously shown to be practical~\citep{Shao2024} for different tasks and with high-dimensional input. Specifically, we experimented with DFIV~\citep{Xu2020}, which is an iterative algorithm that integrates the training of two models that depend on each other, and DeepGMM~\citep{Bennett2019DeepAnalysis}, which solves a minimax game by optimising two models adversarially. Note that DeepIV~\citep{Hartford2017DeepPrediction} can be considered a special case of DML-CMR, so we did not evaluate it.

The additional results for using DFIV and DeepGMM as the CMR estimator are provided in~\cref{fig:additional_toy} and~\cref{fig:additional_ant}. It can be seen from~\cref{fig:additional_toy} that only DFIV achieves good performance in the airline ticket pricing environment, surpassing the performance of ResiduIL. For the Ant Mujoco task in~\cref{fig:additional_ant}, both DFIV and DeepGMM fail to learn good policies, with only slightly lower MSE than BC and BC-SEQ. We conjecture that this is mainly due to the high-dimensional state and action spaces and the inherent instability in the DFIV and DeepGMM algorithms. For DFIV, the interleaving of training of two models causes highly nonstationary training targets for both models, and, for DeepGMM, the adversarial training procedure of two models is similar to that of generative adversarial networks (GANs), which are known to be unstable and difficult to train. In addition, when the CMR problem is weakly identifiable, as in the case of a weak instrument, the algorithms may converge to local minima that are far from the true solution in the face of instabilities in the algorithm.

We conclude that solving the CMR problem for an imitator policy can be sensitive to the choice of CMR estimator as well as to the choice of hyperparameters. In addition, some CMR estimators do not work well with high-dimensional input. Our CMR estimator of choice, DML-CMR, provides a robust base for the DML-IL algorithm that demonstrates good performance across all tasks and environments. This demonstrates the benefit of using double machine learning (recall \cref{chapter:dmliv}), which can debias two-stage estimators and provide good empirical and theoretical convergence.

\begin{figure*}[t]
\begin{subfigure}[c]{0.44\textwidth}
\centering\includegraphics[width=1\textwidth]{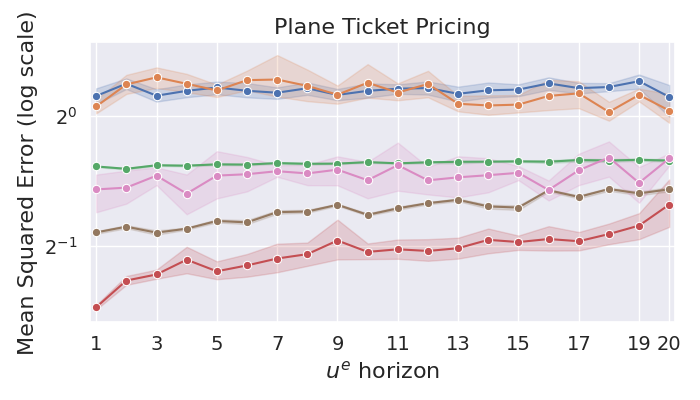}
\end{subfigure}
\centering
\begin{subfigure}[c]{0.44\textwidth}
\centering
\includegraphics[width=1\textwidth]{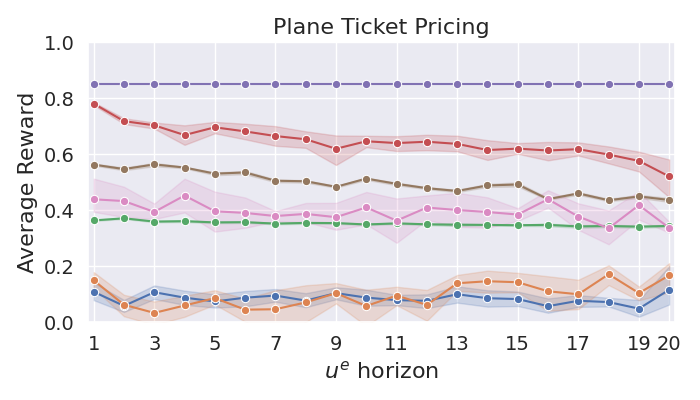}
\end{subfigure}
\begin{subfigure}[c]{0.1\textwidth}
\centering
\includegraphics[width=1\textwidth]{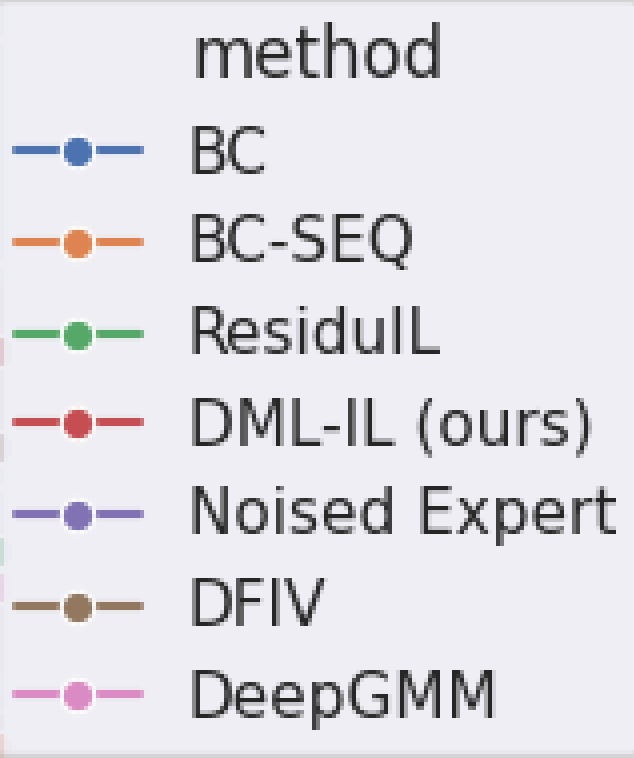}
\end{subfigure}
% \vspace{-8pt}
\caption[Confounded IL using other CMR estimators in the plane ticket environment.]{\textbf{Plane Ticket Environment}: the MSE between learnt policy and expert, and the average reward, with DFIV and DeepGMM as the CMR estimator.}
\label{fig:additional_toy}
\end{figure*}

\begin{figure*}[t]
\begin{subfigure}[c]{0.44\textwidth}
\centering\includegraphics[width=1\textwidth]{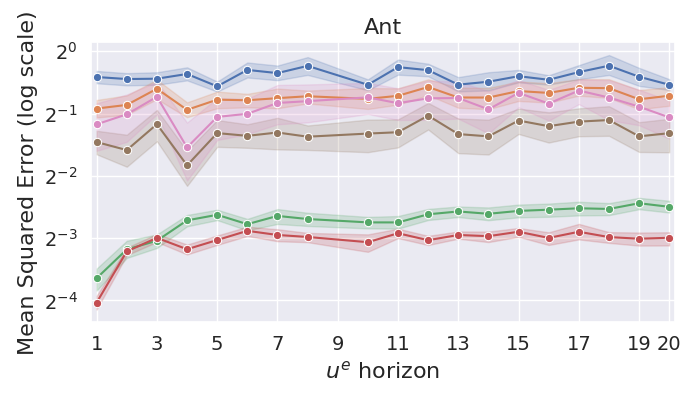}
\end{subfigure}
\centering
\begin{subfigure}[c]{0.44\textwidth}
\centering
\includegraphics[width=1\textwidth]{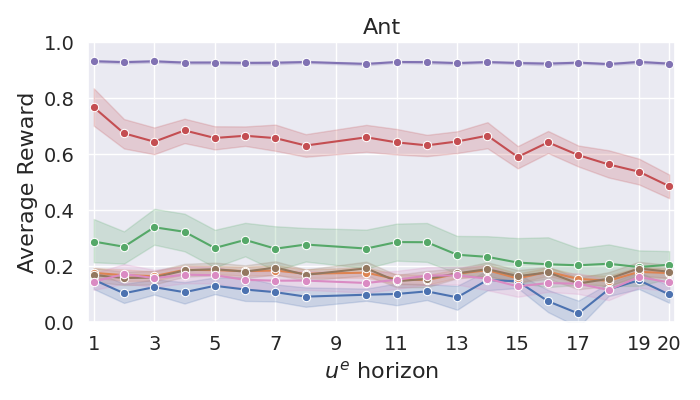}
\end{subfigure}
\begin{subfigure}[c]{0.1\textwidth}
\centering
\includegraphics[width=1\textwidth]{figures/confoundedIL/appendix/additional_legend.png}
\end{subfigure}
% \vspace{-8pt}
\caption[Confounded IL using other CMR estimators in the MuJoCo environment.]{\textbf{MuJoCo}: the MSE between learnt policy and expert, and the average reward, with DFIV and DeepGMM as the CMR estimator.}
\label{fig:additional_ant}
\end{figure*}

\section{Discussion}

In this chapter, we considered the problem of IL with hidden confounders that do not assume explicit IVs are observed in the dataset in addition to the trajectory histories. We proposed a unifying framework for confounded IL that unifies confounded IL settings. Specifically, we considered hidden confounders to be partially observable to the expert and demonstrated that causal IL under this framework can be reduced to a CMR problem with the trajectory histories as instruments. We proposed DML-IL, a novel algorithm that solves this CMR problem to learn an imitator policy, and provided bounds on the imitation gap for the learnt imitator. Finally, we empirically evaluated DML-IL on multiple tasks, including Mujoco environments, and demonstrated state-of-the-art performance against other causal IL algorithms.

One limitation of our proposed framework is the explicit assumptions made in~\cref{sec-il:setting}, which are essential for the expert policy to be identifiable. The first assumption is the confounding noise horizon, but, as discussed in~\cref{sec-il:setting}, it often holds in practice if the effect of the confounding noise on future states and actions diminishes over time or the confounding noise becomes observable at a future time. The second assumption is additive noise, which is a standard assumption in causal inference~\citep{Pearl2000causality,Balke1994} for the causal effect to be identifiable. This assumption holds for confounding noises such as wind noise or any environmental noise in robotics, and supply costs in econometrics settings such as in our ticket pricing scenario. In addition, we assume knowledge of the confounding noise horizon $k$ or an upper bound on it for~\cref{alg:DML-IL} such that the trajectory history $h_{t-k}$ is a valid IV. However, the value of $k$ or the validity of $h_{t-k}$ as an IV generally cannot be verified empirically, but there exist methods that, under certain assumptions, can identify good candidates for IVs (e.g., \citealt{Yuan2021,WuAnpeng2023} Section 6). For DML-IL, we rely on a reasonable choice of $k$ by the user based on the environment, but it would be interesting future work to incorporate methods that test the validity of IVs with increasing $k$ until a good candidate $h_{t-k}$ of IV is found.
\chapter{Learning Policies for High-Level Objectives}\label{chapter:ltl}

\minitoc

\section{Introduction}

Thus far, we have focused on learning optimal and sample-efficient policies that maximise expected reward. In this chapter, we focus on learning policies that satisfy high-level objectives optimally and sample-efficiently using RL. Specifically, we consider the use of linear temporal logic (LTL), which is a temporal logic language that can encode formulae regarding the properties of an infinite sequence of logic propositions, to specify high-level objectives for a system or an agent in the system (recall \cref{prelim:high-level}). Such systems are modelled as Markov decision processes (MDPs), where policies need to be learnt through interactions with the MDP if the states and transitions of the MDP are not known \textit{a priori}.

% LTL is widely used for the formal specification of high-level objectives for robotics and multi-agent systems, and it is desirable for a system or an agent in the system to learn policies with respect to these high-level specifications.
% classic policy synthesis techniques can be adopted if the states and transitions of the MDP are known. However, when the transitions are not known \textit{a priori}, the optimal policy needs to be learnt through interactions with the MDP.

Reinforcement learning (recall~\cref{lit:online_rl}), a powerful method for policy learning that maximises rewards over time in an unknown environment, is a perfect candidate for learning policies that satisfy LTL specifications. However, as discussed in~\cref{chapter:intro}, using RL to learn a policy that maximises the probability of satisfying LTL specifications is nontrivial~\citep{Alur2022}. Most existing work first transforms the LTL specification into an automaton, then builds a product MDP using the original environment MDP and the automaton, on which model-free RL algorithms are applied. However, a crucial obstacle still remains, that is, how to properly define the reward that leads an agent to the optimal satisfaction of the LTL specification with theoretical guarantees. Several algorithms have been proposed~\citep{Hahn2019Omega-regularLearning,Bozkurt2019ControlLearning,Hasanbeig2019ReinforcementGuarantees} for learning LTL specifications with optimality guarantees. However, as discussed in \cref{lit:summary}, they face challenges in determining the key hyperparameters that ensure optimality, which in practice can lead to suboptimal policies or inefficient learning. In addition, it remains unclear how to explicitly select these hyperparameters even with certain knowledge regarding the environment MDP. Furthermore, their proposed reward functions are often inflexible, making them ineffective at encouraging agent exploration, which leads to poor sample efficiency.

% the assumed hyperparameters are evaluated in experiments indirectly, either through the inspection of the value function or through comparisons of the expected reward achieved, making it difficult to tune the hyperparameters for LTL learning in general MDPs.

In this chapter, we propose a novel and more general product MDP with a flexible reward structure and discounting mechanism that, by leveraging model-free reinforcement learning algorithms, efficiently learns the optimal policy that maximises the probability of satisfying the LTL specification with guarantees. We demonstrate improved theoretical results on the optimality of our product MDP and the reward structure, with a more stringent analysis that yields better bounds on the optimality hyperparameters. Moreover, this analysis sheds light on how to explicitly choose the optimality hyperparameters based on the environment MDP. We also adopt counterfactual imagining that takes advantage of the known high-level LTL specification to further improve the performance of our algorithm. Last but not least, we propose to use the PRISM model checker~\citep{Kwiatkowska2011PRISMSystems} to directly evaluate the satisfaction probability of the learnt policies, providing a platform to compare algorithms and tune key hyperparameters. We conduct experiments on several common MDP environments with various challenging LTL tasks and demonstrate the improved sample efficiency and convergence of our methods.

\paragraph{Main Contributions of this Chapter.} 
\begin{itemize}[leftmargin=10pt, topsep=2pt, itemsep=2pt, topsep=2pt]
    \item We propose a novel product MDP that incorporates an accepting states counter with a generalised reward structure (\cref{sec-ltl:product}).
    \item We propose a novel reinforcement learning algorithm that converges to the optimal policy for satisfying LTL specifications (\cref{sec-ltl:q-learning}), with theoretical optimality guarantees and theoretical analysis results for choosing key hyperparameters (\cref{thm:optimality}).
    \item We propose counterfactual imagining (\cref{sec-ltl:imagining}), a method to take advantage of the known structure of the LTL specification by creating imagination experiences through counterfactual reasoning.
    \item We provide a direct evaluation of the proposed algorithms through a novel integration of probabilistic model checkers within the evaluation pipeline, with strong empirical results that demonstrate better sample efficiency and training convergence (\cref{sec-ltl:exps}).
\end{itemize}

Work reported in this chapter first appeared in \citet{Shao2023SampleGuarantees}.

\section{Problem Formulation}\label{sec-ltl:problem}

Our goal is to formulate a model-free reinforcement learning approach (recall \cref{chapter:background}) to efficiently learn an optimal policy that maximises the probability of satisfying an LTL specification with guarantees.

Given an MDP $\MDP=(\states,s_0,\actions,\transitions,\propositions,\labFunc,\reward,\gamma)$ with unknown states and transitions and an LTL objective $\LTL$, for any policy $\policy$ of the MDP $\MDP$, let $\probP_\policy(\state\models\LTL)$ denote the probability of paths from state $\state$ following $\policy$ that satisfy the LTL formula $\LTL$:
\begin{equation*}
\probP_\policy(\state\models\LTL)=\probP_\policy\{\MDPpath\in\paths_\policy\mid\MDPpath[0]=s,\MDPpath\models\LTL\}.
\end{equation*}
Then, we would like to design a model-free RL algorithm that learns a deterministic memoryless optimal policy $\policy_\LTL$ that maximises the probability of $\MDP$ satisfying $\LTL$:
\begin{equation*}
    \probP_{\policy_\LTL}(\state\models\LTL)=\max_\policy\probP_\policy(\state\models\LTL) \quad \forall\state\in\states.
\end{equation*}

\section{$\omega$-Automata and Limit-Deterministic Büchi Automata}\label{sec-ltl:automata}

We first transform the LTL specifications into automata. The most common type of automaton is the finite automaton, which is mathematically equivalent to regular expressions and accepts only finite sequences. However, LTL is strictly more expressive than regular expressions because it can describe properties over infinite sequences, which regular expressions cannot. Fortunately, there exists a more expressive class of automata, called $\omega$-automata, that can be used to represent LTL specifications. Unlike finite automata, $\omega$-automata can process infinite sequences and use specific accepting conditions to determine whether an infinite sequence is accepted. The most common $\omega$-automata include Rabin automata~\citep{Rabin1969} and Büchi automata~\citep{Buchi1966}, which we will introduce next.

\begin{definition}[DRA~\citep{Rabin1969}]
A deterministic Rabin automaton (DRA) is defined as a tuple: $\automaton=(\propositions,\autoStates,\autoState_0,\autoTransitions,\autoAccept)$, where $\propositions$ is the set of atomic propositions, $\autoStates$ is a finite set of states, $\autoState_0\in\autoStates$ is the initial state, and
$\autoAccept=\{(F_i,G_i)\}_{i=1}^k$ is the Rabin acceptance condition, where for each $i$, $F_i\subseteq\autoStates$ is a set of states that must be visited infinitely often, and $G_i\subseteq\autoStates$ is a set of states that must only be visited finitely often. Let $\Sigma=2^\propositions$ be a finite alphabet, and the deterministic transition function is given by $\autoTransitions:\autoStates\times\Sigma\rightarrow\autoStates$.
\end{definition}

An infinite word $\word\in\Sigma^\omega$, where $\Sigma^\omega$ is the set of all infinite words over the alphabet $\Sigma$, is accepted by a DRA $\automaton$ if and only if there exists an infinite automaton run (sequence of states) $\Autorun\in\autoStates^\omega$ from $\autoState_0$, where $\Autorun[t+1]=\autoTransitions(\Autorun[t],\word[t]), \forall t\geq 0$, such that there exists at least one pair $(F_i,G_i)$ where some state in $F_i$ appears infinitely often and all states in $G_i$ appear only finitely often.

\begin{definition}[NBA~\citep{Buchi1966}]\label{defn:NBA}
A nondeterministic Büchi automaton (NBA), or simply a Büchi automaton, is defined as a tuple $\automaton=(\propositions,\autoStates,\autoState_0,\autoTransitions,\autoAccept)$, where $\propositions$ is the set of atomic propositions, $\autoStates$ is a finite set of states, $\autoState_0\in\autoStates$ is the initial state, and $\autoAccept\subseteq\autoStates$ is the set of accepting states that must be visited infinitely often. Let $\Sigma=2^\propositions\cup\{\epsilon\}$ be a finite alphabet, where $\epsilon$ handles the nondeterministic transitions. The nondeterministic transition function is given by $\autoTransitions:\autoStates\times\Sigma\rightarrow 2^{\autoStates}$, which allows the automaton to transition to multiple states due to the inherent nondeterminism.
\end{definition}

An infinite word $\word\in\Sigma^\omega$, where $\Sigma^\omega$ is the set of all infinite words over the alphabet $\Sigma$, is accepted by a Büchi automaton $\automaton$ if and only if there exists an infinite automaton run $\Autorun\in\autoStates^\omega$ from $\autoState_0$, where $\Autorun[t+1]\in\autoTransitions(\Autorun[t],\word[t]), \forall t\geq 0$, such that $\text{inf}(\Autorun)\cap\autoAccept\neq\varnothing$, where $\text{inf}(\Autorun)$ is the set of automaton states that are visited infinitely often in the run $\Autorun$. This means that for an accepting infinite word $\word$, by following the transition function of the NBA, there must exist an infinite automaton run in which at least one of the accepting states is visited infinitely often. Since NBA is nondeterministic, an infinite word $\word$ can produce many infinite automaton runs following the transition function. However, the accepting condition only requires the existence of one such automaton run to visit some accepting states infinitely often.

Both DRA and NBA can be used to translate from LTL specifications, but they each present challenges when applied in reinforcement learning contexts. For DRA, the accepting condition is notably complex, encompassing a set of states that must be visited infinitely often alongside another set of states that must be visited only finitely often. This makes it difficult, if not impossible~\citep{Hahn2022}, to translate the Rabin accepting condition into a reward that the RL agent maximises. For NBA, the accepting condition is simple, but the nondeterminism can significantly increase branching in the search space when identifying an accepting run. This leads to inefficiencies when used to build a product MDP for RL.

In this chapter, we adopt a variant of Büchi automata known as limit-deterministic Büchi automata (LDBA)~\citep{Sickert2016Limit-deterministicLogic}, which has the same accepting condition as a standard Büchi automaton but with limited nondeterminism. Specifically, through the construction of an LDBA, the nondeterminism is restricted to an initial component $\autoStates_\mathcal{N}$ of the automaton, and after reaching any accepting state, the automaton becomes deterministic.

\begin{definition}[LDBA]
A Büchi automaton is limit-deterministic if $\autoStates$ can be partitioned into a deterministic set and a nondeterministic set, that is, $\autoStates = \autoStates_\mathcal{N} \cup \autoStates_\mathcal{D}$, where $\autoStates_\mathcal{N} \cap \autoStates_\mathcal{D}=\varnothing$, such that

\begin{enumerate}
    \item $\autoAccept\subseteq\autoStates_\mathcal{D}$ and $\autoState_0\in\autoStates_\mathcal{N}$;
    \item $\abs{\autoTransitions(\autoState,\alphabet)}\leq1$ for all $\autoState\in\autoStates_\mathcal{N}$ and $\alphabet\neq\epsilon\in\alphabets$;
    \item $\autoTransitions(\autoState,\alphabet)\subseteq\autoStates_\mathcal{D}$ and $\abs{\autoTransitions(\autoState,\alphabet)}\leq1$ for all $\autoState\in\autoStates_\mathcal{D}$ and $\alphabet\in\alphabets$;
\end{enumerate}
\end{definition}

An LDBA starts in a nondeterministic initial component and then transitions into a deterministic accepting component $\autoStates_\mathcal{D}$ through $\epsilon$-transitions after reaching an accepting state, where all transitions after this point are deterministic. We follow the formulation of \citet{Bozkurt2019ControlLearning} to extend the alphabets with $\Sigma$ an $\epsilon$-transition as defined in~\cref{defn:NBA} that handles all nondeterminism, which means only $\epsilon$ can transition the automaton state to more than one state: $\abs{\autoTransitions(\autoState,\epsilon)}>1$. This allows the MDP to synchronise with the automaton, which we will discuss in detail in \cref{sec:product_MDP}.

LDBAs are as expressive as the LTL language, and the satisfaction of any given LTL specification $\LTL$ can be evaluated on the LDBA derived from $\LTL$. We use Rabinizer 4~\citep{Kretinsky2018RabinizerAutomaton} to transform LTL formulae into LDBAs. In \cref{fig:example_task} (left) we present an example of the LDBA derived from the LTL formula \enquote{\textsf{\upshape F}\textsf{\upshape G }a \& \textsf{\upshape G }!c}, where state 1 is the accepting state. LDBAs differ from reward machines~\citep{Icarte2022RewardLearning} in that they can express properties satisfiable by infinite paths, which are strictly more expressive than reward machines, and they have different accepting conditions (see \cref{prelim:ltl} for details).

\begin{figure*}[tb]
\begin{subfigure}[t]{0.37\textwidth}
    \centering
        \includegraphics[width=0.52\textwidth]{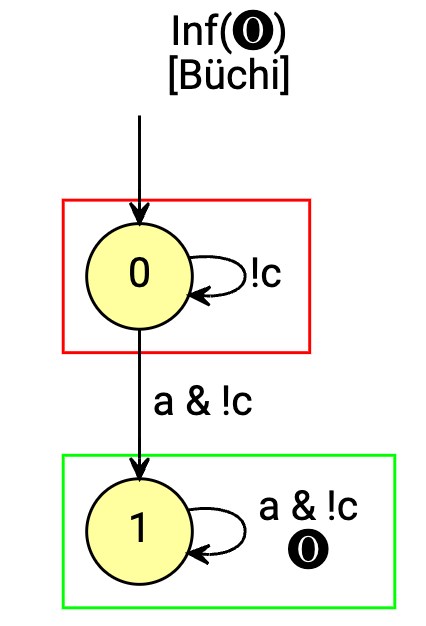}
    % \caption{LDBA for \enquote{\textsf{\upshape F}\textsf{\upshape G} a \& \textsf{\upshape G} !c}.}
\end{subfigure}
\begin{subfigure}[t]{0.62\textwidth}
    \centering
    \includegraphics[width=0.83\textwidth]{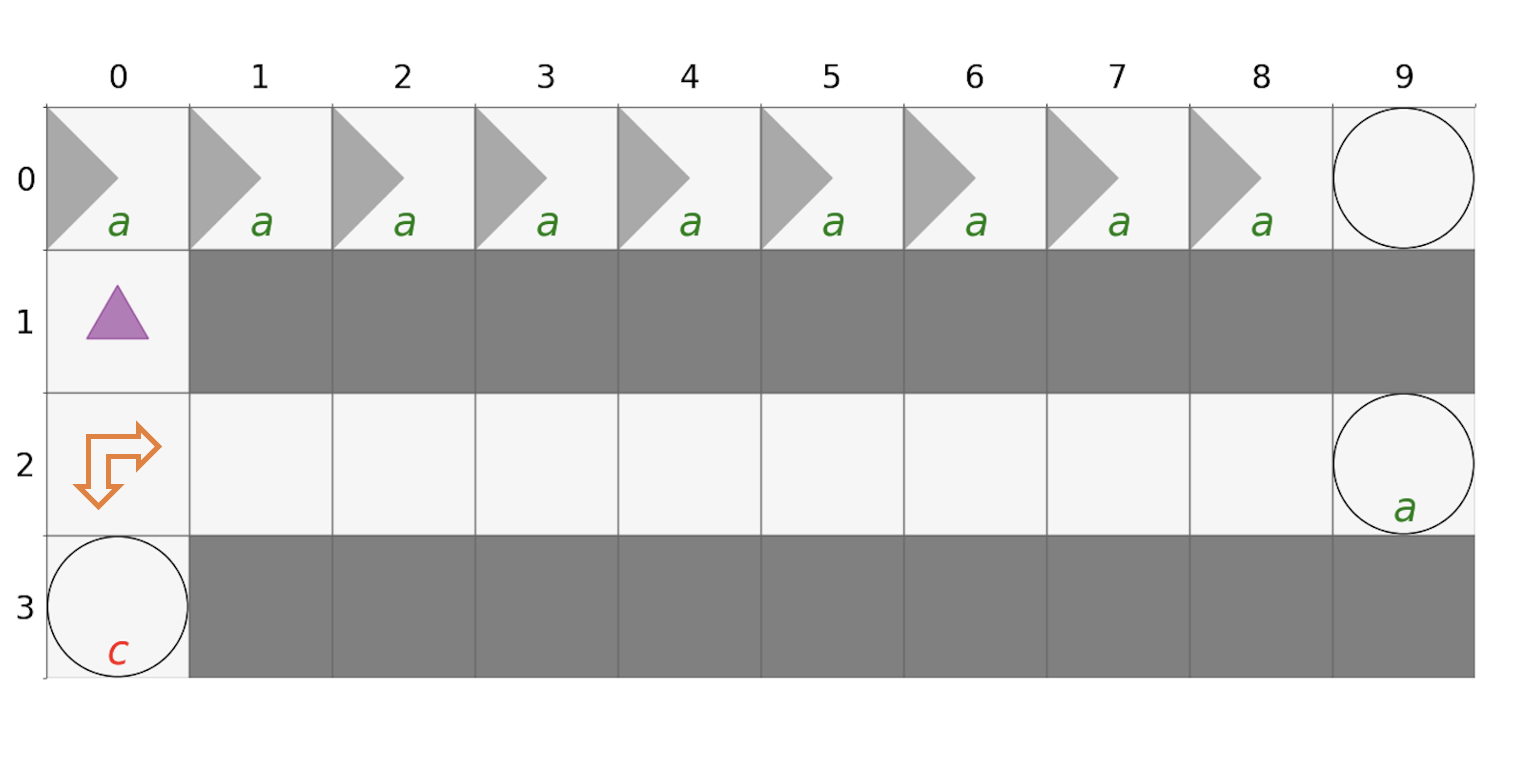}
    % \caption{A motivating MDP example for the $K$ counter, where the purple triangle at (1,0) is the starting point and the bidirectional arrow at (2,0) is a probability gate.}
\end{subfigure}
\caption[An LDBA for \enquote{\textsf{\upshape F}\textsf{\upshape G} a \& \textsf{\upshape G} !c}]{An LDBA for \enquote{\textsf{\upshape F}\textsf{\upshape G} a \& \textsf{\upshape G} !c} (left), and a probabilistic gate MDP (right) motivating the $K$ counter (see \cref{example_mdp}).}
\label{fig:example_task}
\end{figure*}

\section{Methodology}\label{sec-ltl:method}

We now provide an overview of our method. Firstly, we transform the LTL objective $\LTL$ into a limit-deterministic Büchi automaton. Then, we introduce a novel product MDP in \cref{sec-ltl:product} and define a generalised reward structure over it, which serves as a general framework that can be instantiated into specific reward functions. With this reward structure, we propose a Q-learning algorithm that adopts a collapsed Q function to learn the optimal policy with optimality guarantees. Lastly, we enhance our algorithm with counterfactual imagining in \cref{sec-ltl:imagining}, which takes advantage of the automaton structure to improve performance while maintaining optimality.

\subsection{Product MDP}\label{sec-ltl:product}
\label{sec:product_MDP}
In this section, we propose a novel product MDP of the environment MDP, an LDBA, and an integer counter, where the transitions for each component are synchronised. Contrary to the standard product MDP used in the literature~\citep{Bozkurt2019ControlLearning,Hahn2019Omega-regularLearning,Hasanbeig2020DeepLogics}, this novel product MDP incorporates a counter that counts the number of accepting states visited by paths starting at the initial state.

\begin{definition}[Product MDP]
\label{def:productMDP}
Given an MDP $\MDP=(\states,\state_0,\actions,\transitions,\propositions,\labFunc,\reward,\discount)$, an LDBA $\automaton=(\propositions,\autoStates,\autoState_0,\autoTransitions,\autoAccept)$ and $K\in\naturalNumber$, we construct the product MDP as follows: 
$$
\productMDP=\MDP\times\automaton\times[0..K]=(\productStates,\state_0^\times,\productActions,\productTransitions,\productAccept,\productReward,\productDiscount), 
$$
where the product states $\productStates=\states\times\autoStates\times[0..K]$, the initial state $\state_0^\times=(\state_0,\autoState_0,0)$, the product actions $\productActions=\actions\cup\{\epsilon_\autoState\mid\autoState\in\autoStates\}$, the accepting set $\productAccept=\states\times\autoAccept\times[0..K]$, and the product transitions $\productTransitions:\productStates\times\productActions\times\productStates \rightarrow [0,1]$, which are defined as:
\begin{align}
\productTransitions((\state,\autoState,n),\action,(\hat{\state},\hat{\autoState},n))
&=\begin{cases}\transitions(\state,\action,\hat{\state}) &\text{ if } \action\in\actions\text{ and } \hat{\autoState}\in\autoTransitions(\autoState,\labFunc(\state))\setminus\autoAccept;\\
    1 &\text{ if }\action=\epsilon_{\hat{\autoState}}, \hat{\autoState}\in\autoTransitions(\autoState,\epsilon)\text{ and }\hat{\state}=\state;\\
    0 &\text{ otherwise, }
    \end{cases}\nonumber\\
    \productTransitions((\state,\autoState,n),\action,(\hat{\state},\hat{\autoState},\min(n+1,K)))
    &=
    \begin{cases}\transitions(\state,\action,\hat{\state})&\text{ if }\action\in\actions\text{ and } \hat{\autoState}\in\autoTransitions(\autoState,\labFunc(\state))\cap\autoAccept;\\
    0 &\text{ otherwise, }
    \end{cases}\nonumber
\end{align}
where all other transitions are equal to 0. The product reward $\productReward:\productStates\times\productActions\times\productStates\rightarrow\realNumber$ and the product discount function $\productDiscount:\productStates\rightarrow(0,1]$ that are suitable for LTL learning are defined later in \cref{reward_defn}.

Furthermore, an infinite path $\MDPpath$ of $\productMDP$ satisfies the Büchi condition $\LTL_\autoAccept$ if $\textnormal{ inf}(\MDPpath)\cap\productAccept\neq\varnothing$. With a slight abuse of notation, we denote this condition in LTL language as $\MDPpath\models\textsf{\upshape G}\textsf{\upshape F }\LTL_\autoAccept$, meaning that, for all $M\in\naturalNumber$, there always exists $\productState\in\productAccept$ that will be visited in $\MDPpath[M:]$.
\end{definition}

When an MDP action $\action\in\actions$ is taken in the product MDP $\productMDP$, the alphabet used to transition the LDBA is deduced by applying the label function to the current state of the environment MDP: $\labFunc(\state)\in2^{\propositions}$. In this case, the LDBA transition $\delta(\autoState, \labFunc(\state))$ is deterministic. Otherwise, if an $\epsilon$-action $\epsilon_{\hat{\autoState}}\in\{\epsilon_\autoState\mid\autoState\in\autoStates\}$ is taken, LDBA is transitioned with an $\epsilon$-transition, and the nondeterminism of $\delta(q, \epsilon)$ is resolved by transitioning the automaton state to $\hat{q}$. The $K$ counter value is initially set to 0, and each time an accepting state is reached, the counter value increases by one until it is capped at $K$.

\begin{example}
\label{example_mdp}
We motivate our product MDP structure of \cref{def:productMDP} through an example. In \cref{fig:example_task} (right), we have a grid environment where the agent can decide to move up, down, left, or right. The task is to visit states labelled \enquote{a} infinitely often without visiting \enquote{c} as described by the LDBA in \cref{fig:example_task} (left). The MDP starts at (1,0), with walls denoted by solid grey squares. The states in the first row only allow action right, as denoted by the right-pointing triangle, which leads to a sink at (0,9). There is also a probabilistic gate at (2,0) that randomly transitions the agent down or right, and if the agent reaches (2,1), the accepting sink at (2,9) becomes reachable. Therefore, the optimal policy is to move down from the start and stay in (2,9) if the probabilistic gate at (2,0) transitions the agent to the right. The probability of satisfying this task is the probability of the gate sending the agent to the right. Intuitively, this environment has some initial accepting states that are easy to explore but lead to non-accepting sinks, whereas the true optimal path requires more exploration. If we set $K=10$ in the product MDP for this task, we can assign very small rewards to the initially visited accepting states and gradually increase the reward as more accepting states are visited to encourage exploration and guide the agent towards the optimal policy.
\end{example}

Next, we present a theorem which states that the product MDP with Büchi condition $\LTL_\autoAccept$ is equivalent, in terms of the optimal policy, to the original MDP with LTL specification $\LTL$. The proof of this theorem is provided in~\cref{proof:equivalence}.

\begin{theorem}[Satisfiability Equivalence]
For any product MDP $\productMDP$ that is induced from LTL formula $\LTL$, we have that
\begin{equation*}
\sup_\policy{\probP_\policy(\state_0\models\LTL)}=\sup_{\policy^\times}{\probP_{\policy^\times}(\productState_0\models\textsf{\upshape G}\textsf{\upshape F }\LTL_\autoAccept)}.
\end{equation*}
Furthermore, a deterministic memoryless policy that maximises the probability of satisfying the Büchi condition $\LTL_\autoAccept$ on the product MDP $\productMDP$, starting from the initial state, induces a deterministic finite-memory optimal policy that maximises the probability of satisfying $\LTL$ on the original MDP $\MDP$ from the initial state.
\label{thm:equivalence}
\end{theorem}

\subsubsection{Reward Structure for LTL Learning}
We first define a generalised reward structure in $\productMDP$ for learning policies that satisfy LTL specifications, and then prove the equivalence between acquiring the highest expected discounted reward and achieving the highest probability of satisfying $\LTL_\autoAccept$ using this reward structure.

\begin{definition}[Reward Structure]
\label{reward_defn}
Given a product MDP $\productMDP$ and a policy $\policy$, the product reward function $\productReward:\productStates\times\productActions\times\productStates\rightarrow\realNumber$ is suitable for LTL learning if 
\begin{align*}
\productReward((\state,\autoState,n),\productAction,(\state^\prime,\autoState^\prime,n^\prime))=
    \begin{cases} 
    R_n &\text{ if }\autoState^\prime\in\autoAccept;\\
    0  &\text{ otherwise,}
    \end{cases}
\end{align*}
where $R_n\in(0,U]$ are constants for $n\in[0..K]$ and $U\in(0,1]$ is an upper bound on the rewards. The rewards are nonzero only for accepting automaton states and depend on the value of the $K$ counter.

Then, given a discount factor $\discount\in(0,1]$, we define the product discount function $\productDiscount: \productStates\rightarrow(0,1]$ as $$\productDiscount(\productState_j)=\begin{cases}
1-\productReward_{j} &\text{ if } \productReward_{j}>0;\\
\discount &\text{ otherwise,}
\end{cases}$$
and the expected discounted reward following policy $\policy$ starting at $\productState$ and time step $t$ is
\begin{align*}
\ExpReward^\policy_t(\productState)=\expectE_\policy[\sum\limits^\infty_{i=t}(\prod\limits_{j=t}^{i-1}\productDiscount(\productState_j))\cdot\productReward(\productState_{i},\productAction_{i},\productState_{i+1})\mid \productState_t=\productState].
\end{align*}
\end{definition}

The highest $K$ value reached in a path, that is, the number of accepting states visited in the path, acts as a measure of how promising that path is for satisfying $\LTL_\autoAccept$. Depending on the current $K$ value, we can assign varying rewards to accepting states to guide the agent, as discussed in the motivating example in \cref{sec:product_MDP}.

Next, we provide an optimality guarantee that the policy which maximises the expected discounted reward also maximises the probability of satisfying the Büchi condition, and thus the LTL specification. To achieve this, we assume that the state and action spaces of the MDP are discrete and finite.

\begin{assumption}[Finiteness of MDP]\label{sec-ltl:finite}
For the environment MDP $\MDP=(\states,\state_0,\actions,\transitions,\propositions,\labFunc,\reward,\discount)$, we assume that $\states$ is a finite set of states and $\actions$ is a finite set of actions.
\end{assumption}

This assumption is common in the literature~\citep{Bozkurt2019ControlLearning,Hahn2019Omega-regularLearning,Hasanbeig2019ReinforcementGuarantees} of learning LTL objectives with RL to obtain optimality guarantees. It is necessary because we need to analyse the bottom strongly connected components (BSCCs, see~\cref{prelim:mdp}) of the product MDP and the rewards accumulated within them. If we have continuous state and action spaces, it may take an infinite number of steps to reach a BSCC, and no guarantees regarding the behaviour of infinite paths (i.e., LTL satisfaction) can be made based on finite trajectories explored by RL agents. It is an open problem to provide guarantees for learning LTL objectives in continuous state and action spaces. For more discussion, see~\cref{sec-ltl:discussion} and \cref{conclusion:discussion}.

Next, we provide a lemma stating the properties of the product MDP regarding the satisfaction of the Büchi condition $\LTL_\autoAccept$.

\begin{lemma}
\label{useful_lemma}
Given a product MDP $\productMDP$ with its corresponding LTL formula $\LTL$ and a policy $\policy$, we write $\productMDP_\policy$ for the induced Markov chain from $\policy$. Let $B_\productAccept$ denote the set of states that belong to accepting BSCCs of $\productMDP_\policy$, and $B^\times_\varnothing$ denote the set of states that belong to rejecting BSCCs:
\begin{align*}
&B_\productAccept \coloneqq \{\productState \mid \productState\in B \in BSCC(\productMDP_\policy),B\cap\productAccept\neq\varnothing\};\\
&B^\times_\varnothing \coloneqq \{\productState \mid \productState\in B \in BSCC(\productMDP_\policy),B\cap\productAccept=\varnothing\},
\end{align*}
where $BSCC(\productMDP_\policy)$ is the set of all BSCCs of $\productMDP_\policy$. We further define more general accepting and rejecting sets:
\begin{align}
    B_\autoAccept &\coloneqq \{(\state,\autoState,n) \mid \exists n^\prime\in [0..K] : (\state,\autoState,n^\prime)\in B_\productAccept\}\label{B_accept};\\
    B_\varnothing &\coloneqq \{(\state,\autoState,n) \mid \exists n^\prime\in [0..K] : (\state,\autoState,n^\prime)\in B^\times_\varnothing\label{B_reject}\}.
\end{align}
If \cref{sec-ltl:finite} holds, then $B_\varnothing\cap\productAccept=\varnothing$, $\forall \productState\in B_\autoAccept:\probP_\policy(\productState\models\textsf{\upshape G}\textsf{\upshape F }\LTL_\autoAccept)=1 $, and $\forall\productState\in B_\varnothing:\probP_\policy(\productState\models\textsf{\upshape G}\textsf{\upshape F }\LTL_\autoAccept)=0$. Furthermore, $B_\autoAccept$ and $B_\varnothing$ are sink sets, which means that, once the set is reached, no states outside the set can be reached.
\end{lemma}

The proof of this lemma is provided in \cref{proof:useful_lemma}. Using this lemma, we can now state and prove the main theorem of this chapter.

\begin{theorem}[Optimality Guarantee]
Given an LTL formula $\LTL$ and a product MDP $\productMDP$, if \cref{sec-ltl:finite} holds, then there exists an upper bound $U\in(0,1]$ for rewards and a discount factor $\discount\in(0,1]$ such that, for all product rewards $\productReward$ and product discount functions $\productDiscount$ satisfying \cref{reward_defn}, the optimal deterministic memoryless policy $\policy_\reward$ that maximises the expected discounted reward $G^{\policy_\reward}_0(\productState_0)$ is also an optimal policy $\policy_\LTL$ that maximises the probability of satisfying the Büchi condition $\probP_{\policy_\LTL}(\productState_0\models\textsf{\upshape G}\textsf{\upshape F }\LTL_\autoAccept)$ on the product MDP $\productMDP$.
\label{thm:optimality}
\end{theorem}

\begin{proof}[Proof sketch]
We now present a sketch of the proof to provide intuition for the main steps and the selection of key hyperparameters. The complete proof is provided in \cref{proof:optimality}.

To ensure optimality, given a policy $\policy$ with the product MDP $\productMDP$ and the LTL formula $\LTL$, we want to demonstrate a tight bound between the expected discounted reward following $\policy$ and the probability of $\policy$ satisfying $\LTL$, such that maximising one quantity is equivalent to maximising the other.

At a high level, we want to select the two key hyperparameters, the reward upper bound $U\in(0,1]$ and the discount factor $\discount\in(0,1]$, to adequately bound: (i) the rewards given for paths that eventually reach rejecting BSCCs (thus not satisfying the LTL specification); and (ii) the discount of rewards received from rejecting states for paths that eventually reach accepting BSCCs. 

We informally denote by $C^\policy_\varnothing$ the expected number of visits to accepting states before reaching a rejecting BSCC, and (i) can be sufficiently bounded by selecting $U=1/C^\policy_\varnothing$. Next, we informally write $C^\policy_\autoAccept$ for the expected number of rejecting states visited before reaching an accepting BSCC, and denote by $N^\policy$ the expected steps between visits of accepting states in the accepting BSCC. Intuitively, for (ii), we bound the amount of discount before reaching the accepting BSCC using $C^\policy_\autoAccept$, and we bound the discount after reaching the BSCC using $N^\policy$, yielding $\discount=1-1/(C^\policy_\varnothing*N^\policy+C^\policy_\autoAccept)$. 

In practice, using upper bounds of $C^\policy_\varnothing, C^\policy_\autoAccept$ and $N^\policy$ instead also ensures optimality, and these bounds can be deduced from assumptions about, or the knowledge of the MDP.
\end{proof}

As shown in the proof sketch, selecting $U=1/C^\policy_\varnothing$ and $\discount=1-1/(C^\policy_\varnothing*N^\policy+C^\policy_\autoAccept)$ is sufficient to ensure optimality. Using the example of the probabilistic gate MDP in \cref{fig:example_task} (right), we have $C^\policy_\varnothing\approx C^\policy_\autoAccept\leq 10$ and $N^\policy=1$, so choosing $U=0.1$ and $\gamma=0.95$ is sufficient to guarantee optimality. For more general MDPs, under the common assumption that the number of states $\abs{\states}$ and the minimum nonzero transition probability $p_{\min}\coloneqq \min_{s,a,s^\prime}\{\transitions(\state,\action,\state^\prime)>0\}$ are known, $C^\policy_\varnothing$ and $C^\policy_\autoAccept$ can be upper bounded by $\abs{\states}/p_{\min}$, while $N^\policy$ can be upper bounded by $\abs{\states}$.

\subsection{LTL Learning with Q-Learning}\label{sec-ltl:q-learning}

\begin{algorithm}[tb]
   \caption{KC Q-learning from LTL}
\label{alg:Q_learning}
\begin{algorithmic}[1]
   \STATE {\bfseries Input:} environment MDP $\MDP$, LTL formula $\LTL$
\STATE translate $\LTL$ into an LDBA $\automaton$
\STATE construct product MDP $\productMDP$ using $\MDP$ and $\automaton$\\
\STATE initialise Q for each $\state$ and $\autoState$ pair
\FOR {$l\gets 0$ to max\textunderscore episode}
  \STATE initialise $(\state,\autoState,n)\gets(\state_0,\autoState_0,0)$
  \FOR{$t\gets 0$ to max\textunderscore timestep}
  \STATE get policy $\policy$ derived from Q (\textit{e.g.,} $\epsilon$-greedy)
  \STATE take action $\productAction\gets\policy((\state,\autoState,n))$ in $\productMDP$
  \STATE get next product state $(\state^\prime,\autoState^\prime,n^\prime)$
\STATE $r\gets\productReward((\state,\autoState,n),\productAction,(\state^\prime,\autoState^\prime,n^\prime))$
\STATE $\discount\gets\discount^\prime(\state,\autoState,n)$
\STATE $Q((\state,\autoState),\productAction)\overset{\alpha}{\gets}r+\discount\max_{\action\in\productActions}Q((\state^\prime,\autoState^\prime),\action)$
\STATE   update $(\state,\autoState,n)\gets(\state^\prime,\autoState^\prime,n^\prime)$
\ENDFOR
\ENDFOR
\STATE   get greedy policy $\policy_\LTL$ from Q
\STATE {\bfseries Output:} induced policy on $\MDP$ by removing $\epsilon$-actions
\end{algorithmic}
% \vspace{-0.1cm}
\end{algorithm}

% \begin{algorithm}[tb]
% \DontPrintSemicolon
% \SetNlSty{textbf}{}{:}
% \caption{KC Q-learning from LTL}
% \label{alg:Q_learning}
% \BlankLine
% \SetKw{KwIn}{in}
% \SetKwInOut{Input}{Input}
% % \SetKwInOut{Output}{Output}
% % \SetKwFunction{proc}{LTLQ($\LTL$,$\MDP$)}
% \Input{environment MDP $\MDP$, LTL formula $\LTL$}
% % \Output{Game reward at $root$ node: $R(s_0,\sigma)$}
% % \SetKwProg{myproc}{Procedure}{:}{}

% \nl translate $\LTL$ into an LDBA $\automaton$\\
% \nl construct product MDP $\productMDP$ using $\MDP$ and $\automaton$\\
% \nl initialise Q for each $\state$ and $\autoState$ pair\\
% \nl \For {$l\gets 0$ to max\textunderscore episode}{
%   \nl initialise $(\state,\autoState,n)\gets(\state_0,\autoState_0,0)$\\
%   \nl \For{$t\gets 0$ to max\textunderscore timestep}{
%   \nl get policy $\policy$ derived from Q (\textit{e.g.,} $\epsilon$-greedy)\\
%   \nl take action $\productAction\gets\policy((\state,\autoState,n))$ in $\productMDP$\\
%   \nl get next product state $(\state^\prime,\autoState^\prime,n^\prime)$\\
% \nl $r\gets\productReward((\state,\autoState,n),\productAction,(\state^\prime,\autoState^\prime,n^\prime))$\\
% \nl $\discount\gets\discount^\prime(\state,\autoState,n)$ \\
% \nl $Q((\state,\autoState),\productAction)\overset{\alpha}{\gets}r+\discount\max_{\action\in\productActions}Q((\state^\prime,\autoState^\prime),\action)$\\
% \nl   update $(\state,\autoState,n)\gets(\state^\prime,\autoState^\prime,n^\prime)$
%         }
%      }
% \nl   get greedy policy $\policy_\LTL$ from Q\\
% \nl   \Return induced policy on $\MDP$ by removing $\epsilon$-actions
% \end{algorithm}

Using the product MDP $\productMDP$ of \cref{def:productMDP} and its reward structure as in \cref{reward_defn}, we present \cref{alg:Q_learning} (KC), a model-free Q-learning algorithm for LTL specifications using the K counter product MDP. In line 10, the product MDP is constructed on the fly as we explore: for action $\productAction\in\actions$, observe the next environment state $\state^\prime$ by taking action $\productAction$ in environment state $\state$. Then, we obtain the next automaton state $\autoState^\prime$ using the transition function $\autoTransitions(\autoState,\labFunc(\state))$ and counter $n$ depending on whether $n\leq K$ and $\autoState^\prime\in\autoAccept$. If $\productAction\in\{\epsilon_\autoState\mid\autoState\in\autoStates\}$, update $\autoState^\prime$ using the $\epsilon$-transition and leave the environment state $\state$ and counter $n$ unchanged. However, directly adopting Q-learning on this product MDP produces a Q function defined on the entire product state space $\productStates$, which means that the agent needs to learn the Q function for each $K$ value. To improve efficiency, we propose to define the Q function only on the environment states $\states$ and the automaton states $\autoStates$, and for a path $\MDPpath$ of $\productMDP$, the update rule for the Q function at time step $t$ is:
\begin{align*}
Q_{t+1}((\state_t,\autoState_t),\productAction_t)&\overset{\alpha}{\gets}\productReward(\productState_t,\productAction_t,\productState_{t+1})+\discount^\prime(\productState_{t})\max_{\productAction\in\productActions}Q_t((\state_{t+1},\autoState_{t+1}),\productAction)\nonumber,
\end{align*}
where $\productState_t=(\state_t,\autoState_t,n_t)$, $\productAction_t$ is the action taken at time step $t$, ${\productState_{t+1}=(\state_{t+1},\autoState_{t+1},n_{t+1})}$ is the next product state, and $\alpha$ is the learning rate. We assert that, with this collapsed Q function, the algorithm returns the optimal policy for satisfying $\LTL$ because the optimal policy is independent of the K counter, with the proof provided in \cref{proof:q_learning}.
 
\begin{theorem}[Convergence of KC]
\label{thm:q_learning}
Given an environment MDP $\MDP$ and an LTL specification $\LTL$ with an appropriate discount factor $\gamma$ and a reward function $\productReward$ satisfying \cref{thm:optimality}, then if \cref{sec-ltl:finite} holds, Q-learning for LTL described in \cref{alg:Q_learning} converges to an optimal policy $\policy_\LTL$ that maximises the probability of satisfying $\LTL$ in $\MDP$.
\end{theorem}

\subsection{Counterfactual Imagining}\label{sec-ltl:imagining}

\begin{algorithm}[tb]
\caption{CF+KC Q-learning from LTL}
\label{alg:CounterQ}
\begin{algorithmic}[1]
\STATE {\bfseries Input:} environment MDP $\MDP$, LTL formula $\LTL$
% \Output{Game reward at $root$ node: $R(s_0,\sigma)$}
% \SetKwProg{myproc}{Procedure}{:}{}
  \STATE translate $\LTL$ into an LDBA $\automaton$
  \STATE construct product MDP $\productMDP$ using $\MDP$ and $\automaton$
  \STATE initialise Q for each $\state$ and $\autoState$ pair
 \FOR {$l\gets 0$ to max\textunderscore episode}
  \STATE initialise $(\state,\autoState,n)\gets(\state_0,\autoState_0,0)$
  \FOR{$t\gets 0$ to max\textunderscore timestep}
  \STATE get policy $\policy$ derived from Q (e.g., $\epsilon$-greedy)
  \STATE get action $\productAction\gets\policy((\state,\autoState,n))$ in $\productMDP$
  \FOR {$\Bar{\autoState}\in\autoStates$}
  \STATE  get counterfactual imagination $(\state^\prime,\Bar{\autoState}^\prime,n^\prime)$ by taking action $\productAction$ at $(\state,\Bar{\autoState},n)$
  \STATE $r\gets\productReward((\state,\Bar{\autoState},n),\productAction,(\state^\prime,\Bar{\autoState}^\prime,,n^\prime))$
  \STATE $\discount\gets\discount^\prime(\state,\Bar{\autoState},n)$ 
  \STATE $Q((\state,\Bar{\autoState}),\productAction)\overset{\alpha}{\gets}r+\discount\max_{\action\in\productActions}Q((\state^\prime,\Bar{\autoState}^\prime),\action)$
  \ENDFOR
  \STATE
  obtain $(\state^\prime,\autoState^\prime,n^\prime)$ using action $\productAction$ at $(\state,\autoState,n)$
  \STATE   update $(\state,\autoState,n)\gets(\state^\prime,\autoState^\prime,n^\prime)$
\ENDFOR
\ENDFOR
  \STATE   get greedy policy $\policy_\LTL$ from Q
 \STATE {\bfseries Output:} induced policy on $\MDP$ by removing $\epsilon$-actions
\end{algorithmic}
\end{algorithm}

We now propose a method to exploit the structure of the product MDP, specifically the LDBA, to improve sample efficiency. While KC (\cref{alg:Q_learning}) is guaranteed to converge to the optimal policy, it requires the RL agent to explore an expanded state space in the product MDP, which increases the sample complexity. To address this, we leverage counterfactual reasoning to generate synthetic imaginations --- from a given state in the environment MDP, we generate synthetic trajectories by imagining being at each possible automaton state while taking the same actions. This enables more efficient learning by augmenting the training data without additional environment interaction.

If the agent is at product state $(\state,\autoState,n)$ and an action $\productAction\in\actions$ is chosen, for each $\Bar{\autoState}\in\autoStates$, the next state $(\state^\prime,\Bar{\autoState}^\prime,n^\prime)$ by taking an action $\productAction$ from $(\state,\Bar{\autoState},n)$ can be computed by first taking the action in the environment state, and then computing the next automaton state $\Bar{\autoState}^\prime=\autoTransitions(\Bar{\autoState},\labFunc(\state))$ and the next $K$ value $n^\prime$. The reward for the agent is $\productReward(\state,\Bar{\autoState},\Bar{n})$, and therefore we can update the Q function with this enriched set of experiences. These experiences produced by counterfactual imagining are still sampled from $\productMDP$ following the transition function $\productTransitions$, and hence, when used in conjunction with any off-policy learning algorithms, such as Q-learning, the optimality guarantees of the algorithm are preserved.

As shown in \cref{alg:CounterQ} (CF-KC), counterfactual imagining (CF) can be incorporated into our KC Q-learning algorithm by altering a few lines (lines 9-12 in \cref{alg:CounterQ}) of code, and it can also be used in combination with other RL algorithms for LTL learning that use automata-based product MDP. Note that the idea of counterfactual imagining is similar to that proposed by \citet{Icarte2022RewardLearning}, but our approach has adopted LDBAs in the product MDPs for LTL specification learning.

\section{Experimental Results}\label{sec-ltl:exps}

We evaluate our algorithms in various MDP environments, including the more realistic and challenging stochastic MDP environments\footnote{The implementation of our algorithms and experiments can be found on GitHub: \url{https://github.com/shaodaqian/rl-from-ltl}}. We use tabular Q-learning as the core off-policy learning method to implement our three algorithms: Q-learning with $K$ counter reward structure (KC), Q-learning with $K$ counter reward structure and counterfactual imagining (CF+KC), and Q-learning with only counterfactual imagining (CF), in which we set $K=0$. We compare the performance of our methods against the methods proposed by~\citet{Bozkurt2019ControlLearning}, \citet{Hahn2019Omega-regularLearning} and \citet{Hasanbeig2020DeepLogics}. Note that our KC algorithm, in the special case that $K=0$ with no counterfactual imagining, is algorithmically equivalent to \citet{Bozkurt2019ControlLearning} when setting their hyperparameter $\gamma_B=1-U$. Our methods differ from all other existing methods to the best of our knowledge.

\paragraph{Experimental Setup}

The experiments were carried out on a Linux server (Ubuntu 18.04.2) with two Intel Xeon Gold 6252 CPUs and six NVIDIA GeForce RTX 2080 Ti GPUs. We select three stochastic environments for evaluation: the probabilistic gate (\cref{example_mdp}), the frozen lake~\citep{Brockman2016OpenAIGym}, and the office world~\citep{Icarte2022RewardLearning}. We set the learning rate $\alpha=0.1$ and $\epsilon=0.1$ for exploration. We also set a relatively loose upper bound on rewards $U=0.1$ and the discount factor $\gamma=0.99$ for all experiments to ensure optimality. Note that the optimality of our algorithms holds for a family of reward structures defined in \cref{reward_defn}, and for experiments we opt for a specific reward function that linearly increases the reward for accepting states as the value of $K$ increases, namely $r_n=U\cdot n/K \; \forall n\in[0..K]$, to facilitate training and exploration. The Q function is optimistically initialised by setting the Q value for all available state-action pairs at $2U$.  We propose to directly evaluate the probability of satisfying LTL specifications by employing the probabilistic model checker PRISM~\citep{Kwiatkowska2011PRISMSystems}. We build the induced MC $\MDP_\policy$ from the environment MDP and the policy in PRISM format, and adopt PRISM to compute the exact satisfaction probability of the given LTL specification. All experiments are run 100 times, where we plot the average satisfaction probability with half the standard deviation in the shaded area. Further details regarding the experimental setup are provided in \cref{appendix:exp_setup}.
 
\begin{figure*}[t]
\centering
\begin{subfigure}[t]{\textwidth}
    \centering
\includegraphics[width=1\textwidth]{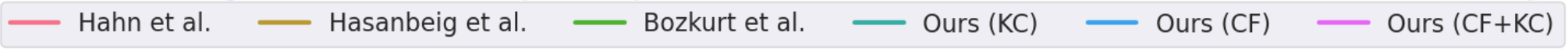}
\end{subfigure}
\begin{subfigure}[t]{0.49\textwidth}
    \centering
    \includegraphics[width=\textwidth]{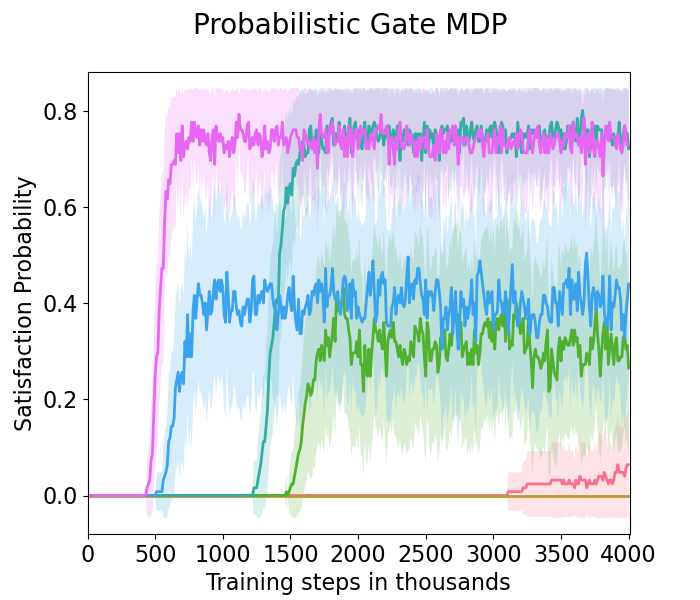}
    \caption{Results on the probabilistic gate MDP}
    \label{fig:hard1_result}
\end{subfigure}
\begin{subfigure}[t]{0.5\textwidth}
    \centering
    \includegraphics[width=\textwidth]{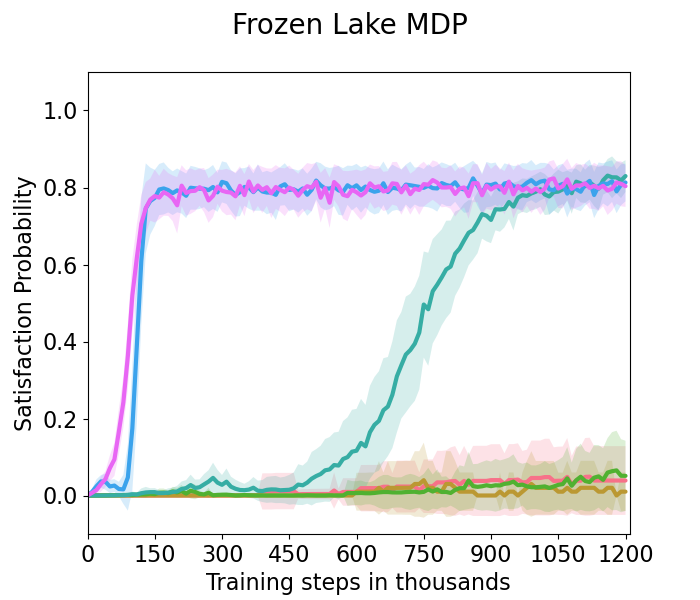}
    \caption{Results on the FrozenLake}
    \label{fig:frozen_result}
\end{subfigure}
\begin{subfigure}[t]{0.6\textwidth}
    \centering
    \includegraphics[width=0.82\textwidth]{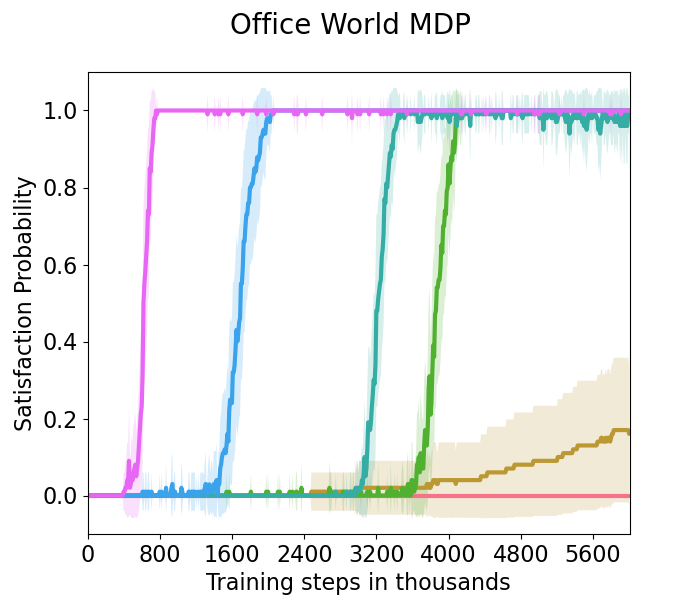}
    \caption{Results on the office world~\citep{Icarte2022RewardLearning}.}
    \label{fig:office_result}
\end{subfigure}
\caption{The experimental results on various MDPs and LTL tasks.}
\label{fig:all_results}
\end{figure*}

\subsection{Probabilistic Gate MDP}

First, we conduct experiments on the probabilistic gate MDP described in \cref{example_mdp} with task \enquote{\textsf{\upshape F}\textsf{\upshape G} a \& \textsf{\upshape G} !c}, which means reaching only states labelled \enquote{a} in the future while never reaching \enquote{c} labelled states. We set $K=10$ for this task, and in \cref{fig:all_results} (left), compared to the other three methods, our method KC achieved better sample efficiency and convergence, and CF demonstrates better sample efficiency while still lacking training stability. The best performance is achieved by CF+KC, while other methods exhibit slower convergence~\citep{Bozkurt2019ControlLearning,Hahn2019Omega-regularLearning} or do not converge~\citep{Hasanbeig2020DeepLogics} due to the lack of theoretical optimality guarantees.

\subsection{Frozen Lake MDP}

The second MDP environment is the $8\times8$ frozen lake environment of OpenAI Gym~\citep{Brockman2016OpenAIGym}. This environment consists of frozen lake tiles, where the agent has a 1/3 chance of moving in the intended direction and 1/3 of going sideways each, with details provided in \cref{appendix:frozenlake}. The task is \enquote{(\textsf{\upshape G}\textsf{\upshape F} a $\mid$ \textsf{\upshape G}\textsf{\upshape F} b) \& \textsf{\upshape G} !h}, meaning to always reach lake camp \enquote{a} or lake camp \enquote{b} while never falling into holes \enquote{h}. We set $K=10$ for this task, and in \cref{fig:all_results} (middle), we observe significantly better sample efficiency for all our methods, especially for CF+KC and CF, which converge to the optimal policy at around 150k training steps. The other three methods, on the other hand, barely start to converge at 1200k training steps. CF performs especially well in this task because the choice of always reaching \enquote{a} or \enquote{b} can be considered simultaneously during each time step, reducing the sample complexity to explore the environment.

\subsection{Office World MDP}
Lastly, we experiment on a slight modification of the more challenging office world environment proposed by \citet{Icarte2022RewardLearning}, with details provided in \cref{appendix:officeworld}. We include patches of icy surfaces in the office world, with the task to either patrol in the corridor between \enquote{a} and \enquote{b}, or write letters at \enquote{l} and then patrol between getting tea \enquote{t} and the workplace \enquote{w}, while never hitting obstacles \enquote{o}. $K=5$ is set for this task for a steeper increase in reward, since the long distance between patrolling states makes visiting many accepting states in each episode time consuming. \cref{fig:all_results} (right) presents again the performance benefit of our methods, with CF+KC performing the best and CF and KC second and third, respectively. For this challenging task in a large environment, the method of \citet{Hahn2019Omega-regularLearning} requires the highest number of training steps to converge.

Overall, the results demonstrate that KC improves both sample efficiency and training stability, especially for challenging tasks. In addition, CF greatly improves the sample efficiency, which, in combination with KC, achieves the best results.

\subsection{Runtime Analysis and Sensitivity Analysis}

\begin{figure}[tb]
    \vspace{-0.5cm}
    \centering
    \includegraphics[width=0.6\textwidth]{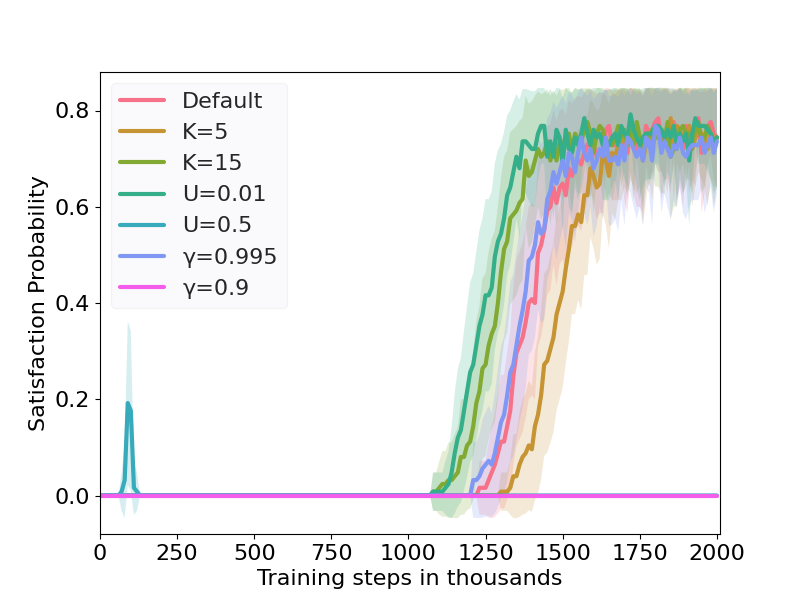}
    \caption{Sensitivity analysis on key hyperparameters.}
    \label{fig:sensitivity_analysis}
\end{figure}

It is worth mentioning that, for counterfactual imagining, multiple updates to the Q function are performed at each step in the environment. This increases computational complexity, but the additional inner loop updates on the Q function will only marginally affect the overall computation time if the environment steps are computationally expensive. Taking the office world task as an example, the average time to perform 6 million training steps is 170.9s, 206.3s, and 236.1s for KC, CF, and CF+KC, respectively. However, the time until convergence to the optimal policies is 96.8s, 70.2s, and 27.5s for KC, CF, and CF+KC, respectively.

For sensitivity analysis of key hyperparameters, we conduct experiments on the probabilistic gate MDP task with different hyperparameters against the default values of $U=0.1, \gamma=0.99$ and $K=10$. As shown in \cref{fig:sensitivity_analysis}, if $U(=0.5)$ is chosen too high or $\gamma(=0.9)$ is chosen too low, the algorithm does not converge to the optimal policy as expected. However, looser hyperparameters $U=0.01$ and $\gamma=0.995$ do not harm performance, which means that, even with limited knowledge of the underlying MDP, our algorithm still performs well with loose hyperparameters. Optimality is not affected by the $K$ value, while performance is only slightly affected by different $K$ values.

\section{Discussion}\label{sec-ltl:discussion}

In this chapter, we considered the problem of learning policies that satisfy high-level LTL objectives. We presented a novel model-free reinforcement learning algorithm to learn the optimal policy for satisfying LTL specifications in an unknown stochastic MDP environment with optimality guarantees. We proposed a novel product MDP, a generalised reward structure, and an RL algorithm that ensures convergence to the optimal policy with the appropriate hyperparameters. Furthermore, we incorporated counterfactual imagining, which exploits the LTL specification to create counterfactual experiences. Lastly, using PRISM~\citep{Kwiatkowska2011PRISMSystems}, we directly evaluated the performance of our methods and demonstrated superior performance in various MDP environments and LTL tasks.

The reward structure we defined with the product MDP is general in the sense that it allows for a family of specific reward functions that depend on the counter value $K$, all of which enjoy the optimality guarantees. In this chapter, we explored a reward function that increases linearly with the $K$ value, since it encourages exploration. It would also be interesting future work to explore other specific reward functions under our generalised reward structure, such as exponentially increasing or decreasing rewards, and to evaluate the different learning dynamics induced by these reward functions.

The main limitation of the method presented in this chapter is the assumption regarding the discrete state and action spaces of the MDP. It is necessary for our analysis to provide theoretical guarantees on the optimality of the learnt policy due to the infinite MDP path required for an LTL specification to be satisfied. It will be a challenging future task to tackle the open problem~\citep{Alur2022} of dropping all assumptions regarding the underlying MDP, and extending the theoretical framework to continuous state and action environment MDPs, which might be addressed through a finite abstraction of the state space. In addition, using potential-based reward shaping~\citep{Ng1999PolicyShaping,Devlin2012DynamicShaping} to exploit the semantic class structures of the LTL specifications and transferring similar temporal logic knowledge of the agent~\citep{Xu2019TransferLearning} between environments are also interesting future directions.

\chapter{Conclusion}\label{chapter:conclusion}

This thesis studied the problem of learning decision policies with optimality and sample efficiency guarantees that improve the practicality of AI-driven decisions in real-world applications. Specifically, we considered the problem of learning policies from offline datasets in the presence of hidden confounders and learning policies that satisfy high-level LTL objectives.

We first considered learning policies from a static offline dataset that is confounded by hidden confounders. Under the RL paradigm, we learn the causal effect of actions on the expected return, which can then be used to learn an optimal policy. Specifically, we used instrumental variables, which are variables independent of the hidden confounders, to learn the causal effect of actions through IV regression. The problem of IV regression is a CMR problem, and we proposed a novel CMR estimator, DML-CMR, using the DML framework that provides fast convergence rate guarantees. The learnt policy using DML-CMR also enjoys optimality guarantees, where strong empirical results are demonstrated. However, this approach requires that IVs are observed in the dataset for each action to allow the causal effect of actions to be identifiable. Although IVs are commonly observed in many applications, they are not guaranteed to exist for all real-world applications, and identifying them in the dataset often requires expert knowledge. To this end, we studied the problem of learning policies without explicit IVs in the confounded dataset and found that it is possible to imitate an expert from a dataset of expert demonstrations, even in the presence of hidden confounders. This describes the problem of confounded imitation learning, where we proposed a framework for confounded IL in~\cref{chapter:il} that unifies and generalises previous work in confounded IL. We showed that, by considering the hidden confounders to be partially observable to the expert, this framework subsumes a wide range of settings in previous works. We then reduced the problem of confounded IL in our framework into a CMR problem, where any CMR estimator in principle can be adopted to solve it. Specifically, we adopted our DML-CMR estimator and proposed DML-IL, a confounded IL algorithm in our framework that inherits the convergence rate guarantees of DML-CMR.

Finally, we studied the problem of learning policies that satisfy high-level objectives under the classic online RL framework. To this end, we considered LTL objectives for their expressiveness and widespread usage in robotics and formal verification. We proposed to first translate the LTL specification into a limit-deterministic Büchi automaton, where an accepting run in the automaton also satisfies the LTL specification. Then, we proposed a product MDP that is the product of the automaton, the environment MDP, and a counter that counts the number of accepting states visited. This counter allows a generalised reward structure to be defined in our product MDP, which enables sample-efficient learning by encouraging the agent to explore more efficiently in the environment. We proved that our algorithm will learn an optimal policy that satisfies the LTL specification with the highest probability and demonstrated significantly better sample efficiency against state-of-the-art algorithms.

\section{Discussion}\label{conclusion:discussion}

We now provide a general discussion of the results derived in this thesis, including the practicality of the proposed algorithms, and the strengths and weaknesses of the methods.

\subsection{Practicality in Real-World Applications}\label{conclusion:practicality}

In this thesis, we proposed policy-learning algorithms that are practical in real-world applications. Specifically, in~\cref{chapter:dmliv} and~\cref{chapter:il}, we considered the problem of learning from offline datasets, which are abundant in practice, as opposed to online interactions, which are often unsafe or impractical. We address the hidden confounder problem, which is present in many real-world datasets and often leads to suboptimal policies, by enabling policy learning from confounded offline datasets. However, the ability to handle hidden confounders comes at a cost, namely, the assumptions required to identify the true causal effect between states and actions. In~\cref{chapter:dmliv}, we assumed that IVs are observed in the offline dataset, such that the causal effect of actions can be estimated. In~\cref{chapter:il}, we weakened the IV assumption and instead assumed that the confounding noise has a maximum horizon of $k$, such that the confounding noise $k$ steps prior is independent from the current confounding noise. This allows us to correctly imitate the intended actions of the expert. Although these assumptions limit practicality, we offered solutions for different types of assumptions such that, depending on the specific real-world scenario, users can choose to adopt the appropriate algorithms. In addition, as discussed in the chapters, these assumptions often hold in real-world applications, for example, in robotics, econometrics, and healthcare.

In~\cref{chapter:ltl}, we considered learning policies that satisfy LTL specifications. This allows policies to be learnt directly from high-level objectives, removing the need to manually design complex reward functions and reducing the risks of reward misspecification. Our proposed algorithm is, in general, very practical since it adopts model-free RL. Although the optimality guarantee requires the state and action space to be discrete, the algorithm and the product MDP can be adapted to environments with continuous states and actions in practice following \citet{Hasanbeig2020DeepLogics}.

Finally, we point out that all the algorithms proposed in this thesis are sample-efficient while providing optimality guarantees. This addresses two important challenges that prevent RL- and DL-driven decision policies from being deployed in the real world. We prove optimality or provide an upper bound on the suboptimality, which guarantees that the algorithms will learn good policies that achieve the desired behaviour. In addition, sample efficiency is important both in offline and online settings, especially because data collection is difficult and costly. This allows a better policy to be learnt given the same amount of data in practice.

\subsection{Strengths and Weaknesses of the Methods}\label{conclusion:strengths}

The main strength of this thesis is the versatile suite of algorithms proposed for various problem settings that improve practicality with provable guarantees, as discussed above. In addition, we proposed novel frameworks and methods that are very general. The method developed in~\cref{chapter:dmliv}, DML-CMR, can be adopted to solve general CMR problems, which include a large family of important problems in econometrics, statistical learning, causal inference, and mathematics. In~\cref{chapter:il}, we proposed a unifying framework for confounded IL that unifies and generalises previous work. The algorithm we proposed for confounded IL, DML-IL, also addresses a wider range of problems described by our unifying framework. In~\cref{chapter:ltl}, we proposed a generalised product MDP and a reward structure admitting a more flexible and general class of reward functions. In addition, each of these reward functions enjoys optimality guarantees thanks to the generalised product MDP and our theoretical framework.

Regarding the weaknesses, we first highlight the assumptions we made for the algorithms and the theoretical analysis in this thesis. In~\cref{chapter:dmliv}, we assumed that IVs can be observed in the dataset as the basis for causal identification, and for the theoretical analysis, we assumed some regularity and identifiability conditions. In~\cref{chapter:il}, we assumed that the confounding noise horizon $k$ is bounded and that an upper bound on it is known for the DML-IL algorithm to imitate the expert. This is also a basic assumption for the expert policy to be identifiable, where we discussed the practicality of these assumptions in detail in~\cref{chapter:il}. In~\cref{chapter:ltl}, for the theoretical analysis, we assumed that the state and action spaces are discrete. This is a strong assumption to make since many real-world applications have continuous state spaces. However, the discrete assumption is required to obtain the theoretical guarantees, and we point out that in practice, discrete abstractions often exist for continuous state and action spaces. The other weakness of this thesis is that the empirical evaluation is done on synthetic academic benchmarks and semi-synthetic datasets. This is due to the nature of academic research, which favours controlled experiments and reproducibility. In addition, it is challenging to handle real-world data due to privacy concerns and to deploy algorithms in real-world scenarios, both of which are beyond the scope of this thesis. Although it would strengthen the thesis to apply the proposed algorithms in the real world, the empirical results in this thesis provide a strong proof-of-concept assurance for practitioners to use them in practice.

% Overall, good progresses are made in this thesis toward decision-making in the real-world, specifically focusing on hidden cnonfounders and high-level objectives. Dispite the limitation and assumptions required, they are necessary 

\section{Open Problems and Future Work}\label{conclusion:open}

Unsurprisingly, the work in this thesis gives rise to open problems. In~\cref{chapter:dmliv}, an interesting future work is to extend our DML framework to nonparametric estimation. A possible way to do this is to follow ~\citet{Foster2019OrthogonalLearning}, who analysed nonparametric estimation with Neyman orthogonal score functions. In~\cref{chapter:il}, an open problem is to explore confounded imitation learning in real-world applications such as robotics. Moreover, the setting in which limited queries can be made to the expert is of interest, as it is realistic. The challenge here would be to find the optimal way to query the expert. In~\cref{chapter:ltl}, an open problem is to provide optimality guarantees for RL with LTL objectives in continuous state and action spaces. In addition, it would also be interesting to explore offline RL with LTL objectives. One challenge is to design effective offline RL algorithms for LTL objectives, while another challenge is to derive guarantees for the learnt policy.

\bill{More broadly and ambitiously, a challenging yet important open problem is to relax the causal identification assumptions when dealing with hidden confounders. Instead, we can consider a setting in which the causal effect can only be partially identified, such as in cases where only the upper and lower bounds of the causal effect can be obtained, which is often considered in the IV literature~\citep{Balke1994,Swanson2018}. Without full identification, we cannot exactly infer the causal effect of actions, but it may be possible to provide pessimistic and robust algorithms based on the uncertainty (e.g.,~\citealt{Pace2024}) arising from partial identification. In addition, it is desirable to develop algorithms that remain robust to imperfect or weak IV, possibly by leveraging partial identification and bounding techniques, using uncertainty quantification techniques and instrument synthesis techniques to infer pseudo-instruments~\citep{Yuan2022}.}

\bill{Furthermore, it would be interesting to develop benchmarking datasets for confounded offline RL problems to standardise evaluation. Currently, algorithms tackling confounded offline RL are typically evaluated on toy causal graphs, MuJoCo or grid-world environments with various types of synthetic confounding, and small semi-synthetic healthcare datasets. Existing publications typically use different confounding mechanisms and dataset setups to showcase various aspects of the problem that suit their algorithm, making it difficult for the researcher to understand all the minor nuanced differences in each experimental setup and compare between the methods. Therefore, it would be beneficial to establish a benchmarking dataset for confounded RL in order to standardise the evaluation, improve reproducibility, and enable fair comparison between algorithms. The potential steps towards this goal would be to formally define a general confounded MDP framework, build confounded simulators which can be adopted from D4RL~\citep{Fu2020D4RL:Learning} environments, establish evaluation metrics, and integrate existing baseline algorithms.}

% \include{chapters/worktodate}

%next line adds the Bibliography to the contents page
\addcontentsline{toc}{chapter}{Bibliography}
%uncomment next line to change bibliography name to references
%\renewcommand{\bibname}{References}
\bibliography{reference.bib}        %use a bibtex bibliography file refs.bib
\bibliographystyle{style}  %use the plain bibliography style

\appendix
\chapter{Learning Policies from Confounded Datasets with Instrumental Variables}

\section{The Score Function for Standard Two-Stage CMR Estimators}\label{appen:score}

In this section, we show that the learning objective, or score function, for standard two-stage CMR estimators~\citep{Angrist1996,Hartford2017DeepPrediction,Singh2019} is not Neyman orthogonal and thus cannot be used to create a DML estimator for the CMR problem.

\begin{repeatprop}{prop:standard_score}
The score (or objective) function for standard two-stage CMR estimators
$\ell=(Y-\widehat{g}(f,c))^2$
is not Neyman orthogonal at $(f_0, g_0)$.
\end{repeatprop}

\begin{proof}
The score $\ell=(Y-\widehat{g}(f,c))^2$ is not Neyman orthogonal because, first of all, $\expectE[(Y-g_0(f_0,c))^2]=\expectE[(Y-\expectE[Y\lvert C])^2]\neq0$ since $\expectE[f_0(X)\lvert C]=\expectE[Y\lvert C]$ and $Y-\expectE[Y\lvert C]\neq0$ due to the noise on $Y$. This violates the basic condition for a Neyman orthogonal score that the score function equals zero with the true functions $f_0$ and $g_0$.

Secondly, the Gateaux derivative against small changes in $g$ for score $\expectE[(Y-g_0(f_0,c))^2]$ at $(f_0, g_0)$ is
\begin{align*}
\frac{\partial}{\partial r}\expectE\Bigl[&(Y-g_0(f_0,C)-r\cdot g(f_0,C))^2\Bigr]\\
=&\frac{\partial}{\partial r}\expectE\Bigl[(Y-g_0(f_0,C))^2-2r\cdot(Y-g_0(f_0,C))g(f_0,C)+r^2\cdot g(f_0,C)^2\Bigr]\\
=&\expectE\Bigl[2(Y-g_0(f_0,C))g(f_0,C)+2r\cdot g(f_0,C)^2
\Bigr],
\end{align*}
and, when $r=0$, this derivative evaluates to
\begin{equation*}
\expectE[2(Y-g_0(f_0,c))g(f_0,c)]=\expectE[2
(Y-\expectE[Y\lvert C])g(f_0,c)],
\end{equation*}
which does not equal 0 for general $g\in\mathcal{G}$ since generally $g(f_0,c)$ and the residual $(Y-\expectE[Y\lvert C])$ are correlated. Therefore, this standard score function for two-stage CMR estimation is not Neyman orthogonal at $(f_0, g_0)$.
\end{proof}

\section{Proofs}
In this section, we restate all the conditions required to prove the $N^{-1/2}$ convergence rate guarantees for the DML-CMR estimator, and provide the omitted proofs in \cref{chapter:dmliv} for \cref{thm:neyman}, \cref{lemma:nuisances}, \cref{thm:dml}, \cref{coro:function_convergence}, and \cref{coro:subopt}.

\subsection{DML-CMR $N^{-1/2}$ Convergence Rate Guarantees}\label{appen:dml}

To obtain $N^{-1/2}$ convergence rate guarantees of the DML-CMR estimator, recall that the following conditions must be satisfied.

\noindent\textbf{\cref{condition:dml} [Conditions for $N^{-1/2}$ convergence of DML, Assumption 3.3 and 3.4 in~\citet{Chernozhukov2018Double/debiased}]}

For sample size $N\geq3$:
\begin{enumerate}[label=(\alph*)]
\item The map $(\theta,(s,g))\mapsto \expectE[\orthoM(\dataset;f_{\theta},(s,g))]$ is twice continuously Gateaux-differentiable.
\item  The score $\orthoM$ obeys the Neyman orthogonality conditions.
\item The true parameter $\theta_0$ obeys $\expectE[\orthoM(\dataset;f_{\theta_0},(s_0,g_0))]=0$ and $\Theta$ contains a ball of radius $c_1 N^{-1/2}\log N$ centered at $\theta_0$. 
\item For all $\theta\in\Theta$, the identification relationship
\begin{align*}
2\norm{\expectE[\orthoM(\dataset;f_{\theta},(s_0,g_0))]}\gtrsim \norm{J_0(\theta-\theta_0)}
\end{align*}
is satisfied, where $J_0\coloneqq\partial_{\theta^\prime}\{\expectE[\orthoM(\dataset;f_{\theta^\prime},(s_0,g_0))]\}|_{\theta^\prime=\theta_0}$ is the Jacobian matrix, with singular values bounded between $c_0>0$ and $c_1>0$.
\item Let $K$ be a fixed integer. Given a random partition $\{I_k\}_{k=1}^K$ of indices $[N]$ each of size $n=N/K$, the nuisance parameter estimator $\widehat{s}_k$ and $\widehat{g}_k$ learnt using data with indices $I^c_k$ belongs to shrinking realisation sets $\mathcal{S}_N$ and $\mathcal{G}_N$ respectively, and the nuisance parameters should be estimated at the $o(N^{-1/4})$ rate, e.g., $\norm{\widehat{s}-s_{0}}_2=o(N^{-1/4})$.
\end{enumerate}

To formalise the convergence rate guarantees in relation to the technical conditions, we present the following proposition as a direct consequence of Lemma 6.3 of~\citet{Chernozhukov2018Double/debiased}.

\begin{proposition}[DML convergence rate of $N^{-1/2}$]\label{prop:thm3.3}
If all conditions in~\cref{condition:dml} hold, then the DML estimator $\widehat{\theta}$ as defined in~\cref{defn:dml} is concentrated in a $1/\sqrt{N}$ neighbourhood of $\theta_0$ with probability $1-o(1)$:
\begin{align*}
    \sqrt{N}\norm{\widehat{\theta}-\theta_0}=O_p(\rho_N)
\end{align*}
where $\{\rho_N\}_{N\geq1}=o(1)$ is some sequence of positive constants converging to zero converges to 0. This implies that, for all $\zeta\in(0,1]$, there exists $K>0$ such that, for all $N\geq K$,  $\norm{\widehat{\theta}-\theta_0}=O(N^{-1/2})$ with probability $1-\zeta$.
\end{proposition}
\begin{proof}
This proposition is a slight variant of Lemma 6.3 from~\citet{Chernozhukov2018Double/debiased}. We will reuse partial results from Lemma 6.3, and specifically the following two technical bounds on some quantity of interest that we will use for this proof. We denote by $\eta=(s,g)$ the nuisance parameters and $\mathcal{T}_N=(\mathcal{S}_N,\mathcal{G}_N)$ the realisation set of the nuisance parameters.

From step 1 of Lemma 6.3~\citep{Chernozhukov2018Double/debiased}, we have that there exists a sequence of positive constants $\{\tau_N\}_{N\geq1}$ that converges to zero, such that $\norm{\widehat{\theta}-\theta_0}\leq\tau_N$
with probability $1-o(1)$.

From step 4 of Lemma 6.3~\citep{Chernozhukov2018Double/debiased}, with probability $1 - o(1)$, there exists $q>2$ such that
\begin{align}
\mathcal{I}_3 := \inf_{\theta \in \Theta} \sqrt{n}\left\| \mathbb{E}_n[\psi(\dataset; \theta, \widehat{\eta}_0)] \right\|
= O\!\left( n^{-1/2 + 1/q} \log n + o(N^{-1/4})^2 \sqrt{n} \right). \label{eq:I3}
\end{align}

From step 3 of Lemma 6.3~\citep{Chernozhukov2018Double/debiased}, with probability $1 - o(1)$, we have that
\begin{align}
\mathcal{I}_5 := \sup_{\|\theta - \theta_0\| \leq \tau_N} 
\left\| \mathbb{G}_n \big( \psi(\dataset; \theta, \widehat{\eta}_0) - \psi(\dataset; \theta_0, \eta_0) \big) \right\|
=O(n^{-1/2 + 1/q} \log n).\label{eq:I5}
\end{align}

For a set of data indices $I$, we define the empirical expectation as:
\[
E_n(\psi(\dataset)) := \frac{1}{n} \sum_{i \in I} \psi(\dataset_i);
\]
and define the empirical process $G_n(\psi(\dataset))$ as a linear operator acting on measurable
functions $\psi : \mathcal{\dataset} \to \mathbb{R}$ such that $\|\psi\|_{P,2} < \infty$:
\begin{align*}
G_n(\psi(\dataset)) &:= \frac{1}{\sqrt{n}} \sum_{i \in I} \psi(\dataset_i) - \int \psi(w) dP(w).\\
&:=E_n(\psi(\dataset))-E_{P_N}(\psi(\dataset)).
\end{align*}

Now, we prove the claim of the proposition. 
First, by definition of $\widehat{\theta}_0$ and the compactness of $\Theta$, we have
\[
\sqrt{n}\left\| \mathbb{E}_n[\psi(\dataset; \widehat{\theta}_0, \widehat{\eta}_0)] \right\|
= \inf_{\theta \in \Theta} \sqrt{n}\left\| \mathbb{E}_n[\psi(\dataset; \theta, \widehat{\eta}_0)] \right\|.\]

Moreover, for any $\theta \in \Theta$ and $\eta \in \mathcal{T}_N$, we have
\begin{align}
\sqrt{n}\mathbb{E}_n[\psi(\dataset; \theta, \eta)]
&= \sqrt{n}\mathbb{E}_n[\psi(\dataset; \theta_0, \eta_0)]
\nonumber\\
&+ G_n[\psi(\dataset; \theta, \eta) - \psi(\dataset; \theta_0, \eta_0)]
+ \sqrt{n}\big( \mathbb{E}_{P_N}[\psi(\dataset; \theta, \eta)] \big),\label{eq:prop_emperical_expect}
\end{align}
% \sqrt{n}\mathbb{E}_n[\psi(\dataset; \theta, \eta)]
% = \sqrt{n}\mathbb{E}_n[\psi(\dataset; \theta_0, \eta_0)]
% \\
% + G_n[\psi(\dataset; \theta, \eta) - \psi(\dataset; \theta_0, \eta_0)]
% + \sqrt{n}\big( \mathbb{E}_{P_N}[\psi(\dataset; \theta, \eta)] \big),
% \tag{A.34}
% \]
where we are using that $\mathbb{E}_{P_N}[\psi(\dataset; \theta_0, \eta_0)] = 0$. 
Finally, by Taylor’s expansion of the function $r \mapsto \mathbb{E}_{P_N}[\psi(\dataset; \theta_0 + r(\theta - \theta_0), 
\eta_0 + r(\eta - \eta_0))]$ and the fact that $\mathbb{E}_{P_N}[\psi(\dataset; \theta_0, \eta_0)] = 0$, we have
\begin{align}
\mathbb{E}_{P_N}[\psi(\dataset; \theta, \eta)] 
&= J_0(\theta - \theta_0) 
+ \partial_\eta \mathbb{E}_{P_N} \psi(\dataset; \theta_0, \eta_0)[\eta - \eta_0]\nonumber\\
&+ 0.5\int_0^1 \partial^2_r \mathbb{E}_{P_N}[\dataset; \theta_0 + r(\theta - \theta_0), 
\eta_0 + r(\eta - \eta_0)] dr. \label{eq:taylor_exp}
\end{align}

Therefore, since $\|\partial_\eta \mathbb{E}_{P_N}[\psi(\dataset; \theta_0, \eta_0)]\|\|\widehat{\eta}_0 - \eta_0\|=0$ by Neyman orthogonality, we apply \eqref{eq:prop_emperical_expect} with $\theta = \tilde{\theta}_0$ and $\eta = \widehat{\eta}_0$, and substitute in \eqref{eq:taylor_exp} to obtain that
\[
\sqrt{n}\left\|\mathbb{E}_n[\psi(\dataset; \theta_0, \eta_0)] + J_0(\tilde{\theta}_0 - \theta_0)\right\|
\leq \epsilon_N \sqrt{n} + \mathcal{I}_3 + \mathcal{I}_4 + \mathcal{I}_5 \quad\text{with probability $1 - o(1)$},
\]
where we have that
\begin{align*}
\mathcal{I}_4 &:= \sqrt{n} \sup_{\|\theta - \theta_0\| \leq \tau_N, \eta \in \mathcal{T}_N}
\left\| \int_0^1 \partial^2_r \mathbb{E}_{P_N}\psi[\dataset; 
\theta_0 + r(\theta - \theta_0), \eta_0 + r(\eta - \eta_0)] dr \right\|\\
&\leq o(\norm{\eta-\eta_0})^2 \sqrt{n}=o(N^{-1/4})^2 \sqrt{n},
\end{align*}
by \cref{condition:dml} (e) and the fact that the curvature of the score function is bounded within $\mathcal{T}_N$.

Therefore, since all singular values of $J_0$ are bounded below from zero by \cref{condition:dml} (d), it follows that
\begin{align}
\left\| J_0^{-1} \sqrt{n}\mathbb{E}_n[\psi(\dataset; \theta_0, \eta_0)] 
+ \sqrt{n}(\tilde{\theta}_0 - \theta_0) \right\| &= O\left( n^{-1/2 + 1/q} \log n+ o(n^{-1/4})^2 \sqrt{n} \right)\\
 &= O\left( o(1)+ o(n^{-1/2})n^{1/2} \right)=o(1).
\end{align}

Since our score function $\psi(\dataset; \theta_0, \eta_0)=\expectE[Y-f_{\theta_0}(X)|C]=0$ almost surely at true parameters $(\theta_0,\eta_0)$ for all $c\in\mathcal{C}$, the empirical expectation $\mathbb{E}_n[\psi(\dataset; \theta_0, \eta_0)]=0$, and thus, with probability $1-o(1)$,
\begin{align}
\left\| \sqrt{n}(\tilde{\theta}_0 - \theta_0) \right\| &=o(1),
\end{align}
as required.
\end{proof}

We will show that all these conditions are satisfied in the proof of~\cref{thm:dml}. To begin with, we prove~\cref{thm:neyman}, which shows that our score function $\orthoM$ is Neyman orthogonal.

\begin{repeatthm}{thm:neyman}[Neyman orthogonality]
The score function $\orthoM(\dataset;f,(s,g))=(s(c)-g(f,c))^2$ obeys the Neyman orthogonality conditions at $(f_0,(s_0,g_0))$.
\end{repeatthm}
\begin{proof}\label{appen:neyman}
Firstly, by~\cref{eq:h_exp}, we have $s_0(C)=g_0(f_0,C)$, thus
\begin{equation*}
\orthoM(\dataset;f_0,(s_0,g_0))=\expectE\Bigl[(s_0(C)-g_0(f_0,C))^2\Bigr]=0.
\end{equation*}
Then, we compute the derivative w.r.t. small changes in the nuisance parameters. For all $s,g\in \mathcal{S}, \mathcal{G}$,
\begin{align*}
\frac{\partial}{\partial r}&\expectE\Bigl[(s_0(C)+r\cdot s(C)-g_0(f_0,C)-r\cdot g(f_0,C))^2\Bigr]\\
=&\frac{\partial}{\partial r}\expectE\Bigl[2r(s_0(C)-g_0(f_0,C))(s(C)-g(f_0,C))+r^2(s(C)-g(f_0,C))^2\Bigr]\\
=&\expectE\Bigl[2(s_0(C)-g_0(f_0,C))(s(C)-g(f_0,C))+2r(s(C)-g(f_0,C))^2\Bigr],
\end{align*}
and, when at $r=0$, the derivative evaluates to
\begin{align*}
    \expectE\Bigl[2(s_0(C)-g_0(f_0,C))(s(C)-g(f_0,C))\Bigr] = \expectE\Bigl[0 \times(s(C)-g(f_0,C))\Bigr] = 0 \; \forall s,g\in \mathcal{S}, \mathcal{G},
\end{align*}
since $s_0(C)=\expectE[Y\lvert C]=\expectE[f_0(X)\lvert C]=g_0(f_0,C)$. Therefore, our moment function $\orthoM$ is Neyman orthogonal at $(f_0,(s_0,g_0))$.
\end{proof}

In turn, we state and prove the convergence of the nuisance parameters $\widehat{s}$ and $\widehat{g}$ with the machinery developed for the analysis of the excess risk in constrained ERM in \cref{appen:critical_radius}.

\begin{repeatlemma}{lemma:nuisances}[Convergence of nuisance parameters]
    Under \cref{assump:dml}, let $\mathcal{S}^*_N$ be the star-hull of the realisation set $\mathcal{S}_N$ of function class $\mathcal{S}$,
    \[
    \mathcal{S}^*_N = \mleft\{ C \mapsto \gamma(s(C) - s_0(C)) \;:\; s \in \mathcal{S}_N, \gamma \in [0, 1] \mright\},
    \]
    $\mathcal{P}^*_N$ be the star-hull of the realisation set $\mathcal{P}_N$ of the function class $\mathcal{P}$,
    \[
    \mathcal{P}^*_N = \mleft\{ C \mapsto \gamma(F(\cdot|C) - F_0(\cdot|C)) \;:\; F \in \mathcal{P}_N, \gamma \in [0, 1] \mright\},
    \]
    and $\mathcal{G}^*_N$ be the star-hull of the realisation set $\mathcal{G}_N$ of the function class $\mathcal{G}$,
    \[
    \mathcal{G}^*_N = \mleft\{C, f \mapsto\gamma(g(C,f)-g_0(C,f))\;:\; g\in\mathcal{G}_N,  \gamma\in[0,1]\mright\},
    \]
     where $\mathcal{S}_N$, $\mathcal{P}_N$ and $\mathcal{G}_N$ are properly shrinking neighbourhoods of the true functions $s_0$, $F_0$ and  $g_0$. Then, there exist universal constants $c_1$ and $c_2$, for which we have that with probability at least $1 - \xi$, the estimation errors are bounded as
    \begin{align*}
    \norm{\widehat{s}-s_0}_2^2 & \leq c_1 \mleft( \delta_N(\mathcal{S}_N^*)^2+\sqrt{\frac{\log(1/\zeta)}{N}} 
    \mright) ; \\
     \norm{\widehat{g} - g_0}_2^2 & \leq c_2 \mleft( \delta_N(\mathcal{P}^*_N)^2+\sqrt{\frac{\log(1/\zeta)}{N}} \mright).
    \end{align*}
\end{repeatlemma}
\begin{proof}
    The bound on estimation error for $\widehat{s}$ is straightforward and follows directly from \cref{theorem:erm_excess}. In order to prove that, we only need to show that the choice of loss function for the ERM estimator $\widehat{s}$ is Lipschitz in its first argument. To this end, we formulate the ERM as
    \[
    \widehat{s} = \argmin_{s \in \mathcal{S}} \mathcal{L}_N (s), \quad \textup{where } \mathcal{L}_N(s) = \frac{1}{N} \sum_{i=1}^N \ell(s; y_i, c_i) \textup{ and } \ell(s; y, c) = \mleft(y - s(c)\mright)^2. 
    \]
    Thus, we have,
    \[
    \mleft| \ell(s_2; y, c_2) - \ell(s_1; y, c_1) \mright| & =  \mleft| \mleft(y - s_2(c_2)\mright)^2 - \mleft(y - s_1(c_1)\mright)^2 \mright| \\
    & = \mleft| \mleft(2 y + s_2(c_2) + s_1(c_1) \mright) \mleft( s_2(c_2) - s_1(c_1) \mright) \mright| \\
    & \leq 4 B \mleft| s_2(c_2) - s_1(c_1) \mright| \numberthis{eq:erm_l2},
    \]
    where we used the fact that $y, \| s \|_\infty \leq B$ by \cref{assump:dml}. Thus, $\ell(s; y,c)$ is $4B$-Lipschitz in its argument and thus, by \cref{theorem:erm_excess}, for a universal constant $c_1$,
    \[
    \norm{\widehat{s}-s_0}_2^2 & \leq c_1 \mleft( \delta_N(\mathcal{S}_{N}^*)^2+\sqrt{\frac{\log(1/\zeta)}{N}} 
    \mright).
    \]
    For the estimation error of $\widehat{g}$, we recall that for any $F(\cdot|C) \in \mathcal{P}$, 
    \[g_{F}(f_\theta, c) \coloneqq \int f_\theta(x) F(x | C = c) dx.
    \]
    Thus, we can connect the estimation error of $\widehat{g}_{\widehat{F}}$ to the estimation error of the conditional density estimator $F$ by showing 
    \[
    \| \widehat{g} - g_0 \|_2 & = \| \widehat{g}_{\widehat{F}} - g_{F_0} \|_2 \\
    &\leq B \| \widehat{F} - F_0 \|_2 \numberthis{eq:est_P_est_G}
    \]
    To prove \eqref{eq:est_P_est_G}, we observe that for any $C$ and any test function $f\in\mathcal{F}$, we have that,
    \[
    \mleft| g_{F_1} (C, f) - g_{F_2} (C, f) \mright| & = \mleft| \int f(x) \mleft[ F_1 - F_2 \mright] (dx | C) \mright| \\
    & \leq \int \mleft| f(x) \mright|  \mleft| F_1 - F_2 \mright| (dx | C) \\
    & \leq B \int \mleft| F_1 - F_2 \mright| (dx | C), \numberthis{eq:temp_bound_on_g}
    \]
    since $\| f\|_\infty \leq B$ by \cref{assump:dml}. Thus, by integrating w.r.t. joint law of $(C, f)$, and the \eqref{eq:temp_bound_on_g}, we can show that \eqref{eq:est_P_est_G} holds since
    \[
    \| g_{F_1} - g_{F_2} \|_2^2 & = \mathbb{E}_{C} \mleft[ \mleft(g_{F_1}(C, f) - g_{F_2}(C, f) \mright)^2 \mright] \\
    & \leq B^2 \mathbb{E}_{C} \mleft[   \mleft(\int \mleft| F_1 - F_2 \mright| (dx | C) \mright)^2  \mright] \\
    & \leq B^2 \mathbb{E}_{C} \mleft[   \int \mleft[ F_1 - F_2 \mright]^2  (dx | C) \mright] \numberthis{eq:temp_bound_on_g_2}
    \\
    & = B^2 \| F_1 - F_2 \|_2^2,
    \]
    for any $F_1,F_2\in\mathcal{P}$,
    where in \eqref{eq:temp_bound_on_g_2}, we used Cauchy–Schwarz. Having \eqref{eq:est_P_est_G} at hand, it suffices to prove the upper bound on the estimation error of $\widehat{F}$. To this end, we observe that Squared CDF error,
    \[
    \ell(F; x, c) = [\mathbbm{1}_{\{X \leq x \;|\;C = c\} } - \mathscr{F}(x|c) ]^2,
    \]
    is Lipschitz by the similar argument as \eqref{eq:erm_l2}, where $\mathscr{F}(x|c)$ is the conditional CDF induced by the conditional density $F$. It is not hard to observe that \emph{clipped} versions of other losses for density estimation such as integrated squared error (ISE), negative log-likelihood, and Hellinger-squared also satisfy the Lipschitz condition in the first argument. Thus, by \cref{theorem:erm_excess}, for a universal constant $c_3$,
    \[
    \norm{\widehat{F}-F_0}_2^2 & \leq c_3 \mleft( \delta_N(\mathcal{P}_{N^*})^2+\sqrt{\frac{\log(1/\zeta)}{N}} 
    \mright).
    \]
    Choosing $c_2 = B^2 c_3$ and \eqref{eq:est_P_est_G} completes the proof.
\end{proof}

Now, we are ready to prove~\cref{thm:dml}, which is the main theorem that states the $N^{-1/2}$ convergence rate guarantees for our DML estimator.

\begin{repeatthm}{thm:dml}[Convergence of the DML estimator for CMRs]
Let $f_{\theta_0}\in\mathcal{F}$ be a solution that satisfies the CMRs in~\cref{eq:cmr}, let $\orthoM$ be the Neyman orthogonal score defined in~\cref{eq:neyman_score} and let $J_0\coloneqq\partial_{\theta^\prime}\{\expectE[\orthoM(\dataset;f_{\theta^\prime},(s_0,g_0))]\}|_{\theta^\prime=\theta_0}$ be the Jacobian matrix of $\expectE[\orthoM]$ w.r.t. $\theta$. Suppose that the upper bound of the critical radius $\delta_N=o(N^{-1/4})$ for $\widehat{s}$, $\widehat{g}$, and $J_0$ has bounded singular values. Then, if \cref{assump:parameter} and \ref{assump:dml} hold, our DML estimator $f_{\widehat{\theta}}$ satisfies that $\widehat{\theta}$ is concentrated in a $N^{-1/2}$ neighbourhood of $\theta_0$ with probability $1-o(1)$. Specifically,
for all $\zeta\in(0,1]$, there exists $K>0$ such that for all integer $N\geq K$, 
\begin{align*}
    \norm{\widehat{\theta}-\theta_0}=O(N^{-1/2})
\end{align*}
with probability $1-\zeta$.
\end{repeatthm}

\begin{proof}\label{appen:dml_guarantee}
Following \cref{prop:thm3.3}, we need to check whether, under \cref{assump:dml}, all of \cref{condition:dml} for the $N^{-1/2}$ convergence rate for DML is satisfied. Condition (a) is satisfied since $(s-g)^2$ is twice continuously differentiable with respect to $s$ and $g$. Condition (b) is satisfied by \cref{thm:neyman}. Condition (c) is satisfied since $f_{\theta_0}$ satisfies the CMRs and, from \cref{thm:neyman}, we have that $\expectE[\orthoM(\dataset;f_{\theta_0},(s_0,g_0))]=0$. In addition, from \cref{assump:parameter}, since the true parameter $\theta_0\in\Theta$ is in the interior of $\Theta$, $\Theta$ contains some neighbourhood centered at the true parameter $\theta_0$.

Condition (d) is a sufficient identifiability condition, which states that the closeness of the score function at point $\theta$ to zero implies the closeness of $\theta$
to $\theta_0$. This condition is standard in condition moment problems and implies that the \textit{ill-posedness} (see \cref{def:ill-posed}) of the CMR problem is bounded, as shown in~\cref{sec:ill-posedness}. To check condition (d), we first point out that, under analytical assumptions for $s, g$, and $h$, we can write down first order Taylor series for the score function $\expectE[\orthoM(\dataset;f_{\theta},(s_0,g_0))]$ around the point $\theta_0$,
\begin{align*}
\expectE[\orthoM(\dataset;f_{\theta},(s_0,g_0))] = \expectE[\orthoM(\dataset;f_{\theta_0},(s_0,g_0))] + J_0 (\theta - \theta_0) + O(\norm{\theta - \theta_0}^2).
\end{align*}
Plugging in the validity of the score function $\orthoM(\dataset;f_{\theta},(s_0,g_0))$, i.e.,  $\expectE[\orthoM(\dataset;f_{\theta_0},(s_0,g_0))] = 0$, we infer that
\begin{align*}
\norm{\expectE[\orthoM(\dataset;f_{\theta},(s_0,g_0))]}\gtrsim \norm{J_0(\theta-\theta_0)}.
\end{align*}
Now for identifiability, we only need to check that $J_0 J_0^T$ is non-singular, which is guaranteed by the bounded singular value of $J_0$ as stated in the Theorem.

% Condition (e) is the non-degeneracy condition for covariance of the score function $\orthoM(\dataset;f_{\theta},(s_0,g_0))$. By definition,
% \begin{align*}
%     \expectE[\orthoM(\dataset;f_{\theta},(s_0,g_0)) \orthoM(\dataset;f_{\theta},(s_0,g_0))^T] = \int \orthoM(\dataset;f_{\theta},(s_0,g_0)) \orthoM(\dataset;f_{\theta},(s_0,g_0))^T d\probP(\dataset).
% \end{align*}
% By trace trick, for each dataset $\dataset$, the only eigenvalue of $\orthoM(\dataset;f_{\theta},(s_0,g_0)) \orthoM(\dataset;f_{\theta},(s_0,g_0))^T $ is $\norm{\orthoM(\dataset;f_{\theta},(s_0,g_0))}^2 \geq 0 $, with $\orthoM(\dataset;f_{\theta},(s_0,g_0))$ as the corresponding eigenvector. Therefore, $\expectE[\orthoM(\dataset;f_{\theta},(s_0,g_0)) \orthoM(\dataset;f_{\theta},(s_0,g_0))^T]$ is positive-definite if for each member $d$ of the support of $\probP$, which is the distribution of $\dataset$, there are at least as many eigenvectors of $d$ as the number of dimensions of $\orthoM(\dataset;f_{\theta},(s_0,g_0))$, which is true in our setting as the co-domain of $\orthoM(\dataset;f_{\theta},(s_0,g_0))$ is $\realNumber$.

Condition (e) is satisfied since we have the critical radius $\delta_N=o(N^{-1/4})$, and together with \cref{lemma:nuisances}, the nuisance parameters converge sufficiently quickly to ensure $\norm{\widehat{s}-s_0}_2\leq O(\delta_N+N^{-1/2})=O(o(N^{-1/4})+ N^{-1/2})=o(N^{-1/4})$ and similarly $\norm{\widehat{g}-g_0}_2\leq O(\delta_N+N^{-1/2})=o(N^{-1/4})$. 

Therefore, all the conditions in \cref{condition:dml} are satisfied, which concludes the proof by \cref{prop:thm3.3}.
\end{proof}

\begin{repeatcoro}{coro:function_convergence}[Estimation error bounds]
Let $f_{\widehat{\theta}}$ be the DML estimator for CMRs. If all assumptions for~\cref{thm:dml} hold and there exists a constant $L>0$ such that $\norm{f_\theta(x)-f_{\theta_0}(x)}_2\leq L \norm{\theta-\theta_0}_2$ for all $x\in\mathcal{X}$ and $\theta\in\Theta$,
then, for all $\zeta\in(0,1]$, we have that, for all $\zeta\in(0,1]$, there exists $K>0$ such that for all integer $N\geq K$,
\begin{align*}
\norm{f_{\widehat{\theta}}-f_{\theta_0}}_2=O\left(L\cdot N^{-1/2}\right),
\end{align*}
with probability $1-\zeta$.
\end{repeatcoro}

\begin{proof}
From \cref{thm:dml}, we have that the parameters $\widehat{\theta}$ for our DML estimator $f_{\widehat{\theta}}$ learnt from a dataset of size $N$ satisfy that, for all $\zeta\in(0,1]$, there exist $M,K>0$ such that for all integer $N\geq K$, 
\begin{align*}
    \probP\left(\norm{\widehat{\theta}-\theta_0}\leq M\cdot N^{-1/2}\right)\geq 1-\zeta.
\end{align*}

If we let $L$ to be a constant such that $\norm{f_\theta(x)-f_{\theta_0}(x)}\leq L \norm{\theta-\theta_0}$ for all $x\in \mathcal{X}$ and $\theta\in\Theta$, we have that, for all $\zeta\in(0,1]$, there exist $M,K>0$ such that for all integer $N\geq K$,
\begin{align*}
\probP\left(\norm{f_{\widehat{\theta}}(x)-f_{\theta_0}(x)}\leq L\cdot M \cdot N^{-1/2}\right)\geq 1-\zeta,
\end{align*}
which implies $\norm{f_{\widehat{\theta}}-f_{\theta_0}}=O\left(L\cdot N^{-1/2}\right)$ with probability 1-$\zeta$ as required.
\end{proof}

\subsection{Ill-posedness and DML Identification}\label{appen:illposed}

\begin{repeatprop}{prop:ill-posed}
For all $\theta\in\Theta$, if there exists a constant $L>0$ such that $\norm{f_\theta(x)-f_{\theta_0}(x)}_2\leq L \norm{\theta-\theta_0}_2$ for all $x\in\mathcal{X}$ and $\theta\in\Theta$, then \cref{condition:dml} (d), which states 
\begin{align*}
2\norm{\expectE[\orthoM(\dataset;f_{\theta},(s_0,g_0))]}\geq \norm{J_0(\theta-\theta_0)}
\end{align*}
and the Jacobian matrix $J_0$ have singular values bounded between $c_0>0$ and $c_1>0$, implies the ill-posedness is bounded by $\ill\leq L/\sqrt{c_0}$.
\end{repeatprop}

\begin{proof}
Recall that our orthogonal score function is $\orthoM(\dataset;f_{\theta},(s,g))=(s(c)-g(f,c))^2$, where $\orthoM(\dataset;f_{\theta},(s_0,g_0))=(s_0(c)-g_0(f,c))^2=(\expectE[Y-f(X)\lvert C])^2$. Under a finite-dimensional parameterised setting, we have that from $2\norm{\expectE[\orthoM(\dataset;f_{\theta},(s_0,g_0))]}\gtrsim \norm{J_0(\theta-\theta_0)}$,
\begin{align}
\norm{\expectE[f_{\theta_0}(X)-f_\theta(X)\lvert C]}_2^2\nonumber
=&\norm{\expectE[(\expectE[f_{\theta_0}(X)-f_\theta(X)\lvert C])^2]}\nonumber\\
=&\norm{\expectE[(\expectE[Y-f_\theta(X)\lvert C])^2]}\nonumber\\
=&\norm{\expectE[\orthoM(\dataset;f_{\theta},(s_0,g_0))]}\nonumber\\
\geq& \frac{1}{2}\norm{J_0(\theta-\theta_0)}\nonumber\\
=&\sqrt{(\theta-\theta_0)^T(J_0^TJ_0)(\theta-\theta_0)}\nonumber\\
\geq& \frac{1}{2}\sqrt{c_0^2 \norm{(\theta-\theta_0)}_2^2}\geq \frac{1}{2}c_0 \norm{(\theta-\theta_0)}_2\geq \frac{1}{2}c_0 \norm{(\theta-\theta_0)}_2^2\label{eq:id_to_ill1}
\end{align}
for $\norm{(\theta-\theta_0)}\leq1$ and the singular value lower bound $c_0>0$ of $J_0$. With a local Lipschitz condition of $f_\theta$ around $\theta_0$: $\norm{f_\theta(x)-f_{\theta_0}(x)}_2\leq L\norm{\theta-\theta_0}_2$ for all $x\in\mathcal{X}$ and $\theta\in\Theta$, we have that 
\begin{align}
    \norm{f_{\theta_0}-f_\theta}_2^2&=\expectE[(f_{\theta_0}(x)-f_\theta(x))^2]\nonumber\\
    &\leq \expectE[(L\norm{\theta_0-\theta})^2]\nonumber\\
    &\leq L^2\norm{\theta_0-\theta}^2;\nonumber\\
    \implies \norm{\theta_0-\theta}^2&\geq \frac{\norm{f_{\theta_0}-f_\theta}_2^2}{L^2}.\label{eq:id_to_ill2}
\end{align}
Therefore, from~\cref{eq:id_to_ill1} and~\cref{eq:id_to_ill2}, we have that
\begin{align*}
\norm{\expectE[f_{\theta_0}(X)-f_\theta(X)\lvert C]}_2^2\geq c_0 \norm{(\theta-\theta_0)}_2^2&\geq \frac{c_0\norm{f_{\theta_0}-f_\theta}_2^2}{L^2}\\
\implies \norm{\expectE[f_{\theta_0}(X)-f_\theta(X)\lvert C]}_2\geq \sqrt{c_0} \norm{(\theta-\theta_0)}_2&\geq \frac{\sqrt{c_0}\norm{f_{\theta_0}-f_\theta}_2}{L},
\end{align*}
which bounds the ill-posedness by
\begin{align}
\ill=\sup_{f\in\mathcal{F}} \frac{\norm{f_{\theta_0}-f_\theta}_{2}}{\norm{\expectE[f_{\theta_0}(X)-f_\theta(X) \lvert C]}_{2}}\leq \frac{L}{\sqrt{c_0}}.
\end{align}
\end{proof}

\subsection{Constrained Empirical Risk Minimisation Bounds} \label{appen:critical_radius}

In this section, we introduce basic concepts from empirical process theory and further discuss bounds on the excess risk of general empirical risk minimizers (ERM) in the style of \citet{Wainwright2019,Foster2019OrthogonalLearning}.

\begin{definition}
    The critical radius denoted by $\delta_N(\mathcal{H}^*)$ is defined as the minimum $\delta$ that satisfies the following upper bound on the local Gaussian complexity of a star-shaped function class $\mathcal{H}^*$\footnote{A function class 
  $\mathcal{H}$ is star-shaped if for every $h \in \mathcal{H}$ and $\alpha \in [0, 1]$, we have $\alpha h \in \mathcal{H}$.}, $\mathcal{G(\mathcal{H}^*, \delta)} \leq {\delta^2}/2$, where local Gaussian complexity is defined as
    \begin{align*}
    \mathcal{G(\mathcal{H}^*, \delta)} = \expectE_{\epsilon}\left[\sup_{h \in \mathcal{H}^*: \norm{h}_N \leq \delta}  \langle \epsilon, h \rangle \right], \numberthis{eq;fixed_point}
    \end{align*}
    with $\epsilon$ being a random i.i.d. zero-mean Gaussian vector.
\end{definition}
The critical radius is a standard notion to bound the estimation error in the regression problem. Since local Gaussian complexity can be viewed as an expected value of a supremum of a stochastic process indexed by $g$, we can apply empirical process theory tools, namely the Dudley's entropy integral~\citep{Wainwright2019,van2014probability}, to provide a bound on the critical radius,
\begin{align*}
    \mathcal{G(\mathcal{H}^*, \delta)} \leq \inf_{\alpha \geq 0} \left \{\alpha + \frac{1}{\sqrt{N}}  \int_{\alpha/4}^{\delta} \sqrt{\log \mathcal{N}(\mathcal{H}^*, L^2(\mathbb{P}_N), \epsilon)}\:d\epsilon\right \},
\end{align*}
where $\mathcal{N}(\mathcal{H}^*, L^2(\mathbb{P}_N), \epsilon)$ is the $\epsilon$-covering number of the function class $\mathcal{H}^*$ in $L^2(\mathbb{P}_N)$ norm. Now, by placing $\alpha = 0$, when the integral is a single scale value of $ \sqrt{\log \mathcal{N}(\mathcal{H}^*, L^2(P_n), \epsilon)}$, we infer that
\begin{align*}
    \mathcal{G(\mathcal{H}^*, \delta)} \leq \frac{\delta}{\sqrt{N}} \sqrt{\log \mathcal{N}(\mathcal{F}^*, L^2(\mathbb{P}_N), \epsilon)}.
\end{align*}
Thus, the critical radius of $\mathcal{H}^*$ will be upper bounded by
\begin{align*}
    \delta_N(\mathcal{H}^*) \lesssim \frac{\sqrt{\log \mathcal{N}(\mathcal{H}^*, L^2(\mathbb{P}_N), \epsilon)}}{\sqrt{N}} = O(d_N (\mathcal{H}^*)^{1/2} N^{-1/2}),
\end{align*}
where \citet{Chernozhukov2022RieszNetForests,Chernozhukov2021AutomaticRegression} referred to 
\[
d_N (\mathcal{H}^*) \coloneqq \inf\mleft\{ d > 0: \log \mathcal{N}(\mathcal{H}^*, L^2(\mathbb{P}_N), \epsilon) \leq d \log \mleft(\frac{C}{\epsilon} \mright)\; \forall \epsilon \in (0, 1) \textrm{ and } C \textrm{ is a constant.}\mright\},
\]
as the effective dimension of the hypothesis space. Note that this rate matches the minimax lower bound of fixed design estimation for this setting~\citep{yang1999information}. 

% We recall the following classic result, regarding convergence of empirical process, induced by realisation of a star-shaped and bounded function class.

Given the dataset $\mathcal{D} = \{z_i \in \mathcal{Z} \}_{i = 1}^N$ consisting of i.i.d. data points $z_i$ drawn from distribution $\mathbb{P}$, and a function class $\mathcal{H}$, we define the realisation of a function space by subscript $N$, e.g., $\mathcal{H}_N$ is the realisation of $\mathcal{H}$ in the $N$ observed data points. Since the definition of critical radius is for star-shaped function classes, we equip ourselves with the star-hull notation, where $\mathcal{H}^*_N$ is the star-hull of the function class $\mathcal{H}_N$ centred at the true function $h_0$, defined as
\[
    \mathcal{H}_N^* \coloneqq \mleft\{ Z \mapsto \gamma \mleft( h(X) - h_0 (Z) \mright)\; : \; f \in \mathcal{H}_N, \gamma \in [0, 1] \mright\},
\]
and denote its critical radius by $\delta_N(\mathcal{H}_N^*)$, or simply $\delta_N(\mathcal{H})$. In statistical learning, we are given a loss function $\ell: \mathcal{H} \times \mathcal{Z} \rightarrow \mathbb{R}$, and we have the risk and empirical risk defined accordingly as:
\[
\mathcal{L} (h) = \mathbb{E}_{z} [\ell(h; z)]  \quad \textrm{ and } \quad \mathcal{L}_N (h) = \frac{1}{N} \sum_{i = 1}^N \ell(h; z_i). 
\]
Let us denote the ground truth function by $h_0$, i.e., the minimizer of the risk, as:
\[
h_0 = \argmin_{h \in \mathcal{H}} \mathcal{L}(h).
\]
The ERM algorithm proposes the estimator $\widehat{h}$ that minimises the empirical risk of the observed $N$ data points,
\[
\widehat{h} = \argmin_{h \in \mathcal{H}} \mathcal{L}_N(h).
\]

% We define the operators $\mathbb{P}_N$ and $\mathbb{P}$ as functionals that takes an input a function $h \in \mathcal{H}$, and returns the empirical average and expected value of $h$ under distribution $\mathbb{P}$. In terms,
% \[
% \mathbb{P}_N h = \frac{1}{N} \sum_{i = 1}^N h(z_i) \quad \textrm{ and } \quad \mathbb{P} h = \mathbb{E}_{z} [h(z)]. 
% \]

\begin{theorem}[{\citet[Lemma 7]{Foster2019OrthogonalLearning}}] \label{thm:foster_emp}
    Consider a function class $\mathcal{H}$ and its star-hull $\mathcal{H}^*$, with $\sup_{h \in \mathcal{H}^*} \| h \|_\infty \leq 1$, critical radius $\delta_N(\mathcal{H}^*)$, and any choice of $\delta$ such that,
    \[ 
    \delta^2 \geq \max \mleft\{ \delta_N(\mathcal{H}^*)^2, \frac{4 \log (41 \log(2c_1 N))}{c_1 N} \mright\},
    \] 
    for a constant $c_1$. Moreover, assume that the loss function $\ell$ is $L$-Lipschitz in its first argument with respect to $\ell_2$ norm. Then there exist universal constants $c_2$ and $c_3$ such that with probability at least $ 1- c_2 \exp\{ c_3 N \delta^2\}$,
    \[
    \mleft| \mleft( \mathcal{L}_N(h) - \mathcal{L}_N(h_0) \mright) - \mleft( \mathcal{L}(h) - \mathcal{L}(h_0)\mright) \mright| \leq 18 L \delta 
 \mleft\{ \| h - h_0 \|_2 + \delta \mright\}, \qquad \forall h \in \mathcal{H}.
    \]
\end{theorem}

We mention an immediate corollary of \cref{thm:foster_emp} for functions $h$ that are bounded by $B$, which follows by rescaling arguments.

\begin{corollary} \label{corr:emp_us}
    Consider a function class $\mathcal{H}$ and its star-hull $\mathcal{H}^*$, with $\sup_{h \in \mathcal{H}^*} \| h \|_\infty \leq B$, critical radius $\delta_N(\mathcal{H}^*)$, and any choice of $\delta$ such that,
    \[ 
    \delta^2 \geq \max \mleft\{ \delta_N(\mathcal{H}^*)^2, \frac{4 \log (41 \log(2c_1 N))}{c_1 N} \mright\},
    \] 
    for a constant $c_1$. Moreover, assume that the loss function $\ell$ is $L$-Lipschitz in its first argument with respect to $\ell_2$ norm. Then there exist universal constants $c_2$ and $c_3$ such that with probability at least $ 1- c_2 \exp\{ c_3 N \delta^2/\max\{1, B\}^2\}$,
    \[
    \mleft| \mleft( \mathcal{L}_N(h) - \mathcal{L}_N(h_0) \mright) - \mleft( \mathcal{L}(h) - \mathcal{L}(h_0)\mright) \mright| \leq \frac{18 L \delta}{\max\{1, B\}} 
 \mleft\{ \| h - h_0 \|_2 + \delta \mright\}, \qquad \forall h \in \mathcal{H}.
    \]
\end{corollary}
\begin{proof}
    If $B \leq 1$, the proof follows trivially from \cref{thm:foster_emp}, thus let us assume otherwise. Define the $B$-scaled function class $\widetilde{\mathcal{H}}$,
    \[
    \widetilde{\mathcal{H}} \coloneqq \mleft\{ \widetilde{h} \;:\; B \widetilde{h} \in \mathcal{H} \mright\}.
    \]
    Then, the loss function $\ell(h,z) = \ell(B \widetilde{h}, z)$ is $LB$-Lipschitz in $\widetilde{h}$ and by homogeneity of the local Gaussian averages, i.e.,
    \[
    \mathcal{G}(\widetilde{\mathcal{H}}^*, r) = \frac{1}{B} \mathcal{G}(\mathcal{H}^*, B r),
    \]
     wee see that $\widetilde{\delta} \coloneqq \delta / B$ satisfies the fixed point condition~\eqref{eq;fixed_point}. Putting these pieces together and invoking \cref{thm:foster_emp} for $\widehat{\mathcal{H}}$ completes the proof.
\end{proof}

We can equivalently write \cref{corr:emp_us} in the failure probability format. That is, for a target failure probability $0 < \xi < 1$, define 
\[
\delta(\xi) \coloneqq \delta_N(\mathcal{H^*}) + \max\{1, B\} \sqrt{\frac{1}{c_3 N} \log (1/\xi)},
\]
then with probability at least $1 - \xi$,
\[
\mleft| \mleft( \mathcal{L}_N(h) - \mathcal{L}_N(h_0) \mright) - \mleft( \mathcal{L}(h) - \mathcal{L}(h_0)\mright) \mright| \leq \frac{18 L \delta(\xi)}{\max\{1, B\}} 
 \mleft\{ \| h - h_0 \|_2 + \delta(\xi) \mright\}, \qquad \forall h \in \mathcal{H}. \numberthis{eq:emp_failure}
\]

Now, we are ready to state and prove the following master theorem for the excess risk of the constrained ERM. For the analysis, in line with \citet{Chernozhukov2021AutomaticRegression}, we require that the population risk has positive curvature for identifiability purposes. Then, the generalisation bound in terms of excess risk can be converted into estimation error.

\begin{theorem}[Estimation Error of Constrained ERM] \label{theorem:erm_excess}
    Assume that the population risk $\mathcal{L}$ has a positive curvature, i.e., for a positive number $\lambda$,
    \[
    \mathcal{L}(h) - \mathcal{L}(h_0) \geq \frac{\lambda}{2} \| h - h_0 \|_2^2 \qquad \forall h \in \mathcal{H}, \numberthis{eq:curve}
    \]
    and the loss function $| \ell(h; z) | \leq M$ is bounded and $L$-Lipschitz in its first argument w.r.t. $\ell_2$ norm.
    Then, the solution to the ERM algorithm:
    \[
    \widehat{h} \coloneqq \argmin_{h \in \mathcal{H}} \mathcal{L}_N(h), \numberthis{eq:erm_min}
    \]
    has the following estimation error with probability at least $1 - \xi$,
    \[
    \| \widehat{h} - h_0 \|_2^2 \leq C \mleft[  \mleft(  \mathcal{L}(h^*) - \mathcal{L}(h_0) \mright) + \delta_N(\mathcal{H}^*)^2 + \sqrt{\frac{\log(1/\xi)}{N}} \mright].
    \]
    for a universal constant $C$, where
    $
    h^* = \arg\inf_{h \in \mathcal{H}} \mathcal{L}(h).
    $
    In addition, if $h_0$ is realisable, i.e., $h_0 \in \mathcal{H}$, then,
    \[
        \| \widehat{h} - h_0 \|_2^2 \leq C \mleft[ \delta_N(\mathcal{H}^*)^2 + \sqrt{\frac{\log(1/\xi)}{N}} \mright].
    \]
\end{theorem}
\begin{proof}
    First, we want to upper bound the population excess risk,
    $
    \mathcal{L} (\widehat{h}) - \mathcal{L} (h_0),
    $
    and then, by the curvature of $\mathcal{L}$, we can convert this bound into an upper bound on the estimation error $\| \widehat{h} - h_0 \|_2$.

    To begin with, we have,
    \[
    \mathcal{L} (\widehat{h}) - \mathcal{L} (h_0) & = \underbrace{{\Big( \mathcal{L}(\widehat{h}) - \mathcal{L}_N (\widehat{h}) \Big)}}_\textup{(I)} + \underbrace{\Big( \mathcal{L}_N (\widehat{h}) - \mathcal{L}_N(h^*) \Big) }_{\textup{(II)}} + \underbrace{\Big( \mathcal{L}_N (h^*) - \mathcal{L}(h_0) \Big)}_{\textup{(III)}}.
    \]
    By ERM algorithm~\eqref{eq:erm_min}, we know that,
    \[
    \mathcal{L}_N(\widehat{h}) \leq \mathcal{L}_N(h^*),
    \]
    Hence, $\textup{(II)} \leq 0$. For the term $\textup{(I)}$,
    \[
    \textup{(I)} & =  \mathcal{L}(\widehat{h}) - \mathcal{L}_N (\widehat{h}) \\
    & = (\mathcal{L} - \mathcal{L}_N) (\widehat{h} - h_0) + (\mathcal{L} - \mathcal{L}_N) (h_0),
    \]
    and for the term $\textup{(III)}$, we know that,
    \[
    \textup{(III)} & = \mathcal{L}_N (h^*) - \mathcal{L}(h_0) \\
    & = ( \mathcal{L}(h^*) - \mathcal{L}(h_0) ) + (\mathcal{L}_N - \mathcal{L}) (h^*) \\
    & = ( \mathcal{L}(h^*) - \mathcal{L}(h_0) ) + (\mathcal{L}_N - \mathcal{L}) (h^* - h_0) + (\mathcal{L}_N - \mathcal{L}) (h_0).
    \]
    Summing these terms from $(\textup{I}), (\textup{II})$ and $(\textup{III})$ yields,
    \[
    \mathcal{L} (\widehat{h}) - \mathcal{L} (h_0) & \leq ( \mathcal{L}(h^*) - \mathcal{L}(h_0) ) + \underbrace{(\mathcal{L} - \mathcal{L}_N) (\widehat{h} - h_0)}_{\epsilon_1(\widehat{h})} + \underbrace{(\mathcal{L}_N - \mathcal{L}) (h^* - h_0)}_{\epsilon_2}.
    \]
    From \cref{corr:emp_us}, especially the formulation in \eqref{eq:emp_failure}, we know that with probability at least $1 - \xi/2$,
    \[
    \sup_{h \in \mathcal{H}} \mleft| (\mathcal{L}_N - \mathcal{L}) (h - h_0)\mright| \leq \frac{18 L \delta(\xi/2)}{\max\{1, B\}} \mleft( \|h - h_0 \|_2 + \delta(\xi/2) \mright),
    \]
    where,
    \[
    \delta(\xi/2) = \delta_N(\mathcal{H^*}) + \max\{1, B\} \sqrt{\frac{1}{c_3 N} \log (2/\xi)}. \numberthis{eq:delta_xi_half}
    \]
    Thus, we can upper bound $\epsilon_1 (\widehat{h})$ as,
    \[
    \epsilon_1 (\widehat{h}) & = (\mathcal{L} - \mathcal{L}_N) (\widehat{h} - h_0) \leq   \frac{18 L \delta(\xi/2)}{\max\{1, B\}} \mleft( \|\widehat{h} - h_0 \|_2 + \delta(\xi/2) \mright).
    \]
    To bound the term $\epsilon_2$, define the random variable,
    \[
    X_i = \ell(h^*; z_i) - \ell(h_0; z_i),
    \]
    where $X_i$ is bounded as $|X_i| \leq 2 M$. Then, by Hoeffding's inequality, with probability at least $1 - \xi/2$, we infer that
    \[
    \epsilon_2 & = (\mathcal{L}_N - \mathcal{L}) (h^* - h_0)  \leq 2 \sqrt{2} M \sqrt{\frac{\log(4/\xi)}{N}}.
    \]
    Thus, by union bound and \eqref{eq:curve}, we conclude that with probability at leat $1 - \xi$
    \[
    \frac{\lambda}{2} \| \widehat{h} - h_0 \|_2^2 & \leq ( \mathcal{L}(h^*) - \mathcal{L}(h_0) ) + \frac{18 L \delta(\xi/2)}{\max\{1, B\}} \mleft( \|\widehat{h} - h_0 \|_2 + \delta(\xi/2) \mright) + 2 \sqrt{2} M \sqrt{\frac{\log(4/\xi)}{N}} \\
    & = ( \mathcal{L}(h^*) - \mathcal{L}(h_0) ) + \frac{18 L \delta(\xi/2)}{\max\{1, B\}}  \|\widehat{h} - h_0 \|_2 + \frac{18 L \delta(\xi/2)^2}{\max\{1, B\}}  + 2 \sqrt{2} M \sqrt{\frac{\log(4/\xi)}{N}} \\
    & \leq ( \mathcal{L}(h^*) - \mathcal{L}(h_0) ) + \frac{\lambda}{4} \|\widehat{h} - h_0 \|_2^2
    + 
    \frac{1}{\lambda} \mleft( \frac{18 L \delta(\xi/2)}{\max\{1, B\}} \mright)^2  \\
    & \qquad \qquad  + \frac{18 L \delta(\xi/2)^2}{\max\{1, B\}} + 2 \sqrt{2} M \sqrt{\frac{\log(4/\xi)}{N}} \numberthis{eq:erm_temp1} \\
    \]
    where \eqref{eq:erm_temp1} follows by Young's inequality, i.e., for any positive number $\lambda > 0$,
    \[
    ax \leq \frac{\lambda}{4} x^2 + \frac{a^2}{\lambda}.
    \]
    Therefore, by cancelling out $\frac{\lambda}{4} \|\widehat{h} - h_0 \|_2^2$ from both sides,
    \[
    \| \widehat{h} - h_0 \|_2^2 \leq C_1 \mleft[ \frac{1}{\lambda} \mleft(  \mathcal{L}(h^*) - \mathcal{L}(h_0)  \mright) + \mleft( \frac{L^2}{\lambda \max\{1, B\}^2} + \frac{L}{\max\{1, B\}}\mright) \delta(\xi/2)^2\mright.\\
    +\mleft. M \sqrt{\frac{\log(1/\xi)}{N}} \mright],
    \]
    for a constant $C_1$.
    In turn, by \eqref{eq:delta_xi_half} and Young's inequality, we know that,
    \[
    \delta(\xi/2)^2 \leq C_2 \mleft[ \delta_N(\mathcal{H}^*)^2 + \frac{\max\{1, B\}^2}{N} \log(1/\xi) \mright],
    \]
    for a constant $C_2$. Hence,
    \[
    \| \widehat{h} - h_0 \|_2^2 & \leq C_3 \Bigg[ \frac{1}{\lambda} \mleft(  \mathcal{L}(h^*) - \mathcal{L}(h_0)  \mright) \\
    & \qquad + \mleft( \frac{L^2}{\lambda \max\{1, B\}^2} + \frac{L}{\max\{1, B\}}\mright) \mleft[ \delta_N(\mathcal{H}^*)^2 + \max\{1, B\}^2 \frac{ \log(1/\xi)}{N} \mright] \\
    & \qquad + M \sqrt{\frac{\log(1/\xi)}{N}} \Bigg] \\
    & \leq C_4 \Bigg[ \frac{1}{\lambda} \mleft(  \mathcal{L}(h^*) - \mathcal{L}(h_0)  \mright) + \frac{\max\{L^2, L\}}{\min\{\lambda, 1\} \max\{1 , B\}}  \delta_N(\mathcal{H}^*)^2 \\
    & \qquad + \frac{\max\{L^2, L\} \max\{1 , B\}}{\min\{\lambda, 1\}} \frac{\log(1/\xi)}{N} + M \sqrt{\frac{\log(1/\xi)}{N}} \Bigg], \numberthis{eq:erm_final}
    \]
    for constants $C_3$ and $C_4$.
    
    Therefore, by hiding dependence on the constants, $\lambda, B, L$ and $M$ in a universal constant $C$, we can summarise the result~\eqref{eq:erm_final} and conclude the proof,
    \[
    \| \widehat{h} - h_0 \|_2^2 \leq C \mleft[  \mleft(  \mathcal{L}(h^*) - \mathcal{L}(h_0) \mright) + \delta_N(\mathcal{H}^*)^2 + \sqrt{\frac{\log(1/\xi)}{N}} \mright].
    \]
\end{proof}

\subsection{Suboptimaltiy of the Decision Policy}

We now present the proof of \cref{coro:subopt}, which bounds the suboptimality of the policy learnt using DML-CMR.

\begin{repeatcoro}{coro:subopt}
Let the learnt policy from a dataset of size $N$ be $\widehat{\pi}(c)\coloneqq\argmax_a f_{\widehat{\theta}}(c,a)$, where $\widehat{\theta}$ is the DML-CMR estimator. Let $L$ be a constant such that $\norm{f_\theta(C,A)-f_{\theta_0}(C,A)}_2\leq L \norm{\theta-\theta_0}_2$ for all $C$ in the support of $\testD$, $A\in\mathcal{A}$, and $\theta\in\Theta$. Then, for all $\zeta\in(0,1]$, we have that there exists $K>0$ such that, for all integer $N\geq K$, the suboptimality of $\widehat{\pi}$ satisfies
\begin{align*}
\textrm{subopt}(\widehat{\policy}_N)=O(L\cdot N^{-1/2})
\end{align*}
with probability $1-\zeta$.
\end{repeatcoro}

\begin{proof}\label{appen:subopt}
From \cref{thm:dml}, we have that the parameters $\widehat{\theta}$ for our DML estimator $f_{\widehat{\theta}}$ satisfy that, for all $\zeta\in(0,1]$, there exist $K,M>0$ such that for all integer $N\geq K$, 
\begin{align*}
    \probP\left(\norm{\widehat{\theta}-\theta_0}\leq M \cdot N^{-1/2}\right)\geq 1-\zeta.
\end{align*}

Since we have that $\norm{f_\theta(C,A)-f_{\theta_0}(C,A)}\leq L \norm{\theta-\theta_0}$ for all $C,A\in \textrm{supp}^M(C,A)$ and $\theta\in\Theta$, we have that, for all $\zeta\in(0,1]$, there exist $M,K>0$ such that for all integer $N\geq K$,
\begin{align}
\probP\left(\norm{h_{\widehat{\theta}}(C,A)-h_{\theta_0}(C,A)}\leq L\cdot M \cdot N^{-1/2}\right)\geq 1-\zeta \quad \forall C,A\in \textrm{supp}^M(C,A).\label{eq:bound_h}
\end{align}

Next, we can show that the suboptimality of $\widehat{\policy}$ satisfies
\begin{align}
\textrm{subopt}(\widehat{\policy})
&=\Value(\policy^*)-\Value(\widehat{\policy})\nonumber \\
&=\expectE_{C\sim\testD}[R\mid C,do(A=\policy^*(c))]-\expectE_{C\sim\testD}[R\mid C,do(A=\widehat{\policy}(c))]\nonumber \\
&=\expectE_{C\sim\testD}[f_r(C,\policy^*(C))-f_r(C,\widehat{\policy}(C))]\nonumber \\
&=\expectE_{C\sim\testD}[h(C,\policy^*(C))-h(C,\widehat{\policy}(C))]\nonumber \\
&\leq \max_{c\in \textrm{supp}(\testD)} \left(h(c,\policy^*(c))-h(c,\widehat{\policy}(c))\right)\nonumber \\
&\leq \max_{c\in \textrm{supp}(\testD)}\abs{h(c,\policy^*(c))-h_{\widehat{\theta}}(c,\policy^*(c))}+(h_{\widehat{\theta}}(c,\policy^*(c))-h_{\widehat{\theta}}(c,\widehat{\policy}(c)))\\
&+\abs{h_{\widehat{\theta}}(c,\widehat{\policy}(c))-h(c,\widehat{\policy}(c))}\nonumber \\
&\leq 2L\cdot M\cdot N^{-1/2} \quad\text{ with probability } 1-\zeta, \label{eq:bound_wp}
\end{align}
by \cref{eq:bound_h} and the fact that $h_{\widehat{\theta}}(C,\policy^*(C))-h_{\widehat{\theta}}(C,\widehat{\policy}(C))\leq0$, where $\textrm{supp}(\testD)$ is the support of $\testD$. This implies that $\textrm{subopt}(\widehat{\policy}_N)=O(L\cdot N^{-1/2})$ with probability $1-\zeta$ as required.
\end{proof}

\section{Datasets Details}\label{appen:DML-IV-dataset}
In this section, we provide details of the datasets considered in this paper for IV regression and proximal causal learning tasks.

\subsection{IV Regression}\label{appen:demand}
We first provide the details for IV regression benchmarking datasets. Recall that we denote $A$ as the action, $Y$ as the outcome, $Z$ as the instrument, and $X$ as additional observed context, and the CMR problem we solve is $\expectE[Y-f(A,X)\lvert Z,X]=0$.

\subsubsection{Ticket Demand Dataset}

Here, we describe the ticket demand dataset for IV regression, first introduced by~\citet{Hartford2017DeepPrediction}. The observable variables are generated by the following model:
\begin{align*}
r&=f_0((t,s),p)+\epsilon, \quad \expectE[\epsilon\lvert t,s,p]=0;\\
p&=25+(z+3)\psi(t)+\omega,
\end{align*}
where $r$ is the ticket sales (as the outcome variable $Y$) and $p$ is the ticket price (as the action variable $A$). $(t,s)$ are observed context variables, where $t$ is the time of year and $s$ is the customer type. The fuel price $z$ is introduced as an instrumental variable, which only affects the ticket price $p$. The noises $\epsilon$ and $\omega$ are correlated with correlation $\rho\in[0,1]$, where in our experiments we set $\rho=0.9$. $f_0$ is the true causal effect function, defined as
\begin{align*}
    f_0((t,s),p)&=100+(10+p)\cdot s \cdot \psi(t)-2p,\\
    \psi(t)&=2\left(\frac{(t-5)^4}{600}+\exp(-4(t-5)^2)+\frac{t}{10}-2\right),
\end{align*}
where $\psi(t)$ is a complex non-linear function of $t$ plotted in~\cref{fig:nonlinear}. The offline dataset is sampled with the following distributions:
\begin{align*}
    s&\sim \text{Unif}\{1,...,7\}\\
    t&\sim \text{Unif}(0,10)\\
    z&\sim \mathcal{N}(0,1)\\
    \omega&\sim \mathcal{N}(0,1)\\
    \epsilon&\sim \mathcal{N}(\rho\omega,1-\rho^2).
\end{align*}
From the observations $(r,p,t,s,z)$, we estimate $\widehat{h}$ using IV regression methods, and the mean squared error between $\widehat{h}$ and the true causal function $f_0$ is computed on 10000 random samples from the above model. For the out-of-distribution test samples, we sample $t\sim \text{Unif}(1,11)$ instead.

We standardise the action and outcome variables $p$ and $r$ to centre the data around a mean of zero and a standard deviation of one following~\citet{Hartford2017DeepPrediction}. This is standard practice for DNN training, which improves training stability and optimization efficiency.

\begin{figure}[tb]
    \centering
\includegraphics[width=0.5\textwidth]{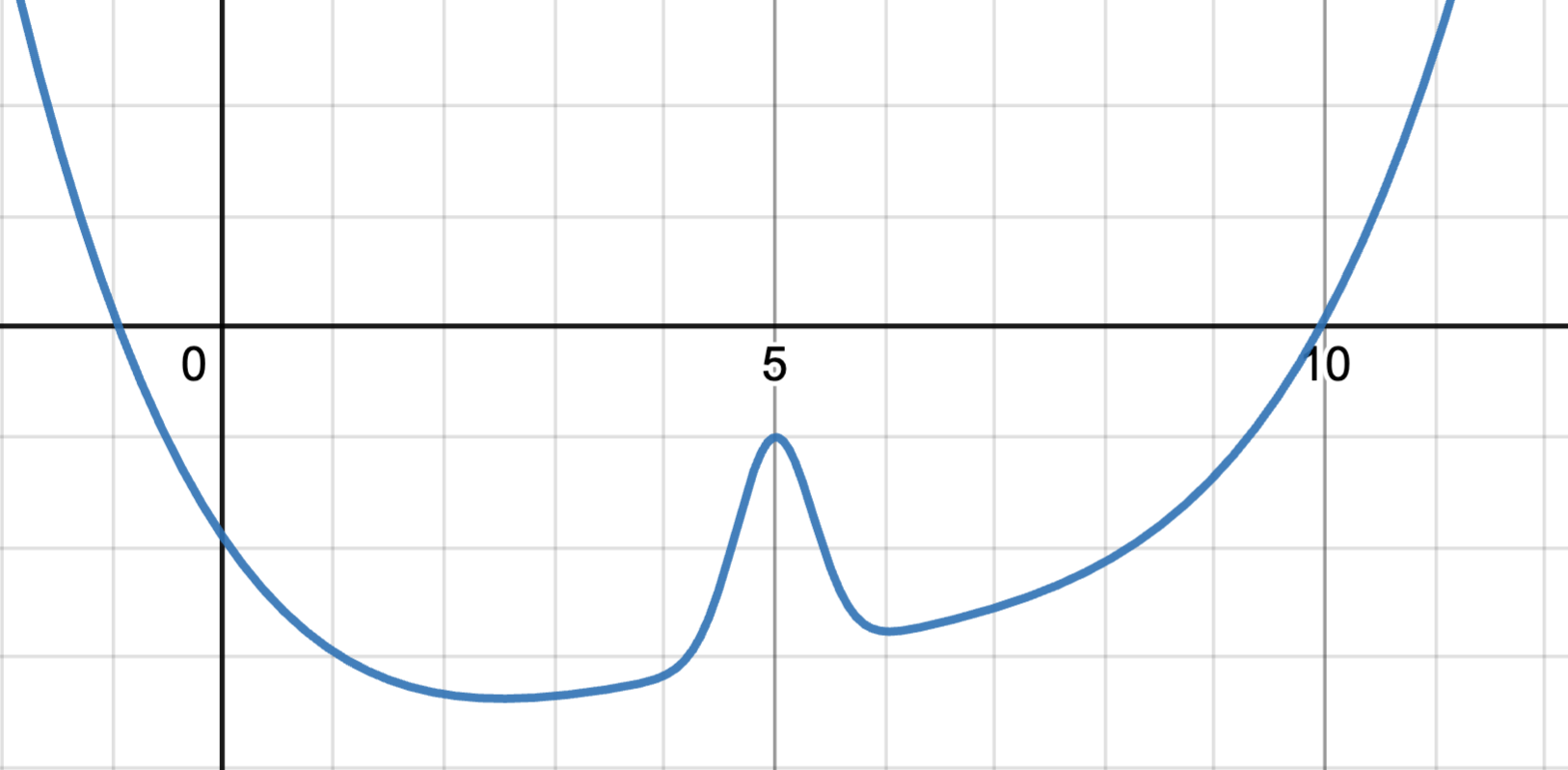}
    \caption[A graph of the causal function in the ticket demand dataset]{A graph of the nonlinear function $\psi(t)$ in the ticket demand dataset for IV regression.}
    \label{fig:nonlinear}
\end{figure}

\subsubsection{Ticket Demand High-Dimensional Setting}

For the high-dimensional setting, we again follow~\citet{Hartford2017DeepPrediction} to replace the customer type $s\in[7]$ in the low-dimensional setting with images of the corresponding handwritten digits from the MNIST dataset~\citep{LeCun2010}. For each digit $d\in[7]$, we select a random MNIST image from the digit class $d$ as the new customer type variable $s$. The images are $28\times28=784$ dimensional.

\subsubsection{Real-World Datasets}

Following previously studied causal inference methods~\citep{Shalit2017,Wu2023,Schwab2019,Bica2020}, we consider two semi-synthetic real-world datasets IHDP\footnote{IHDP: \url{https://www.fredjo.com/}.}~\citep{Hill2011} and PM-CMR\footnote{PM-CMR:\url{https://doi.org/10.23719/1506014}.}~\citep{Wyatt2020} for experiments, since the true counterfactual prediction function is rarely available for real-world datasets. 

IHDP, the Infant
Health and Development Program (IHDP), comprises 747
units with 6 pre-treatment continuous variables, one action variable, and 19 discrete variables related to the children and their mothers, aiming at evaluating the effect of specialist home visits on the future cognitive test scores of premature infants. From the original data, we select all 6 continuous covariance variables as our context variable $X$.

PM-CMR studies the
impact of PM2.5 particle level on the cardiovascular mortality rate (CMR) in 2132 counties in the United States using data provided by the National Studies on Air Pollution and Health~\citep{Wyatt2020}. We use 6 continuous variables
about CMR in each city as our context variable $X$.

Following~\citet{Wu2023}, from the context variables $X$ obtained from real-world datasets, we generate the instrument $Z$, the action $A$, and the outcome $Y$ using the following model:
\begin{align*}
&Z\sim\probP(Z=z)=1/K,\quad z\in[1..K];\\
&A=\sum_{z=1}^K 1_{Z=z} \sum_{i=1}^{d_X}w_{iz}(X_i+0.2\epsilon+f_z(z))+\delta_A, \quad w_{iz}\sim\text{Unif}(-1,1);\\
&Y=9A^2-1.5A+\sum_{i=1}^{d_X} \frac{X_i}{d_X}+\abs{X_1 X_2}-\sin{(10+X_2 X_3)}+2\epsilon+\delta_Y,
\end{align*}
where $X_i$ denotes the $i$-th variable in $X$, $f_z$ is a function that returns different constants depending on the input $z$, $\delta_Y,\delta_A\sim\mathcal{N}(0,1)$ and $\epsilon\sim\mathcal{N}(0,0.1)$ is the unobserved confounder. The fully generated semi-synthetic datasets IHDP and PM-CMR have 747 and 2132 samples respectively, and we randomly split them into training (63\%), validation (27\%), and
testing (10\%) following~\citet{Wu2023}.

\subsection{Proximal Causal Learning}\label{dataset:pcl}
Next, we provide details for benchmarking datasets of proximal causal learning. Recall that we denote $A$ as the treatment, $Y$ as the outcome, $W$ as the outcome proxy, and $V$ as the treatment proxy. The CMR problem we are solving is $\expectE[Y-f(A,W)\lvert A,V]$.

\subsubsection{Ticket Demand Dataset}

The ticket demand dataset~\citep{Hartford2017DeepPrediction} is also extended to the PCL setting, which is first introduced by~\citet{Xu2021}. The data generating process is described by the following model:
\begin{align*}
U&\sim \text{Unif}(0,10)\\
[V_1,V_2]&=[2\sin(2\pi U/10)+\epsilon_1, 2\cos(2\pi U/10)+\epsilon_2]\\
W&= 7g(U)+45+\epsilon_3\\
A&=35+(V_1+3)g(U)+V_2+\epsilon_4\\
Y&=A\cdot \min(\exp(\frac{W-A}{10},5)-5g(U)+\epsilon_5\\
\text{with}\quad g(u)&=2(\frac{(u-5)^4}{600}+\exp(-4(u-5)^2)+u/10-2)\\
\text{and}\quad \epsilon_i&\sim \mathcal{N}(0,1),
\end{align*}
where $U$ is the demand, which acts as the hidden confounder, $V_1, V_2$ are fuel prices, which act as treatment proxy, $W$ is the web page views, which acts as outcome proxy, $A$ is the price, and $Y$ is the sale. Here, we can see that the outcome proxy $W$ and the treatment proxy $V$ are both affected by $U$, where $W$ directly affects the outcome and $V$ directly affects the treatment $A$.

\subsubsection{dSprites high-dimensional Dataset}\label{appen:pcl_highdim}

The dSprites dataset~\citep{dsprites17} is a high-dimensional $(64\times64)$ image dataset described by five latent parameters: \textit{shape}, \textit{scale}, \textit{rotation}, \textit{posX} and \textit{poxY}. It is proposed by~\citet{Xu2021} to adopt it as a benchmark for PCL where the treatment is each figure and the hidden confounder is \textit{posY}. For the experiments, we fix the \textit{shape} to be heart.

The data-generating process can be described by the following steps:
\begin{enumerate}
    \item Randomly generate values for \textit{scale, rotation, posX} and \text{posY}: \\\textit{scale} $\sim\text{Unif}\{0.5,0.6,...,1.0\}$, \textit{rotation} $\sim\text{Unif}(0,2\pi)$, \textit{posX, posY} $\sim\text{Unif}\{0,...,31\}$.
    \item Set $U$=\textit{posY}
    \item Set $V$=\textit{(scale,rotation,posX)}
    \item Set $A$ as the dSprites image with features \textit{(scale, rotation, posX, posY)} and add Gaussian noise $\mathcal{N}(0,0.1)$ to each pixel.
    \item Set $W$ as posY with Gaussian noise $\mathcal{N}(0,1)$.
    \item $Y=\frac{0.1\norm{vec(A)^TB}^2_2-5000}{1000}\times\frac{(31\times U-15.5)^2}{85.25}+\epsilon,\epsilon\sim\mathcal{N}(0,0.5)$, where the matrix $B\in\realNumber^{64\times64}$ is given by $B_{i,j}=\abs{32-j}/32$.
\end{enumerate}

\begin{figure}
    \centering
\includegraphics[width=0.5\linewidth]{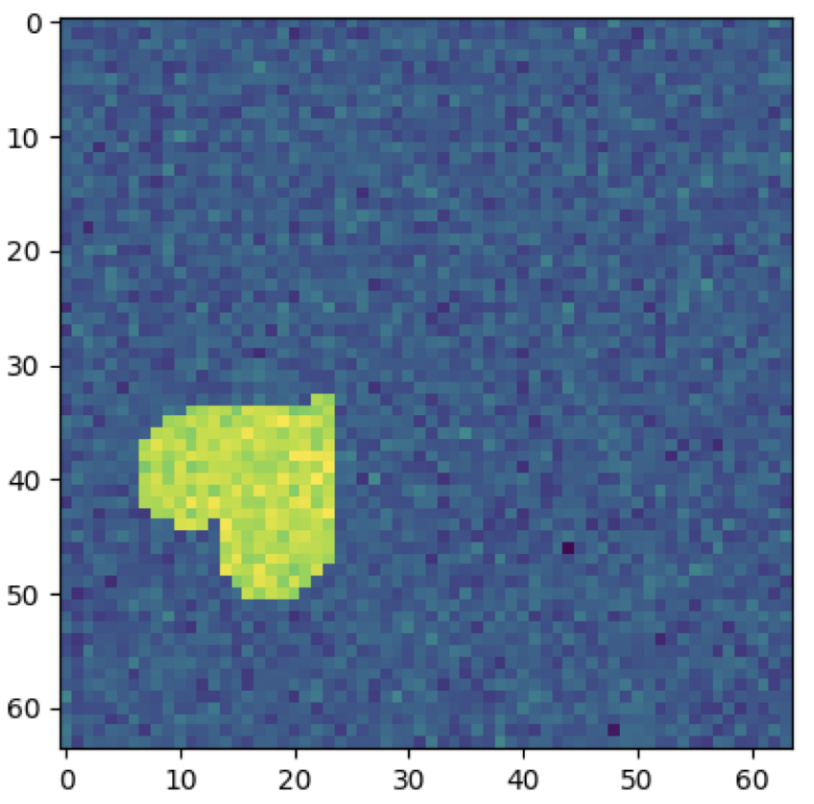}
    \caption{An example of dSprites image, which is used as the treatment $A$ in PCL experiments. Its scale, rotation, $x$ position and $y$ position are randomly generated.}
    \label{fig:dsprite_example}
\end{figure}

For the test dataset, a fixed grid of image parameters is chosen:
\begin{align*}
\textit{posX}&\in[0,5,10,15,20,25,30]\\
\textit{posY}&\in[0,5,10,15,20,25,30]\\
\textit{scale}&\in[0.5,0.8,1.0]\\
\textit{rotation}&\in[0,0.5\pi,\pi,1.5\pi],
\end{align*}
which consists of 588 images to reliably evaluate different PCL algorithms.

\section{Network Architecture and Hyperparameters}\label{appen:networks}

Here, we describe the network architecture and hyperparameters of all experiments. Unless otherwise specified, all neural network algorithms are optimised using AdamW~\citep{Loshchilov2017} with learning rate $= 0.001$, $\beta =(0.9,0.999)$ and $\epsilon=10^{-8}$. In addition, we set $K=10$ for $K$-fold cross-fitting in DML-CMR.
% In addition, all hyperparameter choices for methods and datasets used in this work are available in our code.

\subsection{IV Regression}
We first introduce details for the IV regression experiments.

\subsubsection{Ticket Demand Dataset}

For DML-CMR and CE-DML-CMR, we use the network architecture described in~\cref{tab:demand_arch}. We use a learning rate of $0.0002$ with a weight decay of $0.001$ (L2 regularisation) and a dropout rate of $\frac{1000}{5000+N}$ that depends on the data size $N$. For DeepGMM, we use the same structure as the outcome network of DML-CMR with dropout $=0.1$ and the same learning rate as DML-CMR. For DFIV, we follow the original structure proposed in~\citet{Xu2020} with regularisers $\lambda 1$, $\lambda 2$ both set to 0.1 and weight decay of 0.001. For DeepIV, we use the same network architectures as the action network and stage 2 network for DML-CMR, with the dropout rate in~\citet{Hartford2017DeepPrediction} and weight decay of 0.001. For KIV, we use the Gaussian kernel, where the bandwidth is determined by the median trick as originally described by~\citet{Singh2019}, and we use the random Fourier feature trick with 100 dimensions.

\begin{table}[t]
    \centering
    \subfloat[Action Network for $\widehat{g}$]{
    \begin{tabular}{||c|c||}
    \hline
    \textbf{Layer Type} & \textbf{Configuration}  \\ [0.5ex]
    \hline \hline
    Input & $C$\\      \hline
    FC + ReLU & in:3 out:128\\    \hline
    Dropout & - \\    \hline
    FC + ReLU & in:128 out:64\\    \hline
    Dropout & - \\    \hline
    FC + ReLU & in:64 out:32\\    \hline
    Dropout & - \\    \hline
    MixtureGaussian & 10\\    \hline

    \end{tabular}\label{tab:action_arch}}
    \hspace{30pt}
    \subfloat[Outcome Network for $\widehat{s}$]{
    \begin{tabular}{||c|c||}
    \hline
    \textbf{Layer Type} & \textbf{Configuration}  \\ [0.5ex]
    \hline \hline
    Input & $C$\\      \hline
    FC + ReLU & in:3 out:128\\    \hline
    Dropout & - \\    \hline
    FC + ReLU & in:128 out:64\\    \hline
    Dropout & - \\    \hline
    FC + ReLU & in:64 out:32\\    \hline
    Dropout & - \\    \hline
    FC & in:32 out:1\\    \hline
    \end{tabular}\label{tab:outcome_arch}}

    \vspace{10pt}
    \subfloat[Stage 2 Network for $\widehat{h}$]{
    \begin{tabular}{||c|c||}
    \hline
    \textbf{Layer Type} & \textbf{Configuration}  \\ [0.5ex]
    \hline \hline
    Input & $C,A$\\      \hline
    FC + ReLU & in:3 out:128\\    \hline
    Dropout & - \\    \hline
    FC + ReLU & in:128 out:64\\    \hline
    Dropout & - \\    \hline
    FC + ReLU & in:64 out:32\\    \hline
    Dropout & - \\    \hline
    FC & in:32 out:1\\    \hline
    \end{tabular}\label{tab:stage2_arch}}
    \caption[Network architecture for DML-CMR and CE-DML-CMR for the ticket demand low-dimensional dataset for IV regression]{Network architecture for DML-CMR and CE-DML-CMR for the ticket demand low-dimensional dataset for IV regression. For the input layer, we provide the input variables. For mixture of Gaussians output, we report the number of components. The dropout rate is given in the main text.}
    \label{tab:demand_arch}
\end{table}

\subsubsection{Ticket Demand with MNIST}

For DML-CMR and CE-DML-CMR, we use a convolutional neural network (CNN) feature extractor, which we denote as \textit{ImageFeature}, described in~\cref{tab:mnist_arch}, for all networks. The full network architecture is described in~\cref{tab:mnist_demand_arch}; we use weight decay of 0.05. For DeepGMM, we use the same structure as the outcome network of DML-CMR, with a dropout rate of 0.1 and weight decay of 0.05. For DFIV, we follow the original structure proposed in~\citet{Xu2020} with regularisers $\lambda 1$, $\lambda 2$ both set to 0.1 and weight decay of 0.05. For DeepIV, we use the same network architecture as the action network and stage 2 network for DML-CMR, with the dropout rate in~\citet{Hartford2017DeepPrediction} and weight decay of 0.05. For KIV, we use the Gaussian kernel, where the bandwidth is determined by the median trick as originally described by~\citet{Singh2019}, and we use the random Fourier feature trick with 100 dimensions.

\begin{table}[t]
    \centering
    \begin{tabular}{||c|c||}
    \hline
    \textbf{Layer Type} & \textbf{Configuration}  \\ [0.5ex]
    \hline \hline
    Input & $28\times 28$\\      \hline
    Conv + ReLU & $3\times3\times32$, s:1, p:0\\    \hline
    Max Pooling & $2\times2$, s:2\\    \hline
        Dropout & -\\    \hline
    Conv + ReLU & $3\times3\times64$, s:1, p:0\\    \hline
    Max Pooling & $2\times2$, s:2\\    \hline
        Dropout & -\\    \hline
    Conv + ReLU & $3\times3\times64$, s:1, p:0\\    \hline
    Dropout & -\\    \hline
    FC + ReLU & in: 576, out:64\\    \hline
    \end{tabular}
    \caption[Network architecture of the feature extractor used for the ticket demand dataset with MNIST for IV regression]{Network architecture of the feature extractor used for the ticket demand dataset with MNIST for IV regression. For each convolution
layer, we list the kernel size, input dimension and output dimension, where s stands for stride and p stands for padding. For max-pooling, we provide the
size of the kernel. The dropout rate here is set to 0.3. We denote this feature extractor as \textit{ImageFeature}.}
\label{tab:mnist_arch}
\end{table}

\begin{table}[t]
    \centering
    \subfloat[Action Network for $\widehat{g}$]{
    \begin{tabular}{||c|c||}
    \hline
    \textbf{Layer Type} & \textbf{Configuration}  \\ [0.5ex]
    \hline \hline
    Input & ImageFeature$(C),Z$\\      \hline
    FC + ReLU & in:66 out:32\\    \hline
    Dropout & - \\    \hline
    MixtureGaussian & 10\\    \hline
    \end{tabular}}
    \hspace{30pt}
    \subfloat[Outcome Network for $\widehat{s}$]{
    \begin{tabular}{||c|c||}
    \hline
    \textbf{Layer Type} & \textbf{Configuration}  \\ [0.5ex]
    \hline \hline
    Input & ImageFeature$(C),Z$\\      \hline
    FC + ReLU & in:66 out:32\\    \hline
    Dropout & - \\    \hline
    FC & in:32 out:1\\    \hline
    \end{tabular}}
    \vspace{10pt}
    \subfloat[Stage 2 Network for $\widehat{h}$]{
    \begin{tabular}{||c|c||}
    \hline
    \textbf{Layer Type} & \textbf{Configuration}  \\ [0.5ex]
    \hline \hline
    Input & ImageFeature$(C),A$\\      \hline
    FC + ReLU & in:66 out:32\\    \hline
    Dropout & - \\    \hline
    FC & in:32 out:1\\    \hline
    \end{tabular}}
        \caption[Network architecture for DML-CMR and CE-DML-CMR for the ticket demand dataset with MNIST for IV regression]{Network architecture for DML-CMR and CE-DML-CMR for the ticket demand dataset with MNIST for IV regression. For the input layer, we provide the input variables. For a mixture of Gaussians output, we report the number of components. The dropout rate is given in the main text.}
    \label{tab:mnist_demand_arch}
\end{table}

\subsubsection{IHDP and PM-CMR}
For the two real-world datasets, we use the same network architectures described in~\cref{tab:demand_arch} as in the low-dimensional ticket demand setting, where the input dimension is increased to 7 for all networks. We use a dropout rate of 0.1 and weight decay of 0.001. For DeepGMM, we use the same structure as the outcome network of DML-CMR with dropout $=0.1$. For DFIV, we also use the same network architectures as in the low dimensional ticket demand setting with regularisers $\lambda 1$, $\lambda 2$ both set to 0.1 and weight decay of 0.001. For DeepIV, we use the same network architectures as the action network and stage 2 network of DML-CMR, with a dropout rate of 0.1 and weight decay of 0.001. For KIV, we use the Gaussian kernel where the bandwidth is determined by the median trick as originally described by~\citet{Singh2019}, and we use the random Fourier feature trick with 100 dimensions.

\subsection{Proximal Causal Learning}

Next, we introduce details for the proximal causal learning experiments.

\subsubsection{Ticket Demand Dataset}

For DML-CMR and CE-DML-CMR, we use the network architecture described in~\cref{tab:PCL_demand_arch}. We use a learning rate of $0.0001$ with a weight decay of $0.001$ (L2 regularisation) for the $f$ network and $0.0001$ for the $s$ and $g$ networks. The dropout rate is $\frac{400}{4000+N}$, which depends on the data size $N$. For the comparison methods, we use the default parameter values proposed in their original papers. For CEVAE, we use 1000 epochs, 0.0001 for weight decay, 10 learning samples, 20 hidden dimensions, and 5 for early stop. For DFPV, $\lambda 1$, $\lambda 2$ both set to 0.1, weight decay is 0.01 for all networks, stage 1 iteration is 20, and stage 2 iteration is 1. For KPV, we set $\lambda 1$ and $\lambda 2$ to be 0.001 with data split ratio 0.5. For NMMR, learning rate is 0.003, L2 penalty is 0.000003 with network depth 4 and width 80 for the U statistics estimator. For the V statistics estimator, network depth is 3 and width is 80 while other hyperparameters remain the same. For PKDR, the number of components is 50, gamma is 50, alpha is 35, and cross validation is 5. For PMMR, $\lambda 1$ and $\lambda 2$ are 0.01, with scale 0.5.

\begin{table}[t]
    \centering
    \subfloat[Action Network for $\widehat{g}$]{
    \begin{tabular}{||c|c||}
    \hline
    \textbf{Layer Type} & \textbf{Configuration}  \\ [0.5ex]
    \hline \hline
    Input & $C$\\      \hline
    FC + ReLU & in:3 out:128\\    \hline
    Dropout & - \\    \hline
    FC + ReLU & in:128 out:64\\    \hline
    Dropout & - \\    \hline
    FC + ReLU & in:64 out:32\\    \hline
    Dropout & - \\    \hline
    MixtureGaussian & 10\\    \hline

    \end{tabular}\label{tab:PCL_action_arch}}
    \hspace{30pt}
    \subfloat[Outcome Network for $\widehat{s}$]{
    \begin{tabular}{||c|c||}
    \hline
    \textbf{Layer Type} & \textbf{Configuration}  \\ [0.5ex]
    \hline \hline
    Input & $C$\\      \hline
    FC + ReLU & in:3 out:128\\    \hline
    Dropout & - \\    \hline
    FC + ReLU & in:128 out:64\\    \hline
    Dropout & - \\    \hline
    FC + ReLU & in:64 out:32\\    \hline
    Dropout & - \\    \hline
    FC & in:32 out:1\\    \hline
    \end{tabular}\label{tab:PCL_outcome_arch}}

    \vspace{10pt}
    \subfloat[Stage 2 Network for $\widehat{h}$]{
    \begin{tabular}{||c|c||}
    \hline
    \textbf{Layer Type} & \textbf{Configuration}  \\ [0.5ex]
    \hline \hline
    Input & $C,A$\\      \hline
    FC + ReLU & in:3 out:128\\    \hline
    Dropout & - \\    \hline
    FC + ReLU & in:128 out:64\\    \hline
    Dropout & - \\    \hline
    FC + ReLU & in:64 out:32\\    \hline
    Dropout & - \\    \hline
    FC & in:32 out:1\\    \hline
    \end{tabular}\label{tab:PCL_stage2_arch}}
    \caption[Network architecture for DML-CMR and CE-DML-CMR for the ticket demand dataset for PCL]{Network architecture for DML-CMR and CE-DML-CMR for the ticket demand dataset for PCL. For the input layer, we provide the input variables. For mixture of Gaussians output, we report the number of components. The dropout rate is given in the main text.}
    \label{tab:PCL_demand_arch}
\end{table}

\subsubsection{dSprites dataset}

For the dSprites dataset, we adopt a CNN feature extractor to handle the image inputs for DML-CMR. The architecture of this feature extractor is provided in~\cref{tab:PCL_mnist_arch}. We use 10 components for the mixture of Gaussian model, dropout is 0.2, batch size is 100, weight decay is 0.05, learning rate is 0.001 with Adam, and the number of epochs is the closest integer to $(1000000/N)+100$, which depends on the sample size $N$. For CEVAE, DFPV, KPV, NMMR and PMMR, we follow the hyperparameters and network architecture used in~\citet{Kompa2022} to generate the experimental results. For PKDR, we follow the high-dimensional dataset hyperparameters used in the original paper~\citep{Wu2024} with weight decay 0.0001, 4 layers, learning rate 0.0001, and 500 epochs.

\begin{table}[t]
    \centering
    \begin{tabular}{||c|c||}
    \hline
    \textbf{Layer Type} & \textbf{Configuration}  \\ [0.5ex]
    \hline \hline
    Input & $64\times 64$\\      \hline
    Conv + ReLU & $5\times5\times64$, s:1, p:0\\    \hline
    Max Pooling & $2\times2$, s:2\\    \hline
        Dropout & -\\    \hline
    Conv + ReLU & $5\times5\times128$, s:1, p:0\\    \hline
    Max Pooling & $2\times2$, s:2\\    \hline
        Dropout & -\\    \hline
    Conv + ReLU & $5\times5\times128$, s:1, p:0\\    \hline
    Dropout & -\\    \hline
    Max Pooling & $2\times2$, s:2\\    \hline
    FC + ReLU & in: 2048, out:128\\    \hline
    \end{tabular}
\caption[Network architecture of the feature extractor used for the dSprites image dataset for PCL]{Network architecture of the feature extractor used for the dSprites image dataset for PCL. For each convolution
layer, we list the kernel size, input dimension and output dimension, where s stands for stride and p stands for padding. For max-pooling, we provide the
size of the kernel. The dropout rate here is set to 0.2.}
\label{tab:PCL_mnist_arch}
\end{table}

\subsection{Validation and Hyperparameter Tuning}\label{appen:tune}

Validation procedures are crucial for tuning DNN hyperparameters and optimizer parameters. All the DML-CMR and CE-DML-CMR training stages can be validated by simply evaluating the respective losses on held-out data, as discussed in~\citet{Hartford2017DeepPrediction}. This allows independent validation and hyperparameter tuning of the two first stage networks (the action and the outcome networks), and performs second stage validation using the best network selected in the first stage. This validation procedure guards against the ‘weak instruments’ bias~\citep{Bound1995} that can occur when the instruments are only weakly correlated with the actions variable (see detailed discussion in~\citet{Hartford2017DeepPrediction}).

\section{Failure of Standard Offline Bandit Algorithms}\label{appen:offline_bandit}

\begin{figure}[t]
\centering
\includegraphics[width=0.5\textwidth]{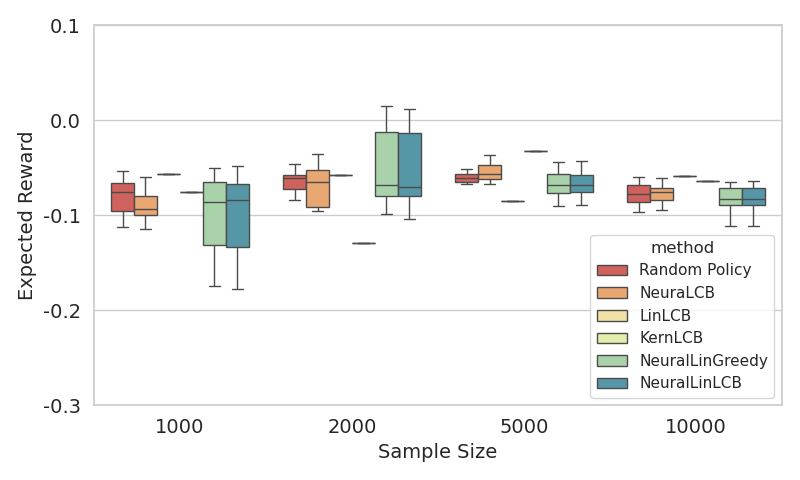}
\caption[Average reward for offline bandit algorithms that do not take IVs into account]{Comparing the average reward obtained by policies learnt using offline bandit algorithms that do not take IVs into account with a random policy on the ticket demand dataset with low-dimensional context.}
\label{fig:offline_bandit}
% \vspace{-0.1cm}
\end{figure}

It has been demonstrated that standard supervised learning that does not %take special consideration of the 
take IVs into account fails to learn the causal function or the counterfactual prediction function from a confounded offline dataset~\citep{Hartford2017DeepPrediction}. Similarly, we demonstrate here that standard offline bandit algorithms also fail to learn meaningful policies from confounded offline datasets. We evaluate PEVI, also called LinLCB~\citep{Jin2021}, NeuraLCB~\citep{Nguyen-Tang2022}, KernLCB~\citep{Valko2013}, NeuralLinLCB~\citep{Nguyen-Tang2022} and NeuralLinGreedy~\citep{Nguyen-Tang2022} algorithms, for which we combine the context $C$ and instrument $Z$ variables together as the new context input for these offline bandit algorithms. For algorithms that only support discrete actions, we discretise the action space $\mathcal{A}$ into 20 discrete actions.

For all methods, we follow the network architecture and hyperparameters from the original papers, and we adopt the implementation\footnote{\url{https://github.com/thanhnguyentang/offline_neural_bandits}} of~\citet{Nguyen-Tang2022}. We evaluate these methods on the ticket demand dataset described in \cref{appen:demand} and compare the average reward obtained by the learnt policies with a random policy in~\cref{fig:offline_bandit}. It can be seen that all the offline bandit algorithms do not outperform a random policy while DML-IV achieves an average reward higher than 1, as shown in~\cref{fig:lowd_r}. This is unsurprising because these bandit methods do not exploit IVs explicitly and are unable to learn the true causal effect of actions.

\section{Additional Experimental Results}

In this section, we provide additional experimental results including the effects of high ill-posedness (e.g., weak IVs), performance with tree-based estimators, and a hyperparameter sensitivity analysis.

\subsection{Effects of Weak Instruments}\label{appen:weak_iv}

When the correlation between instruments and the endogenous variable (the action in our case) is weak, IV regression methods generally become unreliable~\citep{Andrews2019} because the weak correlation induces variance and bias in the first stage estimator thus induces bias in the second stage estimator, especially for non-linear IV regressions. In theory, DML-CMR should be more resistant to biases in the first stage thanks to the DML framework, as long as the causal effect is identifiable under the weak instrument. This identifiability condition is captured in \cref{condition:dml} for DML, and it is connected to the ill-posedness for CMR problems in general as discussed in~\cref{sec:ill-posedness}. With identifiability, \cref{thm:dml} and \cref{coro:function_convergence} all hold, and the convergence rate guarantees still apply. Intuitively, as the ill-posedness increases, worse empirical performance will be observed.

Experimentally, for the ticket demand dataset, we alter the instrument strength by changing how much the instrument z affects the price p. Recall from~\cref{appen:demand} that $p=25+(z+3)\psi(t)+\omega$, where $\psi$ is a nonlinear function and $\omega$ is the noise. We add an IV strength parameter $\varrho$ such that $p=25+(\varrho\cdot z+3)\psi(t)+\omega$. In~\cref{tab:weak_iv}, we present the mean and standard deviation of the MSE of $\widehat{h}$ for various IV strengths $\varrho$ from 0.01 to 1 and sample size $N=5000$. It is very interesting to see that DML-CMR indeed performs significantly better than SOTA nonlinear IV regression methods under weak instruments.

\begin{table}[t]\setlength\extrarowheight{4pt}
\centering
\tiny
\begin{tabular}{||c|c|c|c|c|c|c||}
\hline
\textbf{IV Strength} & \textbf{1.0} & \textbf{0.8}                     & \textbf{0.6}                     & \textbf{0.4}                     & \textbf{0.2}                     & \textbf{0.01}                    \\\hline \hline
DML-CMR              & \textbf{0.0676(0.0116)} & \textbf{0.0984(0.0161)} & \textbf{0.1295(0.0168)} & \textbf{0.1859(0.0376)} & \textbf{0.2899(0.0494)} & \textbf{0.4872(0.1295)} \\\hline
CE-DML-CMR           & \textbf{0.0765(0.0119)} & \textbf{0.1064(0.0120)} & \textbf{0.1514(0.0203)} & \textbf{0.2070(0.0329)} & \textbf{0.3194(0.0572)} & \textbf{0.5302(0.1625)} \\\hline
DeepIV              & 0.1213(0.0209)          & 0.2039(0.0269)         & 0.3051(0.0415)          & 0.4476(0.0656)          & 0.6891(0.1210)          & 0.9293(0.2382)          \\\hline
DFIV                & 0.1124(0.0481)          & 0.1586(0.0320)          & 0.3080(0.1907)          & 0.8117(0.2779)          & 0.9622(0.3892)          & 1.6503(0.6845)          \\\hline
DeepGMM             & 0.2699(0.0522)          & 0.3330(0.1171)          & 0.4762(0.1056)          & 0.8666(0.2248)          & 1.0056(0.4334)          & 2.0218(0.6555)          \\\hline
KIV                 & 0.2312(0.0272)          & 0.3149(0.0218)          & 0.4275(0.0368)          & 0.6646(0.0538)          & 0.8099(0.0657)          & 1.226(0.1014)\\\hline
\end{tabular}
\caption{Results for the low-dimensional ticket demand dataset when the IV is weakly correlated with the action.}
\label{tab:weak_iv}
\end{table}

\subsection{Performance of DML-CMR with Tree-Based Estimators}\label{appen:tree-based}

The DML-CMR framework allows for general estimators following the Neyman orthogonal score function. While deep learning is flexible and widely used in SOTA non-linear IV regression methods, gradient boosting and random forests are all good candidate estimators for DML-CMR. In addition, as shown in \cref{sec:theory}, the convergence rate and suboptimality guarantees in \cref{coro:function_convergence} and \cref{coro:subopt} both hold for these tree-based regressions.

Empirically, we replace the DNN estimators in DML-CMR, CE-DML-CMR, and DeepIV with Random Forests and Gradient Boosting regressors (using scikit-learn implementation). DeepIV is a good baseline for comparison, since it optimizes directly using a non-Neyman-orthogonal score and allows for direct replacement of all DNN estimators with tree-based estimators. We use 500 trees for both regressors, with minimum samples required at each leaf node of 100 for the nuisance parameters and 10 for $\widehat{h}$.

In~\cref{tab:tree_based}, we present the mean and standard deviation of the MSE of $\widehat{h}$ with Random Forests and Gradient Boosting estimators on the ticket demand dataset with various dataset sample sizes. The results demonstrate the benefits of our Neyman orthogonal score function, and interestingly, the performance of Gradient Boosting is comparable to DNN estimators.

\begin{table}[ht]\setlength\extrarowheight{4pt}
\centering
\scriptsize
\begin{tabular}{||c|c|c|c|c||}
\hline
\textbf{IV Strength} & \textbf{Dataset Size} & \textbf{DNN (results in the paper)} & \textbf{Random Forests} & \textbf{Gradient Boosting} \\\hline \hline
DML-CMR    & 2000                  & \textbf{0.1308(0.0206)}             & 0.1689(0.0172)          & \textbf{0.1301(0.0112)}    \\\hline
CE-DML-CMR &    2000               & \textbf{0.1410(0.0246)}             & 0.1733(0.0198)          & \textbf{0.1329(0.0125)}    \\\hline
DeepIV    &   2000               & 0.2388(0.0438)                      & 0.2642(0.0261)          & 0.2052(0.0232)             \\\hline
DML-CMR    & 5000                  & \textbf{0.0676(0.0129)}             & 0.1067(0.0131)          & \textbf{0.0632(0.0107)}    \\\hline
CE-DML-CMR &   5000               & \textbf{0.0765(0.0119)}             & 0.1154(0.0138)          & \textbf{0.0699(0.0069)}    \\\hline
DeepIV    &    5000              & 0.1213(0.0209)                      & 0.1626(0.0128)          & 0.1020(0.0091)             \\\hline
DML-CMR    & 10000                 & \textbf{0.0378(0.0094)}             & 0.0657(0.0062)          & \textbf{0.0482(0.0079)}    \\\hline
CE-DML-CMR &  10000                & \textbf{0.0442(0.0070)}             & 0.0721(0.0039)          & \textbf{0.0523(0.0059)}    \\\hline
DeepIV    &  10000             & 0.0714(0.0140)                      & 0.1106(0.0080)          & 0.1017(0.0075)\\\hline
\end{tabular}
\caption{Results for the low-dimensional ticket demand dataset using tree-based estimators compared to DNN estimators.}
\label{tab:tree_based}
\end{table}

\subsection{Sensitivity Analysis for Different Hyperparameters}\label{appen:sensitivity}

The tunable hyperparameters in DML-CMR are the learning rate, network width, weight decay, and dropout rate (see~\cref{appen:networks}). As a sensitivity analysis, we provide results for the mean and standard deviation of the MSE of the DML-CMR estimator $\widehat{h}$ with different hyperparameter values for both the low-dimensional and high-dimensional datasets with sample size N=5000 in~\cref{tab:ablation_low} and~\cref{tab:ablation_high}. Overall, we see that DML-CMR is not very sensitive to small changes in the hyperparameters.

\begin{table}[t]\setlength\extrarowheight{4pt}
    \centering
    \scriptsize
    \begin{tabular}{||c|c|c|c|c|c||}
    \hline
     \textbf{Learning Rate} & \textbf{Weight Decay}& \textbf{Dropout}& \textbf{DNN Width} & \textbf{DML-CMR} &\textbf{CE-DML-CMR} \\\hline \hline
0.0002 & 0.001  & 0.1  & 128 & \textbf{0.0676(0.0129)} & \textbf{0.0765(0.0119)} \\\hline
0.0005 &        &      &     & 0.0752(0.0122)          & 0.0897(0.0196)          \\\hline
0.0001 &        &      &     & \textbf{0.0703(0.0195)} & \textbf{0.0794(0.0201)} \\\hline
       & 0.0005 &      &     & 0.0794(0.0185)          & 0.0823(0.0149)          \\\hline
       & 0.005  &      &     & 0.0765(0.0135)          & 0.0809(0.0159)          \\\hline
       & 0.01   &      &     & 0.0820(0.0162)          & 0.0865(0.0174)          \\\hline
       &        & 0.05 &     & \textbf{0.0715(0.0074)} & \textbf{0.0813(0.0089)} \\\hline
       &        & 0.2  &     & 0.0836(0.0100)          & 0.0919(0.0157)          \\\hline
       &        &      & 64  & 0.0830(0.0162)          & 0.0924(0.0121)          \\\hline
       &        &      & 256 & 0.0943(0.0179)          & 0.0981(0.0126)          \\\hline
       & 0.0005 & 0.2  &     & 0.0805(0.0133)          & 0.0910(0.0106)          \\\hline
       & 0.005  & 0.05 &     & \textbf{0.0672(0.0116)} & \textbf{0.0742(0.0102)} \\\hline
       & 0.01   & 0.05 &     & 0.0825(0.0152)          & 0.0914(0.0125)          \\\hline
       &        & 0.2  & 256 & 0.0810(0.0129)          & 0.0852(0.0121)          \\\hline
       &        & 0.05 & 64  & 0.0907(0.0149)          & 0.0963(0.0161)          \\\hline
       & 0.005  &      & 256 & 0.0939(0.0146)          & 0.0991(0.0093)\\\hline
    \end{tabular}
    \caption[Results for the low-dimensional ticket demand dataset for a range of hyperparameter values]{Results for the low-dimensional ticket demand dataset for a range of hyperparameter values. The default hyperparameters in this case are: learning rate=0.0002, weight decay=0.001, dropout=0.1 and DNN width 128. The bold results are the best performing hyperparameters.}
    \label{tab:ablation_low}
\end{table}
\begin{table}[ht]\setlength\extrarowheight{4pt}
    \centering
     \scriptsize
    \begin{tabular}{||c|c|c|c|c|c||}
    \hline
\textbf{Learning Rate} & \textbf{Weight Decay} & \textbf{Dropout} & \textbf{CNN Channels} & \textbf{DML-CMR}          & \textbf{CE-DML-CMR}      \\\hline \hline
0.001                  & 0.05                  & 0.2              & 64                    & \textbf{0.3513(0.0125)}  & \textbf{0.3808(0.0150)} \\\hline
0.0005                 &                       &                  &                       & 0.4063(0.0129)           & 0.5008(0.0369)          \\\hline
0.002                  &                       &                  &                       & 0.3659(0.0219)           & 0.4133(0.0267)          \\\hline
0.005                  &                       &                  &                       & \textbf{0.3377(0.0218)}  & \textbf{0.3555(0.0202)} \\\hline
& 0.01                  &                  &                       & 0.3935(0.0176)           & 0.4461(0.0478)          \\\hline
& 0.02                  &                  &                       & \textbf{0.3595(0.03013)} & \textbf{0.3851(0.0293)} \\\hline
& 0.1                   &                  &                       & 0.4066(0.0172)           & 0.5160(0.0329)          \\\hline
&                       & 0.1              &                       & 0.4136(0.0211)           & 0.5386(0.0398)          \\\hline
&                       & 0.3              &                       & 0.3857(0.0171)           & 0.4002(0.0249)          \\\hline
&                       &                  & 128                   & 0.4176(0.01941)          & 0.5129(0.0630)          \\\hline
&                       &                  & 256                   & 0.4942(0.0226)           & 0.6180(0.0396)          \\\hline
& 0.1                   & 0.1              &                       & 0.4163(0.0214)           & 0.5952(0.0343)          \\\hline
& 0.01                  & 0.3              &                       & 0.3636(0.0186)           & 0.3995(0.0250)          \\\hline
&                       & 0.3              & 128                   & 0.4006(0.0187)           & 0.4764(0.0216)          \\\hline
&                       & 0.3              & 256                   & \textbf{0.3429(0.0215)}  & \textbf{0.3971(0.0264)} \\\hline
& 0.1                   &                  & 256                   & 0.4170(0.0283)           & 0.5335(0.0371)\\\hline
                       
\end{tabular}
\caption[Results for the high-dimensional ticket demand dataset for a range of hyperparameter values]{Results for the high-dimensional ticket demand dataset for a range of hyperparameter values. The default hyperparameters in this case are: learning rate 0.001, weight decay=0.05, dropout=0.2 and 64 CNN channels. The bold results are the best performing hyperparameters.}
\label{tab:ablation_high}
\end{table}

\chapter{Imitating Expert Policies from Confounded Datasets}

\section{Proofs of Main Results}\label{appendix:proofs}

In this section, we provide the proofs for the main results and corollaries in~\cref{chapter:il}, which we restate for convenience.

\subsection{Proof of Propositions}\label{appendix:prop}

\begin{repeatprop}{IL-prop:ill-posed}
The ill-posedness $\ill(\Pi,k)$ is monotonically increasing as the confounding horizon $k$ increases.
\end{repeatprop}

\begin{proof}
From definition, we have that,
\begin{align*}
    \ill(\Pi,k)=\sup_{\pi\in\Pi} \frac{\norm{\pi_E-\pi}_{2}}{\norm{\expectE[a_t-\pi(h_t)\lvert h_{t-k}]}_{2}}.
\end{align*}
We would like to show that, for each $\pi\in\Pi$, $\frac{\norm{\pi_E-\pi}_{2}}{\norm{\expectE[a_t-\pi(h_t)\lvert h_{t-k}]}_{2}}$ is increasing as $k$ increases, which would imply that $\ill(\Pi,k)$ is increasing. For each $\pi\in\Pi$, we see that the numerator is constant w.r.t. the horizon $k$. Therefore, it suffices to check that for each $\pi\in\Pi$, the denominator $\norm{\expectE[a_t-\pi(h_t)\lvert h_{t-k}]}_{2}$ decreases as $k$ increases. For any two integer horizons $k_1>k_2$,
\begin{align}
\expectE[a_t-\pi(h_t)\lvert h_{t-k_1}]^2&=\expectE[\expectE[a_t-\pi(h_t)\lvert h_{t-k_2}]\lvert h_{t-k_1}]^2\\
&\leq \expectE[\expectE[a_t-\pi(h_t)\lvert h_{t-k_2}]^2\lvert h_{t-k_1}]\\
&=\expectE[a_t-\pi(h_t)\lvert h_{t-k_2}]^2,
\end{align}
by the tower property of conditional expectation as $\sigma(h_{t-k_1})\subseteq\sigma(h_{t-k_2})$, Jensen's inequality for conditional expectations, and the fact that $\expectE[a_t-\pi(h_t)\lvert h_{t-k_2}]^2$ is $h_{t-k_1}$ measurable, respectively for each line. Therefore, we have that $\expectE[a_t-\pi(h_t)\lvert h_{t-k}]$ is decreasing, which implies that $\norm{\expectE[a_t-\pi(h_t)\lvert h_{t-k}]}_{2}$ is decreasing and $\ill(\Pi,k)$ is increasing as $k$ increases, which completes the proof.
\end{proof}

\subsection{Guarantees on the Imitation Gap}\label{appendix:gap}

\begin{repeatthm}{thm:gap}
    Let $\widehat{\pi}_h$ be the learnt policy with CMR error $\epsilon$ and let $\ill(\Pi,k)$ be the ill-posedness of the problem. Assume that $\delta_{TV}(u^o_t,\expectE_{\pi_E}[u^o_t\lvert h_t])\leq\delta$ for $\delta\in\realNumber^+$, $P(u^\epsilon_t)$ is c-TV stable and $\pi_E$ is deterministic. Then, the imitation gap is upper bounded by
\begin{align*}
    J(\pi_E)-J(\widehat{\pi}_h)\leq T^2(c\epsilon\ill(\Pi,k)+2\delta)=\mathcal{O}(T^2(\delta+\epsilon)).
\end{align*}
\end{repeatthm}

\begin{proof}
Recall that $J(\pi)$ is the expected reward following $\pi$, and we would like to bound the performance gap $J(\pi_E)-J(\widehat{\pi_h})$ between the expert policy $\pi_E$ and the learnt history-dependent policy $\widehat{\pi_h}$. Let $Q_{\widehat{\pi_h}}(s_t,a_t,u^o_t)$ be the Q-function of $\widehat{\pi_h}$. Using the Performance Difference Lemma~\citep{Kakade2002}, we have that, for any Q-function $\tilde{Q}(h_t,a_t)$ that takes in the trajectory history $h_t$ and action $a_t$,
\begin{align}
J(\pi_E)-J(\widehat{\pi_h})&=\expectE_{\tau\sim\pi_E}[\sum_{t=1}^T Q_{\widehat{\pi_h}}(s_t,a_t,u^o_t)-\expectE_{a\sim\widehat{\pi_h}}[Q_{\widehat{\pi_h}}(s_t,a,u^o_t)]]\nonumber\\
&=\sum_{t=1}^T\expectE_{\tau\sim\pi_E}[Q_{\widehat{\pi_h}}(s_t,a_t,u^o_t)-\tilde{Q}(h_t,a_t)+\tilde{Q}(h_t,a_t)-\expectE_{a\sim{\widehat{\pi_h}}}[Q_{\widehat{\pi_h}}-\tilde{Q}+\tilde{Q}]]\nonumber\\
&=\sum_{t=1}^T \expectE_{\tau\sim\pi_E}[\tilde{Q}-\expectE_{a\sim\widehat{\pi_h}}[\tilde{Q}]]+\sum_{t=1}^T \expectE_{\tau\sim\pi_E}[Q_{\widehat{\pi_h}}-\tilde{Q}-\expectE_{a\sim\widehat{\pi_h}}[Q_{\widehat{\pi_h}}-\tilde{Q}]].\label{eq:pdl}
\end{align}

We first bound the second part of~\cref{eq:pdl}. Denote by $\delta_{TV}$ the total variation distance. For two distributions $P,Q$, recall the property of total variation distance for bounding the difference in expectations:
\begin{align*}
\abs{\expectE_P[f(x)]-\expectE_Q[f(x)]}\leq \norm{f}_\infty \delta_{TV}(P,Q).
\end{align*}
In order to bound the second part of~\cref{eq:pdl}, for any $Q$ function, consider inferred $\tilde{Q}$ using the conditional expectation of $u^o$ on the history $h$,
\begin{equation*}
\tilde{Q}(h_t,a_t)\coloneqq Q(s_t,a_t,\expectE_{\tau\sim \pi_E}[u^o_t\lvert h_t]),
\end{equation*}
where note that $s_t\in h_t$. We have that, when the transition trajectory $(s_t,u^o_t,u^\epsilon_t,r_t)\sim \pi_E$ follows the expert policy, for any action $\dot{a}\sim \pi$ following some policy $\pi$ (in our case, it can be $\pi_E$ or $\widehat{\pi_h}$),
% \begin{align}
% \abs{\expectE_{\tau\sim\pi_E}[Q(s_t,u_t;\dot{a})-\tilde{Q}(h_t;\dot{a})]}&=\left\lvert\expectE_{\tau\sim\pi_E}[Q(s_t,u_t;\dot{a})-\expectE_{\tau\sim\pi_E}[Q(s_t,u_t;\dot{a})\lvert h_t]]\right\rvert\\
% &=\left\lvert\expectE_{(s_t,u_t)\sim\pi_E}[Q(s_t,u_t;\dot{a})]-\expectE_{(s_t,u_t\lvert h_t)\sim\pi_E}[Q(s_t,u_t;\dot{a})]\right\rvert\\
% % &\leq \expectE_{\tau\sim\pi_E}[\abs{Q(s_t,a_t,u_t)-\expectE_{u_t\sim\pi_E}[Q(s_t,a_t,u_t)\lvert h_t,a_t]}]\label{eq:jensen}\\
% &\leq \norm{Q}_{\infty}\delta_{TV}(F_{\pi_E}(s_t,u_t),F_{\pi_E}(s_t,u_t\lvert h_t))\label{eq:tv_bound}\\
% % &\leq T \expectE_{\pi_E}[\delta_{TV}(F(u_t),F(u_t\lvert h_t,a_t))]\\
% &\leq T \cdot\delta_{TV}(F_{\pi_E}(s_t,u_t),F_{\pi_E}(s_t,u_t\lvert h_t))\\
% &\leq T\delta \label{eq:second_part_bound}
% \end{align}
\begin{align}
\abs{\expectE_{\tau\sim\pi_E,\dot{a}\sim\pi}[Q(s_t,\dot{a},u_t)-\tilde{Q}(h_t,\dot{a})]}&=\left\lvert\expectE_{\tau\sim\pi_E,\dot{a}\sim\pi}[Q(s_t,\dot{a},u^o_t)-Q(s_t,\dot{a},\expectE_{\tau\sim\pi_E}[u^o_t\lvert h_t]])]\right\rvert\nonumber\\
&=\left\lvert\expectE_{u^o_t\sim\pi_E}[\expectE_{\pi_E,\pi}[Q(s_t,\dot{a},u^o_t)\lvert u^o_t]\right.\nonumber\\
&\left.\quad\quad\quad\quad-\expectE_{u^o_t\lvert h_t\sim\pi_E}[\expectE_{\pi_E,\pi}[Q(s_t,\dot{a},u^o_t)\lvert u^o_t]\right\rvert\label{eq:tower_property}\\
&\leq \norm{\expectE_{\pi_E,\pi}[Q(s_t,\dot{a},u^o_t)\lvert u^o_t]}_{\infty}\delta_{TV}(u^o_t,\expectE_{\pi_E}[u^o_t\lvert h_t])\label{eq:tv_bound}\\
% &\leq T \expectE_{\pi_E}[\delta_{TV}(F(u_t),F(u_t\lvert h_t,a_t))]\\
&\leq T \cdot\delta_{TV}(u^o_t,\expectE_{\pi_E}[u^o_t\lvert h_t])\label{eq:bounded_Q}\\
&\leq T\delta,\label{eq:second_part_bound}
\end{align}
where~\cref{eq:tower_property} uses the tower property of expectations,~\cref{eq:tv_bound} uses the total variation distance bound for bounded functions,~\cref{eq:bounded_Q} uses the fact that the $Q$ function is bounded by $T$ and~\cref{eq:second_part_bound} uses the condition that $\delta_{TV}(u^o_t,\expectE_{\pi_E}[u^o_t\lvert h_t])\leq\delta$ in the theorem statement. Since~\cref{eq:pdl} holds for any choice of $\tilde{Q}$, we choose $\tilde{Q}_{\widehat{\pi_h}}(h_t,a_t)\coloneqq Q_{\widehat{\pi_h}}(s_t,a_t,\expectE_{\tau\sim \pi_E}[u^o_t\lvert h_t])$ such that we can apply~\cref{eq:second_part_bound} twice to bound the second part of~\cref{eq:pdl}:
\begin{align}
\expectE_{\tau\sim\pi_E}[Q_{\widehat{\pi_h}}-\tilde{Q}_{\widehat{\pi_h}}-\expectE_{a\sim\widehat{\pi_h}}[Q_{\widehat{\pi_h}}-\tilde{Q}_{\widehat{\pi_h}}]]&\leq \expectE_{\tau\sim\pi_E}[Q_{\widehat{\pi_h}}-\tilde{Q}_{\widehat{\pi_h}}+\abs{\expectE_{a\sim\widehat{\pi_h}}[Q_{\widehat{\pi_h}}-\tilde{Q}_{\widehat{\pi_h}}]}]\nonumber\\
&=\expectE_{\tau\sim\pi_E}[Q_{\widehat{\pi_h}}-\tilde{Q}_{\widehat{\pi_h}}]+\abs{\expectE_{s_t,u_t\sim\pi_E,a\sim\widehat{\pi_h}}[Q_{\widehat{\pi_h}}-\tilde{Q}_{\widehat{\pi_h}}]}\nonumber\\
&\leq \abs{\expectE_{\tau\sim\pi_E}[Q_{\widehat{\pi_h}}-\tilde{Q}_{\widehat{\pi_h}}]}+T\delta\label{eq:tvd_action}\\
&\leq 2T\delta\nonumber,
\end{align}
where~\cref{eq:tvd_action} holds by applying~\cref{eq:second_part_bound} because the expectation of the trajectories (and their transitions) is over $\pi_E$, and the actions that are used only as arguments into the $Q$ function are sampled from $\widehat{\pi_h}$.

Next, we bound the first part of~\cref{eq:pdl}. Recall that the ill-posedness of the problem for a policy class $\Pi$ is
\begin{align*}
    \ill(\Pi,k)=\sup_{\pi\in\Pi} \frac{\norm{\pi_E-\pi}_2}{\norm{\expectE[a_t-\pi(h_t)\lvert h_{t-k}]}_2},
\end{align*}
where $\norm{\pi_E-\pi}_2$ is the RMSE and $\norm{\expectE[a_t-\pi(s_t)\lvert s_{t-k}]}_2$ is the CMR error from our algorithm. Since the learnt policy $\widehat{\pi_h}$ has a CMR error of $\epsilon$, we have that
\begin{align*}
\norm{\pi_E-\widehat{\pi_h}}_2\leq \ill(\Pi,k){\norm{\expectE[a_t-\widehat{\pi_h}(h_t)\lvert h_{t-k}]}_2} \leq \ill(\Pi,k)\epsilon.
\end{align*}
Next, recall that c-total variation stability of a distribution $P(u^\epsilon)$ where $u^\epsilon\in A$ for some space $A$ implies for two elements $a_1,a_2\in A$,
\begin{align*}
\norm{a_1-a_2}_2\leq\Delta \implies \delta_{TV}(a_1+u^\epsilon,a_2+u^\epsilon)\leq c\Delta.
\end{align*}
Since $P(u^\epsilon_t)$ is c-TV stable w.r.t. the action space $A$, we have that, for all history trajectories $h_t\in H$ (note that $s_t\in h_t$),
\begin{align*}
\delta_{TV}(\pi_E(s_t)+u^\epsilon_t,\widehat{\pi_h}(h_t)+u^\epsilon_t)&\leq c\norm{\pi_E(s_t)-\widehat{\pi_h}(h_t)}_2.
\end{align*}
Then, we have that by Jensen's inequality,
\begin{align*}
\expectE_{h_t\sim \pi_E}[\delta_{TV}(\pi_E(s_t)+u^\epsilon_t,\widehat{\pi_h}(h_t)+u^\epsilon_t)]^2&\leq \expectE_{h_t\sim \pi_E}[\delta_{TV}(\pi_E(s_t)+u^\epsilon_t,\widehat{\pi_h}(h_t)+u^\epsilon_t)^2]\\
\implies\expectE_{h_t\sim \pi_E}[\delta_{TV}(\pi_E(s_t)+u^\epsilon_t,\widehat{\pi_h}(h_t)+u^\epsilon_t)]&\leq \sqrt{\expectE_{h_t\sim \pi_E}[\delta_{TV}(\pi_E(s_t)+u^\epsilon_t,\widehat{\pi_h}(h_t)+u^\epsilon_t)^2]}\\
&\leq \sqrt{c^2\expectE_{h_t\sim \pi_E}[\norm{\pi_E(s_t)-\widehat{\pi_h}(h_t)}^2_2]}\\
&=c \norm{\pi_E-\widehat{\pi_h}}_2\leq c\epsilon\ill(\Pi,k).
\end{align*}

Therefore, by applying the total variation distance bound for expectations of $\tilde{Q}_{\widehat{\pi_h}}$ over different distributions of action $a_t$, we have that
\begin{align*}
\expectE_{\tau\sim \pi_E}[\tilde{Q}_{\widehat{\pi_h}}-\expectE_{a\sim\widehat{\pi_h}}[\tilde{Q}_{\widehat{\pi_h}}]]&=\expectE_{\tau\sim \pi_E}[\tilde{Q}_{\widehat{\pi_h}}(h_t,a_t)-\expectE[\tilde{Q}_{\widehat{\pi_h}}(h_t,\widehat{\pi_h}(h_t))]]\\
&=\expectE_{h_t\sim \pi_E}[\expectE[\tilde{Q}_{\widehat{\pi_h}}(h_t,\pi_E(s_t)+u^\epsilon_t)]-\expectE[\tilde{Q}_{\widehat{\pi_h}}(h_t,\widehat{\pi_h}(h_t)+u^\epsilon_t)]]\\
&\leq \norm{\tilde{Q}_{\widehat{\pi_h}}}_\infty \expectE_{h_t\sim \pi_E}[\delta_{TV}(F(\pi_E(s_t)+u^\epsilon_t),F(\widehat{\pi_h}(h_t)+u^\epsilon_t))]\\
&\leq T c\epsilon\ill(\Pi,k).
\end{align*}

Combining all of the above, we see that from~\cref{eq:pdl}, by selecting $\tilde{Q}_{\widehat{\pi_h}}(h_t,a_t)\coloneqq Q_{\widehat{\pi_h}}(s_t,a_t,\expectE_{\tau\sim \pi_E}[u^o_t\lvert h_t])$, the imitation gap can be bounded by
\begin{align*}
   J(\pi_E)-J(\widehat{\pi_h})&=\sum_{t=1}^T \expectE_{\tau\sim\pi_E}[\tilde{Q}_{\widehat{\pi_h}}-\expectE_{a\sim\widehat{\pi_h}}[\tilde{Q}_{\widehat{\pi_h}}]]\\&+\sum_{t=1}^T \expectE_{\tau\sim\pi_E}[Q_{\widehat{\pi_h}}-\tilde{Q}_{\widehat{\pi_h}}-\expectE_{a\sim\widehat{\pi_h}}[Q_{\widehat{\pi_h}}-\tilde{Q}_{\widehat{\pi_h}}]]\\
    &\leq\sum_{t=1}^T Tc\epsilon\ill(\Pi,k) +\sum_{t=1}^T 2T\delta\\
    &\leq T\cdot (Tc\epsilon\ill(\Pi,k) + 2T\delta)\\&= T^2(c\epsilon\ill(\Pi,k)+2\delta)=\mathcal{O}(T^2(\epsilon+\delta)),
\end{align*}
which concludes the proof.
\end{proof}

\subsection{Proofs of Corollaries}\label{appendix:corollaries}

\begin{repeatcoro}{corollary:noUo}
In the special case that $u^o_t = 0$, meaning that there is no confounder observable to the expert, or $u^o_t=\expectE_{\pi_E}[u^o_t\lvert h_t]$, meaning that $u^o_t$ is $\sigma(h_t)$ measurable (all information regarding $u^o_t$ is represented in the history), the imitation gap bound is $T^2(c\epsilon\ill(\Pi,k))$, which coincides with Theorem 5.1 of~\citet{Swamy2022_temporal}.
\end{repeatcoro}

\begin{proof}
If $u^o_t=0$, then we have $u^o_t=\expectE_{\pi_E}[u^o_t\lvert h_t]$ since $u^o_t$ is a constant. If $u^o_t=\expectE_{\pi_E}[u^o_t\lvert h_t]$, we have that,
\begin{align*}
\delta_{TV}(u^o_t,\expectE_{\pi_E}[u^o_t\lvert h_t])=\delta_{TV}(u^o_t,u^o_t)\leq 0.
\end{align*}
By plugging $\delta=0$ into~\cref{thm:gap}, we have that $J(\pi_E)-J(\widehat{\pi_h})\leq T^2(c\epsilon\ill(\Pi,k))$, which is the same as the imitation gap derived in~\citet{Swamy2022_temporal} and completes the proof.
\end{proof}

\begin{repeatcoro}{corollary:unconfounded}
In the special case that $u^\epsilon_t=0$, if the learnt policy via supervised BC has error $\epsilon$, then the imitation gap bound is $T^2(\frac{2}{\sqrt{\dim(A)}}\epsilon+2\delta)$, which is a concrete bound that extends the abstract bound in Theorem 5.4 of~\citet{Swamy2022}.
\end{repeatcoro}

\begin{proof}
In Theorem 5.4 of~\citet{Swamy2022}, for the offline case, which is the setting we are considering (as opposed to the online settings), they defined the following quantities for bounding the imitation gap in a very general fashion,
\begin{align*}
\epsilon_{\text{off}}&\coloneqq\sup_{\tilde{Q}}\expectE_{\tau\sim\pi_E}[\tilde{Q}-\expectE_{a\sim\widehat{\pi_h}}[\tilde{Q}]];\\
\delta_{\text{off}}&\coloneqq\sup_{Q\times\tilde{Q}}\expectE_{\tau\sim\pi_E}[Q_{\widehat{\pi_h}}-\tilde{Q}-\expectE_{a\sim\widehat{\pi_h}}[Q_{\widehat{\pi_h}}-\tilde{Q}]].
\end{align*}

The imitation gap by Theorem 5.4 in~\citet{Swamy2022} under the assumption that $u^\epsilon_t=0$ is $T^2(\epsilon_{\text{off}}+\delta_{\text{off}})$, which can also be deduced from~\cref{eq:pdl} by naively applying the above supremum. To obtain a concrete bound, we can provide a tighter bound for $\expectE_{\tau\sim \pi_E}[\tilde{Q}_{\widehat{\pi_h}}-\expectE_{a\sim\widehat{\pi_h}}[\tilde{Q}_{\widehat{\pi_h}}]]$, which is the first part of~\cref{eq:pdl}, given that $u^\epsilon_t=0$.

For two elements $a_1,a_2\in A$, we have that by Cauchy–Schwarz,
\begin{align*}
\delta_{TV}(a_1+0,a_2+0)=\frac{1}{2}\norm{a1-a2}_1\leq\frac{\sqrt{\dim(A)}}{2}\norm{a1-a2}_2.
\end{align*}
Then, we have that
\begin{align*}
\norm{a_1-a_2}_2\leq\Delta \implies \delta_{TV}(a_1,a_2)\leq \frac{2}{\sqrt{\dim(A)}}\Delta,
\end{align*}
so that by~\cref{thm:gap},
\begin{align*}
\expectE_{\tau\sim \pi_E}[\tilde{Q}_{\widehat{\pi_h}}-\expectE_{a\sim\widehat{\pi_h}}[\tilde{Q}_{\widehat{\pi_h}}]]&=\expectE_{\tau\sim \pi_E}[\tilde{Q}_{\widehat{\pi_h}}(h_t,a_t)-\expectE[\tilde{Q}_{\widehat{\pi_h}}(h_t,\widehat{\pi_h}(h_t))]]\\
&=\expectE_{h_t\sim \pi_E}[\expectE[\tilde{Q}_{\widehat{\pi_h}}(h_t,\pi_E(s_t))]-\expectE[\tilde{Q}_{\widehat{\pi_h}}(h_t,\widehat{\pi_h}(h_t))]]\\
&\leq \norm{\tilde{Q}_{\widehat{\pi_h}}}_\infty \frac{2}{\sqrt{\dim(A)}}\norm{\pi_E-\pi}_2\\
&\leq T \frac{2}{\sqrt{\dim(A)}}\epsilon,
\end{align*}
since when $u^\epsilon_t=0$ the learning error via supervised learning is $\epsilon:=\norm{\pi_E-\pi}_2$. Therefore, the final imitation bound following~\cref{thm:gap} is
\begin{align*}
   J(\pi_E)-J(\widehat{\pi_h})&=\sum_{t=1}^T \expectE_{\tau\sim\pi_E}[\tilde{Q}_{\widehat{\pi_h}}-\expectE_{a\sim\widehat{\pi_h}}[\tilde{Q}_{\widehat{\pi_h}}]]\\&+\sum_{t=1}^T \expectE_{\tau\sim\pi_E}[Q_{\widehat{\pi_h}}-\tilde{Q}_{\widehat{\pi_h}}-\expectE_{a\sim\widehat{\pi_h}}[Q_{\widehat{\pi_h}}-\tilde{Q}_{\widehat{\pi_h}}]]\\
    &\leq\sum_{t=1}^T T\frac{2}{\sqrt{\dim(A)}}\epsilon+\sum_{t=1}^T 2T\delta\\
   &= T^2(\frac{2}{\sqrt{\dim(A)}}\epsilon+2\delta).
\end{align*}
This bound is a concrete bound, obtained through detailed analysis of the problem at hand, that coincides with the abstract bound $T^2(\epsilon_{\text{off}}+\delta_{\text{off}})$ provided in Theorem 5.4 of~\citet{Swamy2022_temporal}. Note that this bound is independent of the ill-posedness $\ill(\Pi,k)$ and the c-TV stability of $u^\epsilon_t$, which are present in the bound of~\cref{thm:gap}, because of the lack of hidden confounders $u^\epsilon_t$.
\end{proof}

\section{Environments and Tasks}\label{appendix:envs}
\subsection{Dynamic Aeroplane Ticket Pricing}\label{appendix:ticket}
Here, we provide details regarding the dynamic aeroplane ticket pricing environment introduced in \cref{eg:plane}. The environment and the expert policy are defined as follows:
\begin{align*}
\states&\coloneqq\realNumber\\
\actions&\coloneqq[-1,1]\\
s_t&=sign(s)\cdot u^o_t - u^\epsilon_t\\
\pi_E&=clip(-s/u^o_t,-1,1)\\
a_t&=\pi_E+10\cdot u^\epsilon_t\\
u^o_t&=mean(p_t\sim \text{Unif} [-1,1],p_{t-1},....p_{t-M})\\
u^\epsilon_t&=mean(q_t\sim \text{Normal}(0,0.1\cdot\sqrt{k}),q_{t-1},...,q_{t-k+1})
\end{align*}
where $M$ is the influence horizon of the expert-observable $u^o$, which we set to 30. The states $s_t$ are the profits at each time step, and the actions $a_t$ are the final ticket price. $u^o_t$ represent the seasonal patterns, where the expert $\pi_E$ will try to adjust the price accordingly. $u^\epsilon_t$ represent the operating costs, which are additive both to the profit and price. Both $u^o_t$ and $u^\epsilon_t$ are the mean over a set of i.i.d. samples, $q_t$ and $p_t$, and vary across the time steps by updating the elements in the set at each time step. This construction allows $u^\epsilon_t$ and $u^\epsilon_{t-k}$ to be independent since all set elements $q_t$ will be re-sampled from time step $t-k$ to $t$. We multiply the standard deviation of $q_t$ by $\sqrt{k}$ to make sure $u^\epsilon_t$, which is the average over $k$ i.i.d. variables, has the same standard deviation for all choices of $k$.

\subsection{Mujoco Environments}\label{appendix:mujoco}
We evaluate DML-IL on three Mujoco environments: Ant, Half Cheetah, and Hopper. The original tasks do not contain hidden variables, so we modify the environment to introduce $u^\epsilon$ and $u^o$. We use the default transition, state, and action space defined in the Mujoco environment. However, we changed the task objectives by altering the reward function and added confounding noise to both the state and action. Specifically, instead of controlling the Ant, Half Cheetah, and Hopper, respectively, to travel as fast as possible, the goal is to control the agent to travel at a target speed that is varying throughout an episode. This target speed is $u^o$, which is observed by the expert but not recorded in the dataset. In addition, we add confounding noise $u^\epsilon_t$ to $s_t$ and $a_t$ to mimic the environmental noise such as wind noise. In all cases, the target speed $u^o_t$, confounding noise $u^\epsilon_t$, and the action $a_t$ are generated as follows:
\begin{align*}
a_t&=\pi_E+20\cdot u^\epsilon_t\\
u^o_t&=mean(p_t\sim \text{Unif} [-2,4],p_{t-1},....p_{t-M})\\
u^\epsilon_t&=mean(q_t\sim \text{Normal}(0,0.01\cdot\sqrt{k}),q_{t-1},...,q_{t-k+1})
\end{align*}
where $M=30$, the state transitions follow the default Mujoco environment and the expert policy $\pi_E$ is learnt online in the environment. $u^o_t$ and $u^\epsilon_t$ follow the ticket pricing environment to be the average over a queue of i.i.d. random variables. The reward is defined to be the $1_{healthy}-(\text{current velocity}-u^o_t)^2-\text{control loss}$, where $1_{healthy}$ gives reward $1$ as long as the agent is in a healthy state as defined in the Mujoco documentation. The second penalty term penalises deviation between the current agent's velocity and the target velocity $u^o_t$. The control loss term is also as defined in default Mujoco, which is $0.1*\sum(a_t^2)$ at each step to regularise the size of actions.
\subsubsection{Ant}

In the Ant environment, we follow the gym implementation~\footnote{Ant environment: \url{https://www.gymlibrary.dev/environments/mujoco/ant/}} with an 8-dimensional action space and a 28-dimensional observable state space, where the agent's position is also included in the state space. Since the target speed $u^o_t$ is not recorded in the trajectory dataset, we scale the current position of the agent with respect to the target speed, $pos_t^\prime= pos_{t-1}+\frac{pos_t-pos_{t-1}}{u^o_t}$, and use the new agent position $pos_t^\prime$ in the observed states. This allows the imitator to infer information regarding $u^o_t$ from trajectory history, namely from the rate of change in the past positions.

\subsubsection{Half Cheetah}
In the Half Cheetah environment, we follow the gym implementation~\footnote{Half Cheetah environment: \url{https://www.gymlibrary.dev/environments/mujoco/half_cheetah/}} with a 6-dimensional action space and an 18-dimensional observable state space, where the agent's position is also included in the state space. Similarly to the Ant environment, we scale the current position of the agent to $pos_t^\prime= pos_{t-1}+\frac{pos_t-pos_{t-1}}{u^o_t}$ such that the imitator can infer information regarding $u^o_t$ from trajectory history.

\subsubsection{Hopper}
In the Hopper environment, we follow the gym implementation~\footnote{Hopper environment: \url{https://www.gymlibrary.dev/environments/mujoco/hopper/}} with a 3-dimensional action space and a 12-dimensional observable state space, where the agent's position is also included in the state space. Similarly to the Ant environment, we scale the current position of the agent to $pos_t^\prime= pos_{t-1}+\frac{pos_t-pos_{t-1}}{u^o_t}$ such that the imitator can infer information regarding $u^o_t$ from trajectory history.

\section{Implementation Details}\label{appendix:implement}

Experiments are carried out on a Linux server (Ubuntu 18.04.2) with two Intel Xeon Gold 6252 CPUs, and each experiment run uses a single NVIDIA GeForce RTX 2080 Ti GPU for neural network training.

\subsection{Expert Training}

The expert in the ticket pricing environment is explicitly hand-crafted. For the Mujoco environments, we used the Stable-Baselines3~\citep{stable-baselines3} implementation of soft actor-critic (SAC) and the default hyperparameters for each task outlined by Stable-Baselines3. The expert policy is an MLP with two hidden layers of size 256 and ReLU activations, and we train the expert for $10^7$ steps.

\subsection{Imitator Training}

With the expert policy $\pi_E$, we generate 40 expert trajectories, each of 500 steps, following our previously defined environments. Specifically, the confounding noise is added to the state and actions, and crucially $u^o_t$ is not recorded in the trajectories. The naive BC directly learns $\expectE[a_t\mid s_t]$ via supervised learning. ResiduIL mainly follows the implementation of~\citet{Swamy2022_temporal}, where we adapt it to allow a longer confounding horizon $k>1$. For DML-IL and BC-SEQ, a history-dependent policy is used, where we fix the look-back length to be $k+3$, where $k$ is the confounding horizon. BC-SEQ then just learns $\expectE[a_t\mid h_t]$ via supervised learning, and DML-IL is implemented with $K$-fold following~\cref{alg:DML-IL-kfold}. The policy network architectures for BC, BC-SEQ, and ResiduIL are 2-layer MLPs with a 256 hidden size. The policy network $\widehat{\pi}_h$ and the mixture of Gaussians roll-out model $\widehat{M}$ for DML-IL have a similar architecture, with details provided in~\cref{tab:dml-il}. We use the AdamW optimiser with a weight decay of $10^{-4}$ and a learning rate of $10^{-4}$. The batch size is 64, and each model is trained for 150 epochs, which is sufficient for their convergence.

\subsection{Imitator Evaluation}

The trained imitator is then evaluated for 50 episodes, each 500 steps in the respective confounded environments. The average reward and the mean squared error between the imitator's action and the expert's action are recorded.

\begin{table}[t]
    \centering
    \subfloat[Roll-out model $\widehat{M}$]{
    \begin{tabular}{||c|c||}
    \hline
    \textbf{Layer Type} & \textbf{Configuration}  \\ [0.5ex]
    \hline \hline
    Input & state dim $\times$ 3\\      \hline
    FC + ReLU & Out: 256\\    \hline
    FC + ReLU & Out: 256\\    \hline
    MixtureGaussian & 5 components; Out: state dim $\times$ k\\    \hline
    \end{tabular}\label{tab:rollout_arch}}
    \hspace{30pt}
    \subfloat[Policy model $\widehat{\pi}_h$]{
    \begin{tabular}{||c|c||}
    \hline
    \textbf{Layer Type} & \textbf{Configuration}  \\ [0.5ex]
    \hline \hline
    Input & state dim$\times$ (k+3)\\      \hline
    FC + ReLU & Out: 256\\    \hline
    FC + ReLU & Out: 256\\    \hline
    FC & Out: action dim\\    \hline
    \end{tabular}\label{tab:policy_arch}}
        \caption[Network architecture for DML-IL]{Network architecture for DML-IL. For the mixture of Gaussians output, we report the number of components. No dropout is used.}
    \label{tab:dml-il}
\end{table}

\chapter{Learning Policies for High-Level Objectives}
\section{Proofs of Main Results}\label{appendix:ltl-proofs}

In this section, we provide the proofs for the main results in~\cref{chapter:ltl}, which we restate for convenience.

\subsection{Proof of \cref{thm:equivalence}}
\label{proof:equivalence}

\begin{repeatthm}{thm:equivalence}
For any product MDP $\productMDP$ that is induced from LTL formula $\LTL$, we have that
\begin{equation}
\sup_\policy{\probP_\policy(\state_0\models\LTL)}=\sup_{\policy^\times}{\probP_{\policy^\times}(\productState_0\models\textsf{\upshape G}\textsf{\upshape F }\LTL_\autoAccept)}.
\end{equation}
Furthermore, a deterministic memoryless policy that maximises the probability of satisfying the Büchi condition $\LTL_\autoAccept$ on the product MDP $\productMDP$, starting from the initial state, induces a deterministic finite-memory optimal policy that maximises the probability of satisfying $\LTL$ on the original MDP $\MDP$ from the initial state.
\end{repeatthm}

\begin{proof}
We will prove the equality by verifying both sides of the inequality and subsequently constructing the induced policy on $\MDP$.

For $\leq$, it has been proven~\citep{Sickert2016Limit-deterministicLogic} that, for every accepting path of the environment MDP $\MDP$ following a policy $\policy$, there always exists a corresponding accepting run $\Autorun$ of the LDBA that resolves the nondeterminism. Augmenting the resolved nondeterminism as $\epsilon$-actions to the original path yields an accepting path of $\productMDP$, since the counter will eventually reach $K$ and the accepting states $\productAccept$ will be visited infinitely often. Therefore, we have created a policy for $\productMDP$ that is at least as good as $\policy$ on $\MDP$.

For $\geq$, it is clear that any policy $\policy$ on $\productMDP$ can induce a policy for $\MDP$ by eliminating the $\epsilon$-transitions and removing the projection of the automaton and $K$ counter. Therefore, any path following $\policy$ that meets the Büchi condition $\LTL_\autoAccept$ will induce a path of $\MDP$ that is accepting by $\automaton$ induced from $\LTL$, where the nondeterminism of $\automaton$ is resolved by $\epsilon$-transitions of $\policy$, thus satisfying $\LTL$.

\end{proof}

\subsection{Proof of \cref{useful_lemma}}
\label{proof:useful_lemma}

\begin{repeatlemma}{useful_lemma}
Given a product MDP $\productMDP$ with its corresponding LTL formula $\LTL$ and a policy $\policy$, we write $\productMDP_\policy$ for the induced Markov chain from $\policy$. Let $B_\productAccept$ denote the set of states that belong to accepting BSCCs of $\productMDP_\policy$, and $B^\times_\varnothing$ denote the set of states that belong to rejecting BSCCs:
\begin{align*}
&B_\productAccept \coloneqq \{\productState \mid \productState\in B \in BSCC(\productMDP_\policy),B\cap\productAccept\neq\varnothing\};\\
&B^\times_\varnothing \coloneqq \{\productState \mid \productState\in B \in BSCC(\productMDP_\policy),B\cap\productAccept=\varnothing\},
\end{align*}
where $BSCC(\productMDP_\policy)$ is the set of all BSCCs of $\productMDP_\policy$. We further define more general accepting and rejecting sets:
\begin{align}
    B_\autoAccept &\coloneqq \{(\state,\autoState,n) \mid \exists n^\prime\in [0..K] : (\state,\autoState,n^\prime)\in B_\productAccept\}\label{appen_B_accept};\\
    B_\varnothing &\coloneqq \{(\state,\autoState,n) \mid \exists n^\prime\in [0..K] : (\state,\autoState,n^\prime)\in B^\times_\varnothing\label{appen_B_reject}\}.
\end{align}

If \cref{sec-ltl:finite} holds, then $\probP_\policy(\productState\models\textsf{\upshape G}\textsf{\upshape F }\LTL_\autoAccept)=1 \;\forall \productState\in B_\autoAccept$, $\probP_\policy(\productState\models\textsf{\upshape G}\textsf{\upshape F }\LTL_\autoAccept)=0\;\forall\productState\in B_\varnothing$, and $B_\varnothing\cap\productAccept=\varnothing$. Furthermore, $B_\autoAccept$ and $B_\varnothing$ are sink sets, which means that once the set is reached, no states outside the set can be reached.
\end{repeatlemma}

\begin{proof}
To begin with, we recall the property of BSCCs in Markov chains (MC): for any infinite path $\MDPpath$ of a MC, a BSCC will eventually be reached, and once reached it cannot reach any state outside the BSCC and all states within it will be visited infinitely often with probability 1. Therefore, since all $\productState\in B_\productAccept$ belong to BSCCs with accepting states, all paths from $\productState$ will reach accepting states infinitely often with probability 1, so $\probP_\policy(\productState\models\textsf{\upshape G}\textsf{\upshape F }\LTL_\autoAccept)=1 \;\forall \productState\in B_\productAccept$. Similarly, all $\productState\in B^\times_\varnothing$ belong to BSCCs with no accepting states, so $\probP_\policy(\productState\models\textsf{\upshape G}\textsf{\upshape F }\LTL_\autoAccept)=0 \;\forall \productState\in B^\times_\varnothing$.

Now, we observe that, for states in $\productMDP_\policy$, the Markov chain transitions for environment states and automaton states are, in fact, independent of the counter value:
\begin{align}
&\productTransitions_\policy((\state,\autoState,n_1),(\state^\prime,\autoState^\prime,\min(n_1+1,K)))=\productTransitions_\policy((\state,\autoState,n_2),(\state^\prime,\autoState^\prime,\min(n_2+1,K)))\label{eq_lemma}
\end{align} 
for all $n_1,n_2\in[0..K], \state,\state^\prime\in\states$ and $\autoState,\autoState^\prime\in\autoStates$. With this observation, for all $(\state,\autoState,n)\in B_\autoAccept$, there exists $(\state,\autoState,n^\prime)\in B_\productAccept$, which belongs to some BSCC by~\cref{appen_B_accept}. Therefore, any state $(\state_1,\autoState_1,n_1)$ reachable from $(\state,\autoState,n)$ by~\cref{eq_lemma} implies that there exists $n_2$ such that $(\state_1,\autoState_1,n_2)$ is reachable from $(\state,\autoState,n^\prime)$. This,  by the definition of BSCC, means $(\state_1,\autoState_1,n_2)\in B_\productAccept$, which implies $(\state_1,\autoState_1,n_1)\in B_\autoAccept$ by \cref{appen_B_accept}, which shows that $B_\autoAccept$ is a sink set. Furthermore, since $(\state,\autoState,n^\prime)\in B_\productAccept$ belongs to an accepting BSCC, it can reach an accepting state $(\state_\autoAccept,\autoState_\autoAccept,n_\autoAccept)\in B_\productAccept$ in that BSCC with probability 1. By~\cref{eq_lemma}, there exists $(\state_\autoAccept,\autoState_\autoAccept,n_3)\in B_\autoAccept$ reachable from $(\state,\autoState,n)$ with probability 1 and, by the definition of accepting states in \cref{def:productMDP}, $(\state_\autoAccept,\autoState_\autoAccept,n_3)$ is also accepting, which means accepting states can be reached from any state in $B_\autoAccept$ with probability 1. Together with the fact that $B_\autoAccept$ is a sink set, we conclude that accepting states can be reached infinitely often with probability 1 from $B_\autoAccept$, which means $\probP_\policy(\productState\models\textsf{\upshape G}\textsf{\upshape F }\LTL_\autoAccept)=1 \;\forall \productState\in B_\autoAccept$. Another observation is that, since the counter values are increasing for all paths and accepting states will be reached infinitely often in accepting BSCCs, all states in accepting BSCCs must have a counter value equal to $K$.

Using the same argument, we conclude that $B_\varnothing$ is also a sink set. In addition, for $(\state,\autoState,n)\in B_\varnothing$, there exists $(\state,\autoState,n^\prime)\in B^\times_\varnothing$ that belongs to a rejecting BSCC such that no accepting states can be reached from it. By~\cref{eq_lemma} and the accepting states definition of product MDP (\cref{def:productMDP}), we see that no accepting state can be reached from $(\state,\autoState,n)$ either, which implies $\probP_\policy(\productState\models\textsf{\upshape G}\textsf{\upshape F }\LTL_\autoAccept)=0\;\forall\productState\in B_\varnothing$ and $B_\varnothing\cap\productAccept=\varnothing$, which completes the proof.
\end{proof}

\subsection{Proof of \cref{thm:optimality}}
\label{proof:optimality}

\begin{repeatthm}{thm:optimality}
Given an LTL formula $\LTL$ and a product MDP $\productMDP$, if \cref{sec-ltl:finite} holds, then there exists an upper bound $U\in(0,1]$ for rewards and a discount factor $\discount\in(0,1]$ such that, for all product rewards $\productReward$ and product discount functions $\productDiscount$ satisfying \cref{reward_defn}, the optimal deterministic memoryless policy $\policy_\reward$ that maximises the expected discounted reward $G^{\policy_\reward}_0(\productState_0)$ is also an optimal policy $\policy_\LTL$ that maximises the probability of satisfying the Büchi condition $\probP_{\policy_\LTL}(\productState_0\models\textsf{\upshape G}\textsf{\upshape F }\LTL_\autoAccept)$ on the product MDP $\productMDP$.
\end{repeatthm}

\begin{proof}
To give an outline of the proof, we would like to show that, for any deterministic memoryless policy $\policy$, the expected discounted reward $G^\policy_0(\productState)$ for all product states with counter value 0 (i.e., $\productState=(\state,\autoState,0)$) is close to the probability $\probP_\policy(\productState\models\textsf{\upshape G}\textsf{\upshape F }\LTL_\autoAccept)$ of satisfying $\LTL_\autoAccept$ starting from $\productState$ following $\policy$. We show this by upper and lower bounding the difference between the two quantities.

From $\productState=(\state,\autoState,0)$, we consider the expected discounted reward conditioned on whether the infinite path following policy $\policy$ satisfies the Büchi condition $\LTL_\autoAccept$ or not,
\begin{align}
G^\policy_0(\productState)&=G^\policy_0(\productState\mid\productState\models\textsf{\upshape G}\textsf{\upshape F}\LTL_\autoAccept)\probP_\policy(\productState\models\textsf{\upshape G}\textsf{\upshape F}\LTL_\autoAccept)\label{satisfy_condition}\\
&+G^\policy_0(\productState\mid\productState\not\models\textsf{\upshape G}\textsf{\upshape F}\LTL_\autoAccept)\probP_\policy(\productState\not\models\textsf{\upshape G}\textsf{\upshape F}\LTL_\autoAccept).\label{unsatisfy_condition}
\end{align}

We first consider component \eqref{satisfy_condition}. Let the stopping time of first reaching the accepting set $B_\autoAccept$ from $\productState$ be $\tau^{\policy}_{\autoAccept}\coloneqq\inf\{t>0\mid\productState_t\in B_\autoAccept\}$ and let the first reached state in $B_\autoAccept$ be $(\state,\autoState,n)$. Then, let the hitting time between accepting states in the accepting BSCC $B$ containing $(\state,\autoState,n)\in B^\times_\autoAccept$ be $\tau^{\policy}_{BSCC}$. Furthermore, let ${c^{\policy}_{\autoAccept acc}\coloneqq\left\lvert t\leq\tau^{\policy}_\autoAccept : \productState_t\in\productAccept\right\rvert}$ and ${c^{\policy}_{\autoAccept rej}\coloneqq\left\lvert t\leq\tau^{\policy}_\autoAccept : \productState_t\notin\productAccept\right\rvert}$ be the number of accepting and non-accepting states reached before reaching $B_\autoAccept$ from $\productState$ respectively, where ${c^{\policy}_{\autoAccept acc}+c^{\policy}_{\autoAccept rej}=\tau^{\policy}_\autoAccept}$. Note that all the quantities above depend on the starting state $\productState$, but for convenience, we omit it in the notations. Then, we have
\begin{align}
&G^\policy_0(\productState\mid\productState\models\textsf{\upshape G}\textsf{\upshape F}\LTL_\autoAccept)\nonumber\\
&=\expectE_\policy\left[\sum^\infty_{i=0} (\prod_{j=0}^{i-1}\discount(\productState_j))\cdot\productReward_{i}\bigm\vert \productState_0=\productState\models\textsf{\upshape G}\textsf{\upshape F}\LTL_\autoAccept|\right]\nonumber\\
&\geq \expectE_\policy\left[\gamma^{c^{\policy}_{\autoAccept rej}}\left(\sum^{c^{\policy}_{\autoAccept acc}}_{n=0}(\prod_{j=0}^{n-1}(1-R_j))R_n\right)+\sum^\infty_{i=\tau^{\policy}_{\autoAccept}}(\prod_{j=0}^{i-1}\discount(\productState_j))\cdot\productReward_i\bigm\vert \productState_0=\productState\models\textsf{\upshape G}\textsf{\upshape F}\LTL_\autoAccept) \right]\label{sat_line2}\\
&\geq \expectE_\policy\left[\gamma^{c^{\policy}_{\autoAccept rej}}\left(\sum^{c^{\policy}_{\autoAccept acc}}_{n=0}(\prod_{j=0}^{n-1}(1-R_j))R_n\right.\right.\nonumber\\
&\quad\quad\quad\quad+\left.\left.\sum^\infty_{i=\tau^{\policy}_{\autoAccept}}(\prod_{j=0}^{c^{\policy}_{\autoAccept acc}}(1-R_j)\prod_{k=\tau^{\policy}_{\autoAccept}}^{i-1}\discount(\productState_k))\cdot\productReward_i\right)\bigm\vert \productState_0=\productState\models\textsf{\upshape G}\textsf{\upshape F}\LTL_\autoAccept)\right]\label{sat_line4}
\end{align}
\begin{align}
&\geq \expectE_\policy\left[\gamma^{c^{\policy}_{\autoAccept rej}}\left(\sum^{c^{\policy}_{\autoAccept acc}}_{n=0}(\prod_{j=0}^{n-1}(1-R_j))R_n\right.\right.\nonumber\\
&\quad\quad\quad\quad+\left.\left.\gamma^{\tau^{\policy}_{BSCC}/U}\sum^\infty_{n=c^{\policy}_{\autoAccept acc}}(\prod_{j=0}^{n-1}(1-R_j))R_n\right)\bigm\vert \productState_0=\productState\models\textsf{\upshape G}\textsf{\upshape F}\LTL_\autoAccept)\right]\label{sat_line7}\\
    &\geq \gamma^{C^{\policy}_{\autoAccept}+N^\policy/U} \left(\sum^\infty_{n=0}(\prod_{j=0}^{n-1}(1-R_j))R_n\right)\label{sat_line8}\\
    &=\gamma^{C^{\policy}_{\autoAccept}+N^\policy/U}\cdot 1 ,\label{sat_line9}
\end{align}
where $C^{\policy}_{\autoAccept}=\expectE_\policy[c^{\policy}_{\autoAccept rej}\mid\productState_0=\productState\models\textsf{\upshape G}\textsf{\upshape F}\LTL_\autoAccept)]$, $N^\policy=\expectE_\policy[\tau^{\policy}_{BSCC}\mid\productState_0=\productState\models\textsf{\upshape G}\textsf{\upshape F}\LTL_\autoAccept)]$, $R_n$ is the nonzero rewards in \cref{reward_defn} and we slightly abuse the notation to let $R_n=R_K \;\forall n>K$. The inequality in \cref{sat_line2} holds because, for the first $\tau^{\policy}_{\autoAccept}$ elements in the sum, there are at most $c^{\policy}_{\autoAccept rej}$ zero reward discount $\gamma$ terms in the product and taking them out of the sum leaves only the nonzero reward terms. \cref{sat_line4} holds similarly by first taking $c^{\policy}_{\autoAccept rej}$ discount $\gamma$ terms out of the product for the rest of the sum, leaving the $(1-R_j)$ terms and the rest of the discount factors $\discount(\productState_k)$. \cref{sat_line7} holds by \cref{useful_lemma} and the observation that, after reaching the accepting BSCC, each nonzero reward will receive $\gamma^{\tau^{\policy}_{BSCC}}$ additional discount between accepting states. In addition, by summing the infinite sequences we find $U/(1-\gamma^{\tau^{\policy}_{BSCC}}(1-U))\leq \gamma^{\tau^{\policy}_{BSCC}/U}$ for $\gamma\in(0,1)$ and $U\in(0,1)$, upper bounding the additional discount and leaving only the nonzero reward discount factors $1-R_j$. Finally, \cref{sat_line9} holds by induction because the infinite geometric sum $\sum^\infty_{n=0}(1-R_K)^n R_K=1$ and $R_n+(1-R_n)*1=1$ for all $n<K$.

Intuitively, the expected discounted reward for paths with only accepting states is 1, and we lower bound the expected discounted reward for general paths satisfying $\LTL_\autoAccept$ by bounding the amount of discount the reward receives from nonzero reward states.

Next, we consider the second component \eqref{unsatisfy_condition}. Let the stopping time of first reaching the rejecting set $B_\varnothing$ from $\productState$ be
$\tau^{\policy}_\varnothing\coloneqq\inf\{t>0\mid\productState_t\in B_\varnothing\}$, and let the number of accepting states reached before reaching $B_\varnothing$ from $\productState$ be ${c^{\policy}_{\varnothing acc}\coloneqq\left\lvert t\leq\tau^{\policy}_\varnothing: \productState_t\in\productAccept\right\rvert}$. We similarly omit the dependency on $\productState$ in the notation. Then, we have
\begin{align}
G^\policy_0(\productState\mid\productState\not\models\textsf{\upshape G}\textsf{\upshape F}\LTL_\autoAccept)&=\expectE_\policy\left[ \sum^\infty_{i=0}(\prod_{j=0}^{i-1}\discount(\productState_j))\cdot\productReward_{i}\bigm\vert \productState_0=\productState\not\models\textsf{\upshape G}\textsf{\upshape F}\LTL_\autoAccept)\right]\nonumber\\
    &\leq \expectE_\policy\left[\sum^{c^{\policy}_{\varnothing acc}}_{n=0}(\prod_{j=0}^{n-1}(1-R_j))\cdot R_n \bigm\vert \productState_0=\productState\not\models\textsf{\upshape G}\textsf{\upshape F}\LTL_\autoAccept)\right]\label{unsat:line2}\\
    &= \sum^{C^\policy_\varnothing}_{n=0}(\prod_{j=0}^{n-1}(1-R_j))\cdot R_n\label{unsat:line3}\\
    &\leq \sum^{C^\policy_\varnothing}_{n=0}(1-U)^{n}\cdot U\label{unsat:line4}\\
    &=1-(1-U)^{C^\policy_\varnothing},\nonumber
\end{align}
where $C^\policy_\varnothing=\expectE_\policy[c^{\policy}_{\varnothing acc}\mid \productState_0=\productState\not\models\textsf{\upshape G}\textsf{\upshape F}\LTL_\autoAccept]$. In \cref{unsat:line2}, the inequality holds because, by \cref{useful_lemma}, once $B_\varnothing$ is reached, no further accepting states with nonzero rewards can be reached and the total reward can be bounded above by omitting the discount factor $\discount$ from non-accepting states and summing the discounted reward only for the $c^{\policy}_{\varnothing acc}$ accepting states reached before $B_\varnothing$. \cref{unsat:line3} holds by taking expectation on $\tau^{\policy}_\varnothing$ and 
\cref{unsat:line4} holds by induction because $R_n+(1-R_n)*X_1\leq U+(1-U)*X_2$ for all $n\in [0..K]$ if $X_1\leq X_2$, and $U\geq R_n$ is an upper bound for all rewards. Intuitively, we have bounded the amount of nonzero reward received by paths not satisfying $\LTL_\autoAccept$.

Therefore, from \cref{satisfy_condition} and these technical results, we have a lower bound for the expected discounted reward,
\begin{align}
G^\policy_0(\productState)\geq\gamma^{C^{\policy}_{\autoAccept}+N^\policy/U}\cdot\probP_\policy(\productState\models\textsf{\upshape G}\textsf{\upshape F}\LTL_\autoAccept),
\label{lower_bounds}
\end{align}
by assuming $G^\policy_0(\productState\mid\productState\not\models\textsf{\upshape G}\textsf{\upshape F}\LTL_\autoAccept)=0$, and an upper bound of the expected discounted reward:    
\begin{align}
    G^\policy_0(\productState)&\leq\probP_\policy(\productState\models\textsf{\upshape G}\textsf{\upshape F}\LTL_\autoAccept)+(1-(1-U)^{C^\policy_\varnothing})\cdot\probP_\policy(\productState\not\models\textsf{\upshape G}\textsf{\upshape F}\LTL_\autoAccept)\\
    &=\probP_\policy(\productState\models\textsf{\upshape G}\textsf{\upshape F}\LTL_\autoAccept)+(1-(1-U)^{C^\policy_\varnothing})\cdot(1-\probP_\policy(\productState\models\textsf{\upshape G}\textsf{\upshape F}\LTL_\autoAccept))\\
    &=1-(1-U)^{C^\policy_\varnothing}+\probP_\policy(\productState\models\textsf{\upshape G}\textsf{\upshape F}\LTL_\autoAccept)\cdot (1-U)^{C^\policy_\varnothing}.
\label{upper_bounds}
\end{align}

Last but not least, since the deterministic memoryless policy is considered in a finite MDP, there is a finite set of policies and we let the difference in probability of satisfying $\LTL_\autoAccept$ between the optimal policy and the best sub-optimal policy be
$\delta\coloneqq\probP_{\policy_\LTL}(\productState_0\models\textsf{\upshape G}\textsf{\upshape F }\LTL_\autoAccept)-\max_{\{\policy\neq\policy_\LTL\}}\probP_\policy(\productState_0\models\textsf{\upshape G}\textsf{\upshape F }\LTL_\autoAccept)$. We can let the reward upper bound $U\in(0,1)$ be small enough and $\discount\in(0,1)$ large enough such that following the bounds of \cref{lower_bounds} and \cref{upper_bounds}, we have that,
\begin{equation} G^\policy_0(\productState)\leq\max_{\policy\neq\policy_\LTL}\{\probP_\policy(\productState_0\models\textsf{\upshape G}\textsf{\upshape F }\LTL_\autoAccept)\}+\delta=\probP_{\policy_\LTL}(\productState_0\models\textsf{\upshape G}\textsf{\upshape F }\LTL_\autoAccept)\leq G^{\policy_\LTL}_0(\productState) \quad\forall \policy\neq\policy_\LTL.
\end{equation}
This means the expected discounted reward for the optimal policy $\policy_\LTL$ must be greater than the expected discounted reward received by any sub-optimal policy. Therefore, the policy $\policy_\reward$ that maximises the total expected discounted reward $G^{\policy_\reward}_0(\productState)$ is also the optimal policy $\policy_\LTL$ that maximises the probability of satisfying the Büchi condition $\LTL_\autoAccept$ in the product MDP $\productMDP$, which completes the proof.

For choosing the key hyperparameters $U\in(0,1)$ and $\discount\in(0,1)$ to ensure optimality, if we assume a reasonable gap $\delta\approx0.5$ between optimal and suboptimal policies, it generally suffices to let $U=1/C^\policy_\varnothing$ and $\discount=1-1/(C^\policy_\varnothing*N^\policy+C^\policy_\autoAccept)$. For example, with the probabilistic gate MDP in \cref{fig:example_task}, we have $C^\policy_\varnothing\approx C^\policy_\autoAccept\approx 10$ and $N^\policy=1$, so choosing $U=0.1$ and $\gamma=0.95$ is sufficient. If the hitting times and stopping times are not obtainable, under the common assumption for MDPs that the number of states $\abs{\states}$ in $\MDP$ and the minimum nonzero transition probability $p_{\min}\coloneqq \min_{s,a,s^\prime}\{\transitions(\state,\action,\state^\prime)>0\}$ are known, $C^\policy_\varnothing$ and $C^\policy_\autoAccept$ can be upper bounded by $\abs{\states}/p_{\min}$ and $N^\policy$ can be upper bounded by $\abs{\states}$.
\end{proof}

\subsection{Proof of \cref{thm:q_learning}}
\label{proof:q_learning}

\begin{repeatthm}{thm:q_learning}
Given an environment MDP $\MDP$ and an LTL specification $\LTL$ with an appropriate discount factor $\gamma$ and a reward function $\productReward$ satisfying \cref{thm:optimality}, then if \cref{sec-ltl:finite} holds, Q-learning for LTL described in \cref{alg:Q_learning} converges to an optimal policy $\policy_\LTL$ that maximises the probability of satisfying $\LTL$ in $\MDP$.
\end{repeatthm}

\begin{proof}
To begin with, recall from \cref{thm:equivalence} and \cref{thm:optimality} that the policy $\policy_r$ maximising the expected discounted reward of $\productMDP$ induces the optimal policy for satisfying $\LTL$ in $\MDP$ by removing the $\epsilon$-actions.

Next, note that, despite the reward for Q-learning in \cref{alg:Q_learning} being non-Markovian, following the proof of convergence to the optimal Q value for non-Markovian Q-learning~\citep{Majeed2018OnProcesses}, with the $K$ counter in \cref{def:productMDP} and the reward in \cref{reward_defn}, we verify that the Q function in \cref{alg:Q_learning} converges to the optimal Q value for all $(\state,\autoState)$ and action $\productAction$, which is the expected discounted reward by taking action $\productAction$ from $(\state,\autoState,0)$.

With this, recall that the environment states and automaton state transitions do not depend on the counter value by \cref{eq_lemma} and the nonzero reward is the same for accepting states with the same counter value by \cref{reward_defn}. We argue that, for the optimal discounted reward policy $\policy_r$, the new policy $\policy^\prime((\state,\autoState,n))=\policy_r((\state,\autoState,0))\;\forall n\in[0..K]$, where the policy for each counter value is set to be the same, remains optimal. This is because if $\policy^\prime$ is suboptimal, meaning that the satisfiability of $\LTL_\autoAccept$ is lower than that of the optimal policy, it must be that $\policy^\prime$ makes a wrong decision at some counter value such that the reachability probability to some accepting BSCC $B$ is suboptimal. This means $\policy_r$ must also be suboptimal because when the counter value equals 0, by taking the same actions as $\policy^\prime$, the reachability probability to the BSCC $B$ is suboptimal, producing a contradiction. Lastly, since $\policy^\prime$ is independent of the counter values, the policy derived from the collapsed optimal Q function learnt by \cref{alg:Q_learning} is the same as the optimal $\policy^\prime$ by ignoring the $K$ counter, which means that \cref{alg:Q_learning} converges to the optimal policy, completing the proof.
\end{proof}

\section{Additional Experimental Details and Setup}
\label{appendix:exp_setup}

The experiments are carried out on a Linux server (Ubuntu 18.04.2) with two Intel Xeon Gold 6252 CPUs and six NVIDIA GeForce RTX 2080 Ti GPUs. All of our algorithms are implemented in Python. We select three stochastic environments (probabilistic gate, frozen lake, and office world) described below with several difficult LTL tasks. For all environments and tasks, we set the learning rate $\alpha=0.1$ and the exploration rate $\epsilon=0.1$, with the upper bound on the rewards $U=0.1$ and the discount factor $\gamma=0.99$ to ensure optimality. For the method proposed by \citet{Hahn2019Omega-regularLearning}, we adopt their implementation from the Mungojerrie tool\footnote{https://plv.colorado.edu/wwwmungojerrie/docs/v1\_0/index.html}. We set the default hyperparameter values to $\alpha=0.1$ and $\epsilon=0.1$, with $\zeta=0.995$ for the probabilistic gate task and $\zeta=0.99$ for the frozen lake and office world tasks. For the method by \citet{Hasanbeig2020DeepLogics}, we use the implementation in their GitHub repository\footnote{https://github.com/grockious/lcrl} with $\gamma=0.99$, $\alpha=0.1$ and $\epsilon=0.1$.
For the method of \citet{Bozkurt2019ControlLearning}, we also adopt their GitHub repository\footnote{https://github.com/alperkamil/csrl} with $\gamma=0.99$, $\gamma_B=0.9$, $\alpha=0.1$ and $\epsilon=0.1$.

For the environments and tasks used for experimentation, we specifically choose commonly used environments with stochastic transitions that are more realistic and challenging, especially for infinite-horizon LTL tasks. The LTL tasks that we use are standard task specifications for robotics and autonomous systems. The tasks for the frozen lake and the office world environments are more challenging because they involve letting the agent choose between two subtasks, which better showcase the benefits of the $K$ counter and counterfactual imagining.

\subsection{Probabilistic Gate}
\label{appendix:probablistic}

\begin{figure*}[tb]
\centering
\includegraphics[width=0.6\textwidth]{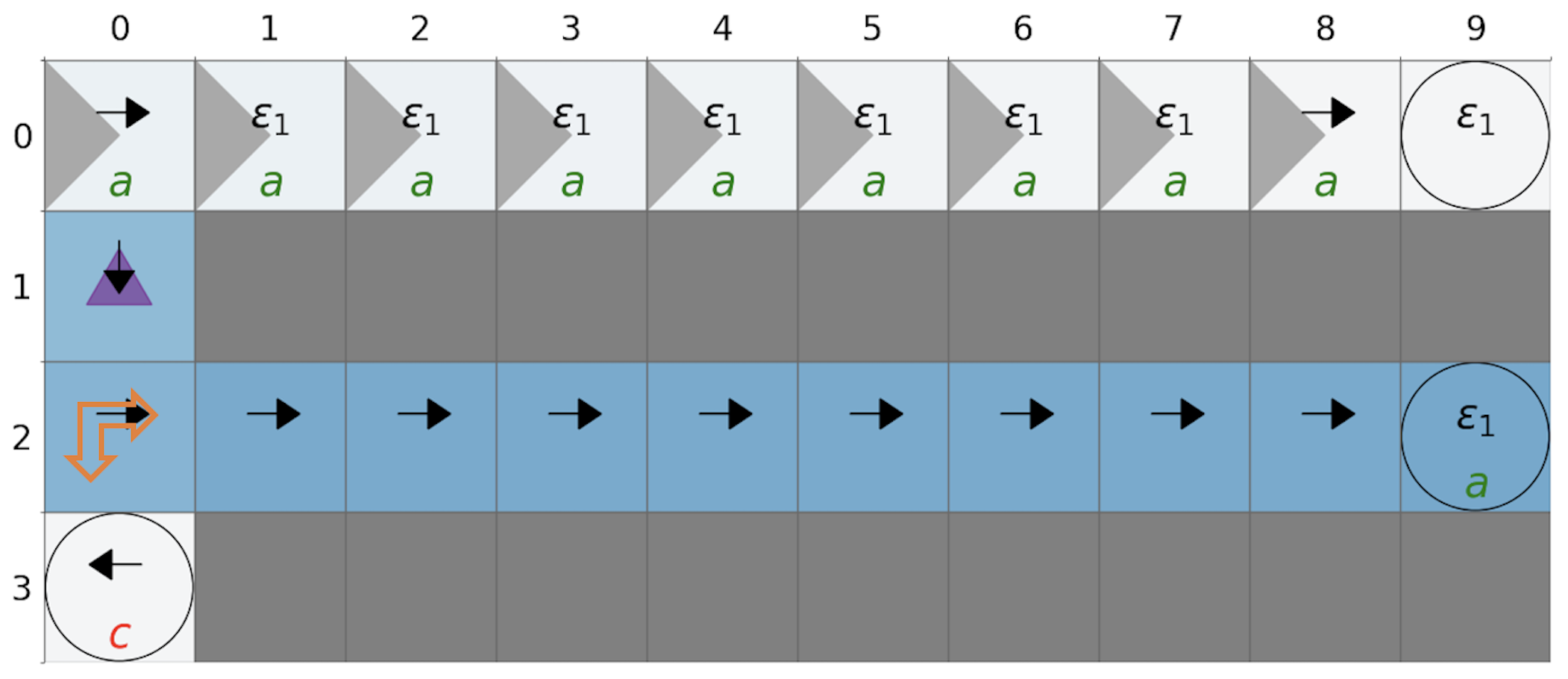}
\caption[The optimal policy for the probabilistic gate MDP task]{The optimal policy for the example probabilistic gate MDP task in \cref{fig:example_task}}
\label{fig:hard1_policy}
\end{figure*}

The probabilistic gate MDP is described in \cref{example_mdp}, where the task is to visit states labelled \enquote{a} infinitely often without visiting states labelled \enquote{c}. We set the episode length to be 100 (which is the maximum length of a path in the MDP) and the number of episodes to be 40000. In \cref{fig:hard1_policy}, the optimal policy is demonstrated by the black arrows, where the agent goes down from the start and goes all the way to the right to (2,9), which is an accepting sink state.

\subsection{Frozen Lake}
\label{appendix:frozenlake}

\begin{figure*}[tb]
\centering
\begin{subfigure}[t]{0.45\textwidth}
\centering
\includegraphics[width=0.8\textwidth]{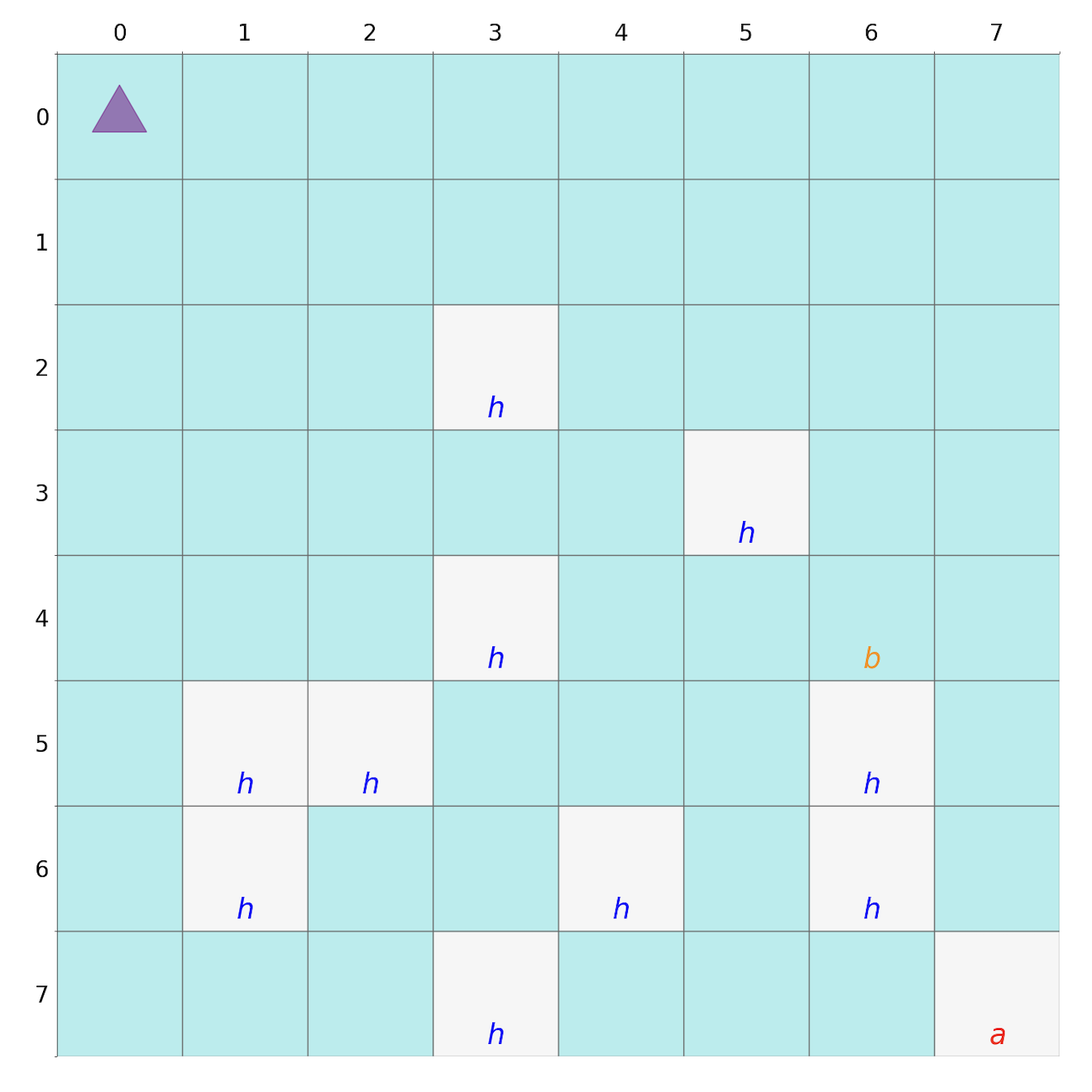}
\caption{The MDP environment for the frozen lake task. Blue represents ice, \enquote{h} are holes, \enquote{a} and \enquote{b} are lake camps, and the purple triangle is the start.}
\label{fig:frozen_env}
\end{subfigure}
\hspace{0.5cm}
\begin{subfigure}[t]{0.5\textwidth}
\centering
\includegraphics[width=\textwidth]{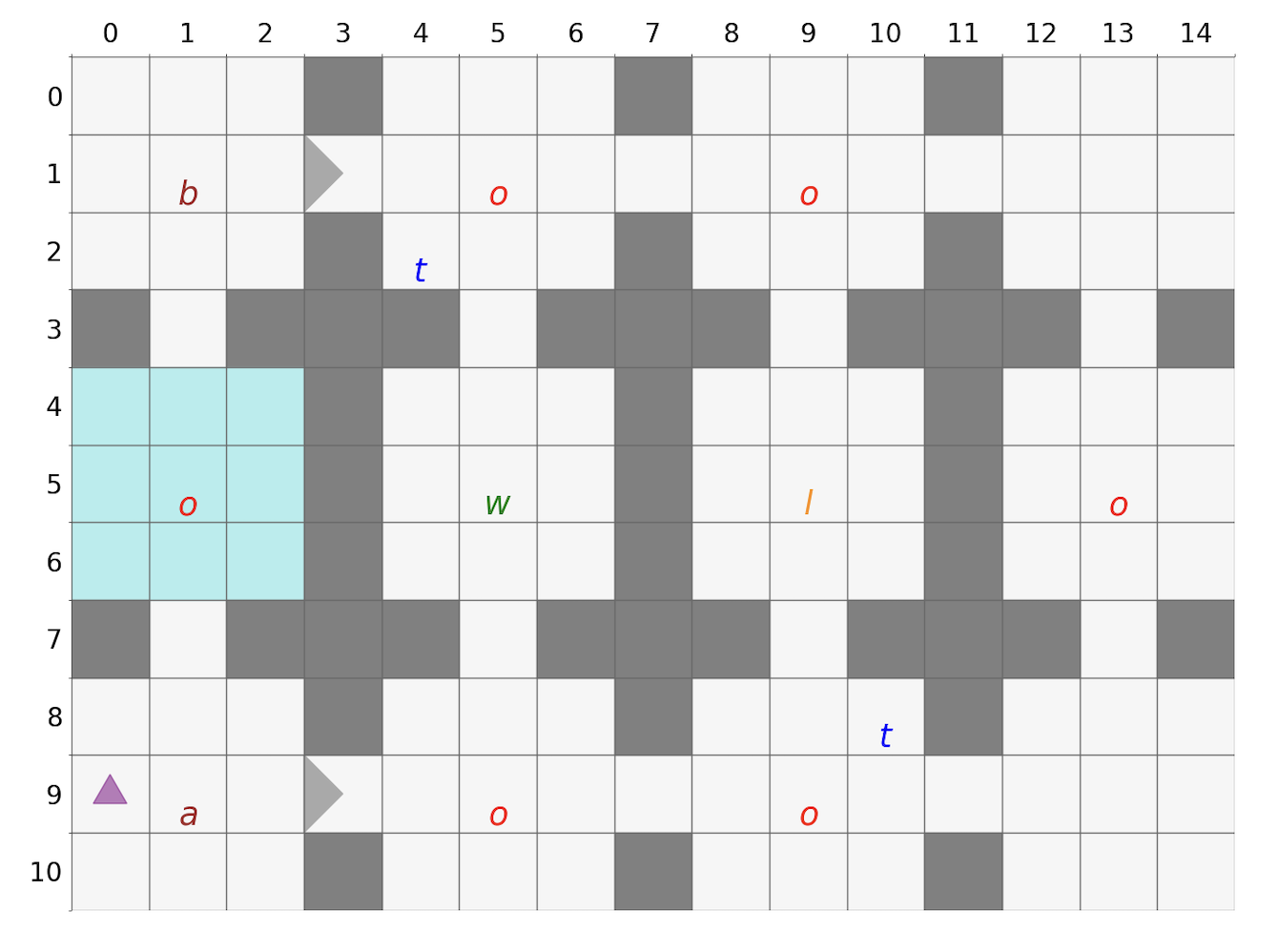}
\caption{The MDP environment for the office world task. Blue represents ice, \enquote{o} are obstacles, \enquote{w}, \enquote{l} and \enquote{t} represent workplace, letter and tea respectively, and the purple triangle is the start.}
\label{fig:office_env}
\end{subfigure}
\caption{MDP environments used in the experiments.}
\end{figure*}

The frozen lake~\citep{Brockman2016OpenAIGym} environment is shown in \cref{fig:frozen_env}. The blue states represent the frozen lake, where the agent has a 1/3 probability of moving in the intended direction and 1/3 each of going sideways (left or right). The white states with label \enquote{h} are holes and states with label \enquote{a} and \enquote{b} are lake camps. The task is \enquote{(\textsf{\upshape G}\textsf{\upshape F} a $\mid$ \textsf{\upshape G}\textsf{\upshape F} b) \& \textsf{\upshape G} !h}, meaning to always reach lake camp \enquote{a} or lake camp \enquote{b} while never falling into holes \enquote{h}. For this task, we set the episode length to 200 and the number of episodes to 6000.

\subsection{Office World}
\label{appendix:officeworld}

The office world~\citep{Icarte2022RewardLearning} environment is demonstrated in \cref{fig:office_env}. We include a patch of ice labelled blue as defined in the frozen lake environment and two one-directional gates at $(1,3)$ and $(9,3)$, where the agent can only cross to the right, as shown by the direction of the grey triangle. The grey blocks are walls, the states labelled \enquote{o}, \enquote{l}, \enquote{t} and \enquote{w} are obstacles, letters, tea, and the workplace, respectively. The task for this environment is also demanding: {\enquote{(\textsf{\upshape G}\textsf{\upshape F} a \& \textsf{\upshape G}\textsf{\upshape F} b) $\mid$ (\textsf{\upshape F} l \& \textsf{\upshape X} (\textsf{\upshape G}\textsf{\upshape F} t \& \textsf{\upshape G}\textsf{\upshape F} w)) \& \textsf{\upshape G} !o}}. This means to either patrol in the corridor between \enquote{a} and \enquote{b}, or go to write a letter at \enquote{l} and then patrol between getting tea \enquote{t} and the workplace \enquote{w}, whilst never hitting obstacles \enquote{o}. For this task, we set the length of the episode to 1000 and the number of episodes to 6000.

\subsection{PRISM}
We use PRISM version 4.7~\citep{Kwiatkowska2011PRISMSystems} to evaluate the learnt policy against the LTL tasks. The PRISM tool and the installation documentation can be obtained from their official website\footnote{https://www.prismmodelchecker.org/download.php}. Each time we would like to evaluate the policy on the environment MDP, our Python code automatically constructs the induced Markov chain from the MDP and the policy as a discrete-time Markov chain in the PRISM language, and evaluates this PRISM model against the LTL task specification through a call from Python. The maximum iteration parameter for PRISM is set to 100,000, and we evaluate the current policy every 10,000 training steps to plot the training graph for all our experiments.

\subsubsection{Example PRISM model}
We provide an example PRISM model of the induced Markov chain from the probabilistic gate MDP and its optimal policy.
\begin{lstlisting}[caption={The PRISM model of the induced Markov chain from the probabilistic gate MDP and its optimal policy.},label={lst:PRISM},language=c++]
dtmc
module ProductMDP
    m : [0..40] init 10;
    a : [0..3] init 0;
    [ep] (m=0)&(a=0) -> (m'=0)&(a'=1);
    [ac] (m=0)&(a=1) -> 1 : (m'=1)&(a'=1);
    [ac] (m=0)&(a=2) -> 1 : (m'=1)&(a'=2);
    [ep] (m=1)&(a=0) -> (m'=1)&(a'=1);
    [ac] (m=1)&(a=1) -> 1 : (m'=2)&(a'=1);
    [ac] (m=1)&(a=2) -> 1 : (m'=2)&(a'=2);
    [ep] (m=2)&(a=0) -> (m'=2)&(a'=1);
    [ac] (m=2)&(a=1) -> 1 : (m'=3)&(a'=1);
    [ac] (m=2)&(a=2) -> 1 : (m'=3)&(a'=2);
    [ep] (m=3)&(a=0) -> (m'=3)&(a'=1);
    [ac] (m=3)&(a=1) -> 1 : (m'=4)&(a'=1);
    [ac] (m=3)&(a=2) -> 1 : (m'=4)&(a'=2);
    [ep] (m=4)&(a=0) -> (m'=4)&(a'=1);
    [ac] (m=4)&(a=1) -> 1 : (m'=5)&(a'=1);
    [ac] (m=4)&(a=2) -> 1 : (m'=5)&(a'=2);
    [ep] (m=5)&(a=0) -> (m'=5)&(a'=1);
    [ac] (m=5)&(a=1) -> 1 : (m'=6)&(a'=1);
    [ac] (m=5)&(a=2) -> 1 : (m'=6)&(a'=2);
    [ep] (m=6)&(a=0) -> (m'=6)&(a'=1);
    [ac] (m=6)&(a=1) -> 1 : (m'=7)&(a'=1);
    [ac] (m=6)&(a=2) -> 1 : (m'=7)&(a'=2);
    [ep] (m=7)&(a=0) -> (m'=7)&(a'=1);
    [ac] (m=7)&(a=1) -> 1 : (m'=8)&(a'=1);
    [ac] (m=7)&(a=2) -> 1 : (m'=8)&(a'=2);
    [ep] (m=8)&(a=0) -> (m'=8)&(a'=1);
    [ac] (m=8)&(a=1) -> 1 : (m'=9)&(a'=1);
    [ac] (m=8)&(a=2) -> 1 : (m'=9)&(a'=2);
    [ep] (m=9)&(a=0) -> (m'=9)&(a'=1);
    [ac] (m=9)&(a=1) -> 1.0 : (m'=9)&(a'=2);
    [ac] (m=9)&(a=2) -> 1.0 : (m'=9)&(a'=2);
    [ac] (m=10)&(a=0) -> 1 : (m'=20)&(a'=0);
    [ac] (m=10)&(a=1) -> 1 : (m'=20)&(a'=2);
    [ac] (m=10)&(a=2) -> 1 : (m'=20)&(a'=2);
    [ac] (m=11)&(a=0) -> 1.0 : (m'=11)&(a'=0);
    [ac] (m=11)&(a=1) -> 1.0 : (m'=11)&(a'=2);
    [ac] (m=11)&(a=2) -> 1.0 : (m'=11)&(a'=2);
    [ac] (m=12)&(a=0) -> 1.0 : (m'=12)&(a'=0);
    [ac] (m=12)&(a=1) -> 1.0 : (m'=12)&(a'=2);
    [ac] (m=12)&(a=2) -> 1.0 : (m'=12)&(a'=2);
    [ac] (m=13)&(a=0) -> 1.0 : (m'=13)&(a'=0);
    [ac] (m=13)&(a=1) -> 1.0 : (m'=13)&(a'=2);
    [ac] (m=13)&(a=2) -> 1.0 : (m'=13)&(a'=2);
    [ac] (m=14)&(a=0) -> 1.0 : (m'=14)&(a'=0);
    [ac] (m=14)&(a=1) -> 1.0 : (m'=14)&(a'=2);
    [ac] (m=14)&(a=2) -> 1.0 : (m'=14)&(a'=2);
    [ac] (m=15)&(a=0) -> 1.0 : (m'=15)&(a'=0);
    [ac] (m=15)&(a=1) -> 1.0 : (m'=15)&(a'=2);
    [ac] (m=15)&(a=2) -> 1.0 : (m'=15)&(a'=2);
    [ac] (m=16)&(a=0) -> 1.0 : (m'=16)&(a'=0);
    [ac] (m=16)&(a=1) -> 1.0 : (m'=16)&(a'=2);
    [ac] (m=16)&(a=2) -> 1.0 : (m'=16)&(a'=2);
    [ac] (m=17)&(a=0) -> 1.0 : (m'=17)&(a'=0);
    [ac] (m=17)&(a=1) -> 1.0 : (m'=17)&(a'=2);
    [ac] (m=17)&(a=2) -> 1.0 : (m'=17)&(a'=2);
    [ac] (m=18)&(a=0) -> 1.0 : (m'=18)&(a'=0);
    [ac] (m=18)&(a=1) -> 1.0 : (m'=18)&(a'=2);
    [ac] (m=18)&(a=2) -> 1.0 : (m'=18)&(a'=2);
    [ac] (m=19)&(a=0) -> 1.0 : (m'=19)&(a'=0);
    [ac] (m=19)&(a=1) -> 1.0 : (m'=19)&(a'=2);
    [ac] (m=19)&(a=2) -> 1.0 : (m'=19)&(a'=2);
    [ac] (m=20)&(a=0) -> 0.8 : (m'=21)&(a'=0) + 0.2 : (m'=30)&(a'=0);
    [ac] (m=20)&(a=1) -> 0.8 : (m'=21)&(a'=2) + 0.2 : (m'=30)&(a'=2);
    [ac] (m=20)&(a=2) -> 0.8 : (m'=21)&(a'=2) + 0.2 : (m'=30)&(a'=2);
    [ac] (m=21)&(a=0) -> 1 : (m'=22)&(a'=0);
    [ac] (m=21)&(a=1) -> 1.0 : (m'=21)&(a'=2);
    [ac] (m=21)&(a=2) -> 1.0 : (m'=21)&(a'=2);
    [ac] (m=22)&(a=0) -> 1 : (m'=23)&(a'=0);
    [ac] (m=22)&(a=1) -> 1.0 : (m'=22)&(a'=2);
    [ac] (m=22)&(a=2) -> 1.0 : (m'=22)&(a'=2);
    [ac] (m=23)&(a=0) -> 1 : (m'=24)&(a'=0);
    [ac] (m=23)&(a=1) -> 1.0 : (m'=23)&(a'=2);
    [ac] (m=23)&(a=2) -> 1.0 : (m'=23)&(a'=2);
    [ac] (m=24)&(a=0) -> 1 : (m'=25)&(a'=0);
    [ac] (m=24)&(a=1) -> 1.0 : (m'=24)&(a'=2);
    [ac] (m=24)&(a=2) -> 1.0 : (m'=24)&(a'=2);
    [ac] (m=25)&(a=0) -> 1 : (m'=26)&(a'=0);
    [ac] (m=25)&(a=1) -> 1.0 : (m'=25)&(a'=2);
    [ac] (m=25)&(a=2) -> 1.0 : (m'=25)&(a'=2);
    [ac] (m=26)&(a=0) -> 1 : (m'=27)&(a'=0);
    [ac] (m=26)&(a=1) -> 1.0 : (m'=26)&(a'=2);
    [ac] (m=26)&(a=2) -> 1.0 : (m'=26)&(a'=2);
    [ac] (m=27)&(a=0) -> 1 : (m'=28)&(a'=0);
    [ac] (m=27)&(a=1) -> 1.0 : (m'=27)&(a'=2);
    [ac] (m=27)&(a=2) -> 1.0 : (m'=27)&(a'=2);
    [ac] (m=28)&(a=0) -> 1 : (m'=29)&(a'=0);
    [ac] (m=28)&(a=1) -> 1 : (m'=27)&(a'=2);
    [ac] (m=28)&(a=2) -> 1 : (m'=27)&(a'=2);
    [ep] (m=29)&(a=0) -> (m'=29)&(a'=1);
    [ac] (m=29)&(a=1) -> 1.0 : (m'=29)&(a'=1);
    [ac] (m=29)&(a=2) -> 1.0 : (m'=29)&(a'=2);
    [ep] (m=30)&(a=0) -> (m'=30)&(a'=1);
    [ac] (m=30)&(a=1) -> 1.0 : (m'=30)&(a'=2);
    [ac] (m=30)&(a=2) -> 1.0 : (m'=30)&(a'=2);
    [ac] (m=31)&(a=0) -> 1.0 : (m'=31)&(a'=0);
    [ac] (m=31)&(a=1) -> 1.0 : (m'=31)&(a'=2);
    [ac] (m=31)&(a=2) -> 1.0 : (m'=31)&(a'=2);
    [ac] (m=32)&(a=0) -> 1.0 : (m'=32)&(a'=0);
    [ac] (m=32)&(a=1) -> 1.0 : (m'=32)&(a'=2);
    [ac] (m=32)&(a=2) -> 1.0 : (m'=32)&(a'=2);
    [ac] (m=33)&(a=0) -> 1.0 : (m'=33)&(a'=0);
    [ac] (m=33)&(a=1) -> 1.0 : (m'=33)&(a'=2);
    [ac] (m=33)&(a=2) -> 1.0 : (m'=33)&(a'=2);
    [ac] (m=34)&(a=0) -> 1.0 : (m'=34)&(a'=0);
    [ac] (m=34)&(a=1) -> 1.0 : (m'=34)&(a'=2);
    [ac] (m=34)&(a=2) -> 1.0 : (m'=34)&(a'=2);
    [ac] (m=35)&(a=0) -> 1.0 : (m'=35)&(a'=0);
    [ac] (m=35)&(a=1) -> 1.0 : (m'=35)&(a'=2);
    [ac] (m=35)&(a=2) -> 1.0 : (m'=35)&(a'=2);
    [ac] (m=36)&(a=0) -> 1.0 : (m'=36)&(a'=0);
    [ac] (m=36)&(a=1) -> 1.0 : (m'=36)&(a'=2);
    [ac] (m=36)&(a=2) -> 1.0 : (m'=36)&(a'=2);
    [ac] (m=37)&(a=0) -> 1.0 : (m'=37)&(a'=0);
    [ac] (m=37)&(a=1) -> 1.0 : (m'=37)&(a'=2);
    [ac] (m=37)&(a=2) -> 1.0 : (m'=37)&(a'=2);
    [ac] (m=38)&(a=0) -> 1.0 : (m'=38)&(a'=0);
    [ac] (m=38)&(a=1) -> 1.0 : (m'=38)&(a'=2);
    [ac] (m=38)&(a=2) -> 1.0 : (m'=38)&(a'=2);
    [ac] (m=39)&(a=0) -> 1.0 : (m'=39)&(a'=0);
    [ac] (m=39)&(a=1) -> 1.0 : (m'=39)&(a'=2);
    [ac] (m=39)&(a=2) -> 1.0 : (m'=39)&(a'=2);
endmodule

label "a" = (m=0) | (m=1) | (m=2) | (m=3) | (m=4) | (m=5) | (m=6) | (m=7) | (m=8) | (m=29);
label "c" = (m=30);
\end{lstlisting}

\subsection{Rabinizer}
We use Rabinizer 4~\citep{Kretinsky2018RabinizerAutomaton} to transform LTL formulae into LDBAs. The tool can be downloaded on their website~\footnote{https://www7.in.tum.de/~kretinsk/rabinizer4.html}, and we use the 'ltl2ldba' script with '-e' and '-d' options to construct the LDBA with $\epsilon$-transitions from the input LTL formulae.

\end{document}